\documentclass[oneside, 12pt]{book}

\usepackage{geometry}
\usepackage{verbatim}
\usepackage{graphicx}
\usepackage{enumerate, enumitem}
\usepackage{subcaption}
\usepackage{booktabs}
\usepackage[
  colorlinks=true,
  citecolor=blue,
  linkcolor=magenta,
  urlcolor=cyan]{hyperref}
\usepackage{url}
\usepackage{xcolor}
\usepackage{amssymb,amsthm,amsmath,amsfonts,amsthm,mathtools}
\newcommand{\eg}{{\it e.g.}}
\newcommand{\ie}{{\it i.e.}}

\newcommand{\ones}{\mathbf 1}

\newcommand{\reals}{{\mbox{\bf R}}}

\newcommand{\symm}{{\mbox{\bf S}}}  %
\newcommand{\sign}{{\mbox{\bf sign}}}  %

\newcommand{\Span}{\mbox{\textrm{span}}}

\newcommand{\diag}{\mathop{\bf diag}}

\newcommand{\dist}{\mathop{\bf dist{}}}

\newcommand{\vect}{\mathop{\bf vec}}

\newcounter{algorithmctr}
\renewcommand{\thealgorithmctr}{\arabic{algorithmctr}}
\newenvironment{algdesc}%
   {\mbox{}\\*[\parskip]\begin{minipage}{\linewidth}%
       \refstepcounter{algorithmctr}\begin{list}{}{%
       \setlength{\rightmargin}{0\linewidth}%
       \setlength{\leftmargin}{.05\linewidth}}%
       \rmfamily\small
       \item[]{\setlength{\parskip}{0ex}\hrulefill\par%
        \nopagebreak{\bfseries\textsf{Algorithm \thealgorithmctr~}}}}%
   {{\setlength{\parskip}{-1ex}\nopagebreak\par\hrulefill\\*[2ex]\par}%
   \end{list}\end{minipage}}

\newcommand{\tr}{\mathop{\bf tr}}
\newcommand{\rank}{\mathop{\bf rank}}
\DeclarePairedDelimiter{\norm}{\|}{\|}

\newcommand{\edges}{\mathcal{E}}
\newcommand{\items}{\mathcal{V}}

\newcommand{\esim}{\mathcal{E}_{\text{sim}}}
\newcommand{\edis}{\mathcal{E}_{\text{dis}}}

\newcommand{\psim}{p_{\text{s}}}
\newcommand{\pdis}{p_{\text{d}}}

\newcommand{\npairs}{p}
\newcommand{\embdim}{m}

\newcommand{\constr}{\mathcal{X}}
\newcommand{\centered}{\mathcal{C}}
\newcommand{\anchored}{\mathcal{A}}
\newcommand{\standardized}{\mathcal{S}}

\newcommand{\anchors}{\mathcal{K}}
\newcommand{\stdemb}{\mathcal{S}}
\newcommand{\dnat}{d_{\mathrm{nat}}}

\newcommand{\tspace}{{\mathcal T}}
\newcommand{\pitspace}{\Pi_{\tspace_{X}}}

\newcommand{\dthresh}{\delta^{\mathrm{thres}}}
\newcommand{\wthresh}{w^{\mathrm{thres}}}

\usepackage{textcomp}
\usepackage{listings}
\definecolor{seagreen}{rgb}{0.18, 0.55, 0.34}
\definecolor{mediumviolet-red}{rgb}{0.78, 0.08, 0.52}
\definecolor{khaki}{rgb}{0.94, 0.9, 0.55}

\lstdefinelanguage{mypython}
{
	keywords=[1]{from, import, assert, not, print},
	keywordstyle=[1]{\color{mediumviolet-red}},
	keywords=[2]{pymde, torch},
	keywordstyle=[2]{\color{seagreen}},
	numbers=none,
	upquote=true,
	showstringspaces=false,
	basicstyle=\ttfamily,
	columns=fullflexible,
	keepspaces=true,
	emph={True,False,as,def,return,float},
	emphstyle={\color{seagreen}},
	frame=trBL,
	belowskip=1em,
	aboveskip=1em,
	captionpos=b
}

\usepackage[
    natbib=true,
    citestyle=alphabetic,
    style=alphabetic,
    maxnames=3,
    minnames=3,
    maxbibnames=99,
    backend=bibtex,
    urldate=iso8601,
    firstinits=true,
    isbn=false,
    doi=false,
    date=year,
    sorting=nty
]{biblatex}
\DeclareFieldFormat{eprint:arXiv}{%
  {\hypersetup{hidelinks}\href{https://www.arXiv.org/abs/#1}{arXiv:#1}}%
}
\DeclareFieldFormat[article,inbook,incollection,inproceedings,patent,thesis,
  unpublished]{citetitle}{#1}
\DeclareFieldFormat[article,inbook,incollection,inproceedings,patent,thesis,
  unpublished]{title}{#1}
\renewbibmacro{in:}{%
  \ifentrytype{article}{}{\printtext{\bibstring{in}\space}}}
\renewcommand*{\labelalphaothers}{\textsuperscript{+}}
\DefineBibliographyStrings{english}{%
  bibliography = {References},
}

\title{Minimum-Distortion Embedding}
\author{
Akshay Agrawal \\ \texttt{\small akshayka@cs.stanford.edu} \and
Alnur Ali \\ \texttt{\small alnurali@stanford.edu} \and
Stephen Boyd \\ \texttt{\small boyd@stanford.edu}
}

\begin{document}
\maketitle

\frontmatter
\clearpage
\tableofcontents
\clearpage

\chapter*{Abstract}
We consider the vector embedding problem. We are given a finite set of items,
with the goal of assigning a representative vector to each one, possibly under
some constraints (such as the collection of vectors being standardized, \ie,
having zero mean and unit covariance).
We are given data indicating that some pairs of items are similar, and optionally,
some other pairs are dissimilar.
For pairs of similar items, we want the corresponding vectors to be
near each other, and for dissimilar pairs, we want the 
vectors to not be near each other, measured in Euclidean distance.
We formalize this by introducing distortion functions, defined
for some pairs of items.
Our goal is to choose an embedding that minimizes the total
distortion, subject to the constraints. We call this the
\emph{minimum-distortion embedding} (MDE) problem.

The MDE framework is simple but general. It includes a wide variety of
specific embedding methods, such as spectral embedding, principal component
analysis, multidimensional scaling, Euclidean distance problems, dimensionality
reduction methods (like Isomap and UMAP), semi-supervised learning, sphere
packing, force-directed layout, and others. It also includes new embeddings,
and provides principled ways of validating or sanity-checking historical and new embeddings alike.

In a few special cases, MDE problems can
be solved exactly. For others, we develop a projected quasi-Newton method that
approximately minimizes the distortion and scales to very large data sets,
while placing few assumptions on the distortion functions and constraints.
This monograph is accompanied by an open-source Python package, PyMDE, for
approximately solving MDE problems. Users can select from a library of
distortion functions and constraints or specify custom ones, making it
easy to rapidly experiment with new embeddings. Because our algorithm is
scalable, and because PyMDE can exploit GPUs, our software scales to problems
with millions of items and tens of millions of distortion functions.
Additionally, PyMDE is competitive in runtime with specialized implementations
of specific embedding methods. To demonstrate our method, we compute embeddings
for several real-world data sets, including images, an academic co-author
network, US county demographic data, and single-cell mRNA transcriptomes. 

\mainmatter
\chapter{Introduction}
An embedding of $n$ items, labeled $1, \ldots, n$, is a function $F$ mapping
the set of items into $\reals^\embdim$.  We refer to $x_i=F(i)$ as the embedding vector
associated with item $i$. In applications, embeddings provide concrete
numerical representations of otherwise abstract items, for use in downstream
tasks. For example, a biologist might look for subfamilies of related cells by
clustering embedding vectors associated with individual cells, while a machine
learning practitioner might use vector representations of words as features for
a classification task. Embeddings are also used for visualizing collections of
items, with embedding dimension $\embdim$ equal to one, two, or three.

For an embedding to be useful, it should be faithful to the known relationships
between items in some way. There are many ways to define faithfulness. A
working definition of a faithful embedding is the following: 
if items $i$ and $j$ are similar, their
associated vectors $x_i$ and $x_j$ should be near each other, as measured by
the Euclidean distance $\|x_i-x_j\|_2$; if items $i$ and $j$ are dissimilar,
$x_i$ and $x_j$ should be distant, or at least not close, in Euclidean distance.
(Whether two items
are similar or dissimilar depends on the application. 
For example two biological cells might
be considered similar if some distance between their mRNA transcriptomes is
small.) 
Many well-known embedding methods like principal component analysis
(PCA), spectral embedding \cite{chung1997spectral, belkin2002laplacian}, and
multidimensional scaling \cite{torgerson1952multidimensional,
kruskal1964multidimensional} use this basic notion of faithfulness, differing
in how they make it precise.

The literature on embeddings is both vast and old. PCA originated over a
century ago \cite{pearson1901lines}, and it was further developed three
decades later in the field of psychology \cite{hotelling1933analysis,
eckart1936approximation}. Multidimensional scaling, a family
of methods for embedding items given dissimilarity scores or
distances between items, was also developed in the field of psychology during the
early-to-mid
20th century \cite{richardson1938multidimensional,
torgerson1952multidimensional, kruskal1964multidimensional}. Methods for
embedding items that are vectors can be traced back to
the early 1900s \cite{menger1928untersuchungen, young1938discussion},
and more recently developed methods use tools from convex optimization and
convex analysis \cite{biswas2004semidefinite, hayden1991cone}. In spectral
clustering, an embedding based on an eigenvector decomposition of the graph Laplacian
is used
to cluster graph vertices \cite{pothen1990partitioning, von2007tutorial}.
During this century, dozens of
embedding methods have been developed for reducing
the dimension of high-dimensional vector data, including Laplacian eigenmaps
\cite{belkin2002laplacian}, Isomap \cite{tenenbaum2000global}, locally-linear
embedding (LLE) \cite{roweis2000nonlinear}, stochastic neighborhood embedding
(SNE) \cite{hinton2003stochastic}, t-distributed stochastic neighbor embedding
(t-SNE) \cite{maaten2008visualizing}, LargeVis \cite{tang2016visualizing} and
uniform manifold approximation and projection (UMAP) \cite{mcinnes2018umap}.
All these methods start with either weights describing the similarity
of a pair of items, or distances describing their dissimilarity.

In this monograph we present a general framework for faithful embedding. The
framework, which we call \emph{minimum-distortion embedding} (MDE), generalizes the
common cases in which similarities between items are described by weights or
distances. It also includes most of the embedding methods mentioned above as
special cases. In our formulation, for some pairs of items, we are given
distortion functions of the Euclidean distance between the
associated embedding vectors. Evaluating a distortion function at the Euclidean
distance between the vectors gives the distortion of the embedding
for a pair of items. The goal is to find an embedding that minimizes the
total or average distortion, possibly subject to some constraints on the
embedding; we call this the MDE problem. We focus on three specific constraints: a centering
constraint, which requires the embedding to have mean zero, an anchoring
constraint, which fixes the positions of a subset of the embedding vectors, and
a standardization constraint, which requires the embedding to be centered
and have identity covariance.

MDE problems are in general intractable, admitting efficiently
computable (global) solutions only in a few special cases like PCA and spectral
embedding. In most other cases, MDE problems can
only be approximately solved, using heuristic methods. 
We develop one such heuristic, a projected quasi-Newton method.
The method we describe works well for a variety of MDE problems.

This monograph is accompanied by an open-source implementation for specifying MDE
problems and computing low-distortion embeddings. Our software package, PyMDE,
makes it easy for practitioners to experiment with different embeddings via
different choices of distortion functions and constraint sets. Our
implementation scales to very large datasets and to embedding dimensions that
are much larger than two or three. This means that our package can be used for both
visualizing large amounts of data and generating features for downstream
tasks. PyMDE supports GPU acceleration and
automatic differentiation of distortion functions by using PyTorch
\cite{paszke2019pytorch} as the numerical backend.

\paragraph{A preview of our framework.}
Here we give a brief preview of the MDE framework, along with a simple example
of an MDE problem. We discuss the MDE problem at length in chapter~\ref{c-mde}.

An embedding can be represented concretely by a matrix $X \in \reals^{n \times
m}$, whose rows $x_1^T, \ldots, x_n^T \in \reals^m$ are the embedding
vectors. We use $\edges$ to denote the set of pairs, and $f_{ij} : \reals_+ \to
\reals$ to denote the distortion functions for $(i, j) \in \edges$. Our goal
is to find an embedding that minimizes the average distortion
\[
  E(X) = \frac{1}{|\mathcal E|} \sum_{(i, j) \in \mathcal E} f_{ij} (d_{ij}),
\]
where $d_{ij} = \|x_i - x_j\|_2$, subject to constraints on the embedding,
expressed as $X \in \constr$, where
$\constr \subseteq \reals^{n \times m}$ is the set of allowable embeddings.
Thus the MDE problem is
\begin{equation*}
\begin{array}{ll}
\mbox{minimize} & E(X)\\
\mbox{subject to} & X \in \mathcal X.
\end{array}
\end{equation*}
We solve this problem, sometimes approximately, to find an embedding.

An important example is the quadratic MDE problem with
standardization constraint.
In this problem the distortion functions
are quadratic $f_{ij}(d_{ij}) = w_{ij}d_{ij}^2$,
where $w_{ij} \in \reals$ is a weight conveying similarity (when $w_{ij} > 0$)
or dissimilarity (when $w_{ij} < 0$) of items $i$ and $j$. 
We constrain the embedding $X$ to be
standardized, \ie, it must satisfy $(1/n)X^TX = I$ and $X^T \ones = 0$,
which forces the embedding vectors to spread out. 
While most MDE problems are
intractable, the quadratic MDE problem is an exception: it admits an analytical
solution via eigenvectors of a certain matrix. Many well-known embedding
methods, including PCA, spectral embedding, and classical multidimensional
scaling, are instances of quadratic MDE problems, differing only in their
choice of pairs and weights. Quadratic MDE problems are 
discussed in chapter~\ref{c-quad-emb}.

\paragraph{Why the Euclidean norm?}
A natural question is why we use the Euclidean norm as our distance measure
between embedding vectors. First, when we are embedding into $\reals^2$ or
$\reals^3$ for the purpose of visualization or discovery, the
Euclidean distance corresponds to actual physical distance, making it a
natural choice. Second, it is traditional, and follows a large number of known
embedding methods like PCA and spectral embedding that also use
Euclidean distance.  Third, the standardization constraint we consider in this
monograph has a natural interpretation when we use the Euclidean distance, but
would make little sense if we used another metric. Finally, we mention that the
local optimization methods described in this monograph can be easily extended
to the case where distances between embedding vectors are measured with a
non-Euclidean metric.

\section{Contributions} The main contributions of this monograph
are the following:
\begin{enumerate}
  \item We present a simple framework, MDE, that unifies and generalizes
    many different embedding methods, both classical and modern. This framework
    makes it easier to interpret existing embedding methods and to create new
    ones. It also provides principled ways to validate, or at least
    sanity-check, embeddings.
  \item We develop an algorithm for approximately solving MDE problems (\ie,
    for computing embeddings) that places very few assumptions on the
    distortion functions and constraints. This algorithm reliably produces good
    embeddings in practice and scales to large problems.
  \item We provide open-source software that makes it easy for users to solve
    their own MDE problems and obtain custom embeddings. Our implementation of
    our solution method is competitive in runtime to specialized algorithms for
    specific embedding methods.
\end{enumerate}

\section{Outline}
This monograph is divided into three parts, \ref{p-mde}~\emph{Minimum-Distortion
Embedding}, \ref{p-algorithms}~\emph{Algorithms}, and
\ref{p-examples}~\emph{Examples}.

\paragraph{Part I: Minimum-distortion embedding.}
We begin part~\ref{p-mde} by describing the MDE problem and some of its
properties in chapter~\ref{c-mde}.  
We introduce the notion of anchored embeddings, in which some of the embedding
vectors are fixed, and standardized embeddings, in which the embedding vectors
are constrained to have zero mean and identity covariance.
Standardized embeddings are favorably scaled for many tasks, such as for use as
features for supervised learning. 

In chapter~\ref{c-quad-emb} we study MDE problems with quadratic distortion,
focusing on the problems with a standardization constraint. This class of problems has
an analytical solution via an eigenvector decomposition of a certain matrix.
We show that many existing embedding methods,
including spectral embedding, PCA, Isomap, kernel PCA, and others, reduce to
solving instances of the quadratic MDE problem.

In chapter~\ref{c-distortion-functions} we describe
examples of distortion functions, showing how different notions of faithfulness
of an embedding can be captured by different distortion
functions. Some choices of the distortion functions (and constraints) lead to
MDE problems solved by well-known methods, while others yield MDE problems
that, to the best of our knowledge, have not appeared elsewhere in the
literature.  

\paragraph{Part II: Algorithms.} In part~\ref{p-algorithms}, we describe 
algorithms for computing embeddings. We begin
by presenting stationarity conditions for the MDE problem in
chapter~\ref{c-stationarity}, which are necessary but not sufficient for an
embedding to be optimal. The stationarity conditions have a simple form:
the gradient of the average distortion, projected onto the set of
tangents of the constraint set at the current point,
is zero. This condition guides our development of algorithms
for computing embeddings.

In chapter~\ref{c-alg}, we present a projected
quasi-Newton algorithm for approximately solving MDE problems. For very
large problems, we additionally develop a stochastic proximal algorithm that
uses the projected quasi-Newton algorithm to solve a sequence of smaller
regularized MDE problems. Our algorithms can be applied to MDE problems
with differentiable average distortion, and any constraint set
for which there exists an efficient projection onto the set and an efficient
projection onto the set of tangents of the constraint set at the current 
point. This includes MDE problems with centering, anchor, or standardization
constraints.

In chapter~\ref{c-num-ex}, we present
numerical examples demonstrating the performance of our algorithms. We also
describe a software implementation of these methods, and briefly describe our
open-source implementation PyMDE.

\paragraph{Part III: Examples.} In part~\ref{p-examples}, we 
use PyMDE to approximately solve many MDE problems involving real datasets,
including images (chapter~\ref{c-images}), co-authorship networks
(chapter~\ref{c-network}), United States county demographics
(chapter~\ref{c-zip-codes}), population genetics (chapter~\ref{c-pop-gen}), and
single-cell mRNA transcriptomes (chapter~\ref{c-scrna-seq}).

\section{Related work}\label{s-rel-work}
\paragraph{Dimensionality reduction.}
In many applications, the original items are associated with high-dimensional
vectors, and we can interpret the embedding into the smaller dimensional space
as \emph{dimensionality reduction}.
Dimensionality reduction can be used to reduce the computational burden of
numerical tasks, compared to carrying them out with the original high-dimensional
vectors.  When the embedding dimension is two or three, dimension
reduction can also be used to visualize the original high-dimensional data
and facilitate exploratory data analysis.
For example, visualization is an
important first step in studying single-cell mRNA transcriptomes, a relatively
new type of data in which each cell is represented by a high-dimensional vector
encoding gene expression \cite{sandberg2014entering, kobak2019art}.

Dozens of methods have been developed for dimensionality reduction.
PCA, the Laplacian eigenmap \cite{belkin2002laplacian}, Isomap
\cite{tenenbaum2000global}, LLE \cite{roweis2000nonlinear},
maximum variance unfolding \cite{weinberger2004unsupervised}, t-SNE
\cite{maaten2008visualizing}, LargeVis \cite{tang2016visualizing}, UMAP
\cite{mcinnes2018umap}, and the latent variable model (LVM) from
\cite{saul2020tractable} are all dimensionality reduction methods. With the
exception of t-SNE and the LVM, these methods can be interpreted as solving
different MDE problems, as we will see in chapters 3 and 4. We exclude t-SNE
because its objective function is not separable in the
embedding distances; however, methods like LargeVis and UMAP have been observed
to produce embeddings that are similar to t-SNE embeddings
\citep{bohm2020unifying}. We exclude the LVM because it fits some parameters
in addition to the embedding.

Dimensionality reduction is sometimes called manifold learning
in the machine learning community, since some of these methods can be motivated
by a hypothesis that the original data lie in a low-dimensional manifold,
which the dimensionality reduction method seeks to recover
\citep{ma2011manifold, cayton2005algorithms, lin2008riemannian,
wilson2014spherical, nickel2017poincare}. 

Finally, we
note that dimensionality reduction methods have been studied under general
frameworks other than MDE \citep{ham2004kernel, yan2006graph,
kokiopoulou2011trace, lawrence2011spectral, wang2020understanding}.

\paragraph{Metric embedding.} Another well-studied class of embeddings are
those that embed one finite metric space into another one. There are many
ways to define the distortion of such an embedding.
One common definition is the maximum fractional error between the embedding distances and
original distances, across all pairs of items.
(This can be done by insisting that the embedding be non-contractive, \ie,
the embedding distances are at least the original distances, and then minimizing the
maximum ratio of embedding distance to original distance.)

An important result in metric embedding is the Johnson-Lindenstrauss Lemma,
which states that a linear map can be used to reduce the dimension of
vector data, scaling distances by no more than $(1 \pm \epsilon)$, when
the target dimension $\embdim$ is $O(\log n / \epsilon^2)$
\cite{johnson1984extensions}. Another important result is due to Bourgain, who
showed that any finite metric can be embedded in Euclidean space with at most a
logarithmic distortion \cite{bourgain1985embedding}. A constructive method
via semidefinite programming was later developed \cite{linial1995geometry}.
Several other results, including impossibility results, have been discovered
\citep{indyk2017embedding}, and some recent research has focused on embedding
into non-Euclidean spaces, such as hyperbolic space
\citep{sala2018representation}.

In this monograph, for some of the problems we consider, all that is required is to
place similar items near each other, and dissimilar items not near each other;
in such applications we may not even have original distances to preserve. In
other problems we do start with original distances. In all cases we are
interested in minimizing an \emph{average} of distortion functions (not
maximum), which is more relevant in applications, especially since real-world
data is noisy and may contain outliers.

\paragraph{Force-directed layout.} Force-directed methods are
algorithms for drawing graphs in the plane in an aesthetically pleasing way. In
a force-directed layout problem, the vertices of the graph are considered to be
nodes connected by springs. Each spring exerts attractive or repulsive forces on
the two nodes it connects, with the magnitude of the forces depending on the
Euclidean distance between the nodes. 
Force-directed methods move the nodes until a
static equilibrium is reached, with zero net force on each node,
yielding an embedding of the vertices into $\reals^2$. 
Force-directed methods, which are also called spring embedders, can
be considered as MDE problems in which the distortion functions give the
potential energy associated with the springs. 
Force-directed layout is a decades-old subject
\cite{tutte1963draw, eades1984heuristic, kamada1989algorithm},
with early applications in VLSI layout \cite{fisk1967accel, quinn1979forced}
and continuing modern interest \cite{kobourov2012spring}.

\paragraph{Low-rank models.} A low-rank model approximates a matrix by one of
lower rank, typically factored as the product of a tall and a wide matrix.
These factors
can be interpreted as embeddings of the rows and columns of the original
matrix. Well-known examples of low-rank models include PCA and non-negative
matrix factorization \cite{lee1999learning}; there are many others
\cite[\S3.2]{udell2016generalized}. PCA (and its kernelized version) can be
interpreted as solving an MDE problem, as we show in
\S\ref{s-historical-examples}.

\paragraph{X2vec.} Embeddings are frequently used to produce features for
downstream machine learning tasks. 
Embeddings for this purpose were popularized
with the publication of word2vec in 2013, an embedding method in which the
items are words \citep{mikolov2013distributed}. 
Since then, dozens of
embeddings for different types of items have been proposed, such as doc2vec
\citep{le2014distributed}, node2vec \citep{grover2016node2vec} and related
methods \citep{perozzi2014deepwalk, tang2015line}, graph2vec \citep{graph2vec},
role2vec \citep{ahmed2020role}, (batter-pitcher)2vec \citep{alcorn20162vec},
BioVec, ProtVec, and GeneVec \citep{asgari2015continuous}, dna2vec
\citep{ng2017dna2vec}, and many others. Some of these methods resemble MDE
problems, but most of them do not. Nonetheless MDE problems generically can be
used to produce such X2vec-style embeddings, where X describes the type of
items.

\paragraph{Neural networks.} Neural networks are commonly used to generate
embeddings for use in downstream machine learning tasks. One generic neural
network based embedding method is the auto-encoder, which starts by
representing items by (usually large dimensional) input vectors, such as
one-hot vectors. These vectors are fed into an encoder neural network, whose
output is fed into a decoder network. The output of the encoder
has low dimension, and will give our embedding.  The
decoder attempts to reconstruct the original input from this low-dimensional
intermediate vector. The encoder and decoder are both trained so the decoder
can, at least approximately, reproduce the original input
\cite[\S14]{goodfellow2016deep}.  

More generally, a neural network may be trained to predict some relevant
quantity, and the trained network's output (or an intermediate activation) can
be used as the input's embedding. For example, neural networks for embedding
words (or sequences of words) are often trained to predict masked words in a
sentence; this is the basic principle underlying word2vec and BERT, two
well-known word embedding methods \cite{mikolov2013distributed,
devlin2019bert}. Similarly, intermediate activations of convolutional neural
networks like residual networks \citep{he2016deep}, trained to classify images,
are often used as embeddings of images. Neural networks have also been used
for embedding single-cell mRNA transcriptomes \citep{szubert2019structure}.

\paragraph{Software.} There are several open-source software libraries for
specific embedding methods. The widely used Python library
sci-kit learn \cite{pedregosa2011scikit} includes implementations of PCA,
spectral embedding, Isomap, locally linear embedding, multi-dimensional
scaling, and t-SNE, among others. The umap-learn package implements UMAP
\cite{umapcode}, the openTSNE package provides a more scalable variant of t-SNE
\cite{policar2019openTSNE}, and GraphVite (which can exploit multiple CPUs
and GPUs) implements a number of embedding methods \cite{zhu2019graphvite}.
Embeddings for words and documents are available in gensim \cite{gensim},
Embeddings.jl \cite{white2019embeddings}, HuggingFace transformers
\cite{huggingface}, and BERT \cite{bertcode}. Force-directed layout methods are
implemented in graphviz \cite{graphviz}, NetworkX \cite{networkx}, qgraph
\cite{qgraph}, and NetworkLayout.jl \cite{networklayout}.

There are also several software libraries for approximately solving
optimization problems with orthogonality constraints (which the 
MDE problem with standardization constraint has).  Some examples include
Manopt (and its related packages PyManopt and Manopt.jl)
\cite{boumal2014manopt, townsend2016pymanopt, manoptjl}, Geoopt
\cite{kochurov2020geoopt}, and McTorch \cite{meghwanshi2018mctorch}.
More generally, problems with differentiable objective and constraint functions
can be approximately solved using solvers for nonlinear programming, such as
SNOPT \cite{gill2002snopt} (which is based on sequential quadratic programming)
and IPOPT \cite{wachter2006implementation} (which is based on an interior-point
method).

\part{Minimum-Distortion Embedding}\label{p-mde}
\chapter{Minimum-Distortion Embedding}\label{c-mde}
In this chapter we introduce the minimum-distortion embedding problem and
explore some of its general properties.

\section{Embedding}

We start with a finite set of items $\mathcal V$, which we label
as $1, \ldots, n$, so $\mathcal{V} = \{1, \ldots, n\}$.
An \emph{embedding} of the set of items $\mathcal{V}$ into
$\reals^\embdim$ is a function $F : \mathcal V \to \reals^\embdim$. We denote
the values of $F$ by $x_i = F(i)$, $i=1, \ldots, n$, and denote the embedding
concretely by a matrix $X \in \reals^{n \times \embdim}$,
\[
X = \begin{bmatrix}
x_1^T \\
x_2^T \\
\vdots \\
x_n^T
\end{bmatrix}.
\]
The rows of $X$ are the transposes of the vectors associated
with the items $1, \ldots, n$. The
columns of $X$ can be interpreted as $m$ features or attributes of the items,
with $X_{ij} = (x_i)_j$ the value of $j$th feature or attribute for item $i$;
the $j$th column of $X$ gives the values of the $j$th feature
assigned to the $n$ items.

Throughout this monograph, the quality of an
embedding $X$ will only depend on the \emph{embedding distances} $d_{ij}$ between
items $i$ and $j$, defined as
\[
d_{ij} = \|x_i-x_j\|_2, \quad i, j = 1, \ldots, n,
\]
\ie, $d_{ij}$ is the Euclidean distance between the vectors $x_i$ and $x_j$.
Embedding distances are evidently not affected by translation of the embedded
points, \ie, replacing each $x_i$ by $x_i+a$, where $a\in \reals^\embdim$, or
by orthogonal transformation, \ie, replacing each $x_i$ by $Qx_i$, where $Q$ is
an orthogonal $\embdim \times \embdim$ matrix. If there are no
constraints on the embedding vectors, then without any loss of generality we
can assume that the average of the $x_i$ is zero, or equivalently, $X^T\ones =
0$, where $\ones$ is the vector with all entries one. This means that each of
the $m$ features (columns of $X$) has mean value zero across our $n$ items.

\section{Distortion}\label{s-distortion}

We express our desires about the embedding distances by \emph{distortion
functions} associated with embedding distances. These have the form
\[
f_{ij} : \reals_+ \to \reals
\]
for $(i, j) \in \mathcal E$, where $\mathcal
E \subseteq \mathcal V \times \mathcal V$
is the set of item pairs for which we have an associated function. We will
assume that $\edges$ is nonempty, and that $i < j$ for every $(i,j)\in \edges$.

The value $f_{ij}(d_{ij})$ is the distortion
for the pair $(i,j) \in \mathcal E$: the smaller $f_{ij}(d_{ij})$ is, the
better the embedding captures the relationship between items $i$ and $j$.
We can interpret these distortion functions as the weights of a
generalized graph $(\items, \edges)$, in which the edge weights are
functions, not scalars.

We will assume that the distortion functions $f_{ij}$ are differentiable.
As in many other applications, however, we have observed that the algorithm
we propose, which assumes differentiability, works well in practice even
when the distortion functions are not differentiable.

\subsection{Examples}\label{s-distortion-functions-examples}
The distortion functions $f_{ij}$ we will encounter typically
derive from given \emph{weights} (or \emph{similarities})
associated with pairs of items, or from associated
\emph{deviations} (or \emph{distances} or \emph{dissimilarities})
between items, or both.  Distortion functions can be also derived
from one or more relations (or undirected graphs) on the items.
Below, we give simple examples of such
distortion functions. We give many more examples in
chapter~\ref{c-distortion-functions}.

\paragraph{Distortion functions derived from weights.}

We start with a set of nonzero weights $w_{ij}\in \reals$, for $(i,j) \in \edges$.
The larger $w_{ij}$, the more similar items $i$ and
$j$ are; negative weights indicate dissimilarity.
We partition the edges into those associated with positive weights
(similar items) and negative weights (dissimilar items), with
\[
\esim = \{ (i,j) \mid w_{ij}>0 \}, \qquad
\edis = \{ (i,j) \mid w_{ij}<0 \},
\]
so $\edges = \esim \cup \edis$ and $\esim \cap \edis = \emptyset$.

Distortion functions derived from weights have the form
\[
f_{ij}(d_{ij}) =
\begin{cases}
  w_{ij} \psim(d_{ij}), & (i,j) \in \esim\\
  w_{ij} \pdis(d_{ij}), & (i,j) \in \edis
\end{cases}
\]
where $\psim$ and $\pdis$ are penalty functions associated with
positive weights (similar items) and negative weights (dissimilar items),
respectively.
The penalty functions are
increasing, so $f_{ij}$ is increasing when $w_{ij}$ is positive and decreasing
when $w_{ij}$ is negative.
Roughly speaking, the closer vectors associated with similar items are,
and the farther vectors associated
with dissimilar items are, the lower the distortion.
The weights $w_{ij}$, which indicate the degree of similarity or 
dissimilarity, scale the penalty functions.

\begin{figure}
\includegraphics{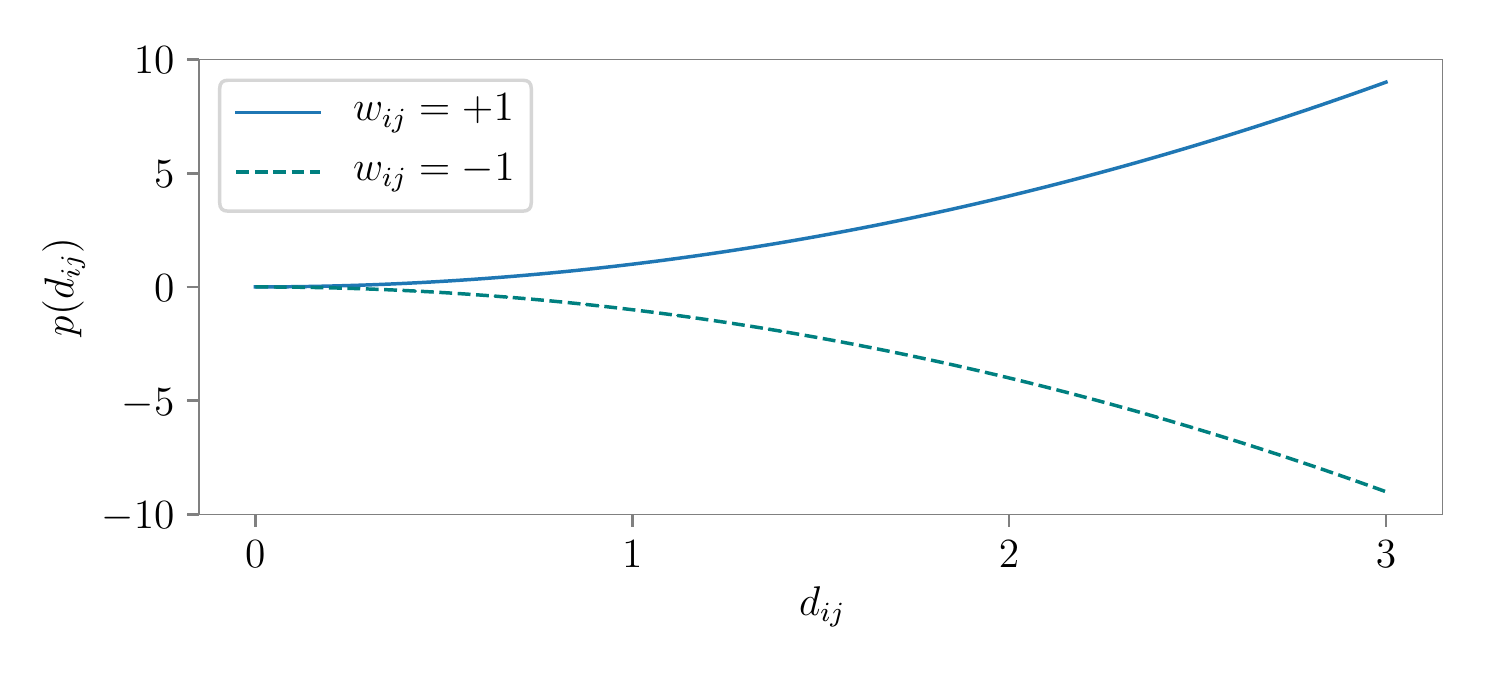}
\caption{Quadratic penalties,
  with $w_{ij} = +1$ or $w_{ij} = -1$ for $(i, j) \in \edges$.}
\label{f-quad-penalties}
\end{figure}

Perhaps the simplest distortion function derived from weights is the quadratic
\[
f_{ij}(d_{ij}) = w_{ij} d_{ij}^2,
\]
for which $\psim(d_{ij}) = \pdis(d_{ij}) = d_{ij}^2$. 
The quadratic distortion is plotted in figure~\ref{f-quad-penalties}, for 
weights $w_{ij}=1$ and $w_{ij}=-1$.
For $w_{ij}=1$, which means items $i$ and $j$ are similar, the distortion rises
as the distance between $x_i$ and $x_j$ increases;
for $w_{ij}=-1$, which means items $i$ and $j$ are dissimilar, the distortion is negative
and becomes more negative as the distance between $x_i$ and $x_j$ increases.
We will study the quadratic penalty in chapter~\ref{c-quad-emb}.  

\paragraph{Distortion functions derived from deviations.}
\begin{figure}
\includegraphics{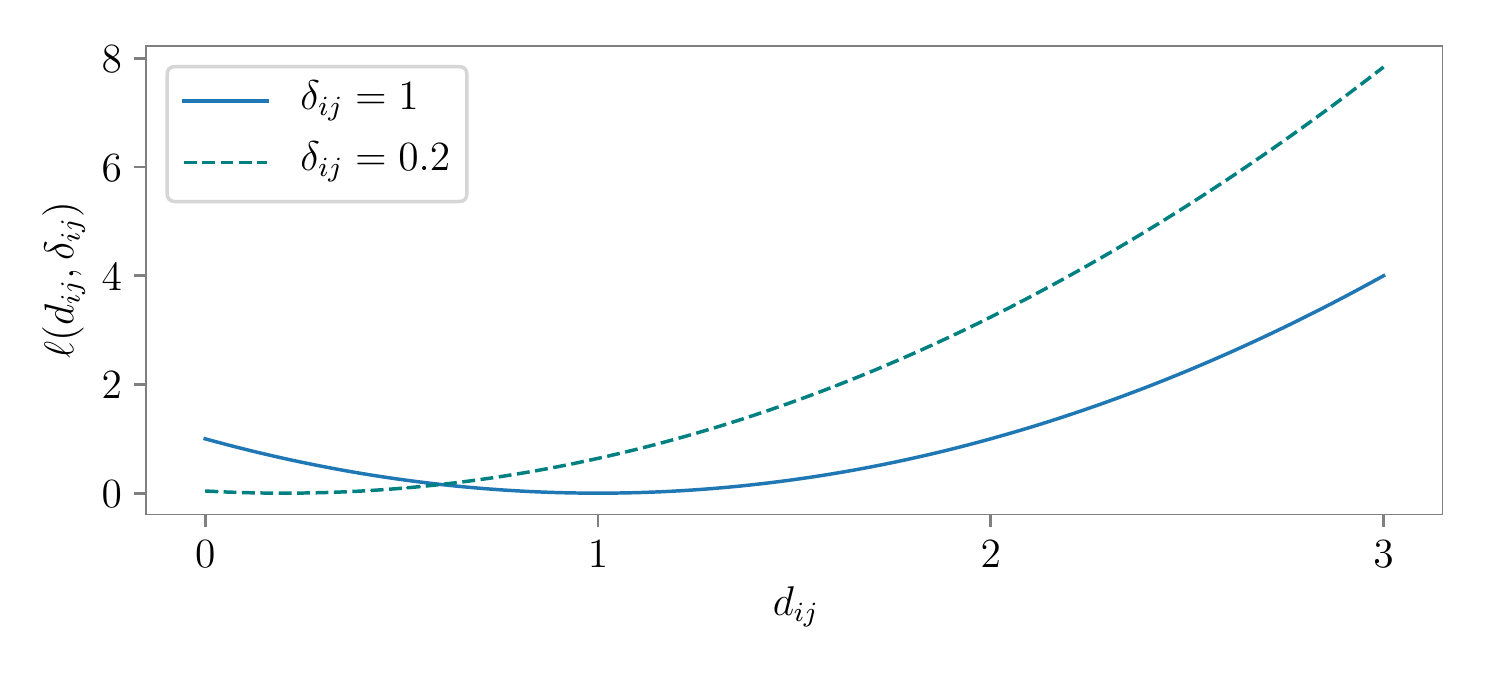}
\caption{Quadratic losses,
  with $\delta_{ij} = 1$ or $\delta_{ij} = 0.2$ for $(i, j) \in \edges$.}
\label{f-deviations}
\end{figure}
We start with nonnegative numbers
$\delta_{ij}$ for $(i,j)\in \edges$ that represent 
deviations or (original) distance between items $i$ and $j$,
with small $\delta_{ij}$ meaning the items are similar and 
large $\delta_{ij}$ meaning the items are dissimilar.
The smaller $\delta_{ij}$, the more similar items $i$ and $j$ are.
The original deviation data $\delta_{ij}$ need not be a metric.

Distortion functions derived from
deviations or distances have the general form
\[
f_{ij} (d_{ij}) = \ell(\delta_{ij}, d_{ij}),
\]
where $\ell$ is a loss function, which is
nonnegative, with $\ell(\delta_{ij},\delta_{ij})=0$,
decreasing in $d_{ij}$ for $d_{ij}<\delta_{ij}$ and increasing
for $d_{ij}>\delta_{ij}$.
We can interpret the given deviation $\delta_{ij}$ as a target
or desired value for the embedding distance $d_{ij}$; the 
distortion $\ell(\delta_{ij}, d_{ij})$ measures the difference
between the target distance and the embedding distance.

The simplest example uses a square loss,
\[
f_{ij}(d_{ij}) = (\delta_{ij}-d_{ij})^2,
\]
the square of the difference between the given deviation and 
the embedding distance.
Distortions derived from the square loss and deviations
are shown in figure~\ref{f-deviations},
for $\delta_{ij}=0.2$ and $\delta_{ij}=1$.
For original deviation $\delta_{ij}=1$, the distortion has its minimum
value at $d_{ij}=1$, \ie, when the distance between $x_i$ and $x_j$ is one.
It rises as $d_{ij}$ deviates from the target $\delta_{ij}=1$.

\paragraph{Distortion functions derived from a graph.}
In some applications we start with an undirected graph or a relation 
on the items, \ie, a set of pairs $\edges$.
The associated distortion function is
\[
f_{ij}(d_{ij}) = p(d_{ij}), \quad (i,j) \in \edges,
\]
where $p$ is a penalty function, which is increasing.
Such distortion functions can be interpreted as
deriving from weights or deviations, in the special case when
there is only one value for the weights or deviations.
The simplest such distortion function is quadratic, 
with $p(d_{ij})=d_{ij}^2$.

\paragraph{Distortion functions derived from multiple graphs.}
As a variation, we can have multiple original graphs or relations 
on the items.
For example suppose we have $\esim$, a set of pairs of items
that are similar, and also $\edis$, a set of pairs of items
that are dissimilar, with $\esim \cap \edis = \emptyset$.
From these two graphs we can derive the distortion functions
\[
f_{ij}(d_{ij}) =
\begin{cases}
  \psim(d_{ij}) & (i,j) \in \esim \\
  -\pdis(d_{ij}) & (i,j) \in \edis ,
\end{cases}
\]
with $\edges = \esim \cup \edis$, where $\psim$ and $\pdis$ are 
increasing penalty
functions for similar and dissimilar pairs, respectively.
This is a special case of distortion functions
described by weights, when the weights take on only two values, $+1$ (for
similar items) and $-1$ (for dissimilar items).

\paragraph{Connections between weights, deviations, and graphs.}
The distinction between distortion functions derived from weights, 
deviations, or graphs is not sharp.
As a specific example, suppose we are given some original deviations
$\delta_{ij}$, and we take
\[
f_{ij}(d_{ij}) = \left\{ \begin{array}{ll}
d_{ij}^2  & \delta_{ij} \leq \epsilon\\
0 & \mbox{otherwise},
\end{array} \right.
\quad (i,j)\in \edges,
\]
where $\epsilon$ is a given positive threshold.
This natural distortion function identifies an unweighted graph
of neighbors (defined by small original distance or deviation), and 
puts unit weight quadratic distortion on those edges.
(We could just as well have started with original similarities or 
weights $w_{ij}$ and constructed the neighbor graph as
$w_{ij} \geq \epsilon$.)
This distortion evidently combines the 
ideas of distances,  weights, and graphs.

\subsection{Average distortion}
We evaluate the quality of an embedding by its \emph{average distortion},
defined as
\[
E(X) = \frac{1}{|\mathcal E|} \sum_{(i, j) \in \mathcal E} f_{ij} (d_{ij}).
\]
The smaller the average distortion, the more faithful the embedding.
In chapter~\ref{c-distortion-functions}, we will see that a
wide variety of notions of faithfulness
of an embedding can be captured by different choices of the distortion
functions $f_{ij}$.
(Our use of the symbol $E$ to denote average distortion is meant to vaguely hint 
at \emph{energy}, from a mechanical interpretation we will see later.)

The average distortion $E(X)$ models our displeasure with the embedding $X$.
With distortion functions defined by weights, small $E(X)$ corresponds to
an embedding with vectors associated with large (positive) weights (\ie, similar items)
typically near each other, and
vectors associated with negative weights (\ie, dissimilar items)
not near each other.
For distortion functions defined by deviations, small $E(X)$ means
that the Euclidean distances between vectors are close to the original
given deviations.

Evidently, $(i,j) \not\in \mathcal E$ means that $E(X)$ does not depend
on $d_{ij}$.
We can interpret $(i,j) \not\in \mathcal E$ in several related ways.
One interpretation is that we are neutral about the distance $d_{ij}$, that is,
we do not care if it is small or large.  Another interpretation is
that the data required to specify the distortion $f_{ij}$, such as a weight
$w_{ij}$ or an original distance $\delta_{ij}$, is not available.

\section{Minimum-distortion embedding}
We propose choosing an embedding
$X \in \reals^{n \times \embdim}$ that minimizes
the average distortion, subject to $X \in \mathcal X$, where $\mathcal X$ is
the set of feasible or allowable embeddings. This gives the optimization
problem
\begin{equation}
\begin{array}{ll}
\mbox{minimize} & E(X)\\
\mbox{subject to} & X \in \mathcal X,
\end{array}
\label{eq-mde-prob}
\end{equation}
with optimization variable $X\in \reals^{n \times m}$.
We refer to this problem as the \emph{minimum-distortion embedding} (MDE) problem,
and we call a solution to this problem a minimum-distortion embedding. 

In this monograph we will focus on three specific constraint sets $\mathcal X$: the
set of embeddings centered at the origin, the set of embeddings in which some
of the embedding vectors are anchored (fixed) to specific values, and the set
of standardized embeddings, which are centered embeddings with unit covariance.
These constraint sets are described in \S\ref{s-constraints}.

MDE problems can be solved exactly only in a few special cases, one of which
is described in chapter~\ref{c-quad-emb}. In most other cases, solving an MDE
problem exactly is intractable, so in chapter~\ref{c-alg} we propose heuristic
methods to solve it approximately. In our experience, these methods often find
low-distortion embeddings and scale to large datasets (chapter \ref{c-num-ex}). With
some abuse of language, we refer to an embedding that is an approximate
solution of the MDE problem as a minimum-distortion embedding, even if it does
not globally solve \eqref{eq-mde-prob}.

With our assumption that the distortion functions are differentiable,
the average distortion $E$ is differentiable, provided the embedding vectors
are distinct, \ie, $x_i \neq x_j$ for $(i,j) \in \edges$, $i \neq j$,
which implies $d_{ij} >0$ for $(i,j)\in \edges$.
It is differentiable even when $d_{ij}=0$, provided $f_{ij}'(0)=0$.
For example, $E$ is differentiable (indeed, quadratic)
when the distortion functions are quadratic.

The MDE problem~(\ref{eq-mde-prob}) admits multiple interesting interpretations.
Thinking of the problem data as a generalized weighted graph, a
minimum-distortion embedding can be interpreted as a representation of the
vertices by vectors that respects the geometry of the graph,
connecting our problem to long lines of work on graph embeddings
\cite{linial1995geometry, yan2006graph, chung1997spectral,
hamilton2017representation}. The MDE problem can also be given a mechanical interpretation,
connecting it to force-directed layout \cite{eades1984heuristic}.

\paragraph{Mechanical interpretation.}
In the mechanical interpretation, we consider $x_i$ to be a point in
$\reals^\embdim$ associated with item $i$. We imagine each pair of points $(i, j)
\in \edges$ as connected by a spring with
potential energy function $f_{ij}$, \ie,
$f_{ij}(d_{ij})$ is the elastic stored energy in the spring when it is 
extended a distance $d_{ij}$.
The spring associated with $(i,j)\in \edges$ has a tension force 
versus extension given by $f'_{ij}$.
A force with this magnitude is applied to both points $i$ and $j$, 
each in the direction of the other.  Thus the
spring connecting $x_i$ and $x_j$ contributes the (vector) forces
\[
  f_{ij}'(d_{ij})\frac{x_j-x_i}{\|x_i-x_j\|_2}, \qquad
  - f_{ij}'(d_{ij})\frac{x_j-x_i}{\|x_i-x_j\|_2},
\]
on the points $x_i$ and $x_j$, respectively.
When $f'_{ij}$ is positive, the force is \emph{attractive}, \ie,
it pulls the points $x_i$ and $x_j$ toward each other.  When
$f'_{ij}$ is negative, the force is \emph{repulsive}, \ie, it pushes the points
$x_i$ and $x_j$ away from each other.

\begin{figure}
\includegraphics{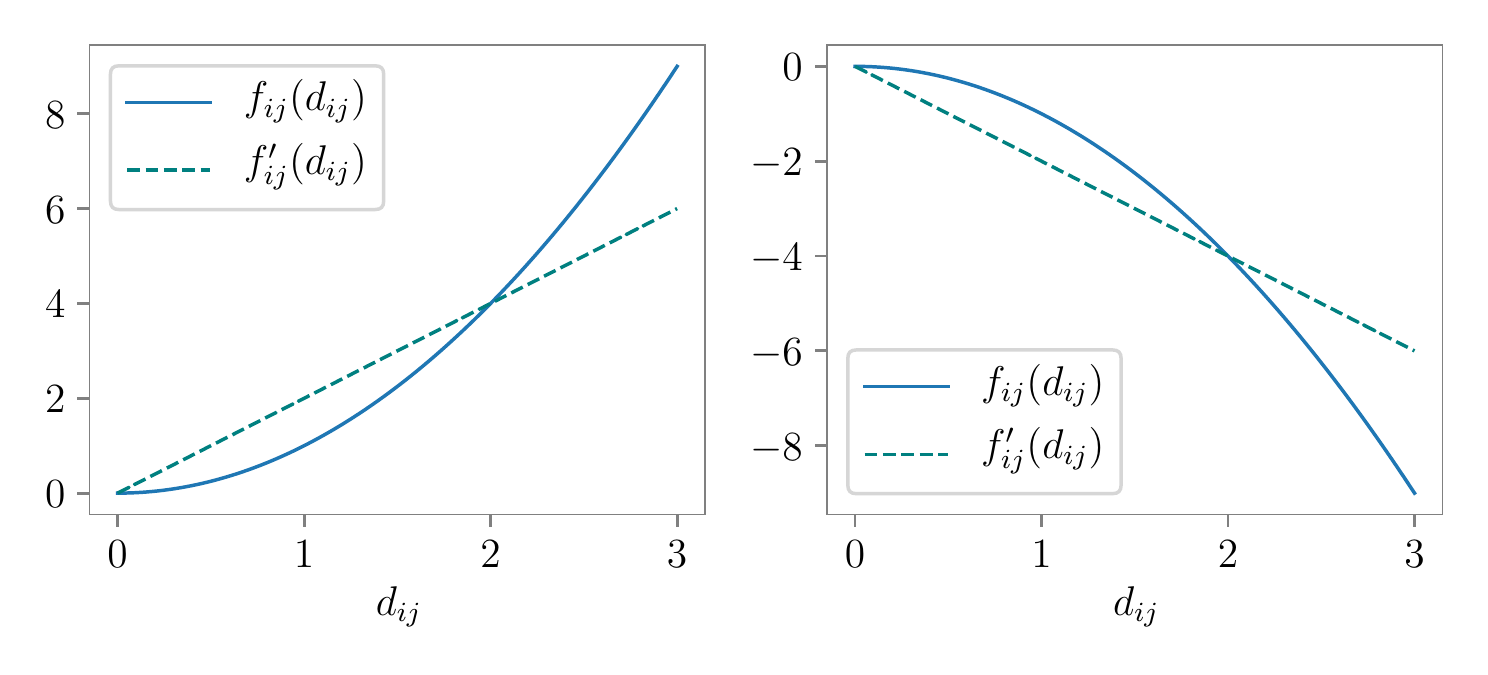}
\caption{Potential energy and force magnitude for quadratic distortion functions. \emph{Left.}
An attractive distortion function. \emph{Right.} A repulsive distortion function.}
\label{f-attractive-repulsive}
\end{figure}

\begin{figure}
\includegraphics{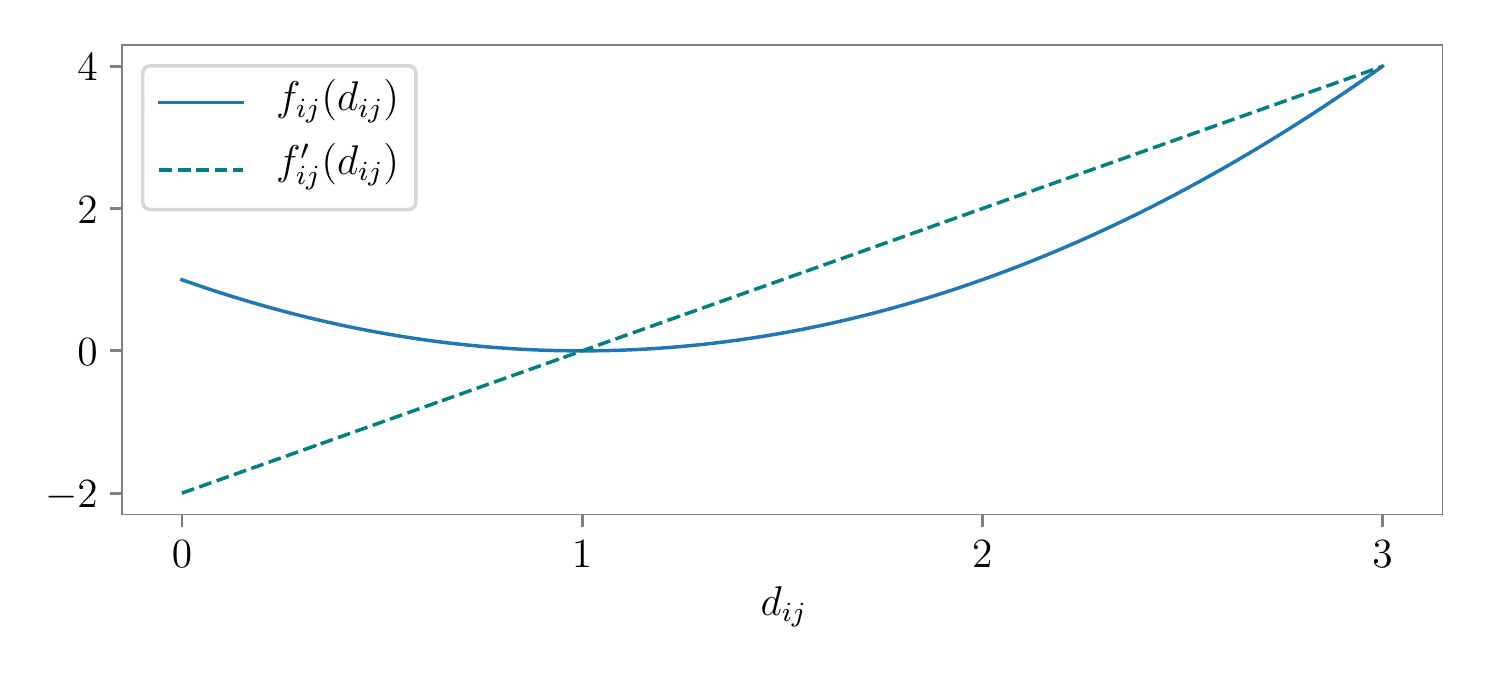}
\caption{Potential energy and force magnitude for a distortion function that is attractive
for $d_{ij} > 1$ and repulsive for $d_{ij} < 1$.}
\label{f-force-loss}
\end{figure}

Distortion function derived from weights are always attractive for 
positive weights (pairs of similar items), and always repulsive for 
negative weights (pairs of dissimilar items).
Distortion functions derived from deviations are attractive when 
$d_{ij}>\delta_{ij}$ and repulsive when $d_{ij}<\delta_{ij}$.
For such distortion functions, we can think of $\delta_{ij}$ as the
natural length of the spring, \ie, the distance at which
it applies no forces on the points it connects.
Figure~\ref{f-attractive-repulsive} shows the potential energy and force
for quadratic distortion functions, with weight $+1$ (left) and $-1$ (right).
Figure~\ref{f-force-loss} shows the potential energy and force
for a function associated with a quadratic loss.

The MDE objective $E(X)$ is the average elastic stored energy
in all the springs. The MDE problem is to find a minimum total potential energy
configuration of the points, subject to the constraints. When the constraint is
only $X^T\ones =0$, such a configuration corresponds to one in which the net
force on each point, from its neighbors, is zero. (Such a point corresponds to
a mechanical equilibrium, not necessarily one of minimum energy.) When there are
constraints beyond $X^T\ones =0$, the mechanical interpretation is a bit more
complicated, since the constraints also contribute an additional force on each
point, related to an optimal dual variable for the constraint.

\paragraph{Single-index notation.}
To simplify the notation, we will sometimes
use a single index $k$ instead of a tuple of indices $(i, j)$ to represent a
pair of items. Specifically, we impose an ordering on the tuples $(i, j) \in
\edges$, labeling them $1, \ldots, \npairs$, where $\npairs = |\edges|$ is the number of
pairs for which distortion functions are given. For a pair $k\in \{1, \ldots,
\npairs\}$, we write $i(k)$ to denote its first item and $j(k)$ to
denote the second item. We use $d_k$ to denote the embedding distance
for the $k$th pair,
\[
d_k = \|x_{i(k)} - x_{j(k)}\|_2, \quad k=1, \ldots, \npairs,
\]
with $d \in \reals^p$ the vector of embedding distances.
Similarly, we define the vector distortion function $f : \reals^\npairs \to \reals^\npairs$,
whose components are the scalar distortion functions $f_k$, $k=1, \ldots, \npairs$.
Using this notation, the average distortion is
\[
E(X) = \frac{1}{\npairs}\sum_{k=1}^{\npairs} f_k(d_k)  = \ones^T f(d) / p.
\]

We note for future reference that $p \leq n(n-1)/2$, with equality when the
graph is full.  For a connected graph, we have $p \geq n-1$, with equality
occurring when the graph is a chain.  In many applications the typical degree is
small, in which case $p$ is a small multiple of $n$.

\paragraph{Incidence matrix.} Since the problem data can be interpreted as
a graph, we can associate with it an incidence matrix $A \in \reals^{n \times
\npairs}$ \citep[\S 7.3]{boyd2018introduction}, defined as
\begin{equation}\label{e-incidence}
A_{ij} = \begin{cases}
1 & (i,j) \in \edges\\
-1 & (j,i) \in \edges \\
0 & \textnormal{otherwise.}
\end{cases}
\end{equation}
(Recall that we assume $i<j$ for each $(i,j)\in \edges$.)
The $k$th column of $A$, which we denote $a_k$,
is associated with the $k$th pair or edge. Each
column has exactly two non-zero entries, with values $\pm 1$, that give the items
connected by the edge.
We can express $a_k$ as $a_k=e_{i(k)}-e_{j(k)}$, where $e_l$ is the $l$th unit
vector in $\reals^n$.

We can express the embedding distances compactly using the incidence matrix.
Using single-index notation, the $k$th distance
$d_k$ can be written as
\[
d_k = \|X^T a_k\|_2,
\]
or equivalently
\[
d_k = \sqrt{(A^TXX^TA)_{kk}}.
\]
We mention briefly that the MDE problem can be alternatively parametrized via
the Gram matrix $G = XX^T$, since $d_k = \sqrt{a_k^T G a_k}$. For some
constraint sets $\constr$, the resulting rank-constrained optimization problem
can be approximately solved by methods that manipulate the square root of $G$,
using techniques introduced in \cite{burer2003nonlinear, burer2005local}. We
will not use this parametrization in the sequel.

\section{Constraints}\label{s-constraints}

\subsection{Centered embeddings}\label{s-unconstrained}
We say that an MDE problem is centered or unconstrained when the constraint set is
$\mathcal X= \mathcal C$, where
\begin{equation}\label{e-center-constraint}
\mathcal C =  \{X \mid X^T\ones = 0\}
\end{equation}
is the set of centered embeddings. 
(Recall that the constraint $X^T \ones = 0$ is without loss of generality.)
An unconstrained MDE problem can fail to be well-posed, \ie, it might not have
a solution or might admit trivial solutions.

When an unconstrained MDE problem does have a solution, it is
never unique, since both the constraint set $\mathcal C$ and the objective $E$
are invariant under rotations. If $X^\star$ is any optimal embedding, then $X =
X^\star Q$ is also optimal, for any orthogonal $m \times m$ matrix $Q$.

\subsection{Anchored embeddings}\label{s-anchored}
In an anchored MDE problem, some of the vectors $x_i$ are known and fixed, with
\begin{equation}\label{e-anchor-constraint}
x_i = x_i^\text{given}, \quad i \in \mathcal K,
\end{equation}
where $\mathcal K \subseteq \items$ is the set of indices of vectors
that are fixed (or anchored), and $x_i^\text{given}\in \reals^m$
are the given values for those vectors. We call \eqref{e-anchor-constraint}
an anchor constraint; the items with vertices in $\mathcal K$ are
\emph{anchored}, and the remaining are \emph{free}.

We will use $\anchored$ to represent a set of anchored embeddings, \ie
\[
  \anchored = \{X \mid x_i = x_i^\text{given}, ~ i \in \mathcal K\}.
\]
(The constraint set $\mathcal A$ depends on $\mathcal K$ and $x_i^\text{given}$,
but we suppress this dependence in our notation to keep it simple.)
Unlike centered embeddings, the constraint set $\anchored$ is not closed under
orthogonal transformations. In particular, MDE problems with this constraint are
neither translation nor rotation invariant.

An anchor constraint has a natural mechanical interpretation: the anchored
vertices are physically fixed to given positions (as if by nails).
We will later describe the force that the nails exert on their associated anchored points.
Not surprisingly, in an optimal anchored embedding, the net force on 
each point (due to its neighbors and its nail, if it is anchored) is zero.

\paragraph{Incremental embedding.}
Anchor constraints can be used to build an embedding \emph{incrementally}.
In an incremental embedding, we start with an embedding of 
$X \in \reals^{n \times \embdim}$ of a set of items $\items = \{1, \ldots,
n\}$. We later 
obtain a few additional items
$\items' = \{n+1, \ldots, n+n'\}$ (with $n' \ll n$), 
along with a set of pairs $\edges' \subseteq
(\items \cup \items') \times (\items \cup \items')$ and their associated
distortion functions relating old items to new
items (and possibly relating the new items to each other),
\ie, $(i, j) \in \edges'$ implies $i \in \items$ and $j \in \items'$, or
$i, j \in \items'$ (and $i < j$). We seek a new embedding $X' \in \reals^{n+n'
\times m}$ in which the first $n$
embedding vectors match the original embedding, and the last $n+1$ embedding
vectors correspond to the new items. Such an embedding is readily found by
anchoring the first $n$ embedding vectors to their original values, leaving
only the last $n+1$ embedding vectors free, and minimizing the average
distortion. 
There are two approaches to handling centering or standardization 
constraints in incremental embedding.  
The simplest is to simply ignore these constraints,
which is justified when $n' \ll n$, so the new embedding $X'$ likely will
not violate the constraints by much.
It is also possible to require that the new embedding $X'$ satisfy the 
constraints.  This approach requires care; for example,
suppose we are to add just one new item, \ie, $n'=1$, and we insist
that the new embedding $X'$ be centered, like the original one $X$.
In this case, the only possible choice for the vector associated with the new 
item is $x_{n+1}=0$.

Incremental embedding can be used to develop a feature map for 
new (unseen) items, given some distortion functions 
relating the new item to the original ones (say, the training items).
In this case we have $n'=1$; we embed the new item $n+1$ so as to 
minimize its average distortion with respect to the original items.

\paragraph{Placement problems.} While MDE problems are usually nonconvex,
anchored MDE problems with nondecreasing convex distortion functions are an
exception. This means that some anchored MDE problems can be efficiently and
globally solved. These MDE problems can be interpreted as (convex) placement
problems, which are described in \cite[\S8.7]{bv2004convex} and have
applications to VLSI circuit design \cite{sherwani2012algorithms}.

\paragraph{Semi-supervised learning.}
Anchored embeddings also arise in graph-based semi-supervised learning problems.
We interpret the vectors $x_i$, $i \in \mathcal K$, as the known labels;
our job is to assign labels for $i \not\in \mathcal K$, using a 
prior graph with positive weights that indicate similarity.
In graph-based semi-supervised learning \citep{xu2010semi,
el2016asymptotic}, this is done by solving an anchored MDE problem with
quadratic distortion functions
$f_{ij}(d_{ij}) = w_{ij}d_{ij}^2$,
which as mentioned above is convex and readily solved.

\subsection{Standardized embeddings}\label{s-standardization-constraint}
The set of standardized embeddings is
\begin{equation}\label{eq-std}
\stdemb = \{X \mid (1/n)X^TX = I, ~ X^T\ones=0\}.
\end{equation}
The standardization constraint $X \in \stdemb$ requires that the collection of
vectors $x_1, \ldots, x_n$ has zero mean and unit covariance.
We can express this in terms of the $\embdim$ feature columns of the embedding 
$X \in \stdemb$: they have zero mean, are uncorrelated,
and have root-mean-square (RMS) value one; these
properties are often desirable when the embedding
is used in downstream machine learning tasks. 
We refer to the MDE problem as standardized when $\mathcal X=\stdemb$.

Because $\stdemb$ is a compact set, the standardized MDE problem
has at least one solution.  Since $\stdemb$ is invariant under
rotations, the standardized MDE problem has a 
family of solutions: if $X^\star$ is an optimal embedding, so is
$X = X^\star Q$, for any $m \times m$ orthogonal matrix $Q$.

\paragraph{Natural length in a standardized embedding.}
We note that for $X \in \stdemb$, the sum of the squared embedding
distances is constant,
\begin{equation}\label{eq-sum-d-sq}
\sum_{1 \leq i < j \leq n} d^2_{ij} = n^2\embdim.
\end{equation} %
(Note that this is the sum over \emph{all}
pairs $(i,j)$ with $i<j$, not just the pairs in $\mathcal E$.)
To see this, we observe that
\[
\sum_{i, j=1}^n d^2_{ij} = 2n\sum_{i=1}^{n} \norm{x_i}^2_2 - 2 \sum_{i, j=1}^{n}x_i^Tx_j.
\]
The first term on the right-hand side equals $2n^2\embdim$, since
\[
\sum_{i=1}^{n} \norm{x_i}^2_2 = \tr (XX^T) = \tr (X^TX) = n\embdim,
\]
while the second term equals $0$, since $X^T\ones = 0$.
(Here $\tr$ denotes the trace of a matrix, \ie, the sum of its diagonal entries.)

From \eqref{eq-sum-d-sq}, we can calculate the RMS value of all
$n(n-1)/2$ embedding distances. We will refer to this value as the
\emph{natural length} $\dnat$ of the embedding, with
\begin{equation}
\dnat = \sqrt{\frac{2n\embdim}{n-1}}.
\label{eq-d-nat}
\end{equation}
For $X \in \stdemb$, $\dnat$ can be interpreted as the typical value
of embedding distances.
We will see later that this number is useful in choosing appropriate
distortion functions.

\subsection{Attraction, repulsion, and spreading}\label{s-attraction}
When all distortion functions are nondecreasing, we say the objective
is \emph{attractive}.  In the mechanical interpretation, this means that
there is a positive (attractive) tension between all pairs of points
$(i,j) \in \edges$.  Roughly speaking, the distortion functions only 
encode similarity between items, and not dissimilarity; the objective encourages
neighboring embedded points to be near each other.
When the only constraint is that the embedding is centered, 
\ie, $\mathcal X= \mathcal C$, a globally optimal embedding is $X=0$,
which is not useful.  Roughly speaking, all forces between points 
are attractive, so the 
points collapse to a single point, which must be $0$ to achieve 
centering.

When the objective is attractive, we need a constraint to enforce
\emph{spreading} the points.  Anchored and standardized constraints both serve 
this purpose, and enforce spreading of the points when the objective is 
attractive. When some distortion functions can be \emph{repulsive}, \ie,
$f'_{ij}(d_{ij})$ can be negative, we can encounter the opposite problem, which
is that some points might spread without bound (\ie, the MDE problem does not
have a solution).  In such situations, an anchoring or standardized constraint
can serve the role of keeping some points from spreading without bound.

The standardization constraint keeps both pathologies from happening.
The points are required to be spread, and also bounded.  Since $\mathcal S$ is
compact, there is always a solution of the standardized MDE problem.

\subsection{Comparing embeddings}\label{s-orth-procrustes}

In some cases we want to compare or evaluate a distance between
two embeddings of the same set of items,
$X$ and $\tilde X$.  
When the embeddings are anchored, a reasonable 
measure of how different they are is the mean-square distance between
the two embeddings,
\[
\frac{1}{n}\|X-\tilde X\|_F^2 = 
\frac{1}{n} \sum_{i=1}^n \|x_i -\tilde x_i\|_2^2.
\]

When the embeddings are centered or standardized, the comparison is
more subtle, since in these cases any embedding can be transformed
by an orthogonal matrix with no effect on the constraints or objective.
As mentioned above, $\tilde X$ and $\tilde XQ$, where $Q^TQ=I$, 
are equivalent for the MDE problem.
A reasonable measure of how different the two embeddings are is then
\[
\Delta = \inf \{ \|X-\tilde X Q \|_F^2/n \mid Q^TQ = I \},
\]
the minimum mean-square distance over all orthogonal 
transformations of the second one (which is same as the minimum over
transforming $X$, or both $X$ and $\tilde X$).

We can work out the distance $\Delta$ analytically, and also find an
optimal $Q$, \ie, one that achieves the minimum.  
This is useful for visualization, if we are plotting $X$ or $\tilde X$
(presumably in the case with $m$ two or three):  we plot both $X$ and 
$\tilde XQ$, the latter being the second embedding,
optimally orthogonally transformed with respect to the first embedding.
We refer to this as \emph{aligning} the embedding $\tilde X$ to another 
target embedding $X$.

Finding the optimal $Q$ is an instance of the orthogonal Procrustes
problem \cite{schonemann1966generalized}.  We first form the $m \times m$ matrix $Z = X^T\tilde X$, and 
find its (full) singular value decomposition (SVD) $Z = U\Sigma V^T$.
The optimal orthogonal matrix is $Q=VU^T$, which gives
\[
\Delta = \frac{1}{n} \left( \|X\|_F^2 + \|\tilde X\|_F^2 - 2\tr \Sigma \right).
\]
For standardized embeddings we have $\|X\|_F^2=\|\tilde X\|_F^2 = mn$, so
$\Delta = m - \frac{2}{n}\tr \Sigma$.

\section{Simple examples}\label{s-simple-examples}
We present some simple examples of synthetic MDE problems to illustrate some of
the ideas discussed thus far. In each example, the original data is a
graph on $n=20$ vertices, and we embed into $m=2$ dimensions. We will see more
interesting examples, using real data, in part~\ref{p-examples} of this
monograph.

\begin{figure}
\begin{centering}
\includegraphics{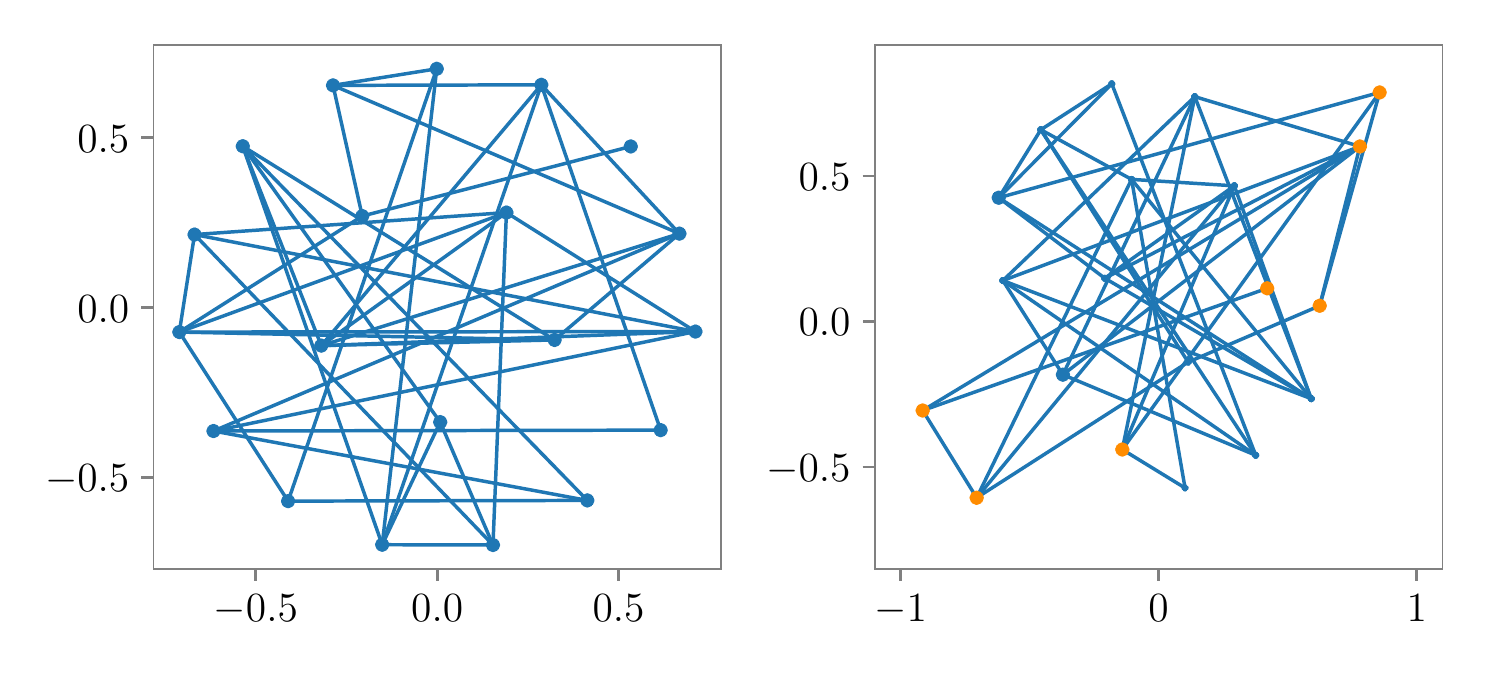}
\caption{Embeddings derived from a graph on $20$ vertices and $40$ edges.
\emph{Left.} Unconstrained. \emph{Right.} Anchored (orange points are
anchors).}
\label{f-equilibrium}
\end{centering}
\end{figure}

\paragraph{An unconstrained embedding.}
In our first example, the original graph has $p=40$ edges, which were
randomly selected. In constructing the MDE problem data, we take $\edges$ to
be all $n(n-1)/2$ pairs of the $n$ vertices, and assign deviations
$\delta_{ij}$ between vertices $i$ and $j$ using the graph distance (\ie, the
length of the shortest path between $i$ and $j$). We use quadratic distortion
functions $f_k(d_k) = (d_k - \delta_k)^2$, and require $X \in \centered$.
The left plot in figure~\ref{f-equilibrium} shows a minimum-distortion
embedding for this problem. 

\paragraph{An anchored embedding.}
We embed the same original graph as the previous example, but in this example
we impose an anchor constraint $X \in \anchored$. Seven of the vertices are
anchors, and the rest are free. The anchors and their positions were selected
randomly.
The right plot in figure~\ref{f-equilibrium} shows a
minimum-distortion embedding for this problem. The anchors are colored orange.

\begin{figure}
\begin{centering}
\includegraphics{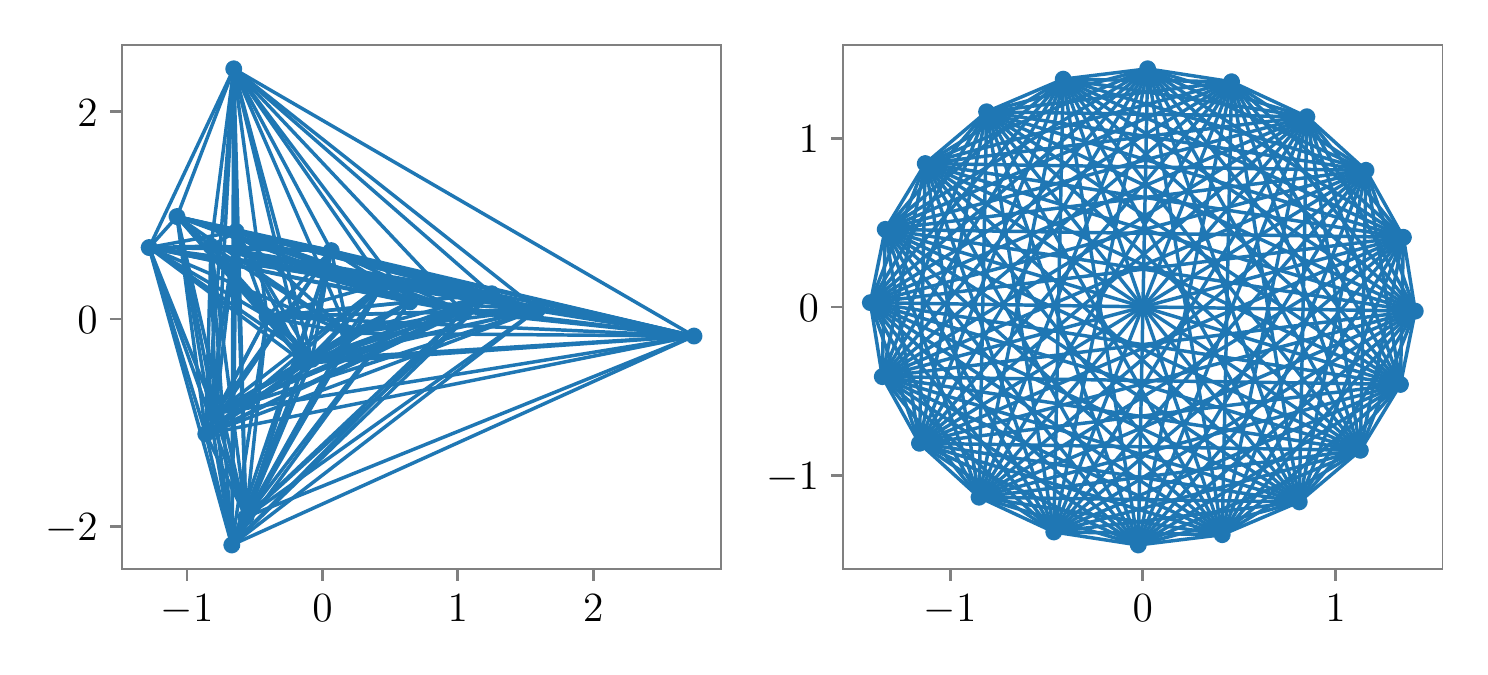}
\caption{Standardized embeddings on a complete graph.
\emph{Left.} Quadratic penalty. \emph{Right} Cubic penalty.}
\label{f-simple-std}
\end{centering}
\end{figure}

\paragraph{A standardized embedding.}
In the third example we produce standardized embeddings $X \in \standardized$
of a complete graph on $20$ vertices. We compute two embeddings: the first is
obtained by solving a standardized MDE problem with quadratic distortion
functions $f_{k}(d_{k}) = d_{k}^2$, and the second uses cubic distortion
functions $f_{k}(d_{k}) = d_{k}^3$. (These MDE problems can be interpreted
as deriving from graphs, or from weights with $w_{ij} = 1$ for all $(i, j) \in
\edges$).

The embeddings are plotted in figure~\ref{f-simple-std}. The
embedding produced using quadratic distortion
functions is known as a spectral layout, an old and widely used method for
drawing graphs \cite{hall1970r, koren2003spectral}.  With cubic penalty,
we obtain the embedding on the right, with the 20 points arranged uniformly
spaced on a circle.
This very simple example
shows that the choice of distortion functions is important. With a quadratic
distortion, most embedding distances are small but some are large; with
a cubic distortion, which heavily penalizes large distances, no single
embedding distance is very large.

\section{Validation}\label{s-validation}
The quality of an embedding ultimately depends on whether it is useful for the
downstream application of interest. Nonetheless, there are simple methods that
can be used to sanity check an embedding, independent of the downstream task.
We describe some of these methods below. (We will use some of these methods to
sanity check embeddings involving real data in part~\ref{p-examples}.)

\paragraph{Using held-out attributes.}
In the original data, some or all of the items may be tagged with
\emph{attributes}. For example, if the items are
images of handwritten digits, some of the images may be tagged with the
depicted digit; if the items are single-cell mRNA transcriptomes collected from
several donors, the transcriptomes may be tagged with their donors of origin;
or if the items are researchers, they may be tagged with their field of study.

If an attribute is \emph{held-out}, \ie, if the MDE problem does not depend on
it, it can be used to check whether the embedding makes sense. In
particular, we can check whether items with the same attribute value are near
each other in the embedding. When the embedding dimension is $1, 2$ or $3$,
this can be done visually by arranging the embedding vectors in a scatter plot,
and coloring each point by its attribute value. When the dimension is greater
than $3$, we can check if a model can be trained to predict the attribute,
given the embedding vectors. 
If items tagged with the same or similar values appear near each other
in the embedding, we can take this a weak endorsement of the embedding.
Of course this requires that the held-out attributes are related to 
whatever attributes were used to create the MDE problem.

\paragraph{Examining high-distortion pairs.}
Thus far we have only been concerned with the average distortion of an
embedding. It can also be instructive to examine the distribution of
distortions associated with an embedding.

Suppose most pairs $(i, j) \in \edges$ have low distortion, but some have much
higher distortion. This suggests that the high-distortion pairs may contain
anomalous items, or that the pairing itself may be anomalous (\eg, dissimilar
items may have been designated as similar). If we suspect that these outlier
pairs have an outsize effect on the embedding (\eg, if their sum of distortions
is a sizable fraction of the total distortion), we might modify the distortion
functions to be more robust, \ie, to not grow as rapidly for large distances,
or simply throw out these outliers and re-embed.

\paragraph{Checking for redundancy.}
A third method is based on the vague idea that a good embedding should not be
too sensitive to the construction of the MDE problem. In particular, it should
not require all the distortion functions.

We can check whether there is some redundancy among the distortion
functions by holding out a subset of them, constructing an embedding without
these functions, and then evaluating the average distortion on the held-out
functions. We partition $\edges$ into training and held-out sets,
\[
\edges_{\mathrm{train}} \subseteq \edges, \qquad
\edges_{\mathrm{val}} = \edges \setminus \edges_{\mathrm{train}},
\]
and find an embedding $X$ using only the pairs in the training set
$\edges_{\mathrm{train}}$. We then compare the average distortions on the
training set and the validation set,
\[
f_{\mathrm{train}} =
  \frac{1}{|\mathcal E_{\mathrm{train}}|}
  \sum_{(i, j) \in \mathcal E_{\mathrm{train}}} f_{ij}(d_{ij}), \qquad
f_{\mathrm{val}} =
  \frac{1}{|\mathcal E_{\mathrm{val}}|}
  \sum_{(i, j) \in \mathcal E_{\mathrm{val}}} f_{ij}(d_{ij}). \quad
\]
While we would expect that $f_{\mathrm{val}} \geq f_{\mathrm{train}}$, if
$f_{\mathrm{val}}$ is much larger than $f_{\mathrm{train}}$, it might suggest
that we do not have enough edges (and that we should obtain more).

\paragraph{Checking for bias.}
\newcommand{\eneut}{\mathcal{E}_{\text{neut}}}
\newcommand{\esens}{\mathcal{E}_{\text{sens}}}
In many applications, it is important to ensure various groups of items are
treated (in some sense) fairly and equitably by the embedding (see, \eg,
\cite{dwork2012fairness,bolukbasi2016man,barocas2018fairness,
corbett2018measure,garg2018word,holstein2019improving}).
For example, consider the task of embedding English words. It has been shown
that some popular embedding methods reinforce harmful stereotypes related to
gender, race, and mental health; \eg, in the BERT language model
\cite{devlin2019bert}, words related to disabilities have been found to be
associated with negative sentiment words \cite{hutchinson2020social}. Moreover,
for embedding methods (like BERT) that are trained to model an extremely large
corpus of natural language documents, it is usually difficult to correct these
biases, even if they have been identified in the embedding;
\citet{bender2021dangers} call this the problem of unfathomable training data.

The MDE framework provides a principled way of detecting and mitigating harmful
bias in an embedding. In our framework, there are at least two ways in which
harmful biases may manifest: the distortion functions may encode harmful
biases, or the embedding may have lower distortion on the edges associated with
one group than another. Along these lines, we propose the following three-step
procedure to mitigate bias. We continue with our running example of embedding
English words, but the procedure applies more broadly.

\begin{enumerate}
\item \emph{Identify groups of interest.} The first step is to identify 
groups of interest. This might correspond to identifying sets of words related to
gender, sex, race, disabilities, and other demographic categories. (These sets
need not be disjoint.)

\item \emph{Audit the distortion functions.} The second step is to audit to
distortion functions in which at least one item is a member of an identified
group, and modifying them if they encode harmful biases.

For example, say the distortion functions are derived from weights, and the
modeler is interested in words related to the female sex. The modeler
should audit  the functions $f_{ij}$ for which $i$ or $j$ is related to
the female sex. In particular, the modeler may want to ensure that the weights
associating these words to certain professions and adjectives are small,
zero, or negative. The modeler can also add or modify distortion functions
between words in different groups, to articulate whether, \eg, the vectors for
``man'' and ``woman'' should be close or not close in the embedding.

This step is extremely important, since the embedding is designed to be
faithful to the relationships articulated by the distortion functions. It is
also difficult, because it requires that the modeler be conscious of the many
kinds of harmful biases that may arise in the embedding.

\item \emph{Compare distortions.} The third step is to compute the embedding,
and check whether it is more faithful to the stated relationships involving
one group than another. This can be done by comparing the distortions for
different groups.

For example, say we have two groups of interest: $G_1 \subseteq \items$
and $G_2 \subseteq \items$. Let $\edges_1$ be the set of pairs $(i, j)$ in
which $i \in G_1$ or $j \in G_1$, and let $\edges_2$ be the same for $G_2$.
We can compute the average distortion of the embedding restricted to these
two sets:
\[
E_{1}(X) = \frac{1}{|\edges_1|} \sum_{(i, j) \in \edges_1} f_{ij}(d_{ij}), \quad
E_{2}(X) = \frac{1}{|\edges_2|} \sum_{(i, j) \in \edges_2} f_{ij}(d_{ij}).
\]
Suppose $E_1(X) \approx E_2(X)$. This means that on average, the embedding
is as faithful to the stated relationships involving group $G_1$ as it is
to the relationships involving group $G_2$. (Importantly, $E_1(X) \approx
E_2(X)$ does not mean that the embedding is free of harmful bias, since the
distortion functions themselves may encode harmful bias if the modeler did
a poor job in step 2.)

Otherwise, if $E_1(X) < E_2(X)$, the embedding is on average more faithful to
the first group than the second, revealing a type of bias. The
modeler can return to step 2 and modify the distortion
functions so that $\edges_2$ contributes more to the total distortion (\eg, by
increasing some of the the weights associated with $\edges_2$, or decreasing
some of the weights associated with $\edges_1$). 
\end{enumerate}

We emphasize that ensuring an embedding is fair and free of harmful bias is
very difficult, since it requires the modelers to be aware of the relevant
groups and biases. All the MDE framework does is provide modelers a procedure
that can help identify and mitigate harmful biases, provided they are
aware of them.

\chapter{Quadratic MDE Problems}\label{c-quad-emb}
In this chapter we consider MDE problems with
quadratic distortion functions,
\[
f_{ij}(d_{ij}) = w_{ij}d_{ij}^2, \quad (i,j) \in \edges,
\]
where $w_{ij}$ is a weight encoding similarities
between items.  Large positive $w_{ij}$ means items $i$ and $j$ are
very similar, small or zero $w_{ij}$ means they are neither similar nor
dissimilar, and large negative $w_{ij}$ means the two items are very dissimilar.
In many cases the weights are all nonnegative, in which case the objective
is attractive.
We refer to an MDE problem with quadratic distortion functions as 
a \emph{quadratic MDE problem}.

\paragraph{Centered embeddings.}
Centered (unconstrained) quadratic MDE problems are not interesting.  
In all cases, either $X=0$ is a global solution, or the problem is unbounded
below, so there is no solution.  

\paragraph{Anchored embeddings.}
Anchored quadratic MDE problems with positive weights are readily solved
exactly by least squares.  This problem is sometimes called 
the \emph{quadratic placement problem} \citep{sigl1991analytical}, and has
applications in circuit design. When there are negative weights, the anchored
quadratic MDE problem can be unbounded below, \ie, not have a solution.

\paragraph{Standardized embeddings.}
For the remainder of this chapter we focus on the standardized quadratic MDE
problem,
\begin{equation}\label{eq-quad-mde}
\begin{array}{ll}
\mbox{minimize} & (1/\npairs) \sum_{(i, j) \in \edges} w_{ij} d_{ij}^2 \\
\mbox{subject to} & X \in \stdemb.
\end{array}
\end{equation}
These problems are also sometimes referred to as quadratic
placement problems, \eg, as in \cite{hall1970r}.

Standardized quadratic MDE problems are special for several reasons. First,
they are tractable: we can solve \eqref{eq-quad-mde} globally, via eigenvector
decomposition, as we will see in \S\ref{s-quad-emb-soln}. Second, many
well-known historical embeddings can be obtained by solving instances of the
quadratic MDE problem, differing only in their choice of weights, as shown in
\S\ref{s-historical-examples}.%

\section{Solution by eigenvector decomposition}\label{s-quad-emb-soln}
We can re-formulate the problem~\eqref{eq-quad-mde} as an eigenproblem.
Note that
\begin{equation}\label{eq-trace-equality}
E(X) = 1/\npairs\sum_{(i, j) \in \edges} w_{ij}d_{ij}^2 = 1/\npairs \tr(X^TLX),
\end{equation}
where the symmetric matrix $L \in \symm^{n}$ has
upper triangular entries (\ie, $i < j$) given by
\[
L_{ij} = \begin{cases}
-w_{ij} & (i, j) \in \edges \\
0 & \text{otherwise,}
\end{cases}
\]
and diagonal entries
\[
L_{ii} = -\sum_{j \neq i} L_{ij}.
\]
(The lower triangular entries are found from $L_{ij}=L_{ji}$.)
We note that $L$ satisfies $L\ones =0$.
If the weights are all nonnegative, the matrix $L$ is a
Laplacian matrix.  But we do not assume here that all the weights
are nonnegative.

The equality~\eqref{eq-trace-equality} follows from the calculation
\begin{align*}
1/\npairs \sum_{(i, j) \in \edges} w_{ij}d_{ij}^2 &= 1/\npairs \sum_{(i, j) \in \edges} w_{ij}\|x_i - x_j\|_2^2 \\
&=  1/\npairs \sum_{(i, j) \in \edges} w_{ij}(\|x_i\|_2^2 + \|x_j\|_2^2 - 2x_i^Tx_j) \\
&=  1/\npairs \left(\sum_{i=1}^{n} L_{ii} \|x_i\|_2^2 + 2\sum_{(i, j) \in \edges} L_{ij} x_i^Tx_j \right)\\
&=  1/\npairs \tr(X^TLX).
\end{align*}
The MDE problem~\eqref{eq-quad-mde} is therefore equivalent to the problem
\begin{equation}\label{eq-eigen}
\begin{array}{ll}
\mbox{minimize} & \tr(X^T L X) \\
\mbox{subject to} & X \in \stdemb,
\end{array}
\end{equation}
a solution to which can be obtained from eigenvectors of $L$. To see this,
we first take an eigenvector decomposition of $L$, letting $v_1, v_2, \ldots, v_n$ be
orthonormal eigenvectors,
\[
Lv_i = \lambda_i v_i, \quad i=1, \ldots, n, \quad
       \lambda_1 \leq \cdots \leq \lambda_n.
\]
Let $\tilde{x}_1, \tilde{x}_2, \ldots, \tilde{x}_m \in \reals^n$ be the columns
of $X$ (which we recall satisfies $X^TX = nI$). Then
\begin{eqnarray*}
  (1/p)\tr(X^T L X) &=&
  (1/p)\sum_{i=1}^{n} \lambda_i
    ((v_i^T \tilde{x}_1)^2 + (v_i^T \tilde{x}_2)^2 + \cdots + (v_i^T \tilde{x}_m)^2)\\
    &\geq& (n/p) \sum_{i=1}^{m} \lambda_i,
\end{eqnarray*}
where the inequality follows from
\[
(v_i^T \tilde{x}_1)^2 + (v_i^T \tilde{x}_2)^2 + \cdots + (v_i^T \tilde{x}_m)^2
\leq n
\]
for $i = 1, \ldots, n$, and 
\[
\sum_{i=1}^{n} \sum_{j=1}^{m} (v_i^T\tilde{x}_j)^2 = nm.
\]
Let $v_j$ be the eigenvector that is a multiple of $\ones$. If $j
> \embdim$, a minimum-distortion embedding is given by $\tilde{x}_i =
\sqrt{n}v_i$, \ie, by concatenating the bottom $\embdim$
eigenvectors,
\[
X = \sqrt{n}[v_1 ~  \cdots ~ v_{\embdim}].
\]
If $j \leq \embdim$, a solution is obtained by removing $v_j$ from the above
and appending $v_{\embdim+1}$,
\[
X = \sqrt{n}[v_1 ~\cdots ~ v_{j-1}~v_{j+1}~ \cdots ~ v_{\embdim+1}].
\]
The optimal value of the MDE problem is
\[
  \frac{n}{p} (\lambda_1+ \cdots +\lambda_\embdim),
\]
for $j>m$, or
\[
  \frac{n}{p} (\lambda_1+ \cdots +\lambda_{\embdim+1}),
\]
for $j\leq m$.

\paragraph{Computing a solution.}
To compute $X$, we need to compute the bottom $m+1$ eigenvectors
of $L$. For $n$ not too big (say, no more than 10,000), it is possible to carry
out a full eigenvector decomposition of $L$, which requires $O(n^2)$ storage
and $O(n^3)$ flops. This na\"ive method is evidently inefficient, since we will
only use $m+1$ of the $n$ eigenvectors, and while storing $X$ requires only
$O(nm)$ storage, this method requires $O(n^2)$. For $n$ larger than around
10,000, the cubic complexity becomes prohibitively expensive. Nonetheless, when
$L$ is dense, which happens when $p$ is on the order of $n^2$, we cannot do
much better than computing a full eigenvector decomposition.

For sparse $L$ ($p \ll n^2$) we can use iterative methods that only require
multiplication by $L$, taking $O(p)$ flops per iteration; these
methods compute just the bottom $\embdim$ eigenvectors that we seek. One
such method is the Lanczos iteration \cite{lanczos1950iteration}, an
algorithm for finding eigenvectors corresponding to extremal eigenvalues of
symmetric matrices. Depending on the spectrum of $L$, the Lanczos method
typically reaches a suitable solution within $O(n)$ or sometimes even $O(1)$
iterations \cite[\S32]{trefethen1997numerical}. Another popular iterative method
is the locally optimal preconditioned conjugate gradient method (LOBPCG), which
also applies to symmetric matrices \cite{knyazev2001toward}. Lanczos iteration
and LOBPCG belong to a family of methods involving projections onto Krylov
subspaces; for details, see \cite[\S 32, \S 36]{trefethen1997numerical} and
\cite[\S 10]{golub2013matrix}.

\section{Historical examples}\label{s-historical-examples}
In this section
we show that many existing embedding methods reduce to solving a quadratic
MDE problem~\eqref{eq-quad-mde}, with different methods using different weights.

\paragraph{Laplacian embedding.} When the weights are all nonnegative, the
matrix $L$ is a Laplacian matrix for the graph $(\items, \edges)$
(with edge weights $w_{ij}$). The bottom eigenvector $v_1$ of $L$ is a multiple
of $\ones$, so a solution is $X = [v_2 \cdots v_{\embdim+1}]$; an embedding
obtained in this way is known as a Laplacian embedding or spectral embedding.
Laplacian embedding is the key computation in spectral clustering, a technique
for clustering the nodes of a graph \cite{pothen1990partitioning,
von2007tutorial}. (We mention that spectral clustering can be extended to
graphs with negative weights \cite{kunegis2010spectral, knyazev2017signed,
knyazev2018spectral}). In machine learning, Laplacian embedding is a popular
tool for dimensionality reduction, known as the Laplacian eigenmap
\cite{belkin2002laplacian}.

A variant of the Laplacian embedding uses the bottom
eigenvectors of the \emph{normalized} Laplacian matrix, $L_{\mathrm{norm}} =
D^{-1/2}LD^{-1/2}$, where $D \in \reals^{n \times n}$ is a diagonal matrix with
entries $D_{ii} = L_{ii}$. This variant can be expressed as a
quadratic MDE problem via the change of variables $Y = D^{-1/2}X$. 

\paragraph{Principal component analysis.}
PCA starts with
data $y_1, \ldots, y_n \in \reals^q$, assumed to be centered (and with $q \geq
\embdim$). The data are assembled into a matrix $Y \in \reals^{n \times q}$ with rows
$y_i^T$. The choice of weights
\[
w_{ij} = y_i^T y_j
\]
for $(i, j) \in \edges = \{(i, j) \mid 1 \leq i < j \leq n\}$ yields an MDE problem that
is equivalent to PCA, in the following sense. The matrix $L$ corresponding to this choice of weights has entries
\[
L_{ij} = \begin{cases}
-y_i^T y_j & i \neq j \\
\sum_{j : i \neq j} y_i^T y_j & i = j.
\end{cases}
\]
Because the data is centered,
\[
\sum_{j : i \neq j} y_i^T y_j = -y_i^T y_i, \quad i=1, \ldots, n,
\]
so the diagonal of $L$ has entries $-y_1^Ty_1, \ldots, -y_n^Ty_n$.
In particular, $L = -YY^T$. Hence, the low-dimensional
embedding of $Y$ obtained by PCA, which takes $X$ to be
the $\embdim$ top eigenvectors of $YY^T$ \cite[\S2]{udell2016generalized}, is
also a solution of the MDE problem.

The MDE formulation of PCA has a natural interpretation, based on the
angles between the data vectors. When the angle between
$y_i$ and $y_j$ is acute, the weight $w_{ij}$ is positive, so items
$i$ and $j$ are considered similar. When the angle is obtuse, the weight
$w_{ij}$ is negative, so items $i$ and $j$ are considered
dissimilar. When the data vectors are orthogonal, $w_{ij}$ is zero,
so the embedding is neutral on the pair $i$, $j$.
If the data vectors each have zero mean, there is an additional
interpretation: in this case, PCA seeks to place vectors associated with
positively correlated pairs near each other, anti-correlated pairs far from
each other, and is neutral on uncorrelated pairs.

\paragraph{Kernel PCA.} Like PCA, kernel PCA starts with vectors $y_1, \ldots,
y_n$, but it replaces these vectors
by nonlinear transformations $\phi(y_1), \ldots, \phi(y_n)$. We define the kernel
matrix $K \in \symm^{n}$ of $\phi(y_1), \ldots, \phi(y_n)$ as
\[
K_{ij} = \phi(y_i)^T \phi(y_j).
\]
Kernel PCA applies PCA to the matrix $K^{1/2}(I - \ones\ones^T/n)$, \ie, it
computes the bottom eigenvectors of the matrix
\[
L = -(I - \ones \ones^T / n)K(I - \ones \ones^T / n).
\]
By construction, $L$ is
symmetric, and its rows sum to $0$ (\ie, $L\ones = 0$); therefore, the choice
of weights
\[
w_{ij} = -L_{ij}, \quad 1 \leq i < j \leq n,
\]
yields an MDE problem equivalent to kernel PCA.

\paragraph{Locally linear embedding.} Locally linear embedding (LLE) seeks
an embedding so that each item is approximately reconstructed by a
linear combination of the embedding vectors associated with its nearest
neighbors \citep{roweis2000nonlinear, saul2001introduction}. As in PCA, the
data is a list of vectors $y_1, \ldots, y_n$. A
linearly-constrained least squares problem is solved to find a matrix $W \in \symm^{n}$
that minimizes the reconstruction error
\[
\sum_{i=1}^{n} \left\|y_i - \sum_{j=1}^{n} W_{ij} y_j\right\|^2_2,
\]
subject to the constraints that the rows of $W$ sum to $1$
and $W_{ij} = 0$ if $y_j$ is not among the $k$-nearest neighbors of $y_i$,
judged by the Euclidean distance ($k$ is a parameter). LLE then obtains an
embedding by computing the bottom eigenvectors of
\[
L = (I - W)^T (I - W).
\]
Once again, $L$ is symmetric with rows summing to $0$, so taking
$w_{ij} = -L_{ij}$ results in an equivalent MDE problem.

\paragraph{Classical multidimensional scaling.} Classical multidimensional
scaling (MDS) is an algorithm for embedding items, given original distances or
deviations $\delta_{ij}$ between all $n(n-1)/2$ item pairs
\cite{torgerson1952multidimensional}. The original distances $\delta_{ij}$ are
arranged into a matrix $D \in \symm^{n}$, with
\[
D_{ij} = \delta_{ij}^2.
\]
Classical MDS produces an embedding by computing the bottom eigenvectors of
\[
L = (I - \ones\ones^T / n)D(I - \ones\ones^T / n)/2.
\]
This matrix is symmetric, and its rows sum to $0$. Therefore,
choosing
\[
w_{ij} = -L_{ij}, \quad 1 \leq i < j \leq n,
\]
results in an MDE equivalent to classical MDS.

When the original distances are Euclidean, the weights $w_{ij}$ are the inner
products between the (centered) points that generated the original distances;
if these points were in fact vectors in $\reals^{\embdim}$, then classical MDS
will recover them up to rotation. But when
the original distances are not Euclidean, it is not clear what $w_{ij}$
represents, or how the optimal embedding distances relate to the original
distances. In \S\ref{s-orig-dist}, we show how distortion functions can be
chosen to preserve original distances directly.

\paragraph{Isomap.} Isomap is a well-known dimensionality reduction method
 that reduces to classical MDS \cite{tenenbaum2000global, bernstein2000graph}
and therefore is also an MDE of the form \eqref{eq-quad-mde}\@. Like classical
MDS, Isomap starts with original distances $\delta_{ij}$ for all pairs of
items. Isomap then constructs a shortest path metric on the items, and
runs classical MDS on that metric. Specifically, it constructs a graph with $n$
nodes, in which an edge exists between $i$ and $j$ if $\delta_{ij} < \epsilon$,
where $\epsilon > 0$ is a parameter. If an edge between $i$ and $j$ exists it
is weighted by $\delta_{ij}$, and the shortest path metric is constructed
from the weighted graph in the obvious way.

\paragraph{Maximum variance unfolding.} Maximum variance unfolding is another
dimensionality reduction method that starts with an original data
matrix and reduces to PCA \cite{weinberger2004unsupervised}. For each item $i$,
maximum variance unfolding computes its $k$-nearest neighbors under the
Euclidean distance, where $k$ is a parameter, obtaining $nk$ original distances
$\delta_{ij}$. Next, the method computes a (centered) Gram matrix $G$ of
maximum variance that is consistent with the $\delta_{ij}$, \ie, such that
$G_{ii} - 2G_{ij} + G_{jj} = \delta^2_{ij}$ for each $\delta_{ij}$; this matrix
can be found by solving a semidefinite program
\cite[\S3.2]{weinberger2004unsupervised}. Finally, like PCA, maximum variance
unfolding takes the $\embdim$ top eigenvectors of $G$ (or the $\embdim$ bottom
eigenvectors of $L = -G$) as the embedding. (When solving the semidefinite
program is difficult, maximum variance unfolding can be approximated by a
method called maximum variance correction \cite{chen2013maximum}.)

\chapter{Distortion Functions}\label{c-distortion-functions}
In this chapter 
we give examples of distortion functions.  We organize these into 
those that derive from given weights,
and those that derive from given original distances or
deviations between items, although the distinction is not sharp.

\section{Functions involving weights}\label{s-weights}
In many MDE problems, the similarities between items are described by nonzero
weights $w_{ij}$ for $(i, j) \in \edges$, 
with positive weights indicating similar items and negative weights
indicating dissimilar items.
In this setting, the edges are partitioned into two sets: $\esim$
(positive weights, \ie, similar items) and
$\edis$ (negative weights, dissimilar items).

We mention some special cases.
In the simplest case, all weights are equal to one. This means that the
original data simply tells us which pairs of items are similar. 
Another common situation is when all weights are positive, which
specifies varying levels of
similarity among the items, but does not specify dissimilarity. 
Another common case is when the weights have only two values,
$+1$ and $-1$.  In this case we are specifying a set of pairs of similar
items, and a (different) set of pairs of dissimilar items.
In the most general case, 
the weights can be positive or negative, indicating similar and
dissimilar pairs of items, and different degrees of similarity and dissimilarity.

The weights are given as, or derive from,
the raw data of an application.  As an example consider
a social network, with $(i, j) \in \edges$
meaning individuals $i$ and $j$ are acquainted.
A reasonable choice for the weight $w_{ij}$ might be
one plus the number of acquaintances individuals $i$ and $j$ have in
common.
(This provides an example of preprocessing, \ie, deriving weights
from the original data, which in this example is the unweighted acquaintance graph.
We describe several other preprocessing methods in \S\ref{s-preprocessing}.)

\paragraph{Distortion functions.}
For problems involving weights, distortion functions have the form
\begin{equation}\label{eq-penalty}
f_{ij}(d_{ij}) =
\begin{cases}
  w_{ij} \psim (d_{ij}) & (i, j) \in \edges_{\text{sim}} \\
  w_{ij} \pdis (d_{ij}) & (i, j) \in \edges_{\text{dis}}
\end{cases}, \qquad (i, j) \in \edges,
\end{equation}
where $\psim: \reals_+ \to \reals$ and $\pdis: \reals_+ \to
\reals$ are increasing penalty functions.
Thus for $(i, j) \in \edges_{\text{sim}}$ ($w_{ij} \geq 0$), the distortion
function $f_{ij}$ is \emph{attractive} (increasing). For $(i, j) \in
\edges_{\text{dis}}$ ($w_{ij}<0$), $f_{ij}$ is \emph{repulsive} (decreasing).
When $\edges_{\text{dis}}$ is empty, the weights are all positive and the
objective $E$ is attractive. In this case the unconstrained problem has the
trivial solution $X=0$; we can use an anchor or standardization constraint to
enforce spreading and avoid this pathology.

We have found that the choice of a penalty function matters much more than the
choice of weights (in many applications, using $+1$ and $-1$ weights suffices).
While there are many possible penalty functions, in practice they
can be characterized by how they treat small distances, and how they treat
large distances. These two vague qualities largely determine the
characteristics of the embedding. For this reason, it suffices to use just a
few simple penalty functions (out of the many possible choices). In the next
two subsections, we describe some attractive and repulsive penalty functions
that we have found to be useful.

\subsection{Attractive distortion functions}
Here we give a few examples of functions that work well as attractive
penalties. We start with powers, the simplest family of penalty functions.
This family is parametrized by the exponent, which
determines the relative contributions of small and large distances to the
distortion. Then we describe three penalties that have gentle slopes for small
distances, and larger (but not large) slopes for large distances; these
typically result in embeddings in which different ``classes'' of items are
well-separated, but items within a single class are not too close.

\begin{figure}
\includegraphics{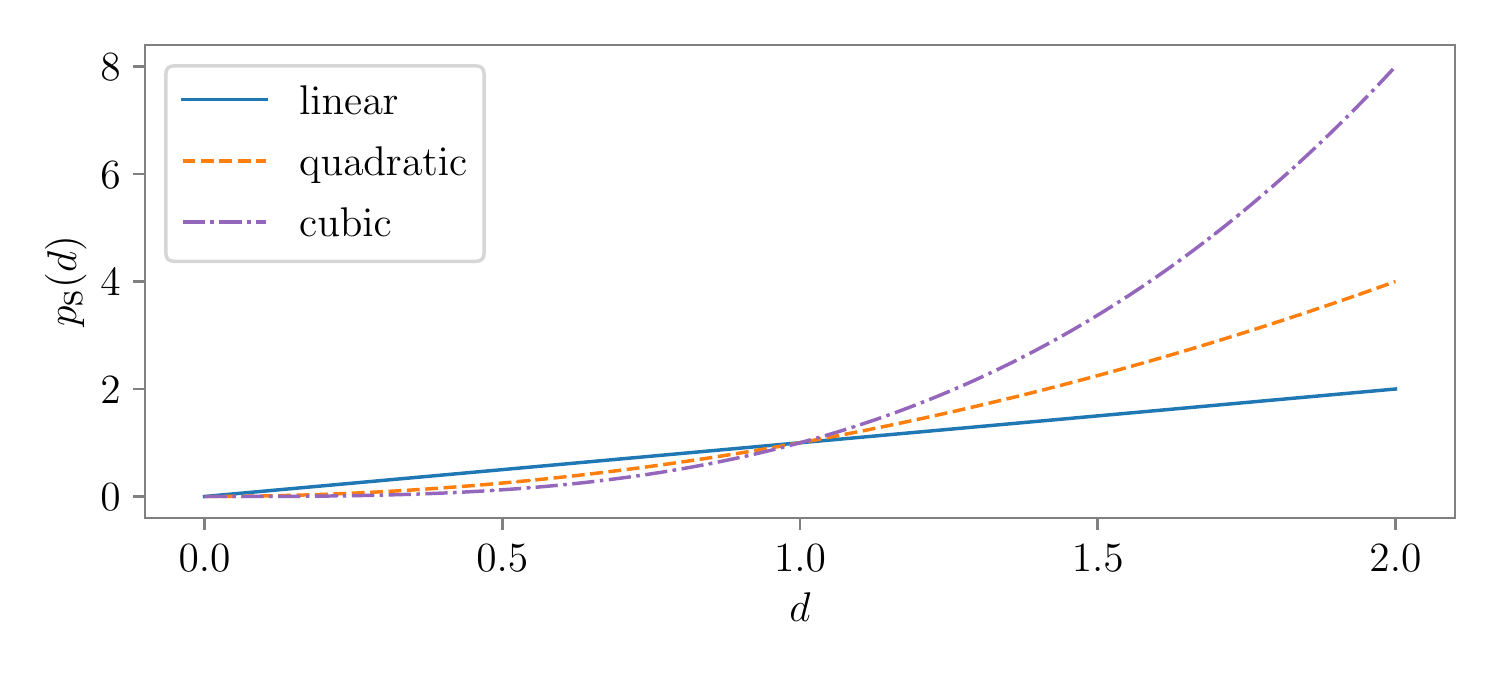}
\caption{\emph{Powers.} A linear, quadratic, and cubic penalty.}
\label{f-power-penalties}
\end{figure}

\paragraph{Powers.} The simplest family of penalty functions is the power
penalty, 
\[
\psim(d)= d^\alpha,
\]
where $\alpha > 0$ is a parameter.
The larger $\alpha$ is, the more heavily
large distances are penalized, and the less heavily small distances are
penalized. This is illustrated in figure~\ref{f-power-penalties}, which
plots linear ($\alpha = 1$), quadratic ($\alpha = 2$), and cubic ($\alpha =
3$) penalties. For $x > 1$, the cubic penalty is larger than than
the quadratic, which is in turn larger than the linear penalty. For $x<1$,
the opposite is true.

Decreasing $\alpha$ increases the penalty for small distances and decreases
the slope for large distances. So smaller values of $\alpha$ causes similar
items to more closely cluster together in the embedding, while allowing for
some embedding distances to be large. In contrast when $\alpha$ is large, the
slope of the penalty for small distances is small and the slope for large
distances is large; this results in embeddings in which the points
are ``spread out'' so that no one embedding distance is too large.

\begin{figure}
\includegraphics{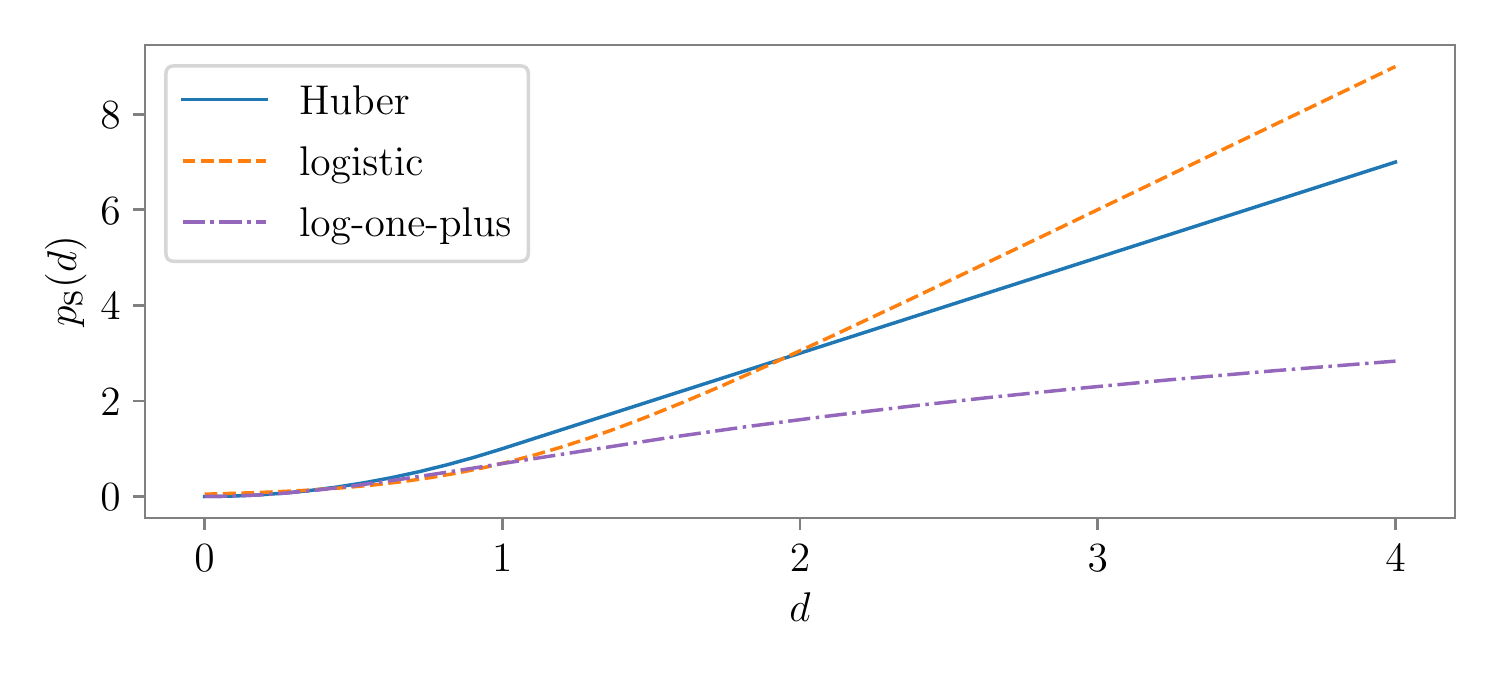}
\caption{\emph{More penalties.} A Huber, logistic, and log-one-plus penalty.}
\label{f-more-penalties}
\end{figure}

\paragraph{Huber penalty.} The Huber penalty is
\begin{equation}\label{eq-huber-penalty}
\psim(d) =
\begin{cases}
d^2 & d < \tau \\
\tau(2d - \tau) & d \geq \tau,
\end{cases}
\end{equation}
where $\tau$ (the threshold) is a positive parameter.
This penalty function is quadratic for $d < \tau$ and affine for $d>\tau$.
Because the penalty is quadratic
for small distances, the embedding vectors do not cluster too closely together;
because it is linear for large distances, some embedding distances may be
large, though most will be small. 

\paragraph{Logistic penalty.} The logistic penalty has the form
\[
\psim (d) = \log (1 + e^{\alpha(d  - \tau)}),
\]
where $\alpha > 0$ and $\tau > 0$ are parameters. The logistic penalty
encourages similar items to be smaller than the threshold $\tau$, while
charging a roughly linear cost to distances larger than $\tau$.

\paragraph{Log-one-plus penalty.} The log-one-plus penalty is
\begin{equation}\label{eq-log1p-penalty}
\psim (d) = \log(1 + d^\alpha),
\end{equation}
where $\alpha > 0$ is a parameter. For small $d$, the log-one-plus penalty
is close to $d^\alpha$, a power penalty with exponent $\alpha$, 
but for large $d$, $\log (1 + d^\alpha) \approx \alpha \log d$, 
which is much smaller than $d^\alpha$.

The dimensionality reduction
methods t-SNE \cite{maaten2008visualizing}, LargeVis
\cite{tang2016visualizing}, and UMAP \cite{mcinnes2018umap} use this function
(or a simple variant of it) as an attractive penalty.

\subsection{Repulsive distortion functions}\label{s-repulsive}
Repulsive distortion functions are used to discourage vectors associated
with dissimilar items from being close.
Useful repulsive penalties are \emph{barrier functions}, with $\pdis(d)$
converging to $-\infty$ as $d \to 0$, and converging to $0$ as $d \to \infty$.
This means that the distortion, which has the form $w_{ij} \pdis(d_{ij})$ with
$w_{ij}<0$, grows to $\infty$ as the distance $d_{ij}$ approaches zero, and
converges to zero as $d_{ij}$ becomes large.

\begin{figure}
\includegraphics{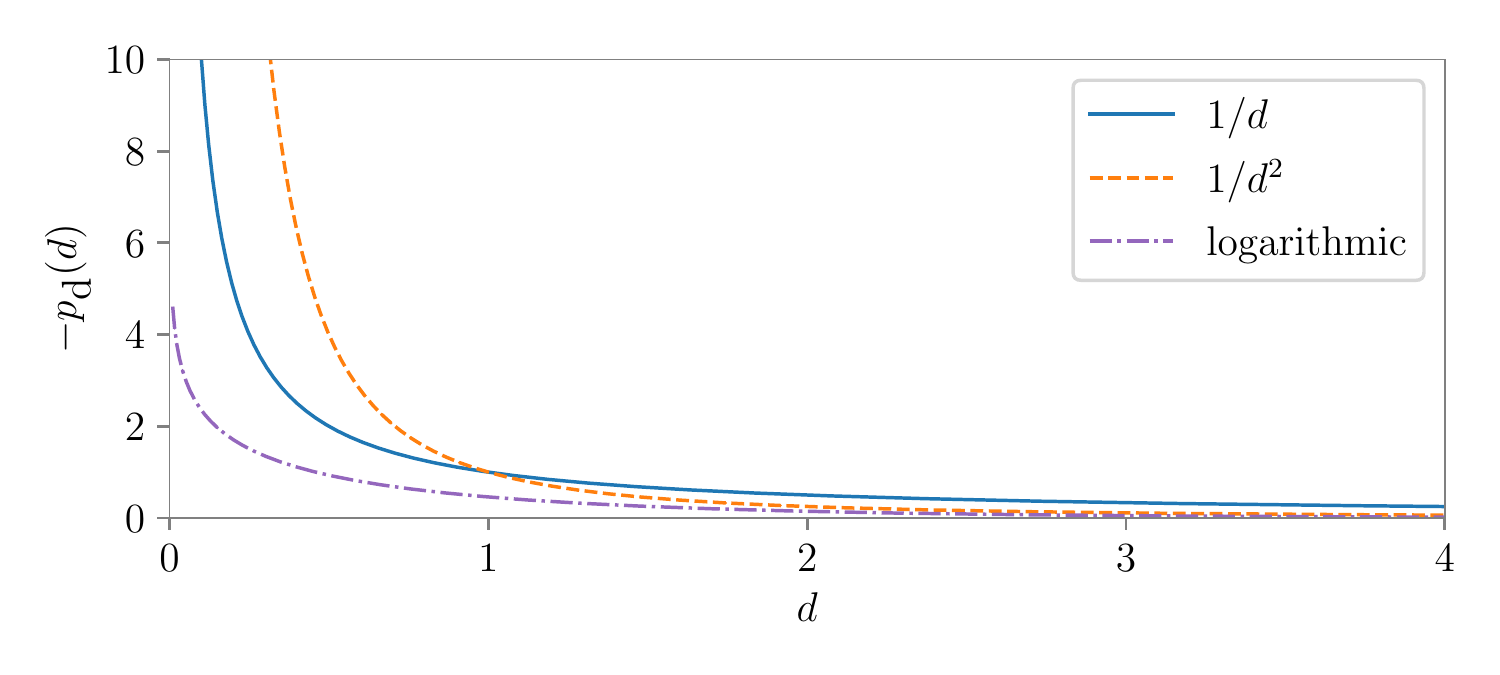}
\caption{\emph{Repulsive penalties.} Inverse power and logarithmic penalties.}
\label{f-repulsive-penalties}
\end{figure}

We give two examples of such penalties below, and plot them in
figure~\ref{f-repulsive-penalties}. 

\paragraph{Inverse power penalty.} The inverse power penalty is
\[
  \pdis(d) = -1/d^{\alpha},
\]
where $\alpha > 0$ is a parameter.
When used in unconstrained MDE problems, this penalty sometimes results
in a few embedding vectors being very far from the others. This pathology
can be avoided by imposing a standardization constraint.

\paragraph{Logarithmic penalty.}
The logarithmic penalty is
\begin{equation}\label{eq-log-penalty}
  \pdis(d) = \log(1 - \exp(-d^\alpha)),
\end{equation}
where $\alpha > 0$ is a parameter. We have found that the logarithmic penalty
is effective with or without the standardization constraint.  We will see some
examples of embeddings produced using logarithmic penalties in
part~\ref{p-examples} of this monograph.

Many other repulsive penalties can and have been used, including for example
the barrier 
\[
\pdis(d) = \log\left(\frac{d^\alpha}{1+d^\alpha}\right),
\]
with $\alpha > 0$, used by
LargeVis \cite{tang2016visualizing} and UMAP \cite[appendix C]{mcinnes2018umap}.
In fact, LargeVis and UMAP are equivalent to unconstrained MDE problems based
on the log-one-plus attractive penalty and this repulsive penalty. The
dimensionality reduction method t-SNE is almost equivalent to solving an MDE
problem, except its objective encourages spreading via a non-separable
function of the embedding distances. In practice, however, these three methods
produce similar embeddings when their hyper-parameters are suitably adjusted
\cite{bohm2020unifying}. While there are many choices for repulsive (and
attractive) penalties, in our experience, just a few simple functions, such as
the ones described above, suffice.

Finally, we remark that it is possible to design distortion functions
derived from weights that combine attractive and repulsive penalties. For
example, the function
\[
  f_{ij}(d_{ij}) = w_{ij}d_{ij} - \log (d_{ij} - \rho_{ij}),
\]
where $w_{ij} > 0$ and $\rho_{ij} > 0$, combines a linear penalty that attracts
items $i$ and $j$ with a logarithmic barrier that repulses them. MDE problems
using distortion functions like this one can be used to solve Euclidean
placement and sphere-packing problems \citep[\S8.7]{bv2004convex}. 

\section{Functions involving original distances}\label{s-orig-dist}
In another class of MDE problems, for each item pair $(i, j) \in \edges$, we
are given a nonnegative number $\delta_{ij}$ representing the
distance, deviation, or dissimilarity between items $i$ and $j$.  A large value
of $\delta_{ij}$ means $i$ and $j$ are very dissimilar, while $\delta_{ij} = 0$
means items $i$ and $j$ are the same, or at least, very similar.
We can think of $\delta_{ij}$ as the target distance for our embedding.

If the items can be represented by vectors (\eg, images,
time-series data, bitstrings, and many other types
of items), the original distances might be generated by a standard distance
function, such as the Euclidean, Hamming, or earth-mover's distance.
But the data $\delta_{ij}$ need not be metric; for example, they
can fail to satisfy the triangle inequality $\delta_{ij} \leq
\delta_{ik}+ \delta_{kj}$, for $i<k<j$, $(i,j),(i,k), (k,j)\in \edges$. This
often happens when the $\delta_{ij}$ are scored by human experts.

\paragraph{Distortion functions.}
For problems involving distances, we consider distortion functions of the form
\[
f_{ij}(d_{ij}) = \ell(\delta_{ij}, d_{ij}), \quad (i, j) \in \edges,
\]
where $\ell : \reals_+ \times \reals_{+} \to \reals$ is a loss function.
The loss function is nonnegative, zero for $d=\delta$, 
decreasing for $d<\delta$, and increasing for $d>\delta$.
The embedding objective $E(X)$ is a measure of how closely the 
embedding distances match the original deviations.
Perfect embedding corresponds to $E(X)=0$, which means $d_{ij}=\delta_{ij}$,
\ie, the embedding exactly preserves the original distances.
This is generally impossible to achieve, for example when the original
deviations are not a metric, and also in many cases even when it is.
(For example, it is impossible to isometrically embed a four-point
star graph with the shortest path metric into $\reals^m$, for any $m$.)
Instead, as in all MDE problems, we seek embeddings with low average
distortion.

In these problems, we do not require the standardization constraint
to encourage the embedding to spread out, though we may choose to enforce it
anyway, for example to obtain uncorrelated features.
When the standardization
constraint is imposed, it is important to scale the original distances
$\delta_{ij}$ so that their RMS value is not too far from the RMS value $\dnat$
\eqref{eq-d-nat} of the embedding distances $d_{ij}$.

When the original deviations $\delta_{ij}$ are Euclidean distances, it is
sometimes possible to compute a perfect embedding (see
\cite[\S8.3.3]{bv2004convex} and \cite{liberti2014euclidean,
dokmanic2015euclidean}). In general, however, MDE problems involving original
distances are intractable. Nonetheless, the methods described in
chapter~\ref{c-alg} often produce satisfactory embeddings.

\subsection{Examples}
\paragraph{Quadratic loss.}
Perhaps the most obvious choice of loss function is the squared loss,
\[
\ell(\delta, d) = (\delta - d)^2.
\]
The squared loss places more emphasis on preserving the
distance between pairs with large original distances than small ones.
In particular,
when $\edges$ contains all pairs
and the feasible set is $\stdemb$, a simple calculation shows
that the least-squares MDE is equivalent to the problem
\begin{equation*}
\begin{array}{ll}
\mbox{minimize} & (1/p)\sum_{(i, j) \in \mathcal E} -\delta_{ij} d_{ij}\\
\mbox{subject to} & X \in \mathcal \stdemb,
\end{array}
\end{equation*}
which we recognize as a linear distortion function, with coefficient
the negative of the original distance.
The objective of this problem gives outsized rewards to matching large original
distances with large embedding distances.

The problem of finding an embedding that minimizes the least squares distortion (or
simple variants of it) has been discussed extensively in the literature. A
whole family of methods called multidimensional scaling is almost entirely
concerned with (approximately) solving it \cite{kruskal1964multidimensional,
kruskal1964nonmetric, groenen1996least, cox2000mds, borg2003modern}.

\begin{figure}
\includegraphics{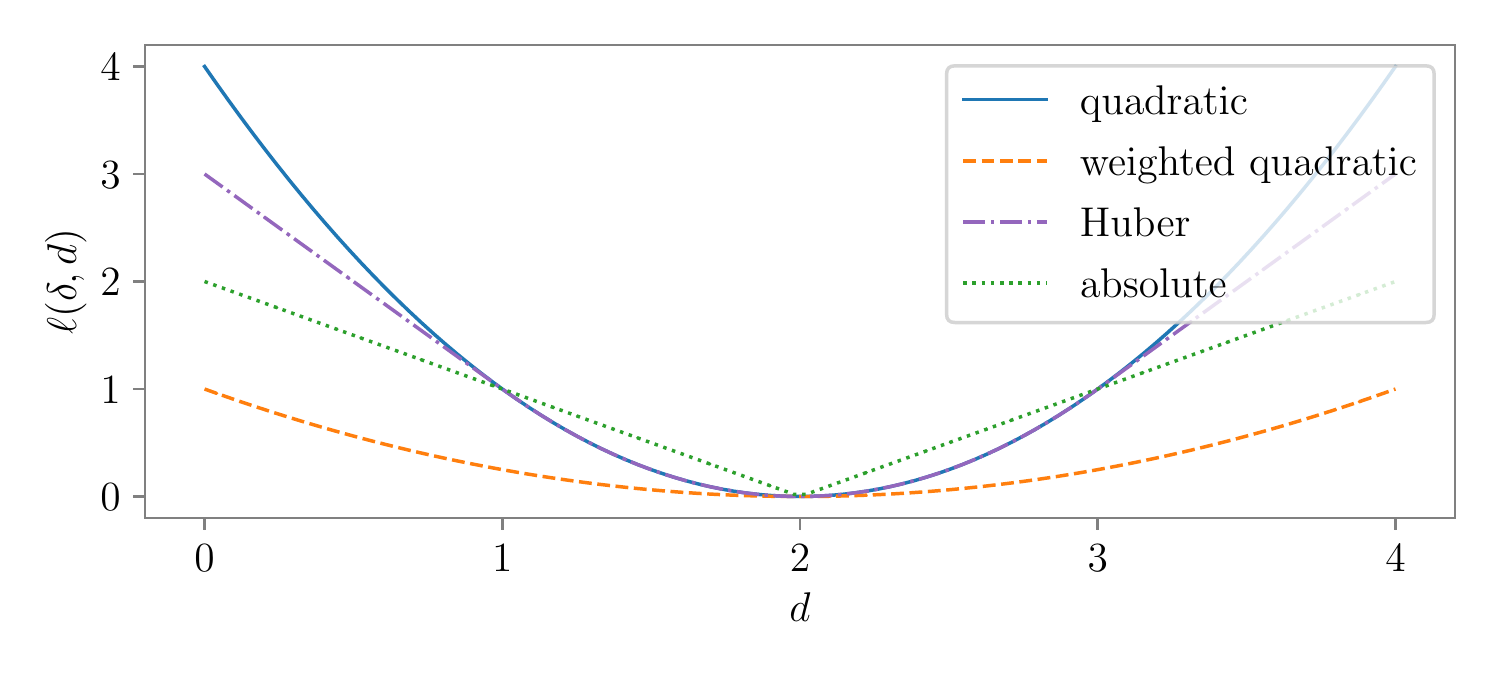}
\caption{\emph{Basic losses involving distances.} A quadratic loss, weighted
quadratic (with $\kappa(\delta) = 1/\delta^2$), and a Huber loss (with threshold
$\tau = 1$), evaluated at $\delta = 2$.}
\label{fig-basic-losses-from-dist}
\end{figure}

\paragraph{Weighted quadratic loss.} A weighted quadratic loss
is
\[
\ell(\delta, d) = \kappa(\delta)(\delta - d)^2,
\]
where $\kappa$ is a real function. If $\kappa$ is decreasing, the weighted
quadratic loss places less emphasis on large distances $\delta$, compared
to the quadratic loss.
The choice $\kappa(\delta)
= 1/\delta$ and $\mathcal X = \reals^{n \times \embdim}$ yields a method known
as Sammon's mapping \cite{sammon1969nonlinear}, while $\kappa(\delta) =
1/\delta^2$ gives the objective function for the Kamada-Kawai algorithm
for drawing graphs \citep{kamada1989algorithm}.

\paragraph{Huber loss.} The Huber loss is
\[
\ell(\delta, d) =
\begin{cases}
(\delta - d)^2 & |\delta - d| \leq \tau \\
\tau(2|\delta - d| - \tau) & |\delta - d| > \tau,
\end{cases}
\]
where $\tau \in \reals$ is a parameter \cite[\S6.1.2]{bv2004convex}. This loss can
be thought of as a robust version of the quadratic loss, in that it is less
sensitive to large residuals. This is illustrated in figure~\ref{fig-basic-losses-from-dist}.

\paragraph{Absolute loss.} The absolute loss is
\begin{equation}\label{eq-absolute-loss}
\ell(\delta, d) = |\delta - d|.
\end{equation}
MDE problems with this loss function have been
referred to as robust MDS and robust EDM in the literature
\cite{cayton2006robust, zhou2019robust}.

\paragraph{Logistic loss.}
Composing the logistic function with the absolute distortion gives
a soft version of the absolute distortion,
\[
\ell(\delta, d) = \log \left( \frac{1 + \exp|\delta - d|}{2}\right).
\]
(Dividing the argument of the log by two has no effect on the MDE problem, but ensures that
$\ell(\delta,\delta)=0$.)

\begin{figure}
\includegraphics{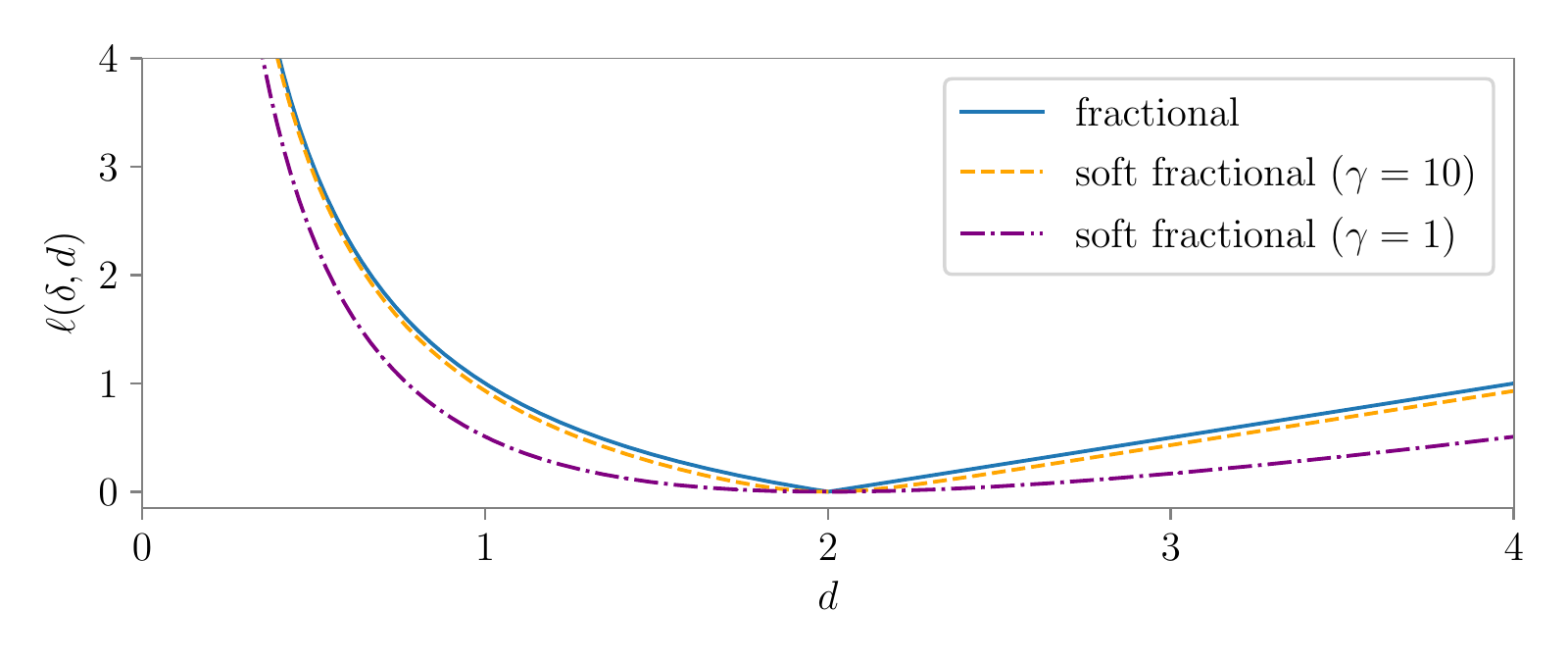}
\caption{\emph{Fractional loss.} The fractional and soft fractional losses,
with $\delta=2$.}
\label{fig-fractional}
\end{figure}

\paragraph{Fractional loss.}
The fractional loss is defined as
\[
\ell(\delta, d) = \max\{\delta/d, d/\delta\} - 1.
\]
It is a barrier function: it approaches $+\infty$ as $d \to 0$, 
naturally encouraging the embedding vectors to spread out.

\paragraph{Soft fractional loss.}
A differentiable approximation of the fractional loss can be obtained
by replacing the max function with the so-called softmax function,
\[
\ell(\delta, d) = \frac{1}{\gamma} \log\left(
\frac{\exp(\gamma\delta/d) + \exp(\gamma d/\delta)}{2 \exp \gamma}\right),
\]
where $\gamma>0$ is a parameter.  For large $\gamma$, this distortion function
is close to the fractional distortion; for small
$\gamma$ it is close to $\delta/d + d/\delta-2$.
The soft fractional distortion is compared to the fraction distortion
in figure~\ref{fig-fractional}.

\section{Preprocessing}\label{s-preprocssing}
In this section we discuss methods that can be 
used to create distortion functions, given some raw 
similarity or dissimilarity data on pairs of items.
These preprocessing methods are analogous to feature engineering
in machine learning, where raw features are converted to feature
vectors and possibly transformed before further processing.
As in feature engineering, preprocessing the raw data about items
can have a strong effect on the final result, \ie, the embedding
found.

Below, we discuss two preprocessing methods. The first method, based on
neighborhoods, can be used to create distortion functions based on weights (and
to choose which pairs to include in the sets $\esim$ and $\edis$). The second
method, based on graph distances, is useful in creating distortion
functions based on original deviations. These methods can also be combined,
as explained later.

\subsection{Neighborhoods}
In some applications we have an original deviation $\delta_{ij}$ or 
weight $w_{ij}$ for every pair $(i,j)$, but we care mostly about
identifying pairs that are very similar.  These correspond 
to pairs where the deviation is small, or the weight is large.
For raw data given as deviations, the \emph{neighborhood} of an item $i$
is
\[
  \mathcal{N}(i; \dthresh_i) = \{ j \mid \delta_{ij} \leq \delta^\text{thres}_{i} \},
\]
and for raw data given as weights, the neighborhood is
\[
  \mathcal{N}(i; \wthresh_i) = \{ j \mid w_{ij} \geq w^\text{thres}_{i} \},
\]
where $\delta^{\mathrm{thres}}_i$ and $w^{\mathrm{thres}}_i$ are threshold values.

If item $j$ is in the neighborhood of item $i$ (and $j \neq i$), we say that
$j$ is a \emph{neighbor} of $i$. Note that this definition is not symmetric: $j
\in \mathcal N(i)$ does not imply $i \in \mathcal N (j)$. The graph whose edges
connect neighbors is called a \emph{neighborhood graph}, with edges
\[
  \esim = \{(i, j) \mid 1 \leq i < j \leq n, ~
    j \in \mathcal{N}(i) \textnormal{ or } i \in \mathcal{N}(j) \}.
\]

\paragraph{Building the neighborhood graph.}
The neighborhood graph depends on the threshold values
$\delta^{\mathrm{thres}}_i$ and $w^{\mathrm{thres}}_i$. A simple option is to
use the same threshold $\dthresh$ or $\wthresh$ for all items, 
but we can also
use different thresholds for different items. An effective way to do this is
based on the \emph{nearest neighbors} of each item. For original data given as
deviations, we choose
\[
  \dthresh_i = \sup\{\delta \mid |\mathcal{N}(i; \delta)| \leq k\},
\]
and analogously for weights; that is, we choose the thresholds so that each
item has $k$ neighbors, where $k$ is a parameter that is much smaller than $n$.
More sophisticated methods of building neighborhood graphs are possible,
such as the one proposed in \citep{carreira2005proximity}, but this simple
choice often works very well in practice.

Computing the $k$-nearest neighbors of each item is expensive. In almost all
applications, however, it suffices to carry out this computation approximately.
Approximate nearest neighbors can be computed cheaply using algorithms such as
the one introduced in \cite{dong2011efficient}, which has a reported empirical
complexity of $O(n^{1.14})$, or the methods surveyed in
\cite{andoni2018approximate}. Python libraries such as \texttt{pynndescent}
\cite{pynndescent} and \texttt{annoy} \cite{annoy} make these computations
straightforward in practice.

\paragraph{Choosing dissimilar items.} 
A natural choice for the set of dissimilar items is
\[
  \edis = \{(i, j) \mid (i, j) \notin \esim\},
\]
\ie, all pairs of non-neighbors. This says that all items that are not similar
are dissimilar. If $n$ is large, this choice is untenable because
$\esim \cup \edis$ will include all $n(n-1)/2$ pairs. A practical alternative
is to randomly select a subset of non-neighbors, and include only these edges
in $\edis$. The size of $\edis$ affects how spread out the embedding is; in our
experience, an effective choice is to sample $|\esim|$ non-neighbors uniformly
at random. (If a standardization constraint or anchor constraint is imposed, we
can also take $\edis = \emptyset$.) Several well-known
embedding methods, such as word2vec, LargeVis, and UMAP, choose dissimilar
pairs via random sampling (these methods call this ``negative sampling'')
\cite{mikolov2013distributed, tang2016visualizing, mcinnes2018umap, bohm2020unifying}.

Other choices are possible. Instead of including all pairs
not in $\esim$, we can take $\edis$ to only include pairs of items
with a large original distance (or small original weight). This can yield
embeddings in which similar items are tightly clustered together, and are far
away from dissimilar items.

\paragraph{Choosing weights.}
After constructing $\esim$ and $\edis$, we assign distortion functions to
edges. We have already discussed how to choose penalty functions, in
\S\ref{s-weights}. Here we discuss the choice of weights $w_{ij}$ in the
distortion functions.

The simplest choice of weights is $w_{ij} = +1$ for $(i,j)\in \esim$ and
$w_{ij} = -1$ for $(i,j) \in \edis$. Though exceedingly simple, this weighting
often works very well in practice. Another effective choice takes
$w_{ij} = +2$ if $i \in \mathcal N(j)$ and $j \in \mathcal N(i)$, $w_{ij} = +1$
if $i \in \mathcal N (j)$ but $j \not\in \mathcal N(i)$ (or vice versa), and
$w_{ij} = -1$ if $(i, j) \in \edis$.

The weights can also depend on the raw weights or deviations, as in
\[
  w_{ij} = \exp(-\delta_{ij}^2/\sigma^2),
\]
where $\sigma$ is a positive parameter.
In this preprocessing step, the thresholds and parameters such as $\sigma$ are
chosen, usually by experimentation, to get good results.

\paragraph{Sparsity.} Preprocessing methods based on neighborhoods focus on
the \emph{local} structure in the raw data. Roughly speaking, we trust the
original data to
indicate pairs of items that are similar, but less so which items are
dissimilar. These preprocessing steps typically yield a sparse graph, with
many fewer than $n(n-1)/2$ edges.  In many cases the average degree of the
vertices in the graph is modest, say, on the order of 10, in which case number
of edges $p$ is a modest multiple of the number of items $n$. This makes
problems with $n$ large, say $10^6$, tractable; it would not be practical to
handle all $10^{12}$ edges in a full graph, but handling $10^7$ edges is quite
tractable.

\subsection{Graph distance}
The previous preprocessing step focuses on the local structure of 
the raw data, and generally yields a sparse graph;
we can also use sparse original data to create a full graph.
We consider the case with original deviations $\delta_{ij}$, for some or
possibly all pairs $(i, j)$. We replace these with the graph distance $\tilde
\delta_{ij}$ between vertices $i$ and $j$, defined as the minimum sum of
deviations $\delta_{ij}$ over all paths between $i$ and $j$. (If the original
graph is complete and the original
deviations satisfy the triangle inequality, then this step does nothing, and we
have $\tilde \delta_{ij} = \delta_{ij}$.)

We mention that the graph distances can be computed efficiently, using
standard algorithms for computing shortest paths (such as Dijkstra's algorithm,
or breadth-first search for unweighted graphs): simply run the shortest-path
algorithm for each node in the graph. While the time complexity of computing all
$n(n-1)/2$ graph distances is $\tilde O(n^2 + np)$, the computation is
embarrassingly parallel across the nodes, so in practice this step does not
take too long. (For example, on a standard machine with 8 cores, 300 million
graph distances can be computed in roughly one minute.)

\paragraph{Sparsity.}
This operation yields a complete graph with $n(n-1)/2$ edges (assuming the
original graph is connected). When $n$ is large, the complete graph might be
too large to fit in memory. To obtain sparser graphs, as a variation, we can
randomly sample a small fraction of the $n(n-1)/2$ graph distances, and use
only the edges associated with these sampled graph distances. This variation can
be computed efficiently, without storing all $n(n-1)/2$ graph distances in
memory, by iteratively sampling the graph distances for each node.

Another sparsifying variation is to limit the length of paths used to determine the
new deviations. For example, we can define $\tilde \delta_{ij}$ as the minimum
sum of original deviations over all paths between $i$ and $j$ of length up to
$L$. The parameter $L$ would be chosen experimentally.

\begin{figure}
\includegraphics{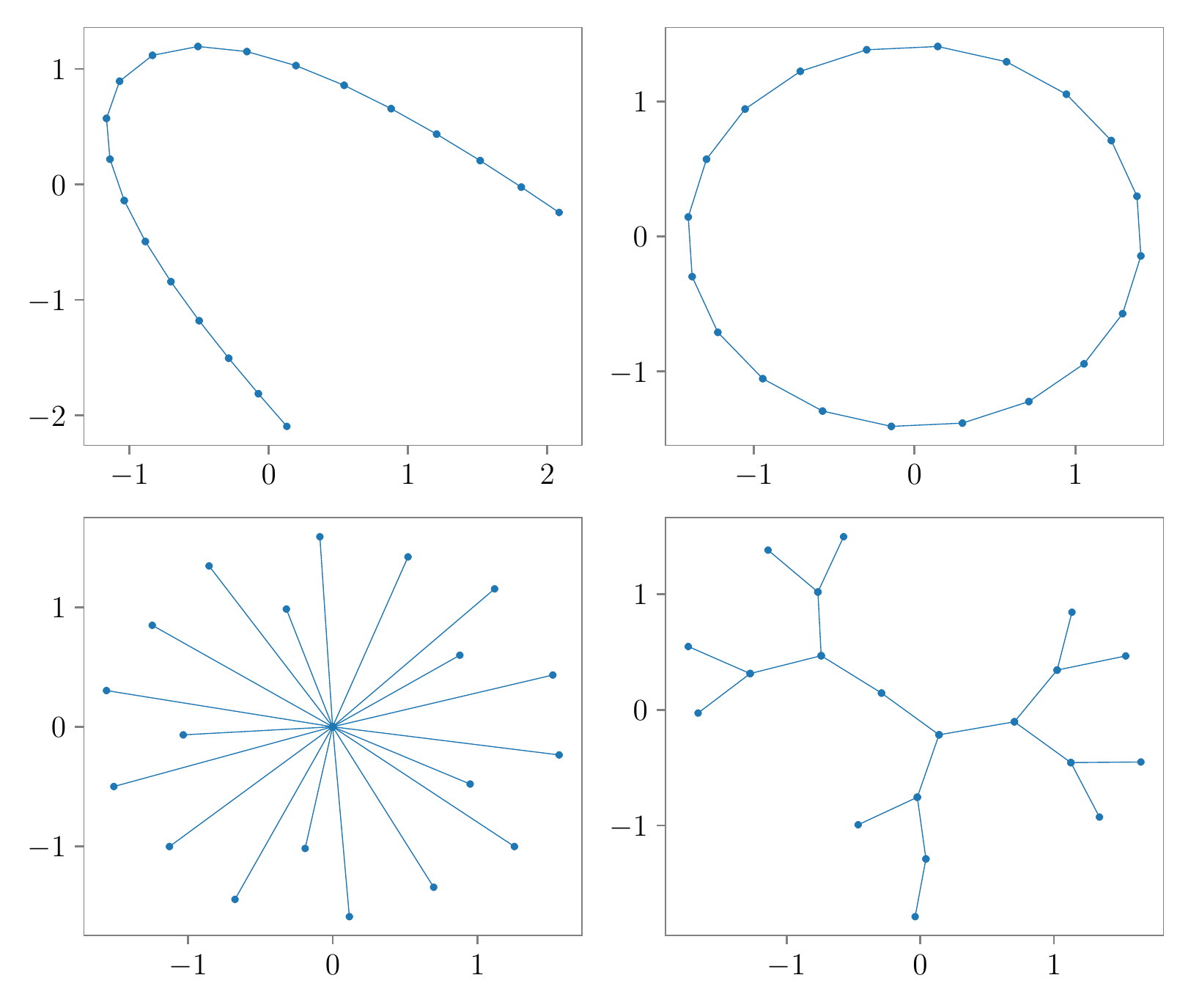}
\caption{\emph{Graph layout with weighted quadratic loss.} Standardized
embeddings of a chain graph, a cycle graph, a star graph, and a binary tree,
using the weighted quadratic loss derived from graph distances, with
$\kappa(\delta) = 1/\delta^2$.}
\label{f-graph-layout-kamada-kawai}
\end{figure}

\paragraph{Example.} As a simple example, we use this preprocessing step
to create MDE problems for visualizing graphs (similar to the examples
in \S\ref{s-simple-examples}). Figure~\ref{f-graph-layout-kamada-kawai} plots
standardized embeddings of a chain graph, a cycle graph, a star graph, and a
binary tree on $n = 20$ vertices, obtained by solving MDE problems with the
weighted quadratic loss and original distances $\delta_{ij}$ given by the graph
distance.

\subsection{Other methods}
Preprocessing based on shortest path distances and neighborhoods are readily
combined. As an example, we might first sparsify the original data, forming a
neighborhood graph. Then we define distortion functions as $f_{ij}(d_{ij}) =
d_{ij}^2$ if $(i,j)$ is a pair of neighbors, and $f_{ij} = (1/2) d_{ij}^2$ if
$(i,j)$ have distance two on the neighborhood graph.

Another natural preprocessing step is based on preference data.  Here, we are
given a set of comparisons between items.  This kind of data is often readily
available (\eg, surveys), though we mention the literature on preference
elicitation \citep{schouten2013optimizing} and aggregation
\citep{arrow1950difficulty,bradley1952rank,plackett1975analysis,luce2012individual,fligner1986distance,}
is rich and goes back centuries \citep{easley2010networks}, with a number of
advances continuing today
\citep{joachims2002optimizing,dwork2001rank,von2008designing}.  As a simple
example, imagine we have a set of rankings, expressing whether item $i$ is
preferred to item $j$.  In this case, it can make sense to set the deviations
$\delta_{ij}$ to any increasing function of the average distance between the
ranks of items $i$ and $j$.  As another example, we might have a set of survey
responses, and some of these responses might indicate items $i$ and $j$ are
more similar to each other than they are to item $k$.  There are many ways to
create weights from these kinds of comparisons.  A simple one is to set the
weights $w_{ij}$ to the number of times items $i$ and $j$ are indicated as the
most similar pair, minus the number of times they are not.  Thus, $\esim$
contains all pairs with $w_{ij} \geq 0$, while $\edis$ contains those with
$w_{ij} < 0$.

Evidently there are many reasonable preprocessing steps that can be used. As
in feature engineering for machine learning, there is some art in the choice of
preprocessing. But also as in feature engineering, a few simple preprocessing
steps, such as the ones described above, often work very well.

\part{Algorithms}\label{p-algorithms}

\chapter{Stationarity Conditions}\label{c-stationarity}
In this chapter we give conditions for an embedding $X \in \mathcal X \subseteq
\reals^{n \times m}$ to be stationary for MDE problems with differentiable
average distortion; these conditions are necessary, but not sufficient, for an
embedding to be optimal. The stationarity conditions presented here will guide
our development of algorithms for computing minimum-distortion embeddings, as
we will see in the
next chapter.

The stationarity conditions can be expressed as a certain matrix $G \in
\reals^{n \times m}$ being zero. This matrix $G$ is the projection of the
gradient $\nabla E(X)$ onto the (negative) tangent cone of $\mathcal X$ at $X$,
and it will play a crucial rule in the algorithms presented in
chapter~\ref{c-alg}. We give explicit expressions for $G$ for the constraint
sets $\mathcal C$ \eqref{e-center-constraint}, $\anchored$
\eqref{e-anchor-constraint}, and $\stdemb$ \eqref{eq-std}, in
\S\ref{s-unconstr-opt}, \S\ref{s-anchor-opt}. \S\ref{s-std-opt}.

Below, we develop and explain this simple stationarity condition, interpreting
$-G$ as the steepest feasible descent direction for $E$ at $X$ for the
constraint $X \in \constr$. We start by giving an expression for the gradient
of $E$.

\paragraph{Gradient.}
The gradient of the average distortion is
\begin{equation}\label{e-gradient}
\nabla E(X) = (1/\npairs)ACA^TX, \quad
C = \diag(f_1'(d_1)/d_1, \ldots, f_\npairs '(d_\npairs)/d_\npairs),
\end{equation}
when $d_k>0$ for all $k$, \ie, $x_i \neq x_j$ for $(i,j)\in \edges$; here,
$A \in \reals^{n \times p}$ is the incidence matrix \eqref{e-incidence}.
The average distortion is differentiable for all embeddings if $f_k'(0)=0$ for
$k=1, \ldots, p$. In this case we replace $f_k'(d_k)/d_k$ with $f_k''(0)$ in
our expression for $C$ when $d_k=0$. In the sequel we will ignore these points
of nondifferentiability, which do not arise in practice with reasonable choices
of distortion functions.

We observe that if the distortion functions are nondecreasing,
\ie, the objective is attractive, 
the matrix $C$ has nonnegative entries, and the matrix $ACA^T$ is a
Laplacian matrix for a graph $(\items, \edges)$ with edge weights
given by the diagonal of $C$.

\paragraph{Tangents.}
A matrix $V \in \reals^{n \times m}$ is \emph{tangent} to
$\constr$ at $X \in \constr$ if for small $h$,
\[
  \dist(X + hV, \constr) = o(h),
\]
where $\dist$ is the Euclidean distance of its first argument to its second.
We write $\mathcal T_X(\constr)$ to denote the set of tangents to $\constr$ at
$X$. For smooth constraint sets, such as $\centered$, $\anchored$, and
$\standardized$, this set is a subspace; more generally, it is a cone. In the
context of optimization algorithms, a direction $V \in \mathcal T_X(\constr)$
is called a \emph{feasible direction} at $X$, for the constraint $X \in
\constr$. 

\paragraph{Feasible descent directions.}
The first-order Taylor approximation of $E(X + V)$ at $X$ is
\[
\hat E(X + V ;X) = E(X) + \tr (\nabla E(X)^TV),
\]
where the term $\tr (\nabla E(X)^T V)$ is the directional derivative of $E$ at
$X$ in the direction $V$. The directional derivative gives the approximate
change in $E$ for a small step $V$; 
the direction $V$ is a \emph{descent direction }if
$\tr(\nabla E(X)^T V) < 0$, in which case 
$\hat E(X + V; X) < E(X)$ for small $V$ \cite[\S 9.4]{bv2004convex}.
A direction $V$ is called a \emph{feasible descent direction} if it is 
a feasible direction and a descent direction, \ie,
\[
  V \in \mathcal{T}_{X}(\constr), \quad \tr (\nabla E(X)^TV)<0.
\]
This defines a cone of feasible descent directions. It is empty if $X\in
\constr$ is a stationary point for the MDE problem. (We mention that the
negative gradient is evidently a descent direction for $E$, but it is not
necessarily feasible for the constraint.)

\paragraph{Steepest feasible descent direction.} The \emph{steepest feasible
descent direction} (with respect to the Frobenius norm) is defined as the 
(unique) solution of the problem
\begin{equation}\label{e-steepest-desc-dir-def}
\begin{array}{ll}
  \mbox{minimize} & \tr(\nabla E(X)^T V) + \frac{1}{2}\|V\|_F^2 \\
  \mbox{subject to} & V \in \tspace_X(\constr)
\end{array} \end{equation}
(see \cite[chapter~VIII, \S2.3]{urruty1993convex}).
When the solution is $0$, the cone of feasible descent
directions is empty, \ie, $X$ is a stationary point of the MDE problem.
We will denote the steepest feasible descent direction as $-G$.
(We use the symbol $G$ since when $\constr = \reals^{n \times m}$, 
it coincides with the gradient $\nabla E(X)$.)

\paragraph{Projected gradient.}
We can express the (negative) steepest feasible descent direction as
\[
G = \Pi_{\mathcal{T}_X}(\nabla E(X)),
\]
the Euclidean projection of the gradient $\nabla E(X)$ onto the 
(negative) tangent cone of $\constr$
at $X$. That is, $G$ is the solution of
\[
  \begin{array}{ll}
    \mbox{minimize} & \frac{1}{2}\|\nabla E(X) - V\|_F^2 \\
    \mbox{subject to} & V \in -\tspace_{X}(\constr),
  \end{array}
\]
We will refer to $G$ as the \emph{projected gradient} of $E$ at $X \in
\constr$. When the set of tangents to $\constr$ at $X$ is a subspace, the
constraint can be written more simply as $V \in \tspace_{X}(\constr)$.

\paragraph{Stationarity condition.}
The stationarity condition for a constrained MDE problem is simply
\begin{equation}\label{e-stationarity}
  G = 0.
\end{equation}
This condition is equivalent to the well-known requirement that the negative
gradient of the objective lie in the normal cone of the constraint set at $X$
\cite[\S 12.7]{nocedal2006numerical}.

\paragraph{Stopping criterion.}
The stationarity condition \eqref{e-stationarity} leads to a natural stopping
criterion for iterative optimization algorithms for the MDE problem. We can
interpret $G$ as a \emph{residual} for the optimality condition; 
algorithms should seek to make the
Frobenius norm of the residual very small, if not exactly $0$. 
This suggests the termination condition
\begin{equation}\label{e-stopping-criterion}
\|G\|_F \leq \epsilon,
\end{equation}
where $\epsilon$ is a given threshold or tolerance.
The threshold $\epsilon$ can be a
small constant, such as $10^{-5}$, or it may depend on problem parameters 
such as $n$, $m$, or $p$.

We use the stopping criterion \eqref{e-stopping-criterion} for the algorithm we
present in chapter~\ref{c-alg}. Of course, a stationary embedding need not be a
global or even local minimum, but we find that using this stopping criterion
(along with our algorithm) consistently yields good embeddings in practice, as
long as the MDE problem is well-posed.\bigskip

\noindent
In the following sections we derive explicit expressions for the
tangent space $\tspace_{X}(\mathcal X)$, and the projected gradient $G$, at a
point $X$, for our three specific constraint sets $\centered$, $\anchored$, and
$\standardized$. We also derive expressions for the optimal dual variables,
though we will not use these results in the sequel. We start each section
by recalling the definition of the constraint set (the definitions were given
in \S\ref{s-constraints}).

\section{Centered MDE problems}\label{s-unconstr-opt}
The set of centered embeddings is
\[
\mathcal C =  \{X \mid X^T\ones = 0\}.
\]
The tangent space of $\mathcal C$ is the set of centered directions,
\[
  \tspace_{X}(\mathcal C) = \{ V \mid V^T \ones = 0\}.
\]
From the expression \eqref{e-gradient} for the gradient, it can be seen that
$\nabla E(X) \in \tspace_{X}(\mathcal C)$ for $X \in \mathcal C$. Therefore,
the projection onto the tangent space is the identity mapping, \ie,
\begin{equation}\label{e-proj-tspace-centered}
  \pitspace (\nabla E(X)) = \nabla E(X),
\end{equation}
yielding the familiar stationarity condition
\begin{equation}\label{eq-unc-stationary}
G = \nabla E(X) =0.
\end{equation}

In the mechanical interpretation, the
rows of the gradient $\nabla E(X) = 1/p(ACA^T X)$ are the net force on the
points due to the springs, and
the stationarity condition \eqref{eq-unc-stationary} says that the
points are at equilibrium. To see this, note that the matrix $CA^T X$ gives the
forces between items
incident to the same edge.
The $p$ rows of $CA^T X$ are
\[
  f'_k(d_k)\frac{x_{i(k)} - x_{j(k)}}{d_k}, \quad k=1, \ldots, p
\]
\ie, $x_{i(k)}$ experiences a force of magnitude $|f'_k(d_k)|$ in the
direction $\sign(f'_k(d_k))(x_{i(k)} - x_{j(k)})/d_k$, due to its connection to
$x_{j(k)}$.

\section{Anchored MDE problems}\label{s-anchor-opt}
The set of anchored embeddings is
\[
  \anchored = \{X \mid x_i = x_i^\text{given}, ~ i \in \anchors \},
\]
where $\anchors \subseteq \items$ is the index set of anchored vertices.
The tangent space of an anchor constraint set $\anchored$ at a
point $X \in \anchored$ is the set of directions that preserve the values of
the anchors, \ie,
\[
  \tspace_{X}(\anchored) = \{V \mid v_i = 0, ~ i \in \anchors\},
\]
where $v_1^T, v_2^T, \ldots, v_n^T$ are the rows of $V$. The projection
onto the tangent space zeros out the rows corresponding to the
anchors:
\begin{equation}\label{e-proj-tspace-anchored}
  \pitspace(\nabla E(X)) = P \nabla E(X), \quad
  P = \begin{bmatrix} p_1^T \\ p_2^T \\ \vdots \\ p_n^T \end{bmatrix}, \quad
  p_i = \begin{cases}
    e_i & i \not\in \anchors \\
    0 & \text{otherwise,} 
  \end{cases}
\end{equation}
where $P \in \reals^{n \times n}$ is a masking matrix and $e_i \in \reals^{n}$
is the $i$th standard basis vector. Therefore, the stationarity condition
\begin{equation}\label{eq-anchor-stationary}
  G = P\nabla E(X) = 0
\end{equation}
says that the rows of the gradient corresponding to the free vertices must be
zero.

\subsection{Dual variables} 
The stationarity condition \eqref{eq-anchor-stationary} can also be seen via
Lagrange multipliers. A simple calculation shows that for a stationary
embedding $X \in \anchored$, the optimal dual variables $\lambda_i \in
\reals^{m}$ (associated with the constraints $x_i = x_i^\text{given}$) must
satisfy
\[
  \lambda_i = - \nabla E(X)_i.
\]
In the mechanical interpretation of an MDE problem, the dual variables
$\lambda_i$ are the forces on the anchored vertices due to the constraint $X
\in \anchored$.

\section{Standardized MDE problems}\label{s-std-opt}
The set of standardized embeddings is
\[
\stdemb = \{X \mid (1/n)X^TX = I, ~ X^T\ones=0\}.
\]
The tangent space to $\stdemb$ at $X \in \stdemb$ is the subspace
\[
  \tspace_{X}(\stdemb) = \{ V \mid V^T \ones  = 0, ~ X^T V + V^T X = 0 \},
\]
as can be seen by differentiating the constraint $(1/n)X^TX = I$.
The projected gradient $G$ of $E$ at $X \in \stdemb$ is therefore the solution
to the linearly constrained least squares problem
\begin{equation}\label{e-proj-grad-std-prob}
  \begin{array}{ll}
    \mbox{minimize} & \frac{1}{2} \|\nabla E(X) - V\|_F^2 \\
    \mbox{subject to} & V^T \ones =0, \quad V^TX+X^TV = 0,
  \end{array}
\end{equation}
with variable $V \in \reals^{n \times m}$.
This problem has the simple solution
\begin{equation}\label{e-proj-tspace-standardized}
  \pitspace(\nabla E(X)) = \nabla E(X) - (1/n)X \nabla E(X)^TX.
\end{equation}
This means that $X \in \stdemb$ is stationary for a standardized MDE problem if
\begin{equation}\label{e-std-stationary}
G = \nabla E(X) - (1/n)X \nabla E(X)^TX = 0.
\end{equation}

The solution~\eqref{e-proj-tspace-standardized} is readily derived with Lagrange multipliers. The
Lagrangian of~\eqref{e-proj-grad-std-prob} is
\[
  \mathcal{L}(V, \Lambda) = \frac{1}{2}(\|\nabla E(X)\|_F^2 -
  2\tr(\nabla E(X)^T V) + \|V\|_F^2) + \tr(\Lambda^T(V^TX + X^TV)),
\]
where $\Lambda \in \reals^{m \times m}$ is the multiplier (we do not need a
multiplier for the centering constraint $X^T \ones = 0$, since it can be
imposed without loss of generality). The optimal $V$ and $\Lambda$ must satisfy
\[
\nabla_V \mathcal{L}(V, \Lambda) = -\nabla E(X) + V + X(\Lambda^T + \Lambda) = 0.
\]
Using $V^TX + X^TV = 0$ and $(1/n) X^TX = I$, we obtain $\Lambda^T + \Lambda =
(1/n)\nabla E(X)^T X$, from which~\eqref{e-proj-tspace-standardized} follows. From the
expression of the gradient \eqref{e-gradient} and the fact that $X^T \ones =
0$, it is easy to check that $G$ satisfies $G^T \ones = 0$.

\subsection{Dual variables} We can directly derive the stationarity
condition~\eqref{e-std-stationary} using Lagrange multipliers, without
appealing to the projected gradient. The Lagrangian of the MDE problem with
standardization constraint is
\[
  \mathcal{L}(X, \Lambda) = E(X)+ (1/n)\tr(\Lambda^T(I - X^TX)),
\]
where $\Lambda = \Lambda^T \in \reals^{m \times m}$
is the Lagrange multiplier associated with the constraint $I - (1/n)X^TX=0$
(again, we omit a multiplier for the centering constraint).
The optimal multiplier $\Lambda$ must satisfy
\begin{equation}\label{e-std-dl}
  \nabla_X \mathcal{L}(X, \Lambda) =\nabla E(X) - (2/n)X\Lambda = 0.
\end{equation}
Multiplying by $X^T$ on the left and using $(1/n)X^TX = I$ yields
\begin{equation}\label{e-std-dl-var}
  \Lambda = \frac{1}{2}\nabla E(X)^T X = \frac{1}{2p}X^TACA^TX.
\end{equation}
Substituting this expression for $\Lambda$
into \eqref{e-std-dl}, we obtain the stationarity condition
\eqref{e-std-stationary}.

We can interpret the rows of $-(2/n)X\Lambda$ as the forces 
put on the points by the constraint $X\in \stdemb$.
The stationarity condition \eqref{e-std-stationary} is again
that the total force on each point is zero, including both the 
spring forces from other points, and the force due to the 
constraint $X\in \stdemb$.

Finally, we observe that the multiplier $\Lambda$ satisfies the interesting property
\[
  \tr(\Lambda) = \frac{1}{p} \tr(X^T ACA^T X) /2
               = \frac{1}{p} \sum_{k=1}^{p} f'_k(d_k)d_k/2.
\]
In other words, $\tr(\Lambda)$ equals the average distortion of
an MDE problem with distortion functions $\tilde f_k(d_k) = f'_k(d_k)d_k/2$ (and
the same items and pairs as the original problem). When
the distortion functions are weighted quadratics, this MDE problem coincides
with the original one, \ie, $\tr(\Lambda) = E(X)$.

\paragraph{Quadratic problems.} For quadratic MDE problems, which were
studied in chapter~\ref{c-quad-emb}, the matrix $(1/2)ACA^T$ is the matrix $L$, 
so from \eqref{e-std-dl-var} the optimal dual variable satisfies
\[
  LX = (p/n)X\Lambda.
\]
Substituting the optimal value of $X$ into this equation, we see that the
optimal dual variable is diagonal, and its entries are eigenvalues of $L$. In
particular,
\[
  \Lambda = (n/p)\diag(\lambda_1, \ldots, \lambda_m)
\]
for $j > m$, or
\[
  \Lambda = (n/p)\diag(\lambda_1, \ldots, \lambda_{j-1}, \lambda_{j+1}, \ldots, \lambda_{m+1}).
\]
for $j \leq m$. Notice that in both cases, $\tr(\Lambda) = E(X)$.

\chapter{Algorithms}\label{c-alg}
MDE problems are generally intractable: there are no efficient algorithms for
finding exact solutions, except in special cases such as when the distortion
functions are quadratic, as described in chapter~\ref{c-quad-emb}, or when
the distortion functions are convex and nondecreasing 
and the embedding is anchored.  
In this chapter we describe two heuristic algorithms for approximately solving MDE
problems. The first algorithm is a projected quasi-Newton algorithm, suitable
for problems that fit in the memory of a single machine. The second algorithm
is a stochastic method that builds on the first to scale to much larger
problems. While we cannot guarantee that these algorithms find
minimum-distortion embeddings, in practice we observe that they reliably find
embeddings that are good enough for various applications.

We make a few assumptions on the MDE problem being solved. We assume that the
average distortion is differentiable, which we have seen occurs when the
embedding vectors are distinct. In practice, this is almost always the case.
We have also found that the methods described in this chapter work well
when this assumption does not hold, even in the case when 
the distortion functions themselves are nondifferentiable.

Our access to $E$ is mediated by a first-order oracle: we assume that we can
efficiently evaluate the average distortion $E(X)$ and its gradient $\nabla
E(X)$ at any $X \in \constr$. Additionally, we also assume that projections
onto the tangent cone and constraint set exist, and that the associated
projection operators $\pitspace(Z)$ and $\Pi_\constr(Z)$ can be efficiently
evaluated for $Z \in \reals^{n \times m}$. Our algorithms can be applied to any
MDE problem for which these four operations, evaluating $E$, $\nabla E$,
$\pitspace$, and $\Pi_\constr$, can be efficiently carried out. We make no
other assumptions.

We describe the projected quasi-Newton algorithm in \S\ref{s-projected-lbfgs}.
The algorithm is an iterative feasible descent method, meaning that it produces
a sequence of iterates $X_k \in \mathcal X$, $k=0,1, \ldots $, with $E(X_k)$
decreasing. In \S\ref{s-batch-method} we describe a stochastic proximal method
suitable for very large MDE problems, with billions of items and edges. This
method uses the projected quasi-Newton algorithm to approximately solve a
sequence of smaller, regularized MDE subproblems, constructed by sampling a
subset of the edges. Unlike traditional stochastic methods, we use very large
batch sizes, large enough to fill the memory of the hardware used to solve the
subproblems. We will study the performance of our algorithms when applied to a
variety of MDE problems, using a custom software implementation, in the
following chapter.

\section{A projected quasi-Newton algorithm}\label{s-projected-lbfgs}
In this section we describe a projected quasi-Newton method for approximately
solving the MDE problem
\[
  \begin{array}{ll}
    \mbox{minimize} & E(X) \\
    \mbox{subject to} & X \in \constr,
  \end{array}
\]
with matrix variable $X \in \reals^{n \times m}$.

Each
iteration involves computing the gradient $\nabla E(X_k)$
and its projection $G_k = \pitspace(\nabla
E(X_k))$ onto the tangent space (or cone) of the constraint set at the current
iterate. A small history of the projected gradients is stored, which is used to
generate a quasi-Newton search direction $V_k$.  After choosing a suitable step
length $t_k$ via a line search, the algorithm steps along the search direction,
projecting $X_k + t_kV_k$ onto $\constr$ to obtain the next iterate. We use the
stopping criterion \eqref{e-stopping-criterion}, terminating the algorithm when
the residual norm $\|G_k\|_F$ is sufficiently small (or when the iteration number
$k$ exceeds a maximum iteration limit). We emphasize that this algorithm can be
applied to any MDE problem for which we can efficiently evaluate the average
distortion, its gradient, and the operators $\pitspace$ and $\Pi_{\constr}$.

Below, we explain the various steps of our algorithm. Recall that we have
already described the computation of the gradient $\nabla E(X)$, in
\eqref{e-gradient}; likewise, the projected
gradients for these constraint sets are given in
\eqref{e-proj-tspace-centered}, \eqref{e-proj-tspace-anchored}, and
\eqref{e-proj-tspace-standardized}. We first describe the computation of the
quasi-Newton search direction $V_k$ and step length $t_k$ in each iteration.
Next we describe the projections onto the constraint sets $\centered$,
$\anchored$, and $\standardized$. In all three cases the overhead of computing
the projections is negligible compared to the cost of computing the average
distortion and its gradient. Finally we summarize the algorithm
and discuss its convergence.

\subsection{Search direction}\label{s-search-direction}
In each iteration, the quasi-Newton search direction $V_k$ is computed using a
procedure that is very similar to the one used in the limited-memory
Broyden-Fletcher-Goldfarb-Shanno (L-BFGS) algorithm \cite{nocedal1980updating}, except we use projected gradients instead of gradients. For completeness, we
briefly describe the algorithm here; a more detailed discussion (in terms of
gradients) can be found in \cite[\S7.2]{nocedal2006numerical}. In what follows,
for a matrix $X \in \reals^{n \times m}$, we write $\vect{X}$ to denote its
representation as a vector in $\reals^{nm}$ containing the entries of $X$ in
some fixed order.

In iteration $k$, the search direction $V_k$ is computed as
\begin{equation}\label{e-lbfgs-search}
\vect V_{k} = -B_{k}^{-1} \vect G_k,
\end{equation}
where $B_{k} \in \symm^{nm}_{++}$ is a positive definite matrix.
In particular, $B_{k}^{-1}$ is given by the recursion
\begin{equation}\label{e-Binv}
B_{j+1}^{-1} =
	\left(I - \frac{s_jy_j^T}{y_j^Ts}\right)
	B_{j}^{-1}
	\left(I - \frac{y_js_j^T}{y_j^Ts_j}\right) +
	\frac{s_{j}s_{j}^T}{y_{j}^Ts_{j}}, \quad
j=k-1, \ldots, k-M,
\end{equation}
using $B_{k-M} = \gamma_{k} I$;
the positive integer $M$ is the \emph{memory} size, and
\[
  s_j = \vect(X_{j+1} - X_{j}), \qquad y_j = \vect(G_{j+1} - G_j), \qquad \gamma_k = \frac{y_{k-1}^Ts_{k-1}}{y_{k-1}^Ty_{k-1}}.
\] 
When $M \geq k$ and $\constr = \reals^{n \times m}$, the matrix $B_{k}$ is
the BFGS approximation to the Hessian of $E$ (viewed as a function from
$\reals^{nm}$ to $\reals$), which satisfies the secant condition $B_{k} s_{k-1} =
y_{k-1}$ \cite{broyden1970convergence, fletcher1970new, goldfarb1970family,
shanno1970conditioning}. More generally, $B_k$ approximates the curvature
of the restriction of $E$ to $\constr$. In practice the matrices $B_{k}^{-1}$
are not formed explicitly, since the matrix-vector product $B_{k}^{-1} \vect
G_k$ can be computed in $O(nmM) + O(M^3)$ time
\cite[\S7.2]{nocedal2006numerical}.
(The memory $M$ is typically small, on the order of $10$, so the second term is
usually negligible.)

After computing the search direction, we
form the intermediate iterate
\[
  X_{k+1/2} = X_k + t_k V_k,
\]
where $t_k > 0$ is a step length. The next iterate is obtained by projecting
the intermediate iterate onto $\constr$, \ie,
\[
  X_{k+1} = \Pi_{\constr} (X_{k+1/2}).
\]
We mention for future reference that the intermediate iterates satisfy
\begin{equation}\label{e-nesterov}
  X_{k + 1/2} \in X_k + \Span \{G_k, G_{k-1}, \ldots, G_{k-m}, s_{k-1}, \ldots, s_{k-m}\}.
\end{equation}
(see \cite[\S A2]{jensen2017approach}). Note in particular that if $G_1, G_2,
\ldots, G_k$ and $X_1, X_2, \ldots, X_k$ are all centered, then $X_{k+1/2}$ is
also centered (\ie, $X_{k+1/2}^T \ones = 0$).

We choose the step length $t_k$ via a line search, to
satisfy the modified strong Wolfe conditions
\begin{equation}\label{e-wolfe}
\begin{array}{l}
  E(\Pi_{\constr}(X_{k+1/2})) \leq E(X_k) + c_1 t_k \tr(G_k^T V_k), \\[1ex]
|\tr(G_{k+1}^T V_k))| \leq c_2 |\tr(G_k^T V_k)|,
\end{array}
\end{equation}
where $0 < c_1 < c_2 < 1$ are constants (typical values are $c_1=10^{-4}$ and $c_2=0.9$)
\cite[chapter 3]{nocedal2006numerical}.
Since $B_k^{-1}$ is positive definite, 
$\tr(G_k^T V_k) = \tr (G_k^T B_k^{-1} G_k) < 0$; 
this guarantees the descent condition, \ie, $E(X_{k+1})< E(X_k)$. There are
many possible ways to find a direction satisfying these conditions; for
example, the line search described in \cite[\S 3.5]{nocedal2006numerical} may
be used. Typically, the step length $t_k = 1$ is tried first. After a few
iterations, $t_k = 1$ is usually accepted, which makes the line search
inexpensive in practice.

Finally, we note that unlike $G_k$, the search direction $V_k$ is not
necessarily a tangent to $\constr$ at $X_k$. While algorithms similar to ours
require the search directions to be tangents (\eg, \cite{huang2017intrinsic,
hosseini2018line}), we find that this is unnecessary in practice.

\subsection{Projection onto constraints}\label{s-projection}
The next iterate $X_{k+1}$ is obtained by projecting $X_{k+1/2} = X_k + t_kV_k$
onto the constraint set $\constr$, \ie,
\[
  X_{k+1} = \Pi_\constr(X_{k+1/2}).
\]
We will see below that the projections onto $\centered$ and $\anchored$ are
unnecessary, since $X_{k+1/2} \in \constr$ when $\constr = \centered$ or
$\constr = \anchored$. We will then describe how to efficiently compute the
projection $\Pi_\standardized$.

\paragraph{Centered embeddings.}
As long as the first iterate $X_0$ satisfies $X_0^T \ones = 0$,
then the intermediate iterates $X_{k+1/2}$ also satisfy $X_{k+1/2}^T \ones =
0$. To see this, first note that if $X^T \ones = 0$, then $\nabla E(X)^T \ones
= 0$; this can be seen from the expression~\eqref{e-gradient} for the gradient.
Since $G = \nabla E(X)$ for unconstrained embeddings, if $X_k$ is
centered, from \eqref{e-nesterov} it follows that $X_{k+1/2} \in
\centered$. For this reason, for the constraint $X \in \centered$, we can
simply take
\[
  X_{k+1} = X_{k+1/2}.
\]

\paragraph{Anchored embeddings.}
As was the case for centered embeddings, provided that $X_{k} \in \anchored$,
$X_{k+1/2} \in \anchored$ as well, so the projection is unnecessary. This is
immediate from the expression \eqref{e-proj-tspace-anchored} for the projected
gradient $G$, which is simply the gradient with rows corresponding to the
anchors replaced with zero.

\paragraph{Standardized embeddings.}
A projection of $Z \in \reals^{n \times \embdim}$ onto 
$\stdemb$ is any  $X\in \stdemb$ that minimizes
$\|X-Z\|_F^2$. (Since $\stdemb$ is nonconvex, the projection can be
non-unique; a simple example is $Z=0$, where any $X\in \stdemb$ is a projection.)
In the sequel, we will only need to compute a projection
for $Z$ satisfying $Z^T \ones = 0$ and $\rank Z = m$.
For such $Z$, the projection onto $\stdemb$ is unique and can be computed from
the singular value decomposition (SVD), as follows.

Let $Z=U\Sigma V^T$ denote the SVD of
$Z$,
with $U\in \reals^{n\times m}$,
$U^TU=I$, $\Sigma \in \reals^{m \times m}$ diagonal with nonnegative
entries, and $V\in \reals^{m \times m}$, $V^TV=I$.
The projection is given by
\begin{equation}
  \Pi_{\standardized}(Z) = \sqrt n UV^T.
\label{eq-proj}
\end{equation}
To see this, we first observe that the matrix $\Pi_{\stdemb} (Z)$ is the
projection of $Z$ onto the set $\{X \in \reals^{n \times m} \mid (1/n)X^TX =
I\}$; this fact is well known (see, \eg, \cite{fan1955some},
\cite[\S4]{higham1989matrix} or \cite[\S3]{ absil2012projection}), and it is
related to the orthogonal Procrustes problem \cite{schonemann1966generalized}.
To show that $\Pi_{\stdemb}(Z)$ is in fact the projection onto $\stdemb$, it
suffices to show that $\Pi_{\stdemb}(Z)^T \ones = 0$. From $Z^T \ones = 0$, it
follows that the vector $\ones$ is orthogonal to the range of $Z$.
Because $\rank Z = m$, the $m$ columns of $U$ form
a basis for the range of $Z$. This means
$U^T \ones = 0$ and in particular $\Pi_{\stdemb}(Z)^T \ones = \sqrt n V U^T
\ones = 0$. It is worth noting that the projection \eqref{eq-proj} is
inexpensive to compute, costing only $O(nm^2)$ time; in most applications, $m
\ll n$.

We now explain why just considering centered and full rank $Z$ is sufficient.
Note from \eqref{e-proj-tspace-standardized} that $G_k^T \ones = 0$, for  $k=1,
2, \ldots$; therefore, from \eqref{e-nesterov}, we see that $X_{k+1/2}$ is
centered as long as the initial iterate is centered. Since the rank
of a linear transformation is preserved under small perturbations and $\rank X
= m$ for any $X \in \stdemb$, the intermediate iterate $X_{k+1/2}$ will be full
rank provided that the step size $t_k$ is not too large.

From \eqref{eq-proj}, it follows that Frobenius distance of $Z$ to $\stdemb$ is
given by
\[
\dist(Z,\stdemb) = \| Z-\Pi(Z)\|_F = \left(
\sum_{i=1}^m (\sigma_i-\sqrt n)^2 \right)^{1/2},
\]
where $\sigma_i$ are the diagonal entries of $\Sigma$, \ie, the singular
values of $Z$.

\subsection{Algorithm summary}\label{s-projected-lbfgs-summary}
Algorithm~\ref{alg-projected-lbfgs} below summarizes our projected L-BFGS method.

\begin{algdesc}{\sc Projected L-BFGS method} \label{alg-projected-lbfgs}

\textbf{given} an initial point $X_0 \in \constr$, maximum iterations $K$,
memory size $M$, tolerance $\epsilon>0$.

\textbf{for} $k=0, \ldots, K$
\begin{enumerate}
\item \emph{Projected gradient.} Compute projected gradient
    $G_k = \Pi_{\mathcal{T}_{X_k}}(\nabla E(X_k))$.
\item \emph{Stopping criterion.} Quit if $\|G_k\|_F \leq \epsilon$.
\item \emph{Search direction.} Compute the quasi-Newton search direction $V_k$, as in \eqref{e-lbfgs-search}.
\item \emph{Line search.} Find step length $t_k$ satisfying the modified Wolfe conditions \eqref{e-wolfe}.
\item \emph{Update.} $X_{k+1} := \Pi_{\constr}(X_k + t_k V_k)$.
\end{enumerate}
\end{algdesc}

Our method is nearly parameter-free, except for the memory size $M$, which
parametrizes the computation of $V_k$ in step 3, and possibly some parameters
in the particular line search algorithm used.  We have empirically found that a small
$M$, specifically $M=10$, works well across a wide variety of problems,
and that the line search parameters have very little effect.
Additionally, the algorithm appears to be fairly robust to the choice of the
initial iterate $X_0$, which may be chosen randomly or by a heuristic, such as
initializing at a spectral embedding (projected onto $\constr$).

Practical experience with our algorithm suggests that the use of projected
gradients $G_k$, instead of the gradients $\nabla E(X_k)$, is important.
In our experiments, our algorithm substantially decreases the time (and
iterations) to convergence compared to standard gradient and L-BFGS
implementations equipped with only $\Pi_\constr$ (but not $\pitspace$).
We show an example of this in \S\ref{s-num-ex-quad},
figure~\ref{f-projected-gradient-comparison}.

\paragraph{Convergence.}
The MDE problem is in general nonconvex, so we cannot
guarantee that algorithm~\ref{alg-projected-lbfgs} will converge to a global solution.
Nonetheless, we have found that it reliably finds good embeddings in practice
when applied to the constraint sets $\centered$, $\anchored$, and
$\standardized$. When we have applied the method to quadratic MDE problems, it
has always found a global solution.

There is a large body of work studying optimization with orthogonality
constraints (\eg, over the Stiefel manifold), which are closely related to the
standardization constraint. See, for example, \cite{edelman1998geometry,
manton2002optimization, absil2009optimization, absil2012projection,
jiang2015framework, huang2015broyden, huang2018riemannian, hu2019structured,
chen2020proximal}. In particular, methods similar to
algorithm~\ref{alg-projected-lbfgs} have been shown to converge to stationary
points of nonconvex functions, when certain regularity conditions are satisfied
\cite{ji2007optimization, ring2012optimization, huang2015broyden,
huang2018riemannian}.

\paragraph{Computational complexity.} Each iteration of
algorithm~\ref{alg-projected-lbfgs} takes $O(mp + nmM) + O(M^3)$ work, plus any
additional work required to compute the projected gradient and the projection
onto $\constr$. Computing the average distortion and its gradient contributes
$O(mp)$ work, and computing the search direction contributes $O(nmM) + O(M^3)$.
The overhead due to the projections is typically negligible; indeed, the
overhead is zero for centered and anchored embeddings, and $O(nm^2)$ for
standardized embeddings.

In most applications, the embedding dimension $m$ is much smaller than $n$ and
$p$. Indeed, when an MDE problem is solved to produce a visualization, $m$ is
just $2$ or $3$; when it is used to compute features, it is usually on the
order of $100$. In these common cases, the cost of each iteration is dominated
by the cost of computing the average distortion and its gradient, which scales
linearly in the number of pairs $p$. Suppressing the dependence on $m$ and $M$
(which we can treat as the constant $10$), the cost per iteration is just
$O(p)$. For some special distortion functions, this cost can be made as small
as $O(n\log n)$ or $O(n)$, even when $p = n(n-1)/2$, using fast multipole
methods \cite{greengard1987fast, beatson1997short}.

\section{A stochastic proximal algorithm}\label{s-batch-method}
Building on the projected L-BFGS method described in the previous section, here
we describe a simple scheme to approximately solve MDE problems with a number
of distortion functions $p$ too large to handle all at once.  It can be
considered an instance of the stochastic proximal iteration method
(see, \eg, \cite{ryu2014stochastic, asi2019stochastic}), which uses much larger
batch sizes than are typically used in stochastic optimization methods.
(Stochastic proximal iteration is itself simply a stochastic version of the
classical proximal point algorithm \cite{martinet1970breve,
rockafellar1976monotone}.)
In each iteration, or round, we randomly choose a subset of
$p^\text{batch}$ of the original $p$ distortion functions and (after the first
round) approximately solve a slightly modified MDE problem.

In the first round, we approximately solve the MDE problem corresponding to the
sampled distortion functions using algorithm~\ref{alg-projected-lbfgs}, and we
denote the resulting embedding as $X^0$.  (We used superscripts
here to denote rounds, to distinguish them from $X_k$, which are iterates in
the methods described above.) In round $k$, $k=1,2, \ldots$, we randomly choose
a set of $p^\text{batch}$ distortion functions, which gives us average
distortion $E^k$, and approximately solve the MDE problem with an additional
proximal term added to the objective,
\ie \begin{equation}\label{e-prox-obj}
E^k(X) + \frac{ck}{2} \|X-X^{k-1}\|_F^2,
\end{equation}
where $c$ is a hyper-parameter (we discuss its selection below). We denote the
new embedding as $X^k$. The second term in the objective, referred to as a
proximal term \cite{parikh2014proximal}, encourages the next embedding to be
close to the one found in the previous round. In computing $X^k$ in round $k$,
we can warm-start the iterative methods described above by initializing $X$ as
$X^{k-1}$. This can result in fewer iterations required to compute $X^k$.

When the objective and constraints are convex and satisfy some additional
conditions, this algorithm can be proved to converge to a solution of the
original problem. However, in an MDE problem, the objective is generally not
convex, and for the standardized problem, nor is the constraint.  Nevertheless
we have found this method to reliably find good embeddings for problems with
more edges than can be directly handled in one solve. The algorithm is outlined
below.

\begin{algdesc}{\sc Stochastic method for MDE problems with large $p$.}
\label{alg-serial-batch}

\textbf{given} a batch size $p^\text{batch} < p$, iterations $K$.

\textbf{for} $k=0, \ldots, K$
\begin{enumerate}
\item \emph{Select a batch of edges.} 
Randomly select a batch of $p^\text{batch}$ distortion functions,
which defines average distortion $E^k$.
\item \emph{First iteration.}  If $k=0$, solve the MDE problem with objective $E^k$
to get embedding $X^0$.
\item \emph{Subsequent iterations.}
If $k>0$, solve the MDE problem using the batch, with additional objective term
as in~(\ref{e-prox-obj}), to get $X^{k}$.
\end{enumerate}
\end{algdesc}

\paragraph{Hyper-parameter selection.} The empirical convergence rate of the
algorithm is sensitive to the choice of hyper-parameter $c$. If $c$ is too
big, we place too much trust in the iterate $X^{0}$ obtained by the first
round, and the algorithm can make slow progress. If $c$ is too small, the
algorithm oscillates between minimum-distortion embeddings for each batch,
without converging to a suitable embedding for all $p$ distortion functions.
For this reason, the selection of $c$ is crucial.

The hyper-parameter can be chosen in many ways, including manual
experimentation. Here we describe a specific method for automatically choosing
a value of $c$, based on properties of the MDE problem being solved. We find
this method works well in practice. The method chooses
\[
c = \frac{\tr (\nabla^2 E^0(X^0) )}{\lambda nm}
\]
where $\lambda \in \reals$ is a parameter (we find that the choice $\lambda=10$
works well). This choice of $c$ can be interpreted as using a diagonal
approximation of the Hessian of $E^0$ at $X^0$, \ie,
\[
\nabla^2 E^0(X^0) \approx cI.
\]
In our implementation, instead of computing the exact Hessian $\nabla^2
E^0(X^0)$, which can be intractable depending on the size of $n$ and $m$, we
approximate it using randomized linear algebra and Hessian-vector products. We
specifically use the Hutch$++$ algorithm \cite{meyer2020hutch++}, a recent
method based on the classical Hutchinson estimator
\cite{hutchinson1989stochastic}. This algorithm requires a small number $q$ of
Hessian-vector products, and additionally $O(mq^2)$ time to compute a QR
decomposition of an $m \times q/3$ matrix. Using modern systems for automatic
differentiation, these Hessian-vector products can be computed efficiently,
without storing the
Hessian matrix in memory. For our application, $q=10$ suffices, making the cost
of computing $c$ negligible.

\chapter{Numerical Examples}\label{c-num-ex}
In this chapter we present numerical results for several MDE problems. The
examples are synthetic, meant only to illustrate the performance and scaling of
our algorithms. The numerical results in
this chapter were produced using a custom implementation of our solution
algorithms. We first present the numerical examples; the implementation, which
we have packaged inside a high-level Python library for specifying and solving
MDE problems, is presented in \S\ref{s-impl}.

In the first set of examples we apply the projected quasi-Newton algorithm
(algorithm~\ref{alg-projected-lbfgs}) to quadratic MDE problems. Using a
specialized solver for eigenproblems, which can solve quadratic MDE problems
globally, we verify that our method finds a global solution in each example. We
also compare our algorithm to standard methods for smooth constrained
optimization, which appear to be much less effective. The problems in this
section are of small to medium size, with no more than one million items
and ten million edges.

Next, in \S\ref{s-num-ex-other}, we apply algorithm~\ref{alg-projected-lbfgs} to
small and medium-size (non-quadratic) MDE problems derived from weights, with
constraint sets $\centered$, $\anchored$, and $\standardized$. We will see that
the overhead incurred in handling these constraints is negligible.

In \S\ref{s-stoch-example} we apply the stochastic proximal method
(algorithm~\ref{alg-serial-batch}) to two MDE problems, one small and one
very large. We verify that the method finds nearly optimal embeddings for the
small problem, makes progress on the very large one, and is robust across
different initializations for both.

\paragraph{Experiment setup.}
All examples were solved with the default parameters of our
implementation. In particular, we used a memory size of $10$, and terminated
the projected quasi-Newton algorithm when the residual norm fell below
$10^{-5}$. Experiments were run on a computer with an Intel i7-6700K CPU (which
has 8 MB of cache and four physical cores, eight virtual, clocked at 4 GHz), an
NVIDIA GeForce GTX 1070 GPU (which has 8 GB of memory and 1920 cores at 1.5
GHz), and 64 GB of RAM.

\section{Quadratic MDE problems}\label{s-num-ex-quad}
In this first numerical example we solve quadratic MDE problems with
weights $w_{ij} = 1$ for $(i, j) \in \edges$. (Recall that a quadratic
MDE problem, studied in chapter~\ref{c-quad-emb}, has distortion functions
$f_{ij}(d_{ij}) = w_{ij}d_{ij}^2$, and the constraint $X \in \stdemb$.) The
edges are chosen uniformly at random by
sampling $p = |\edges|$ unique pairs of integers between $1$ and $n(n-1)/2$,
and converting each integer to a pair $(i, j)$ ($1 \leq i < j \leq
n$) via a simple bijection. We solve two problem instances, a medium size
one with dimensions
\[
n=10^5, \quad p=10^6, \quad m=2
\]
and a large one with dimensions
\[
n=10^6, \quad p=10^7, \quad m=2.
\]
The medium size problem is solved on a CPU, and the large one is solved on a GPU.

\begin{figure}
\includegraphics{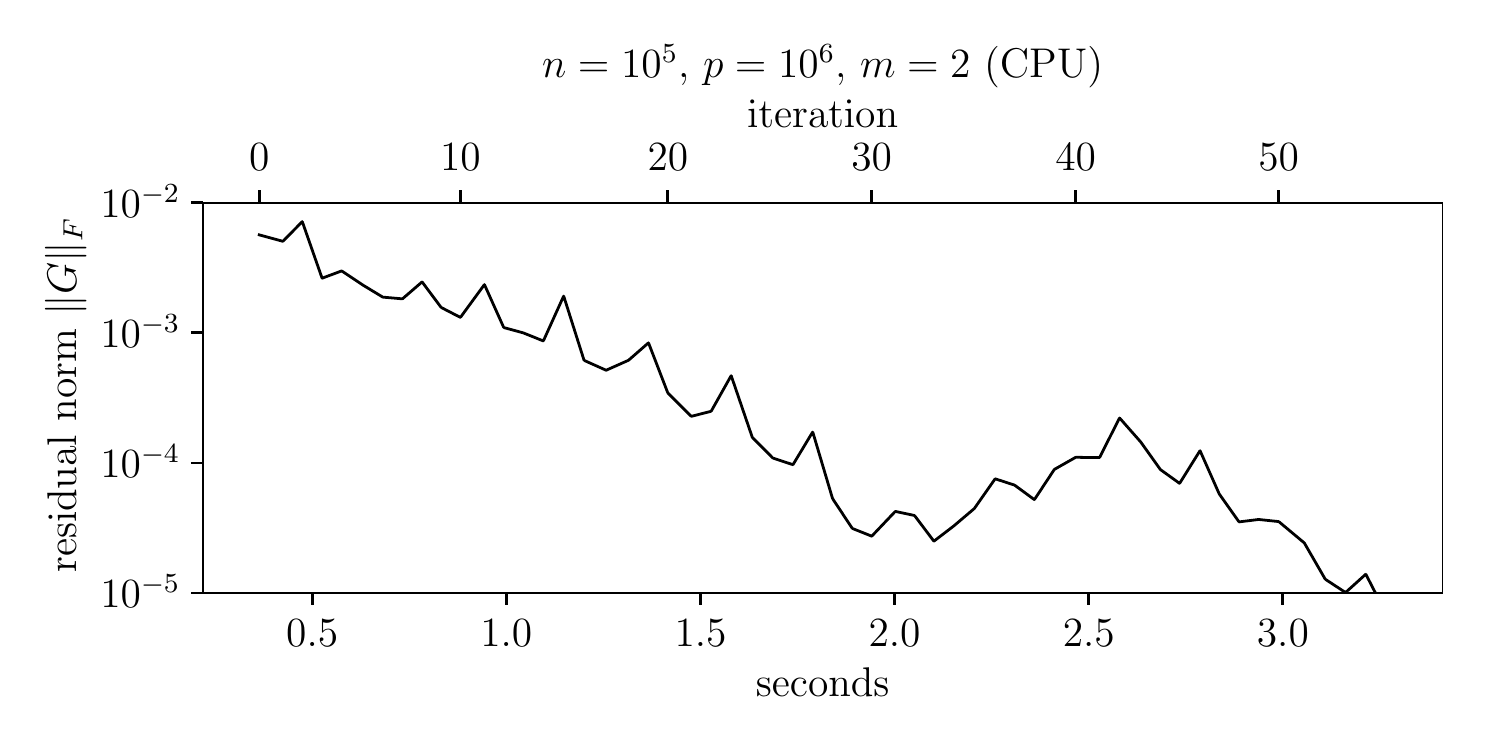}
\includegraphics{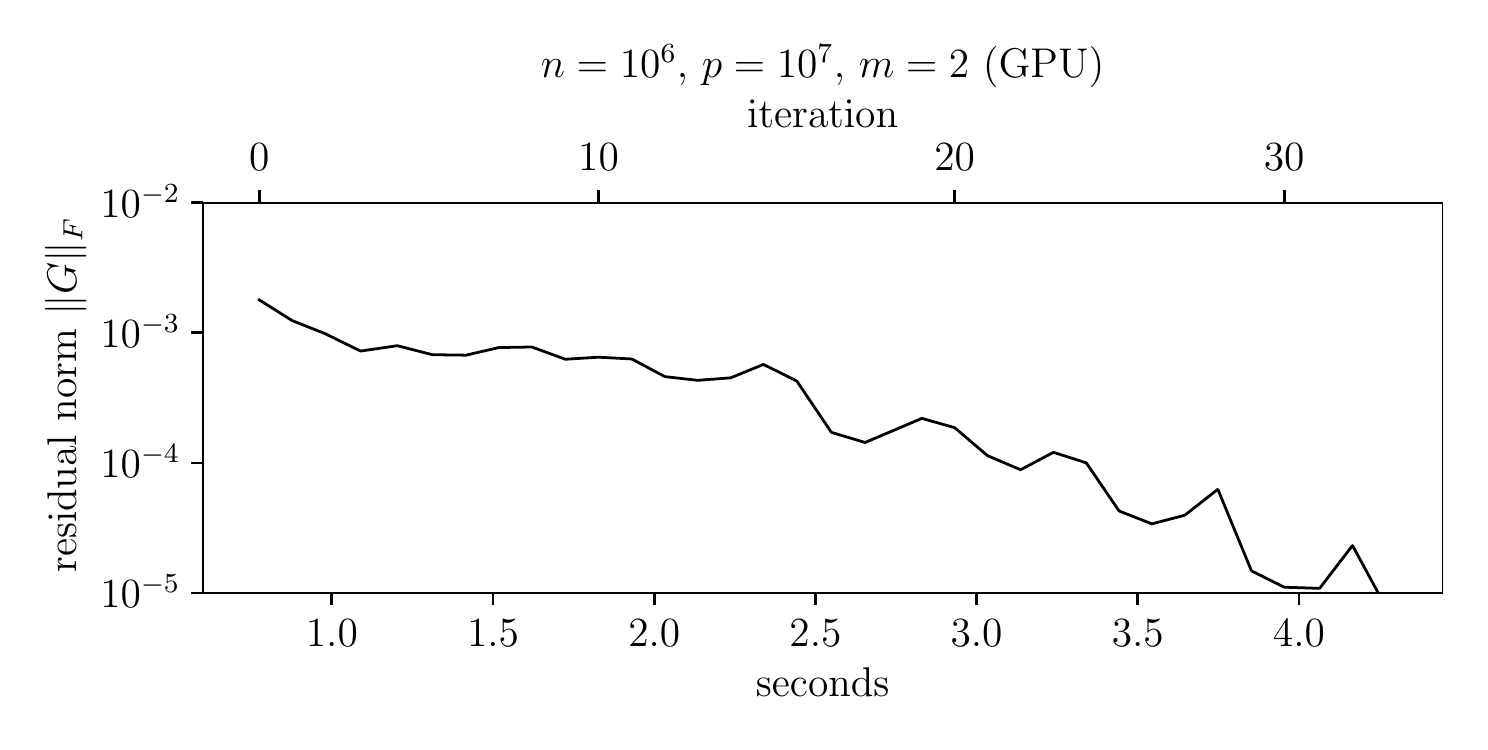}
\caption{Residual norm versus iterations (top horizontal axis) and time 
(bottom horizontal axis) for the projected L-BFGS algorithm.
\emph{Top.} Medium size problem instance.
\emph{Bottom.} Large problem instance.}
\label{f-residual-norm-quad}
\end{figure}

Figure~\ref{f-residual-norm-quad} plots the residual norm $\|G_k\|_F$ of the projected
L-BFGS method (algorithm~\ref{alg-projected-lbfgs}) when applied to these problems,
against both elapsed seconds and iterations.
On a CPU, it takes 55 iterations and 3 seconds to reach a residual norm of
$10^{-5}$ on the first problem, over which the average distortion is decreased
from an initial value of 4.00 to 0.931. On a GPU, it takes 33 iterations and 4
seconds to reach the same tolerance on the second problem, over which the
average distortion is decreased from 4.00 to 0.470.

These quadratic MDE problems can be solved globally as eigenvalue problems,
as described in chapter~\ref{c-quad-emb}.
We solve the two problem instances using LOBPCG, a
specialized algorithm for large-scale eigenproblems \cite{knyazev2001toward}
We specifically used \texttt{scipy.sparse.linalg.lobpcg}, initialized with a
randomly selected standardized matrix, restricted to search in the orthogonal
complement of the ones vector, and limited to a maximum iteration count of 40.
For the medium size problem, LOBPCG obtained a solution with average distortion 
0.932 in 2 seconds; for the large problem, it gives a solution with
average distortion of 0.469 in 17 seconds.
We conclude that our algorithm found global solutions for both problem
instances.  (We have not encountered a quadratic MDE problem for which our
method does not find a global solution.)
We also note that the compute time of our method is comparable to,
or better than, more specialized methods based on eigendecomposition.

\subsection{Comparison to other methods}
\begin{figure}
\includegraphics{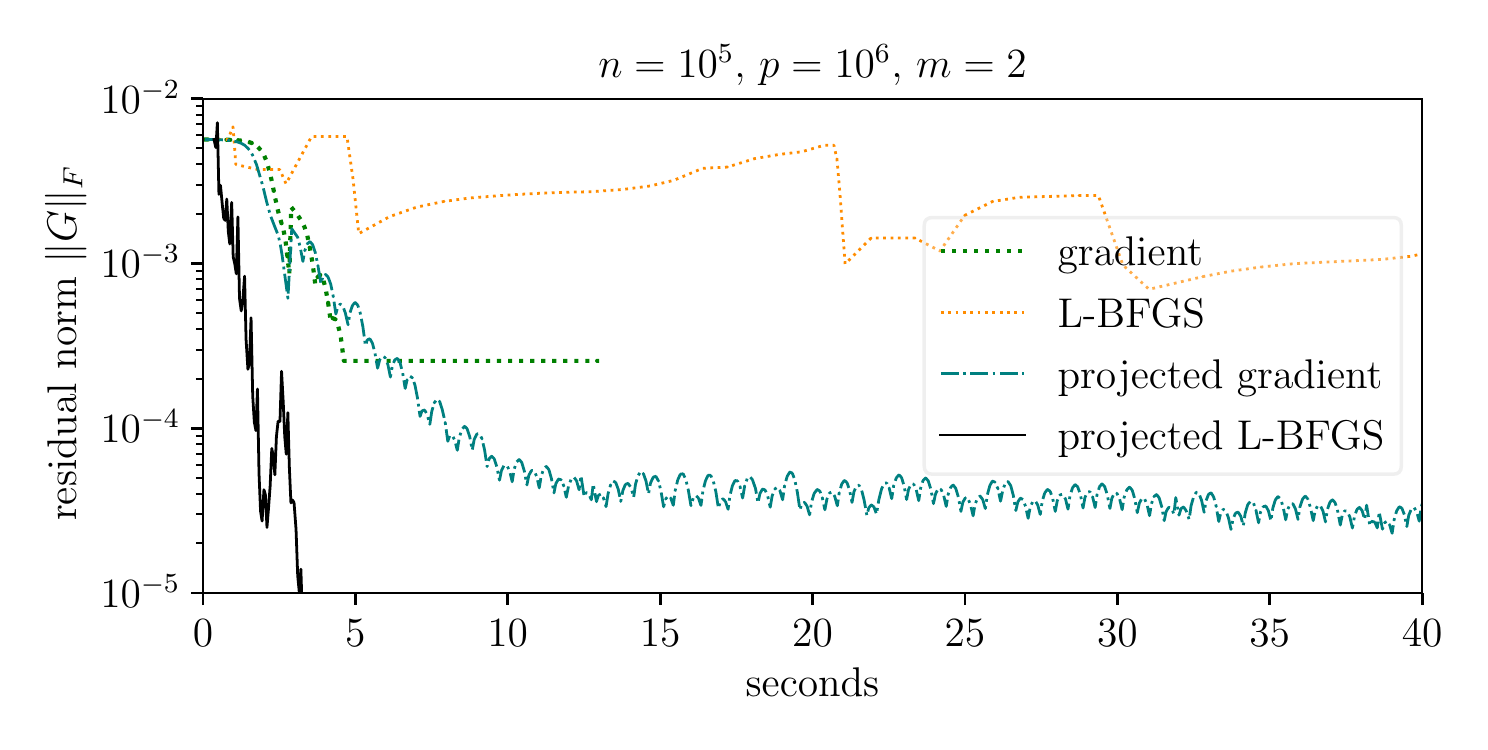}
\caption{Residual norm versus seconds for our algorithm (projected-LBFGS)
and other algorithms, on a medium size quadratic problem instance solved on CPU.}
\label{f-projected-gradient-comparison}
\end{figure}

The projected L-BFGS tends to converge to approximate solutions much faster than
simpler gradient-based methods, such as standard gradient descent, which
performs the update $X_{k+1} = \Pi_{\constr}(X_k - t_k \nabla E(X_k))$, and
projected gradient descent, which performs the update 
$X_{k+1} = \Pi_{\constr}(X_k - t_k G_k)$. 
Similarly, our method is much more effective than a standard
L-BFGS method that only projects iterates onto the constraint set (but does not
project the gradients).

To compare these methods, we solved the medium size quadratic problem instance
on a CPU using each method, logging the residual norm versus elapsed
seconds. (We used a standard backtracking-Armijo linesearch for the gradient
methods.) Figure~\ref{f-projected-gradient-comparison} shows the results. The
gradient method fails to decrease the residual norm below roughly $3 \times
10^{-4}$, getting stuck after 66 iterations. The projected gradient method
decreases the residual norm to just below $10^{-4}$ after about 8 seconds, but
makes very slow progress thereafter. The standard L-BFGS method performs the
worst, taking nearly 30 seconds to reach a residual norm of $10^{-3}$. The
projected L-BFGS method finds a global solution, with $10^{-5}$ residual norm,
in 3 seconds.

\subsection{Scaling}
To get some idea of how our implementation scales, we solved instances with a
variety of dimensions $n$, $m$, and $p$. We terminated the algorithm after
$K=40$ iterations for each solve, noted the solve time, and compared the
average distortion $E(X_K)$ with the average distortion of a global solution
$X^{\star}$
obtained using LOBPCG. The results are listed in table~\ref{t-quad-scaling}. We
can see that the scaling of our method is roughly $O(mp)$, which
agrees with our previous discussion on computational complexity in
\S\ref{s-projected-lbfgs}. Additionally, for sufficiently large problems,
GPU-accelerated solves are over 10 times faster than solving on CPU. After
just $40$ iterations, our method is nearly optimal for each problem instance,
with optimality gaps of $0.4$ percent or less. 

\begin{table}[]
  \centering
  \begin{tabular}[]{lll ll ll}
    \toprule
    \multicolumn{3}{l}{Problem instance dimensions} &
    \multicolumn{2}{l}{Embedding time (s)} &
    \multicolumn{2}{l}{Objective values} \\
    \cmidrule(lr){1-3}
    \cmidrule(lr){4-5}
    \cmidrule(lr){6-7}
    $n$ & $p$ & $m$ & CPU & GPU &  $E(X_K)$ & $E(X^\star)$ \\
    \midrule
    \midrule
    $10^3$ & $10^4$ & $2$ &0.1& 0.4  & 1.432 & 1.432  \\
    $10^3$ & $10^4$ & $10$ &0.2& 0.4 & 7.797 & 7.795  \\
    $10^3$ & $10^4$ & $100$ &0.8& 1.5  &  104.5  & 104.5 \\
    $10^4$ & $10^5$ & $2$ &0.4& 0.4 & 1.007 & 1.007 \\
    $10^4$ & $10^5$ & $10$ &0.7& 0.4 & 6.066 & 6.065 \\
    $10^4$ & $10^5$ & $100$ &20.8& 2.9 &  81.12  & 81.08 \\
    $10^5$ & $10^6$ & $2$ &2.5& 0.6 & 0.935 & 0.931 \\
    $10^5$ & $10^6$ & $10$ &10.5& 1.2 & 5.152 & 5.137 \\
    $10^5$ & $10^6$ & $100$ & 334.7& 15.8 & 63.61  & 63.51 \\
    \bottomrule
  \end{tabular}
  \caption{Embedding time and objective values when solving quadratic
    problem instances via algorithm~\ref{alg-projected-lbfgs}, for $K=40$ iterations.
  }\label{t-quad-scaling}
\end{table}%

\section{Other MDE problems}\label{s-num-ex-other} As a second example, we
solve MDE problems involving both positive and negative weights. The edges are
chosen in the same way as the previous example. We randomly partition $\edges$
into $\esim$ and $\edis$, and choose weights $w_{ij} = +1$ for $(i, j) \in
\esim$ and $w_{ij} = -1$ for $(i, j) \in \edis$.

We use distortion functions based on
penalties, of the form \eqref{eq-penalty}. We choose the log-one-plus penalty
\eqref{eq-log1p-penalty} for the attractive penalty, and the
logarithmic penalty \eqref{eq-log-penalty} for the repulsive penalty.

As before, we solve problems of two sizes, a medium size one with
\[
n=10^5, \quad p=10^6, \quad m=2
\]
and a large one with
\[
n=10^6, \quad p=10^7, \quad m=2.
\]
We solve three medium problems and three large problems, with constraints $X
\in \centered$, $X \in \anchored$, and $X \in \standardized$. For the anchor
constraints, a tenth of the items are made anchors, with associated values
sampled from a standard normal distribution. The medium size problems are
solved on a CPU, and the large problems are solved on a GPU.

\begin{figure}
\includegraphics{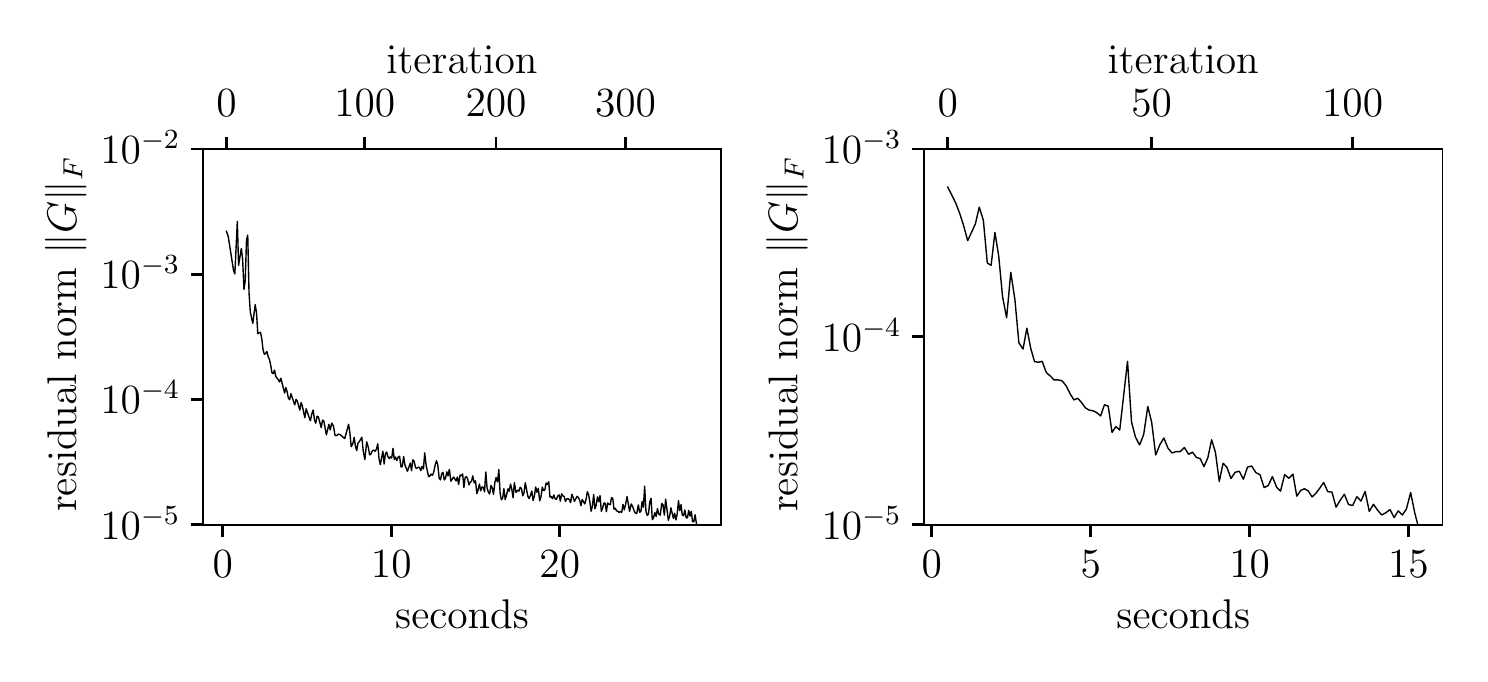}
\includegraphics{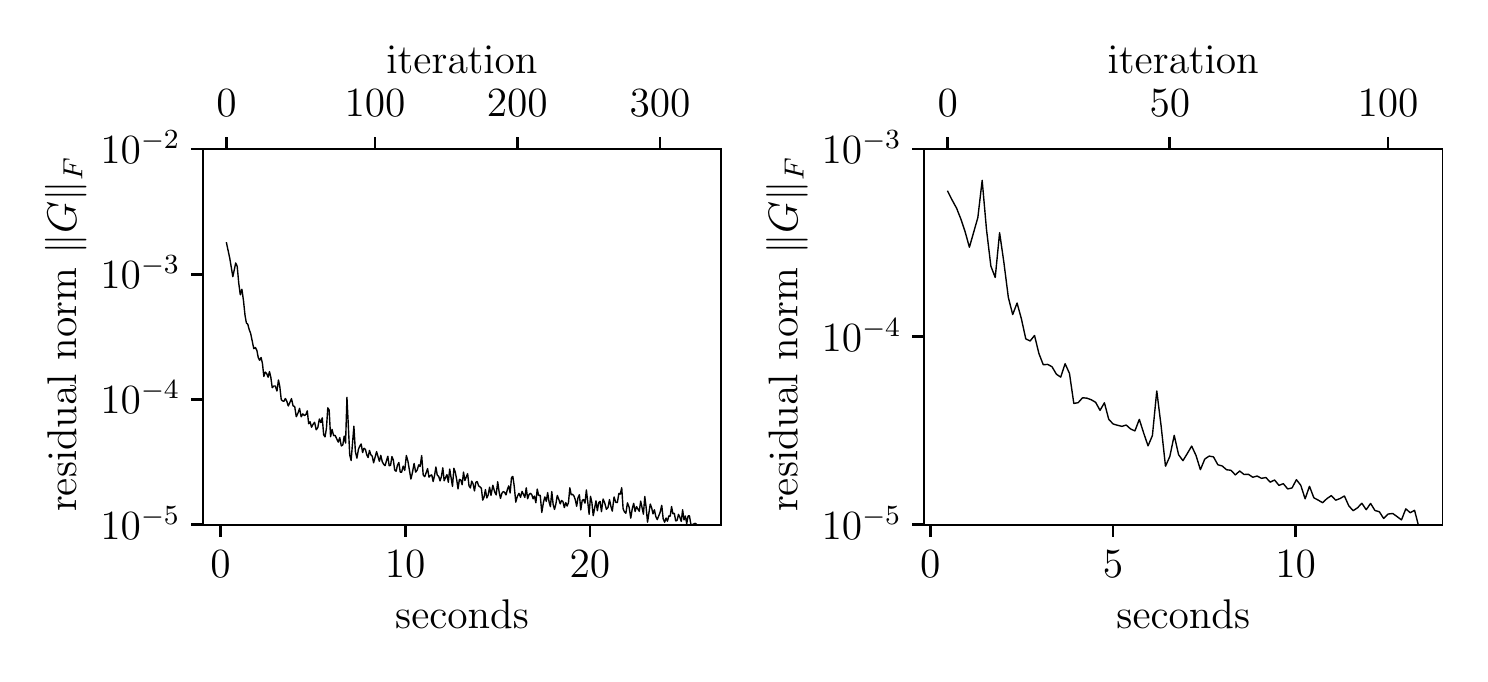}
\includegraphics{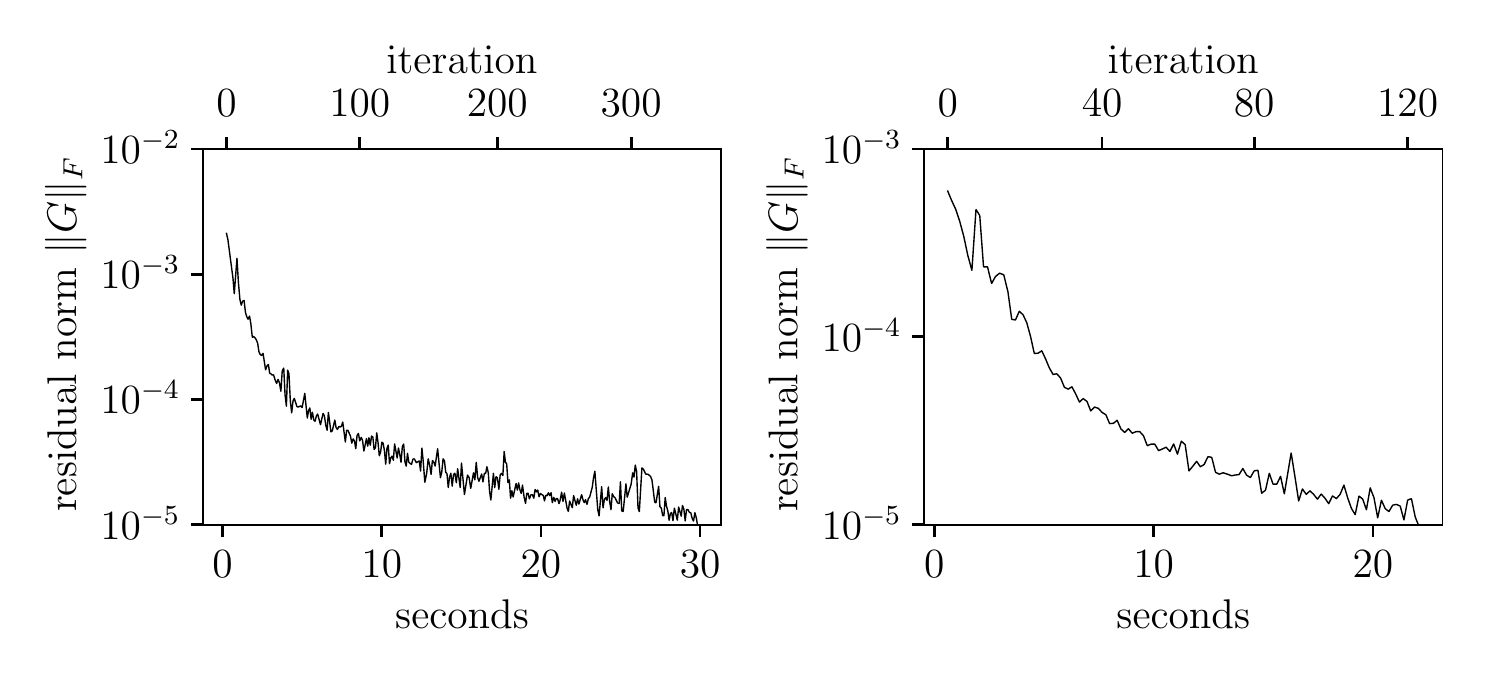}
\caption{Residual norm versus iterations (top horizontal axis) and time 
(bottom horizontal axis) for the projected L-BFGS algorithm, applied
to MDE problems derived from weights.
\emph{Top.} Centered.
\emph{Middle.} Anchored.
\emph{Bottom.} Standardized.
\emph{Left.} Medium problem instances (CPU).
\emph{Right.} Large problem instances (GPU).
}
\label{f-residual-norm-log}
\end{figure}

Figure~\ref{f-residual-norm-log} plots the residual norm of the 
L-BFGS methods when applied to these problems,
against both seconds elapsed and iterations; all problems are solved to
a residual norm of $10^{-5}$. Note the embedding times for centered and
anchored embeddings are comparable, and the per-iteration overhead of computing
the projections associated with the standardization constraint is small.

\clearpage
\section{A very large problem}\label{s-stoch-example}
In this third set of examples we use the stochastic proximal method 
(algorithm~\ref{alg-serial-batch}) to approximately solve some quadratic MDE
problems. To show that our method is at least reasonable, we first apply our
method to a small quadratic MDE problem; we solve this MDE problem
exactly, and compare the quality of the approximate solution, obtained by the
stochastic method, to the global solution. Next we apply the stochastic method
to a problem so large that we cannot compute a global solution, though we can
evaluate its average distortion, which the stochastic method is able to
steadily decrease.

In our experiments, we set the hyper-parameter $c$ as
\[
c \approx \frac{\tr(\nabla^2 E^0(X^0))}{10nm},
\]
using a randomized method to approximate the Hessian trace, as described in
\S\ref{s-batch-method}.

\paragraph{A small problem.}
We apply our stochastic proximal method to a small quadratic MDE
problem of size
\[
n=10^3, \quad p=10^4, \quad m=10,
\]
generated in the same fashion as the other quadratic problems. We approximately
solve the same instance 100 times, using a different random initialization for
each run. For these experiments, we use $p^\text{batch}=p/10$, sample
the batches by cycling through the edges in a fixed order, and run the
method for 300 rounds. This means that we passed through the full set of
edges $30$ times.

\begin{figure}
\includegraphics{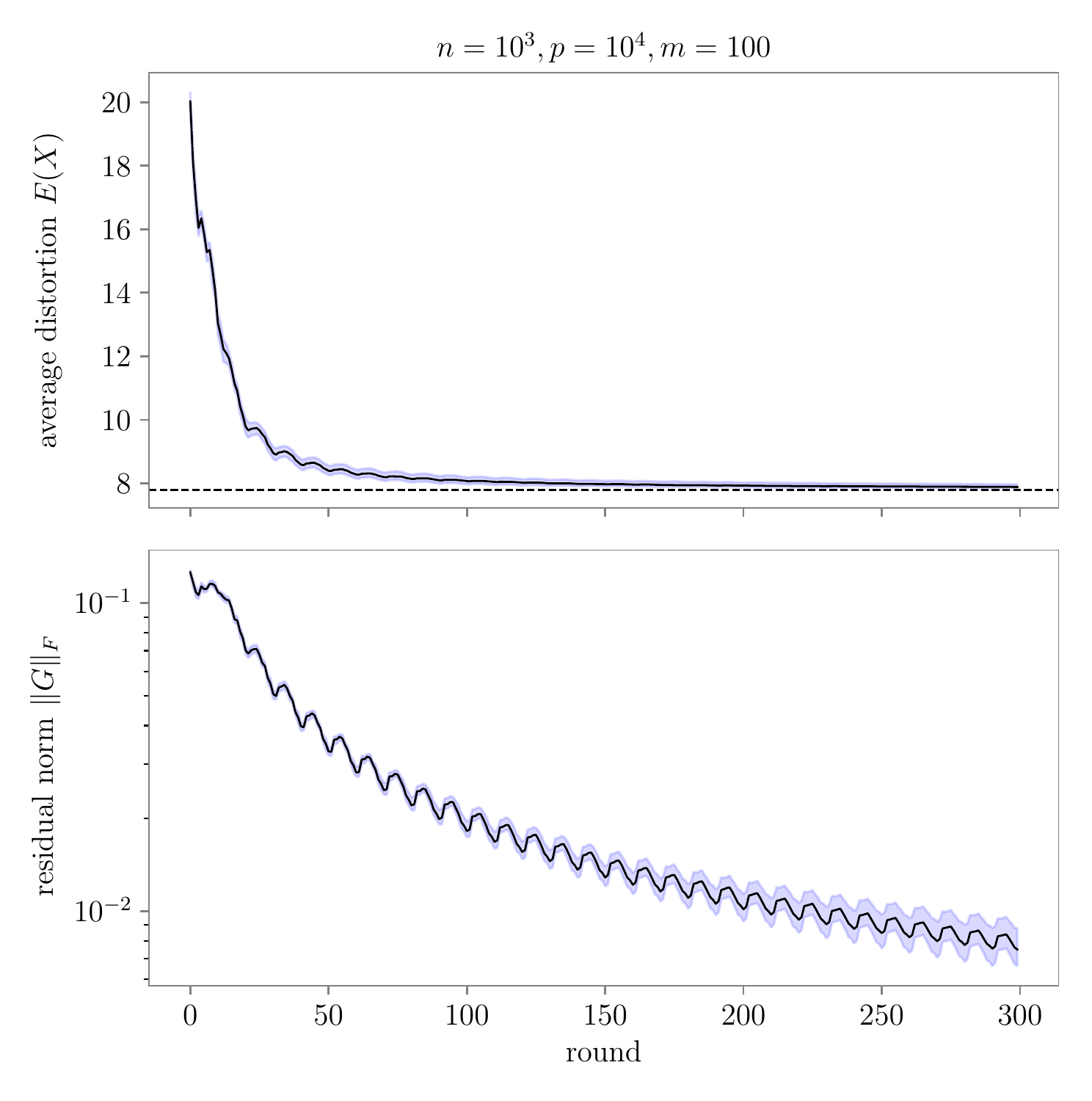}
\caption{The stochastic method (algorithm~\ref{alg-serial-batch}) applied to a
small quadratic MDE problem, with 100 different initializations. \emph{Top.}
Median average distortion versus rounds, optimal value plotted as a dashed
horizontal line. \emph{Bottom.} Median residual norm versus rounds.}
\label{f-spi-small}
\end{figure}

Figure~\ref{f-spi-small} shows the average distortion and
residual norms versus rounds (the solid lines are the medians across all 100
runs, and the cone represents maximum and minimum values). The mean distortion
of the final embedding was 7.89, the standard deviation was 0.03, the max
was 7.97, and the min was 7.84; a global solution has an average
distortion of 7.80.

\clearpage
\paragraph{A very large problem.}
We approximately solve a very large quadratic
MDE problem, with
\[
  n = 10^5, \quad p = 10^7, \quad m=100.
\]
This problem is too large to solve using off-the-shelf specialized solvers for
eigenproblems such as \texttt{scipy.sparse.linalg.lobpcg}. It is also too
large to fit on the GPU we use in our experiments, since storing the $p \times
m$ matrix of differences alone takes roughly $4$ GB of memory (and forming it
can take up to $8$ GB of memory). While this problem does fit in our
workstation's 64 GB of RAM, evaluating the average distortion and its gradient
takes 12 seconds, which makes running the full-batch projected quasi-Newton
algorithm on CPU impractical.

\begin{figure}
\includegraphics{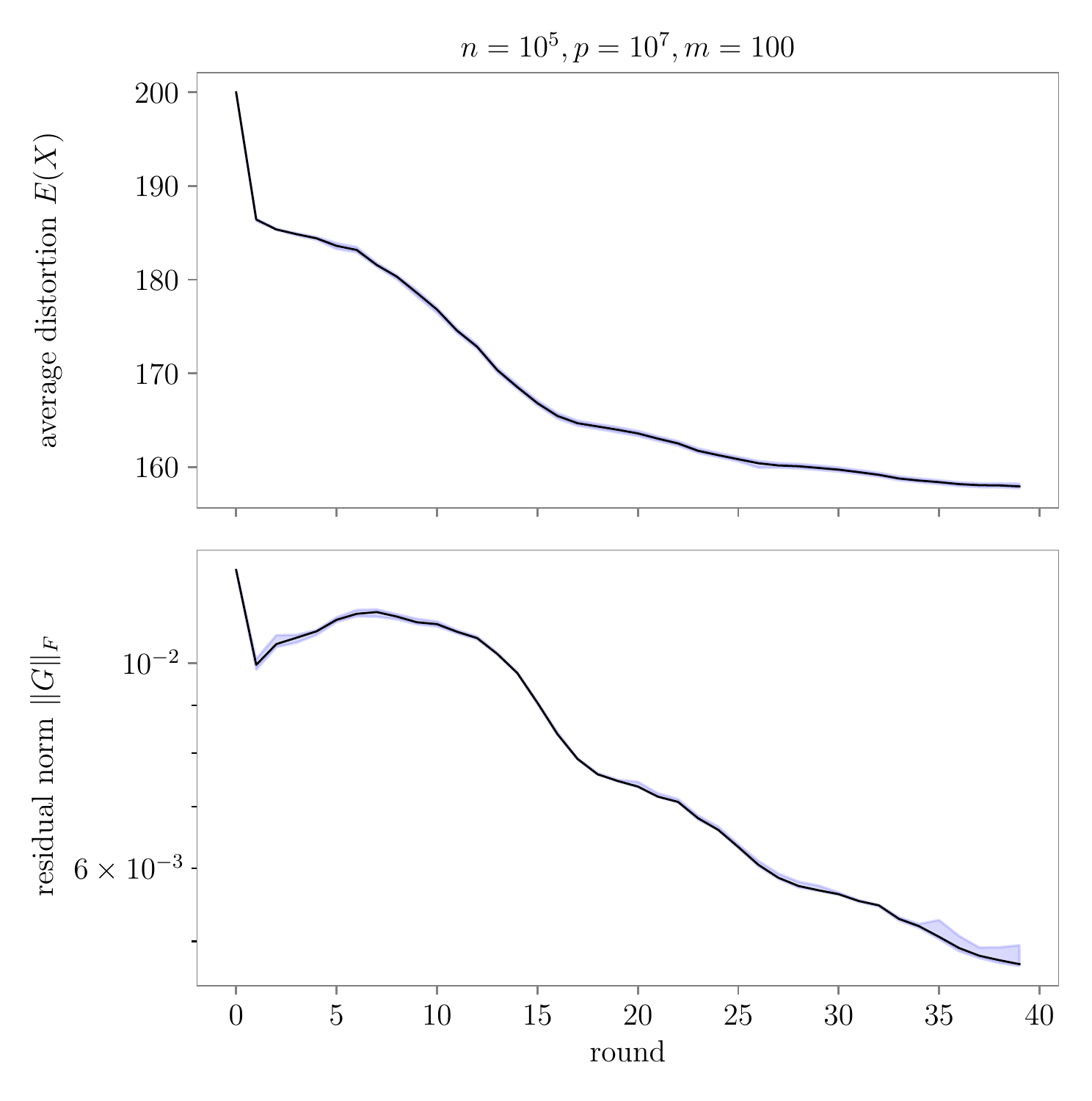}
\caption{The stochastic method applied to a very large quadratic MDE problem,
with 10 different initializations.}\label{f-spi}
\end{figure}

We solved a specific problem instance, generated in the same way as the other
quadratic MDE problems. We used $p^\text{batch} = p/10$, sampled the batches by
cycling through the edges in a fixed order, and ran the stochastic proximal
iteration for $40$ rounds. We ran 10 trials, each initialized at a randomly
selected iterate. Figure~\ref{f-spi} shows the median average distortion and
residual norm at each round, across all $p$ pairs. On our GPU, the 40 rounds
took about 10 minutes to complete. The mean average distortion after 40 rounds
was 157.94, the standard deviation was 0.14, the maximum was 158.21, and the min
was 157.71.

\clearpage
\section{Implementation}\label{s-impl}
We have implemented our solution methods in PyMDE, an object-oriented
Python package for specifying and approximately solving MDE problems. In
addition, PyMDE provides higher-level interfaces for preprocessing data,
visualizing embeddings, and other common tasks. PyMDE is designed to be simple
to use, easily extensible, and performant. 

Our software is open-source, and available at
\begin{center}
  \url{https://github.com/cvxgrp/pymde}.
\end{center}

\subsection{Design}
In this section we briefly describe the design of PyMDE.

\paragraph{Embedding.}
In our package, users instantiate an \texttt{MDE} object by specifying the
number of items $n$, the embedding dimension $m$, the list of edges $\edges$,
the vector distortion function $f$, and the constraint set ($\mathcal C$,
$\anchored$, or $\stdemb$). Calling the \texttt{embed} method on this object
runs the appropriate solution method and returns an embedding,
represented as a dense matrix. 

The distortion function may be selected from a library of functions, including
the functions listed in chapter~\ref{c-distortion-functions}, but it can also
be specified by the user. We use automatic differentiation (via PyTorch
\cite{paszke2019pytorch}) to cheaply differentiate through distortion
functions. For performance, we manually implement the gradient of the average
distortion, using an efficient implementation of the analytical
expression~\eqref{e-gradient}.

\paragraph{Extensibility.}
PyMDE can be easily extended, at no performance cost. Users can simply write
custom distortion functions in PyTorch, without modifying the internals of our
package. Likewise, users can implement custom constraint sets $\mathcal X$ by
subclassing the \texttt{Constraint} class.

\paragraph{Performance.}
On modern CPUs, PyMDE can compute embeddings for hundreds of thousands of items
and millions of edges in just a few seconds (if the embedding 
dimension is not too large). On a modest GPU, in the same amount of time, PyMDE
can compute embeddings with millions of items and tens of millions of edges,
as we have seen in the previous sections.

The most expensive operation in an iteration of
algorithm~\ref{alg-projected-lbfgs} is evaluating the average distortion,
specifically computing the differences $x_i - x_j$ for $(i, j) \in \edges$. On
a CPU, this can become a bottleneck if $p$ is greater than, say, $10^6$. A GPU
can greatly accelerate this computation. For cases when $p$ is very large and a
GPU is unavailable (or has insufficient memory), we provide an implementation
of the stochastic proximal algorithm~\ref{alg-serial-batch}.

By default we use single-precision floating point numbers, since we typically
do not need many significant figures in the fractional part of our embedding.
To minimize numerical drift, for centered and standardized embeddings, we
de-mean the iterate at the end of each iteration of
algorithm~\ref{alg-projected-lbfgs}.

\subsection{Examples}
Here we demonstrate PyMDE with basic code examples.

\paragraph{Hello, world.}
The below code block shows how to construct and solve a very simple MDE
problem, using distortion functions derived from weights. This specific example
embeds a complete graph on three vertices using quadratic distortion functions,
similar to the simple graph layout examples from \S\ref{s-simple-examples}.

\lstset{language=mypython}
\begin{lstlisting}
import pymde
import torch

n, m = 3, 2
edges = torch.tensor([[0, 1], [0, 2], [1, 2]])
f = pymde.penalties.Quadratic(torch.tensor([1.0, 2.0, 3.0]))
constraint = pymde.Standardized()

mde = pymde.MDE(
  n_items=n,
  embedding_dim=m,
  edges=edges,
  distortion_function=f,
  constraint=constraint)
mde.embed()
\end{lstlisting}

In this code block, after importing the necessary packages, we specify the
number of items $n$ and the embedding dimension $m$. Then we construct the list
\texttt{edges} of item pairs; in this case, there are just three edges, $(0, 1)$, $(0,
2)$, and $(1, 2)$. Next we construct the vector distortion function \texttt{f},
a quadratic penalty with weight $w_1 = 1$ across the edge $(0, 1)$, $w_2 = 2$
across $(0, 2)$, and $w_3 = 3$ across $(1, 2)$. Then we create a
standardization constraint. In the last two lines, the MDE problem is
constructed and a minimum-distortion embedding is computed via the
\texttt{embed} method. This method populates the attribute \texttt{X} on the
\texttt{MDE} object with the final embedding. It also writes the distortion and
the residual norm to the attributes \texttt{value} and \texttt{residual\_norm}:
\lstset{language=mypython, numbers=none}
\begin{lstlisting}
print('embedding matrix:\n', mde.X)
print(f'distortion: {mde.value:.4f}')
print(f'residual norm: {mde.residual_norm:.4g}')

embedding matrix:
 tensor([[ 1.4112,  0.0921],
        [-0.6259, -1.2682],
        [-0.7853,  1.1761]])
optimal value: 12.0000
residual norm: 1.583e-06
\end{lstlisting}

The \texttt{MDE} class has several other methods, such as
\texttt{average\_distortion}, which computes the average distortion of a
candidate embedding, \texttt{plot}, which plots the embedding when
the dimension is one, two, or three, and \texttt{play}, which generates
a movie showing the intermediate iterates formed by the solution method.
Additionally, the \texttt{embed} method takes a few optional arguments, through
which (\eg) an initial iterate, the memory size, iteration limit, and solver
tolerance can be specified.

\paragraph{Distortion functions.}
PyMDE's library of distortion functions includes all the
functions listed in \S\ref{s-weights} and \S\ref{s-orig-dist}, among others. 
Users simply construct a vector distortion function by passing weights or
original deviations to the appropriate constructor. Every parameter appearing
in a distortion function has a sensible default value (which the user can
optionally override). PyMDE uses single-index notation for distortion
functions, meaning the $k$th component of the embedding distances
corresponds to the $k$th row of the list of edges supplied to the \texttt{MDE}
constructor. 
For example, a quadratic penalty with $f_k(d_k) = w_kd_k^2$ can be
constructed with 
\texttt{pymde.penalties.Quadratic(weights)},
while a quadratic loss with $f_k(d_k) = (d_k - \delta_k)^2$ can be constructed with
\texttt{pymde.losses.Quadratic(deviations)}. 

For functions derived
from both positive and negative weights, we provide a generic distortion
function based on attractive and repulsive penalties, of the form
\eqref{eq-penalty}. As an example, a distortion function with a log-one-plus
attractive penalty and a logarithmic repulsive penalty can be constructed as
\begin{lstlisting}
f = pymde.penalties.PushAndPull(weights,
  attractive_penalty=pymde.penalties.Log1p,
  repulsive_penalty=pymde.penalties.Log)
\end{lstlisting}

Users can implement their own distortion functions. These simply need
be Python callables mapping the vector of embedding distances to the vector
of distortions via PyTorch operations. For example, a quadratic penalty can be
implemented as 
\begin{lstlisting}
def f(distances):
  return weights * distances**2
\end{lstlisting}
and a quadratic loss can be implemented as
\begin{lstlisting}
def f(distances):
  return (deviations - distances)**2
\end{lstlisting}
(note the weights and deviations are captured by lexical closure). More
generally, the distortion function may be any callable Python object,
such as a \texttt{torch.nn.Module}.

\paragraph{Constraints.} Each \texttt{MDE} object is parametrized by
a constraint. PyMDE currently implements three constraint sets, which
can be constructed using \texttt{pymde.Centered()}, \texttt{pymde.Standardized()},
and \texttt{pymde.Anchored(anchors, values)}, where \texttt{anchors} is a
\texttt{torch.Tensor} enumerating the anchors and \texttt{values} enumerates their
values. Custom constraints can be implemented by subclassing the \texttt{Constraint}
class and implementing the methods \texttt{project\_tspace} and \texttt{project}.

\paragraph{Preprocessors.} We have implemented all the preprocessors
described in \S\ref{s-preprocessing}, in the module \texttt{pymde.preprocess}.
These preprocessors can be used to transform original data and eventually
obtain an MDE problem.

Several of these preprocessors take as input either a data matrix $Y$ with $n$
rows, with each row a (possibly high dimensional) vector representation of an
item, or a sparse adjacency matrix $A \in \reals^{n \times n}$ of a graph
(wrapped in a \texttt{pymde.Graph} object), with $A_{ij}$ equal to an original
deviation $\delta_{ij}$ between items $i$ and $j$. For example, the below code
constructs a $k$-nearest neighborhood graph from a data matrix $Y$.
\begin{lstlisting}
graph = pymde.preprocess.k_nearest_neighbors(Y, k=15)
\end{lstlisting}
The same function can be used for a graph object, with adjacency matrix $A$.
\begin{lstlisting}
graph = pymde.preprocess.k_nearest_neighbors(pymde.Graph(A), k=15)
\end{lstlisting}
The associated edges and weights are attributes of the returned object.
\begin{lstlisting}
edges, weights = graph.edges, graph.weights
\end{lstlisting}

We have additionally implemented a preprocessor for computing some or all
of the graph distances (\ie, shortest-path lengths), given a graph. Our
(custom) implementation is efficient and can exploit multiple CPU cores: the
algorithm is implemented in C, and we use multi-processing for
parallelism. As an example, computing roughly 1 billion graph distances
takes just 30 seconds using 6 CPUs.

\paragraph{Other features.} PyMDE includes several other features, including
a stochastic solve method, high-level interfaces for constructing
MDE problems for preserving neighborhoods or distances, visualization tools and
more. The full set of features, along with code examples, is presented in our
online documentation at \begin{center}
  \url{https://pymde.org}.
\end{center}

\part{Examples}\label{p-examples}

\chapter{Images}\label{c-images}
In this chapter we embed the well-known MNIST collection
of images \cite{lecun1998mnist}, 
given by their grayscale vectors, into $\reals^2$ or $\reals^3$.
Embeddings of images in general, into three or fewer dimensions,
provide a natural user interface for 
browsing a large number of images; \eg, an interactive application might
display the underlying image when the cursor is placed over an
embedding vector. 
All embeddings in this chapter (and the
subsequent ones) are computed using the experiment setup detailed in
chapter~\ref{c-num-ex}.

\section{Data}
The MNIST dataset consists of $n=70,000$, 28-by-28
grayscale images of handwritten digits, represented as vectors in
$\reals^{784}$, with components between 0 and 255.
Each image is tagged with the digit it depicts. 
The digit is a \emph{held-out
attribute}, \ie, we do not use the digit in our embeddings, which are
constructed using only the raw pixel data. We use the digit 
attribute to check whether the embeddings are sensible, as 
described in \S\ref{s-validation}.

\section{Preprocessing}\label{s-preprocessing}
We use distortion functions derived from weights. The set
of similar pairs $\esim$ holds the edges of a $k$-nearest neighbor graph, using
$k=15$; in determining neighbors, we use the Euclidean distance between image
vectors. (The Euclidean distance is in general a poor global metric on images,
but an adequate local one. Two images with large Euclidean distance might be
similar \eg, if one is a translation or rotation of the other, but
two images with small Euclidean distance are surely similar.)

To compute the nearest neighbors efficiently, we use the approximate nearest
neighbor algorithm from \cite{dong2011efficient}, implemented in the Python
library \texttt{pynndescent} \cite{pynndescent}. Computing the neighbors takes
roughly 30 seconds. The resulting neighborhood graph is connected and has
774,746 edges. We sample uniformly at random an additional $|\esim|$ pairs
not in $\esim$ to obtain the set $\edis$ of dissimilar image pairs.
We assign $w_{ij} = +2$ if $i$ and $j$ are both neighbors of
each other; $w_{ij} = +1$ if $i$ is a neighbor of $j$ but $j$ is not a neighbor
of $i$ (or vice versa); and $w_{ij} = -1$ for $(i, j) \in \edis$.

\section{Embedding}\label{s-image-embedding}
We consider two types of MDE problems: (quadratic) MDE problems based on the
neighborhood graph $\esim$ (which has 774,746 edges), and MDE problems based on
the graph $\esim \cup \edis$ (which has 1,549,492 edges).

\subsection{Quadratic MDE problems}
We solve two quadratic MDE problems on the graph $\esim$ (\ie, MDE problems
with quadratic penalties and a standardization constraint; the resulting
embeddings can be interpreted as Laplacian embeddings, since the weights are
nonnegative). The first problem has embedding dimension $m=2$ and the second
has dimension $m=3$.

We solve these MDE problems using algorithm~\ref{alg-projected-lbfgs}. Using
PyMDE, embedding into $\reals^2$ takes $1.8$ seconds on our GPU
and 3.8 seconds on our CPU; embedding into $\reals^3$ takes $2.6$ seconds
on GPU and $7.5$ seconds on CPU. The solution method terminates after 98
iterations when $m=2$ and after 179 iterations when $m=3$.

\begin{figure}
\centering
\includegraphics{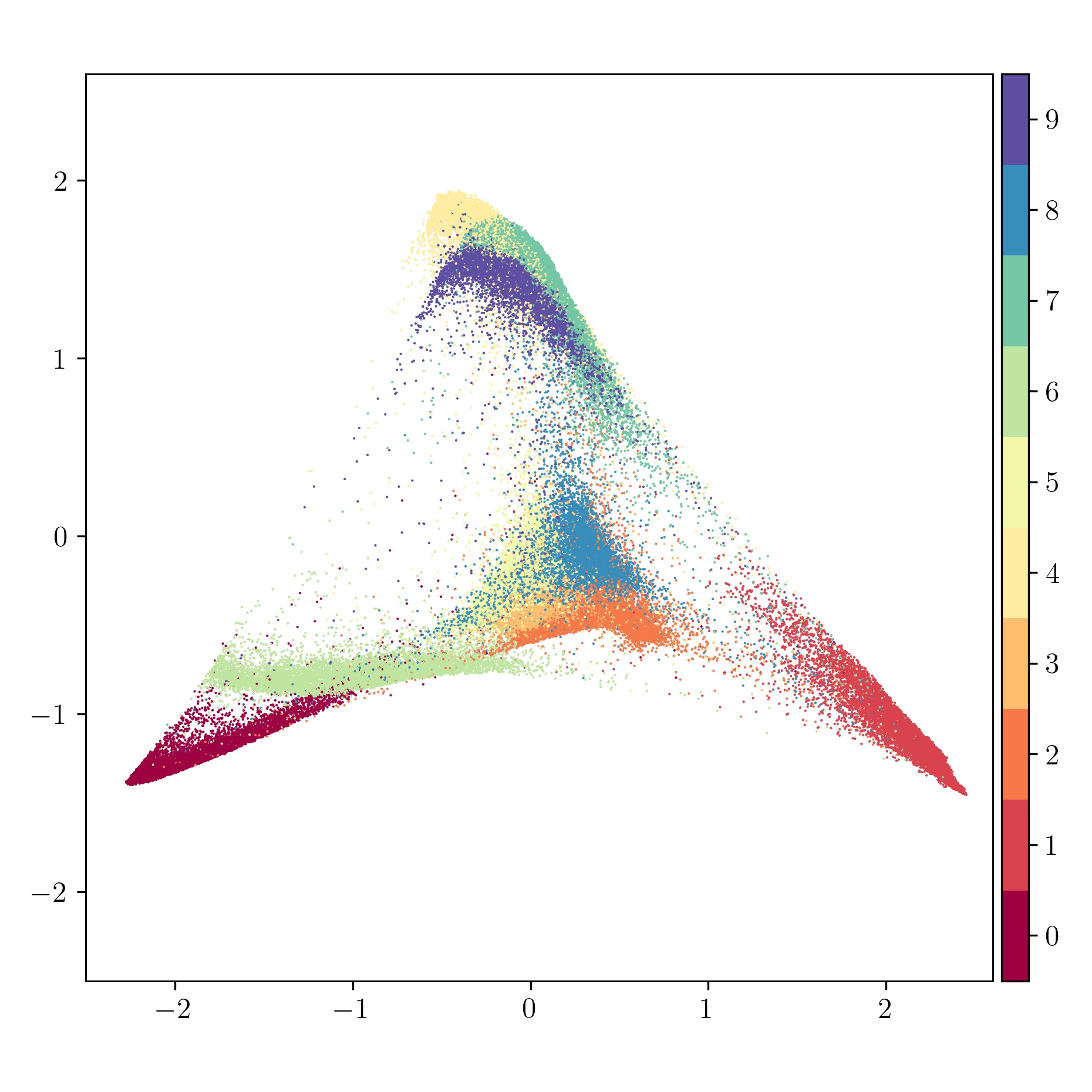}
\caption{\emph{Embedding based on} $\esim$ $(m=2)$.
An embedding of MNIST, obtained by solving a quadratic MDE problem derived
from a neighborhood graph.}\label{f-mnist-quad}
\end{figure}

\begin{figure}
\centering
\includegraphics{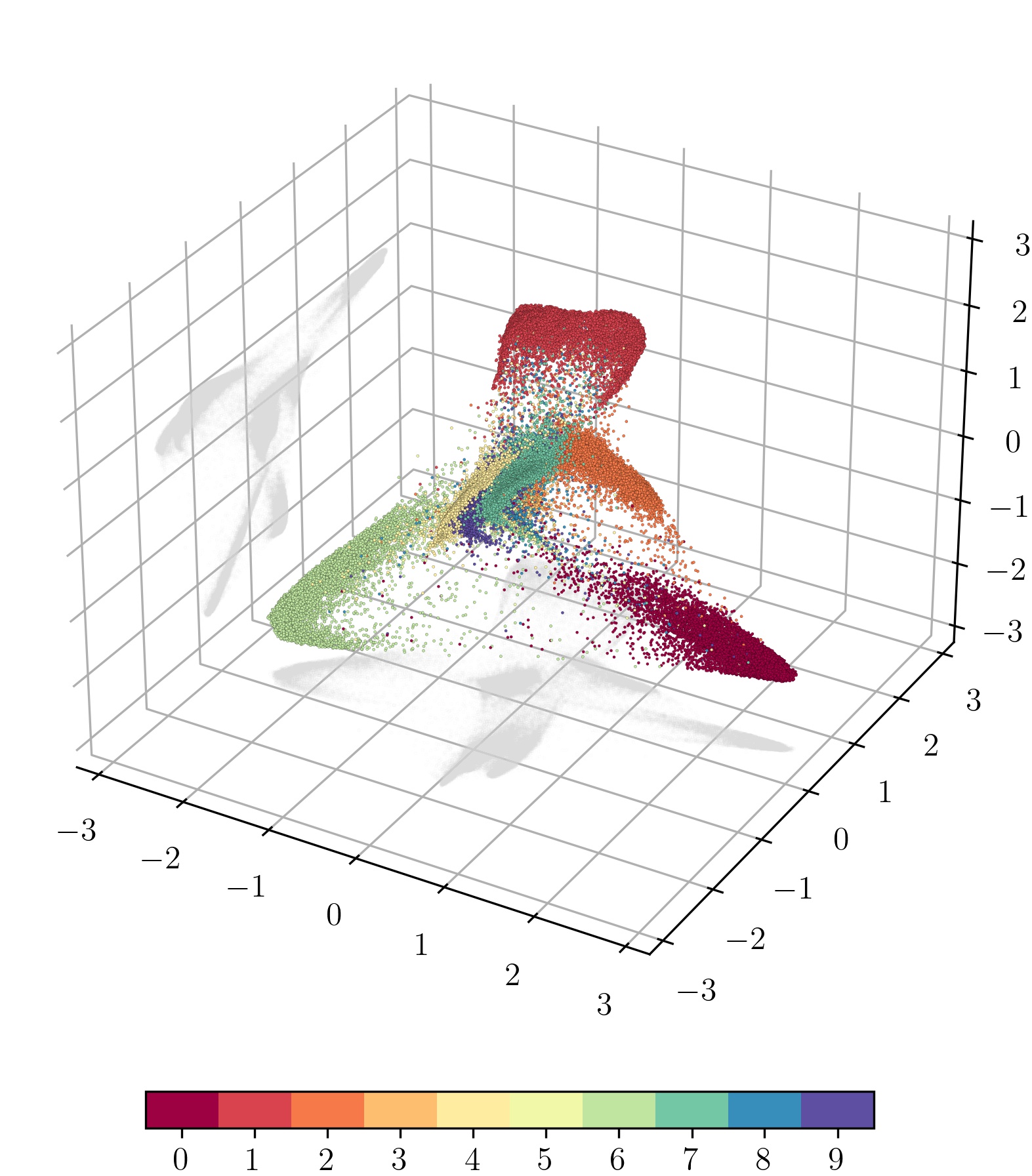}
\caption{\emph{Embedding based on} $\esim$ $(m=3)$.
An embedding of MNIST, obtained by solving a quadratic MDE problem derived
from a neighborhood graph.}\label{f-mnist-quad-3d}
\end{figure}

The embeddings are plotted in figures~\ref{f-mnist-quad} and
\ref{f-mnist-quad-3d}, with the embedding vectors colored by digit. In both
cases, the same digits are clustered near each other in the embeddings.

\clearpage
\paragraph{Outliers.}
\begin{figure}
\centering
\includegraphics{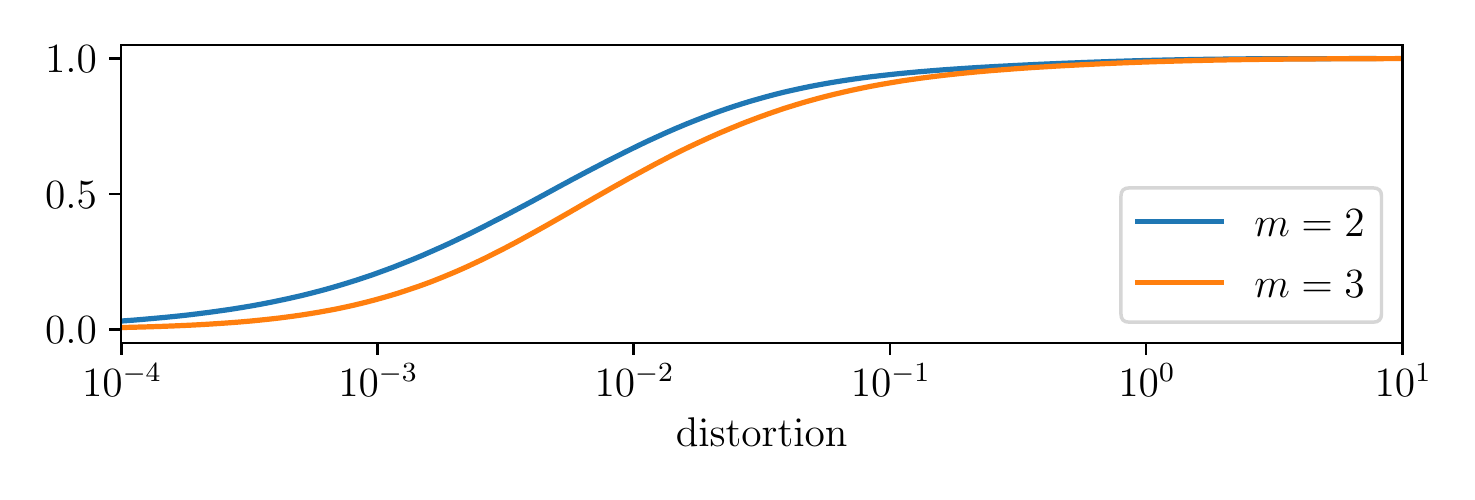}
\caption{\emph{Distortions.} CDF of distortions for the quadratic MDE problem.}
\label{f-mnist-distortions}
\end{figure}

\begin{figure}
\centering
\includegraphics{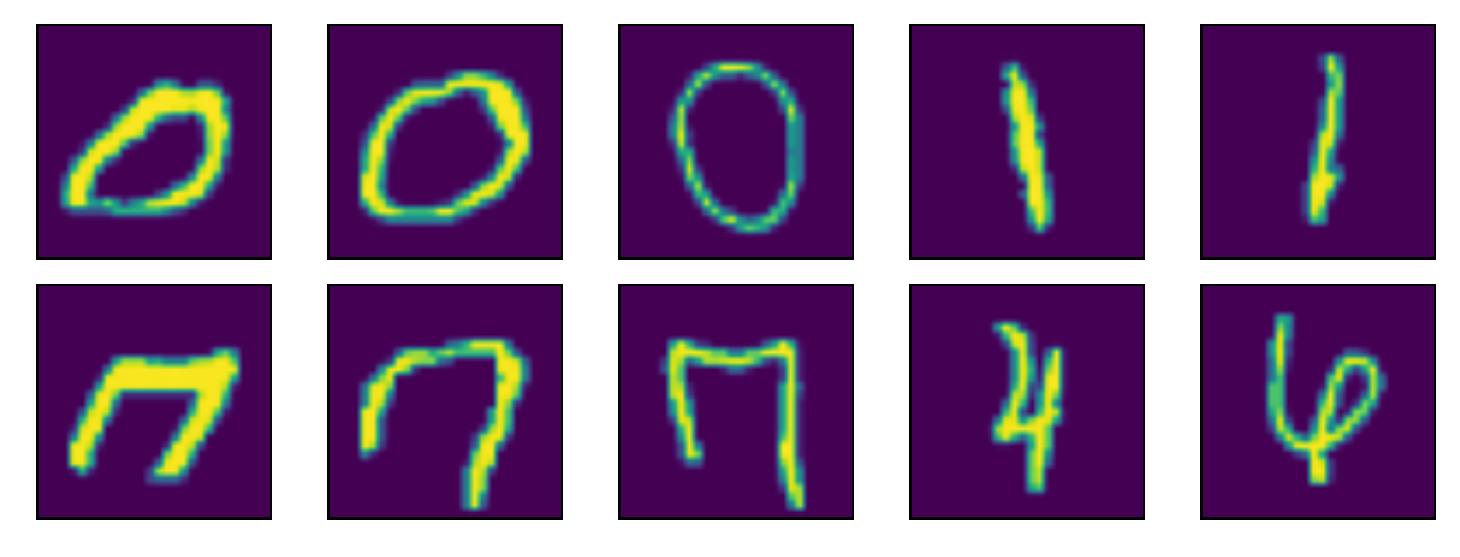}
\caption{\emph{Outliers.} Five pairs of images associated with
item pairs with large distortion.}
\label{f-mnist-outliers}
\end{figure}

The embeddings obtained by solving the quadratic MDE problem have low average
distortion, but some pairs $(i, j)$ have much larger distortion
$f_{ij}(d_{ij})$ than others. The cumulative distribution functions (CDFs)
of distortions are shown in figure~\ref{f-mnist-distortions}.
Figure~\ref{f-mnist-outliers} plots the images associated with the five pairs
in $\esim$ that incurred the highest distortion, for the embedding into
$\reals^2$. Each column shows a pair of images. Some of
these pairs do not appear to be very similar, \eg, a $1$ is paired with a $4$.
Other pairs include images that are nearly illegible.

\subsection{Other embeddings}\label{s-mnist-log}
Next we solve MDE problems with distortion functions derived from both positive
and negative weights.
In each MDE problem, the distortion functions are based on two penalty
functions, as in \eqref{eq-penalty}: $\psim$, which measures the distortion for
pairs in $\esim$, and $\pdis$, which measures the distortion for pairs in
$\edis$. All the MDE problems take $\pdis$ to be the logarithmic penalty
\eqref{eq-log-penalty}, with hyper-parameter $\alpha=1$. For each problem, we
initialize the solution method at the solution to the quadratic MDE problem
from the previous section.

All embeddings plotted in the remainder of this chapter were aligned
to the embedding in figure~\ref{f-mnist-quad}, via orthogonal transformations,
as described in \S\ref{s-orth-procrustes}.

\paragraph{Standardized embeddings.}
The first problem is derived from the graph $\edges = \esim \cup \edis$
(with $|\esim| = |\edis| = 774,776$).
We impose a standardization constraint and use the log-one-plus penalty
\eqref{eq-log1p-penalty} for $\psim$, with hyperparameter $\alpha=1.5$.

\begin{figure}
\includegraphics{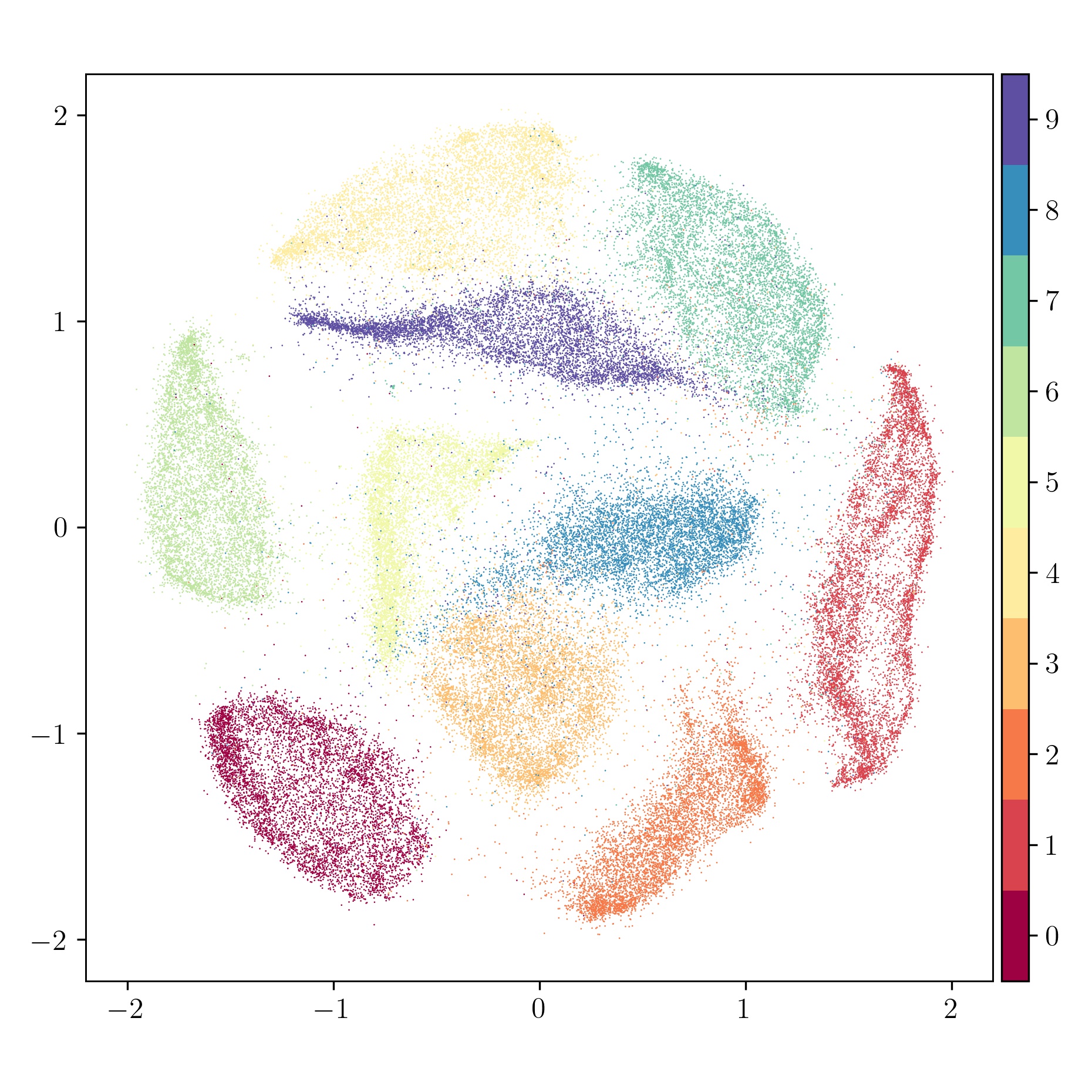}
\caption{\emph{Embedding based on $\esim \cup \edis$.} A standardized
  embedding of MNIST, based on the union of a neighborhood graph
  and a dissimilarity graph, with log-one-plus attractive penalty and
  logarithmic repulsive penalty.}
\label{f-mnist-log1p}
\end{figure}

The solution method ran for 177 iterations, taking 6 seconds on a GPU and 31
seconds on a CPU. Figure~\ref{f-mnist-log1p} shows an embedding obtained by
solving this MDE problem. Notice that this embedding is somewhat similar to the
quadratic one from figure~\ref{f-mnist-quad}, but with more separation between
different classes of digits.

\begin{figure}
\includegraphics{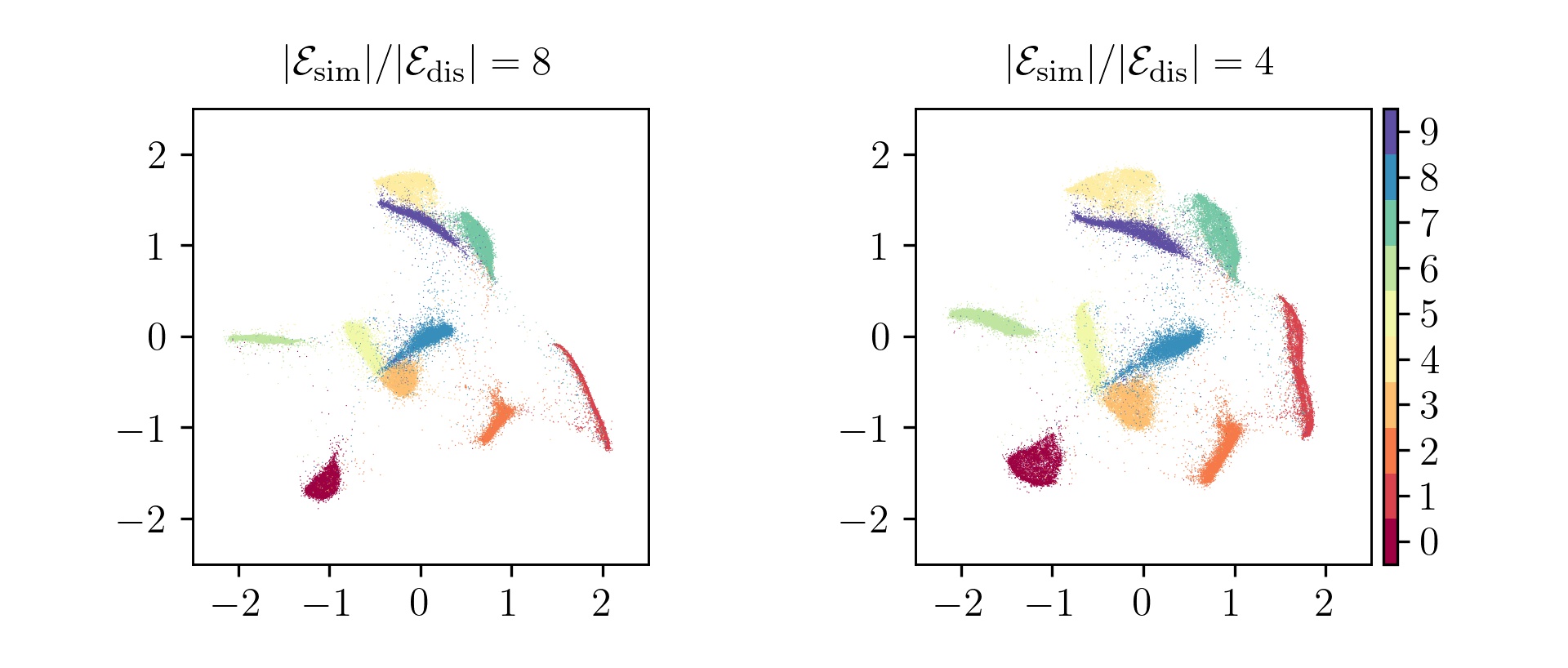}
\includegraphics{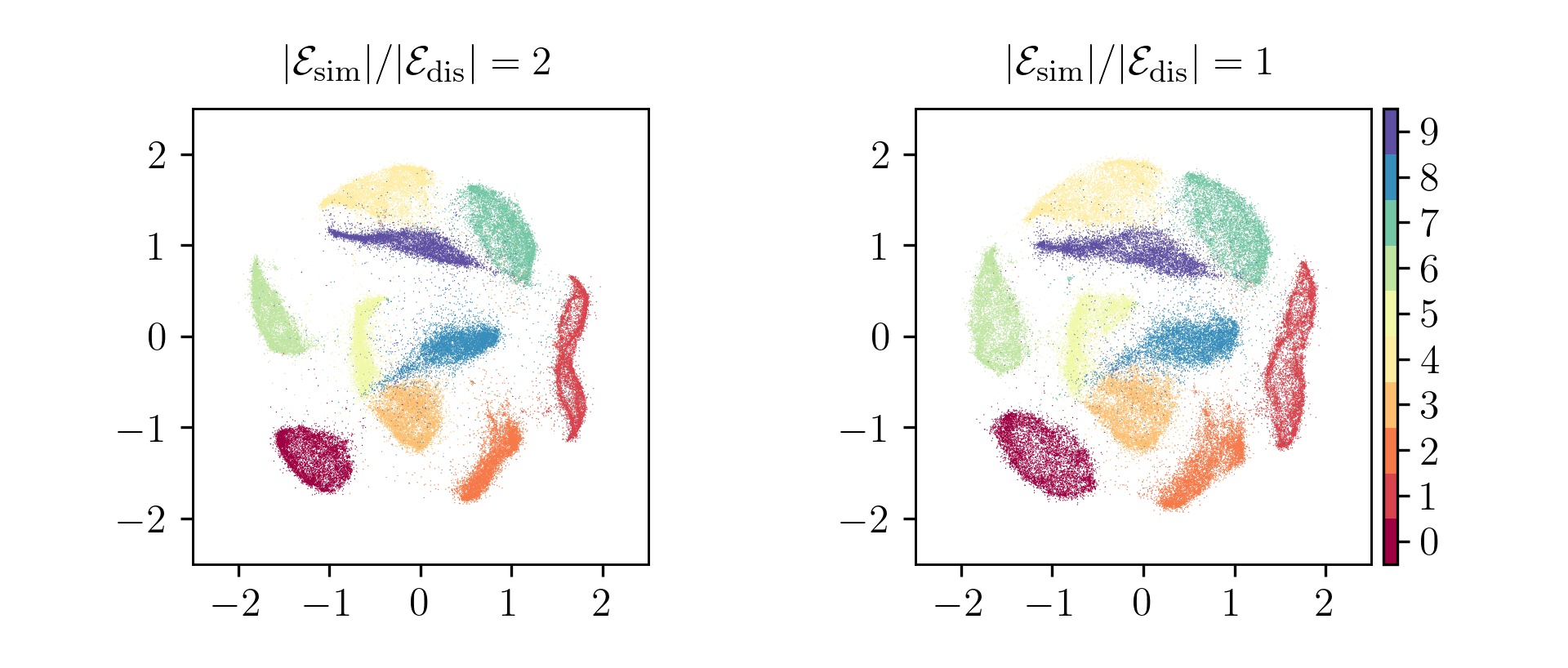}
\includegraphics{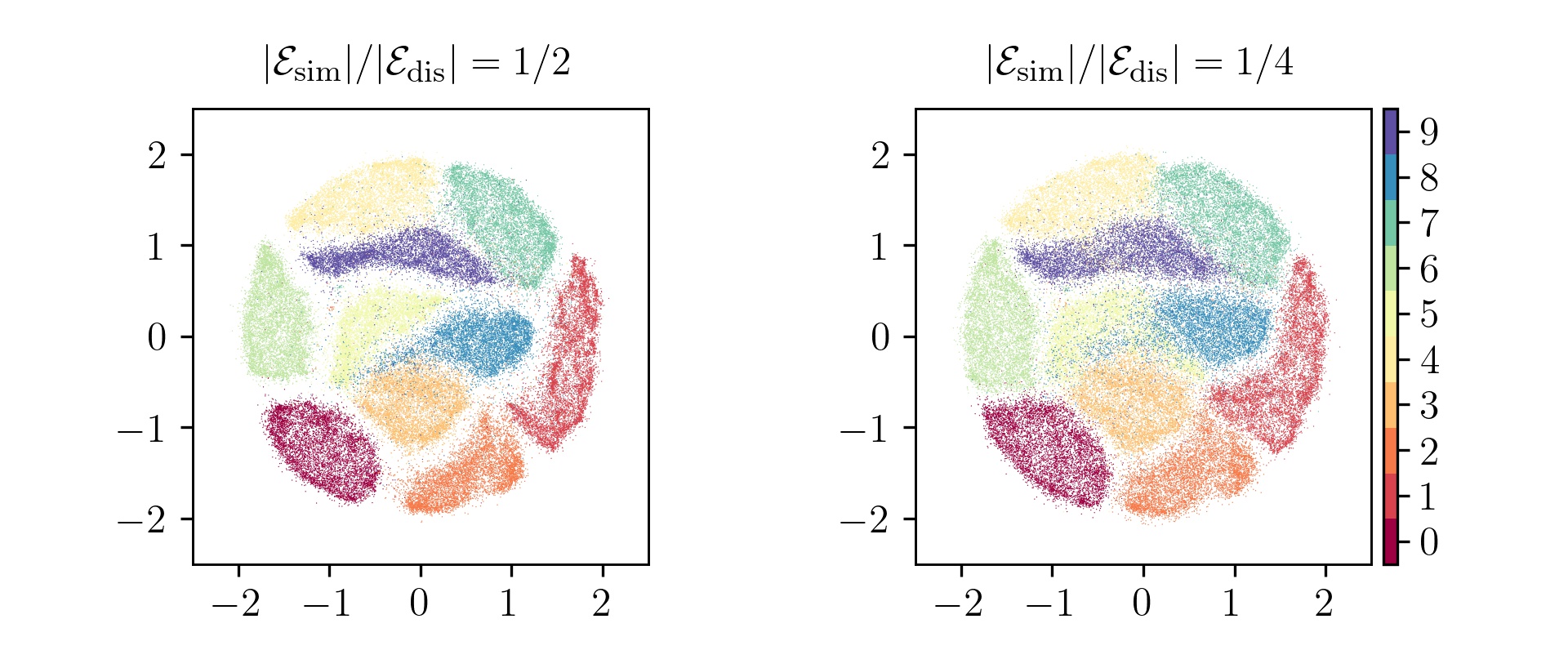}
\caption{\emph{Varying $|\edis|$.} Standardized embeddings of MNIST,
varying $|\edis|$.}
\label{f-mnist-edis}
\end{figure}

\paragraph{Varying the number of dissimilar pairs.}
We solve five additional MDE problems, using the same penalties and constraints
as the previous one but varying the number of dissimilar pairs, to show how the
embedding depends on the ratio $|\esim| / |\edis|$. The results are plotted in
figure~\ref{f-mnist-edis}. When $|\edis|$ is small, items belonging to similar
digits are more tightly clustered together; as $|\edis|$ grows larger, the
embedding becomes more spread out.

\begin{figure}
\includegraphics{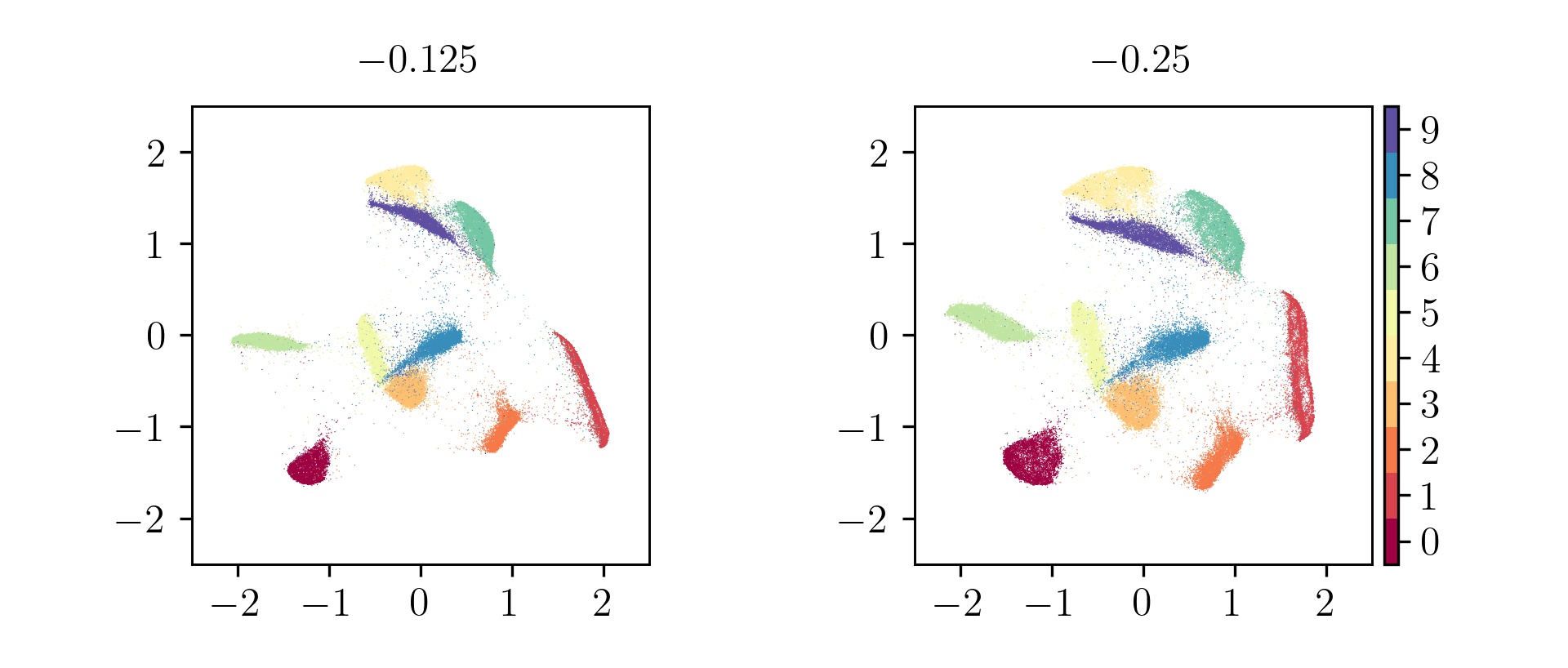}
\includegraphics{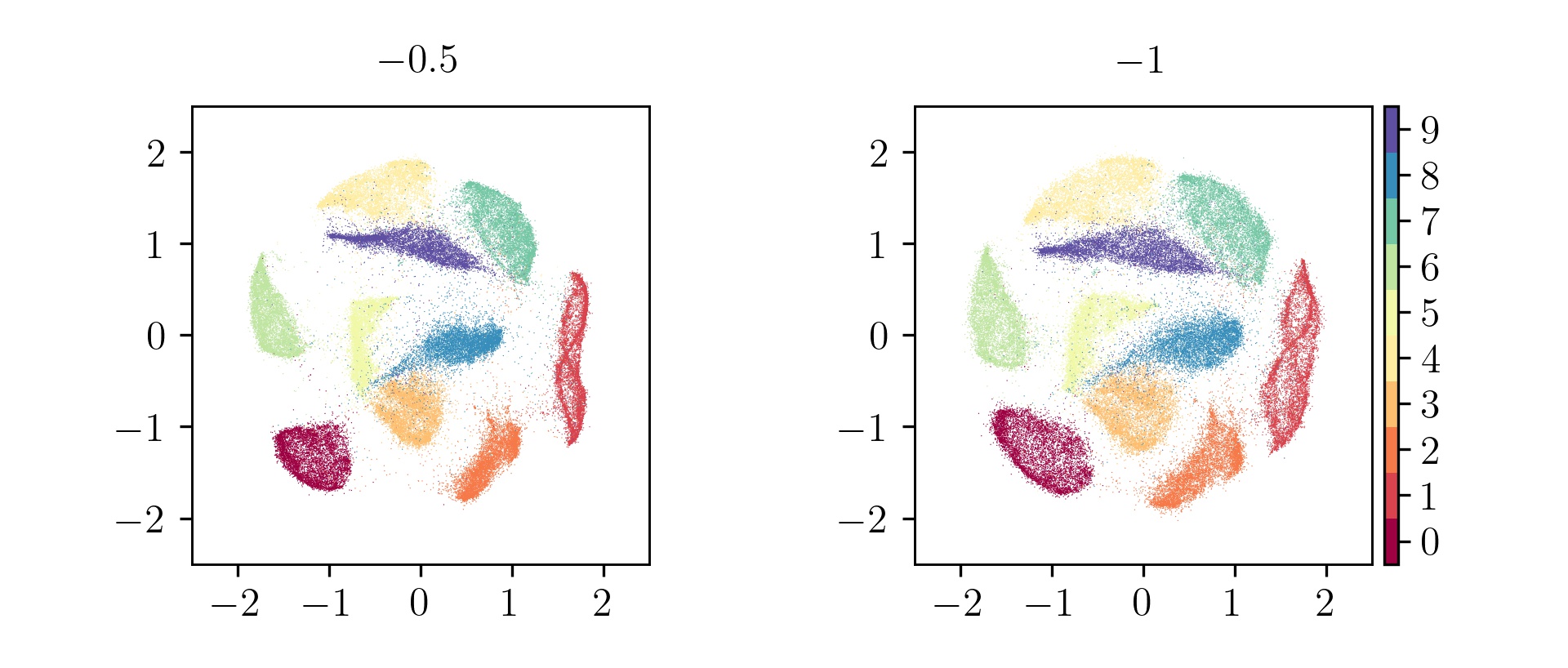}
\includegraphics{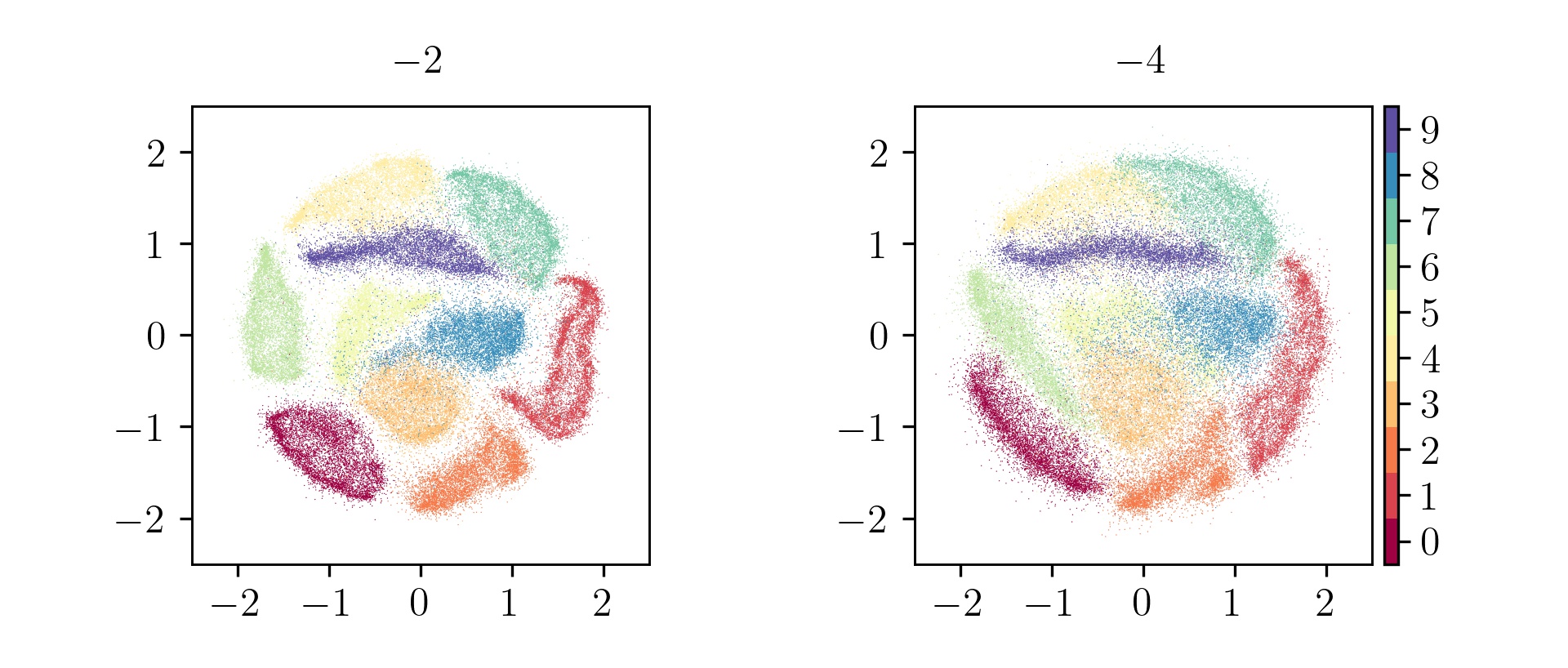}
\caption{\emph{Varying negative weights.} Standardized
embeddings of MNIST, varying negative weights.}
\label{f-mnist-neg-weights}
\end{figure}

\paragraph{Varying the weights for dissimilar pairs.}
Finally, we solve another five MDE problems of the same form, this time
varying the magnitude of the weights $w_{ij}$ for $(i, j) \in \edis$
but keeping $|\edis| = |\esim|$ fixed. The embeddings are plotted
in figure~\ref{f-mnist-neg-weights}. Varying the weights (keeping
$|\edis|$ fixed) has a similar effect to varying $|\edis|$ (keeping the weights
fixed).

\clearpage
\paragraph{Centered embeddings.}
We solve two centered MDE problems and compare their solutions. The first
MDE problem has a log-one-plus attractive penalty \eqref{eq-log1p-penalty}
($\alpha = 1.5$), and the second has a Huber attractive penalty
\eqref{eq-huber-penalty} ($\tau = 0.5$). These problems were solved in about 4
seconds on a GPU and 32 seconds on a CPU. The first problem was solved in 170
iterations, and the second was solved in 144 iterations.

\begin{figure}
\includegraphics{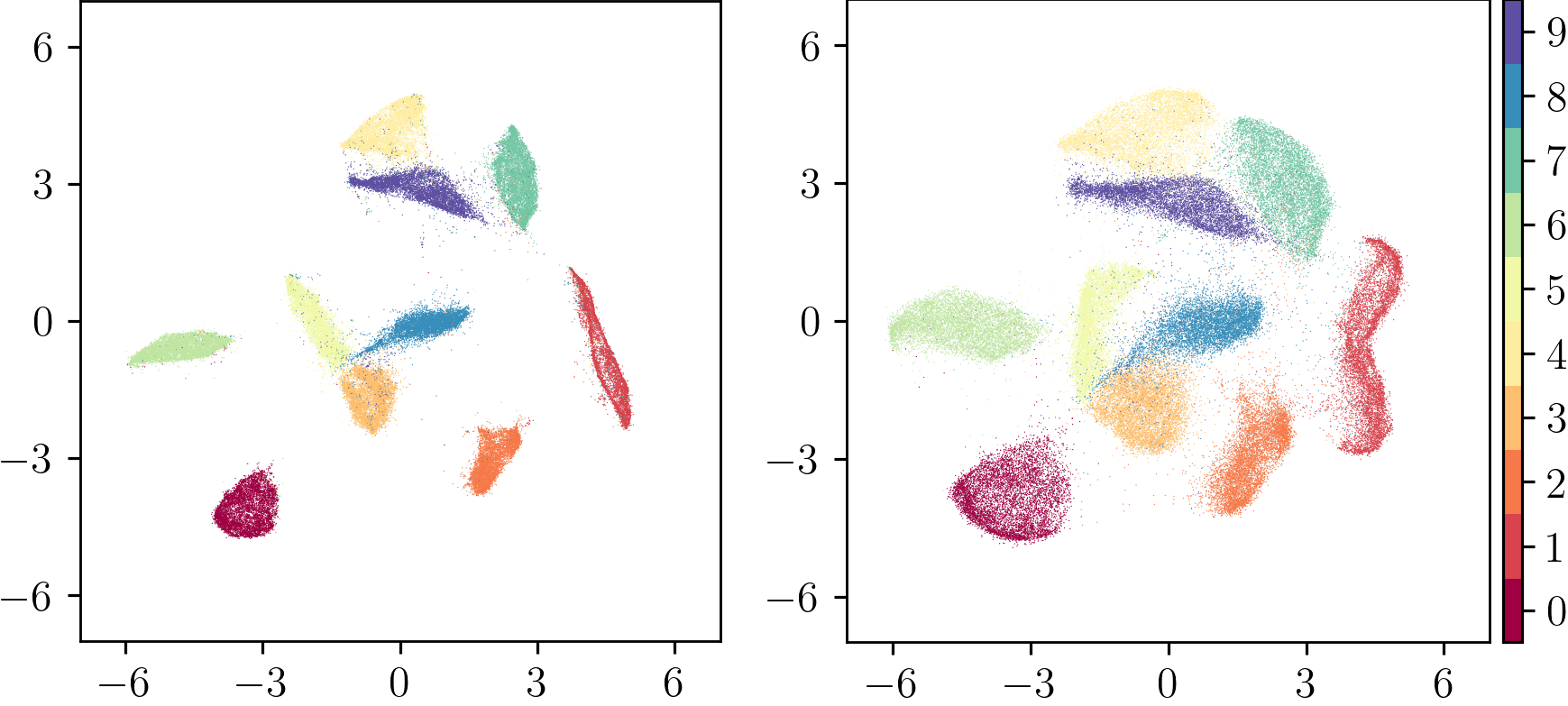}
\caption{\emph{More embeddings.} Centered embeddings of
  MNIST with log-one-plus (\emph{left}) and Huber (\emph{right}) attractive
  penalties, and logarithmic repulsive penalties.}\label{f-mnist-log1p-huber}
\end{figure}

The embeddings, shown in figure~\ref{f-mnist-log1p-huber}, are very similar.
This suggests that the preprocessing (here, the choice of $\esim$ and $\edis$)
has a strong effect on the embedding. The embedding produced using the Huber
penalty has more points in the negative space between classes of digits. This
makes sense, since for large $d$, the Huber penalty more heavily discourages
large embedding distances (for similar points) than does the log-one-plus
penalty. Moreover, the neighborhood graph is connected, so some points must be
similar to multiple classes of digits.

\clearpage
\subsection{Comparison to other methods}
For comparison, we embed MNIST with UMAP \cite{mcinnes2018umap}
and t-SNE \cite{hinton2003stochastic}, two popular embedding methods,
using the umap-learn and openTSNE \cite{policar2019openTSNE} implementations.
Both methods start with neighborhood graphs on the original data, based on the
Euclidean distance; umap-learn uses the nearest 15 neighbors of each image, and
openTSNE uses the nearest 90 neighbors of each image. The UMAP optimization
routine takes 20 seconds on our CPU, and the openTSNE optimization routine
takes roughly one minute. (Neither implementation supports GPU acceleration.)
On similar problems (the centered embeddings from \S\ref{s-mnist-log}), PyMDE
takes roughly 30 seconds on CPU.

\begin{figure}
\centering
\includegraphics{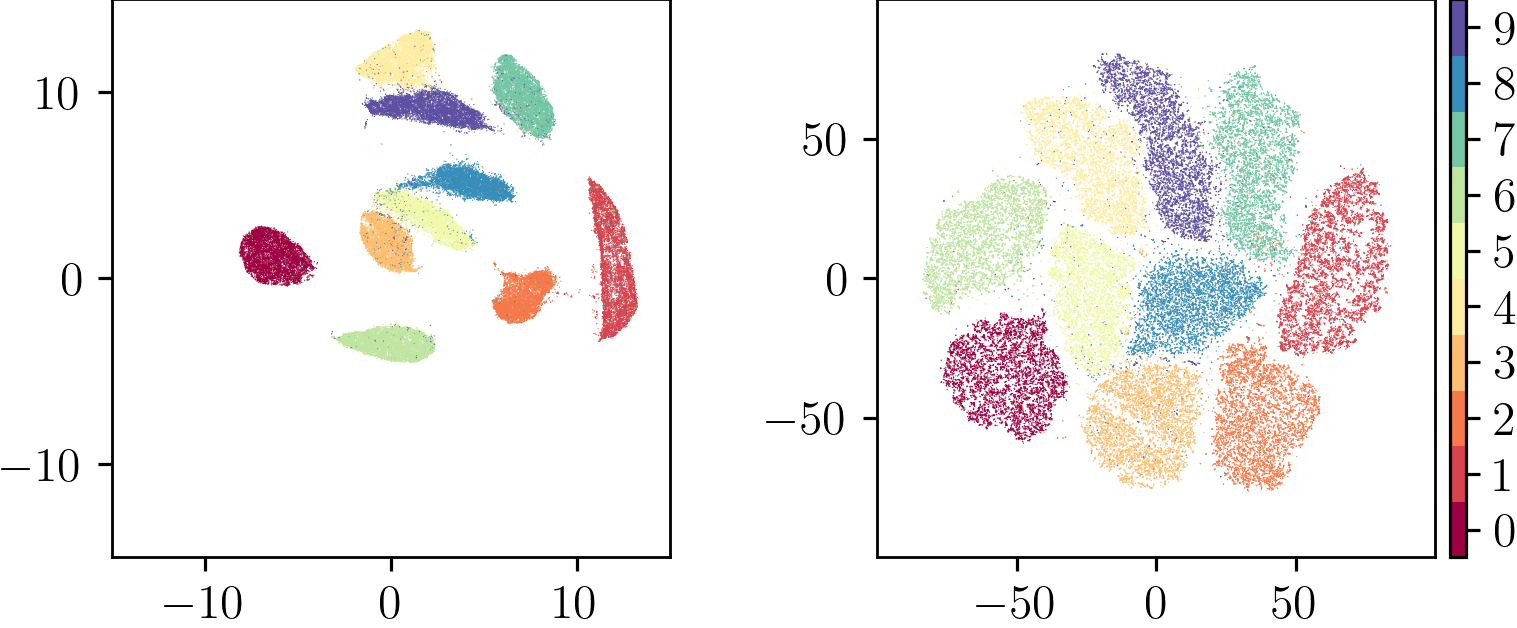}
\caption{UMAP (\emph{left}) and t-SNE (\emph{right})
embeddings of MNIST.}
\label{f-mnist-umap-tsne}
\end{figure}

Figure~\ref{f-mnist-umap-tsne} shows the embeddings. Note that the UMAP
embedding bears a striking resemblance to the previously shown centered
embeddings (figure~\ref{f-mnist-log1p-huber}). This is not surprising, since
UMAP solves an unconstrained optimization problem derived from a similar graph
and with similar distortion functions.

\chapter{Networks}\label{c-network}
The MDE problems we formulate in this chapter are derived from networks. We
start with an undirected (possibly weighted) network on $n$ nodes, in which the
length of the shortest path between nodes $i$ and $j$ encodes an original
distance between them. Such networks frequently arise in practice:
examples include social networks, collaboration networks, road networks, web
graphs, and internet networks. Our goal is to find an embedding of the nodes
that preserves the network's global structure.

The specific network under consideration in this chapter is a
\emph{co-authorship network}: each node corresponds to a unique author, and two
nodes are adjacent if the corresponding authors have co-authored at least one
publication.

\section{Data}
We compiled a co-authorship network by crawling Google Scholar \cite{gscholar},
an online bibliography cataloging research papers from many different fields.
Each cataloged author has a profile page, and each page contains metadata
describing the author's body of work. We crawled these pages to create our
network. Our network contains 590,028 authors and 3,812,943 links. We
additionally collected each author's unique identifier, affiliation,
self-reported interests (such as \emph{machine learning}, \emph{renewable
energy}, or \emph{high energy physics}), h-index \cite{hirsch2005index},
and cumulative number of citations. Our dataset is publicly available, at
\url{https://pymde.org}.

We will compute two embeddings in this chapter: an embedding on the full
network, and an embedding on a network containing only \emph{high-impact
authors}, which we define as authors with h-indices 50 or higher. The
subgraph induced by high-impact authors has 44,682 vertices and 210,681 edges.

When constructing the embeddings, we use only the graph distances between
nodes. Later in this section we describe how we categorize authors into
academic disciplines using their interests. An author's academic discipline
is the held-out attribute we use to check if the embedding makes sense.

\paragraph{Crawling.}
The data was collected in November 2020 by exploring Google
Scholar with a breadth-first search, terminating the search after one week.
The links were obtained by crawling authors' listed collaborators. Google
Scholar requires authors to manually add their coauthors
(though the platform automatically recommends co-authors); as a result, if two
authors collaborated on a paper, but neither listed the other as a co-author,
they will not be adjacent in our network. As a point of comparison, a
co-authorship network of Google Scholar from 2015 that determined co-authorship
by analyzing publication lists contains 409,392 authors and 1,234,019 links
\cite{chen2017building}.

While our co-authorship network faithfully represents Google Scholar, it
deviates from reality in some instances. For example we have found that a small
number of young researchers have listed long-deceased researchers such as
Albert Einstein, John von Neumann, Isaac Newton, and Charles Darwin as
co-authors. Compared to the size of the network, the number of such anomalies
is small; \eg, there are only thirty authors who are adjacent to Einstein but
not among his true co-authors. As such we have made no attempt to remove
erroneous links or duplicated pages (Einstein has two).
For a more careful study of academic collaboration, see
\cite{chen2017building}.

\paragraph{Some basic properties.}
The diameter of the full network (the length of the longest shortest path) is
10. The diameter of the network induced by high-impact authors is 13. 

\begin{figure}
\includegraphics{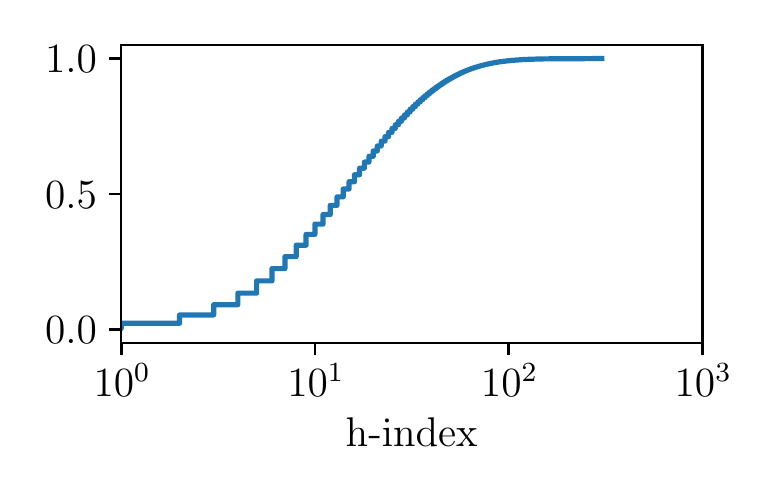}
\includegraphics{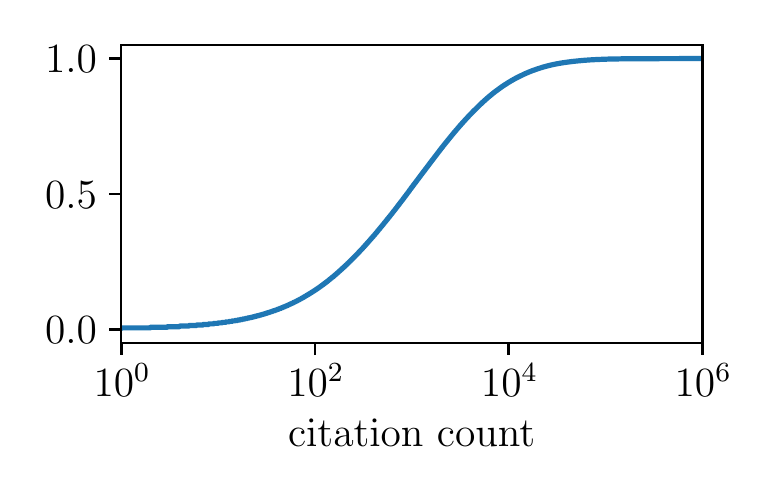}
\caption{CDFs of author h-indices and citation counts}\label{f-scholar-metrics}
\end{figure}

Figure~\ref{f-scholar-metrics} shows the CDFs of h-indices and citation
counts. The median h-index is 14, the 95th percentile is 58, and the 
maximum is
303;  the median citation count is 891, the 95th percentile is 1616, and the
maximum is 1,107,085. 
The maximum h-index and maximum citation count both belong to the
late philosopher Michel Foucault.

\begin{figure}
\includegraphics{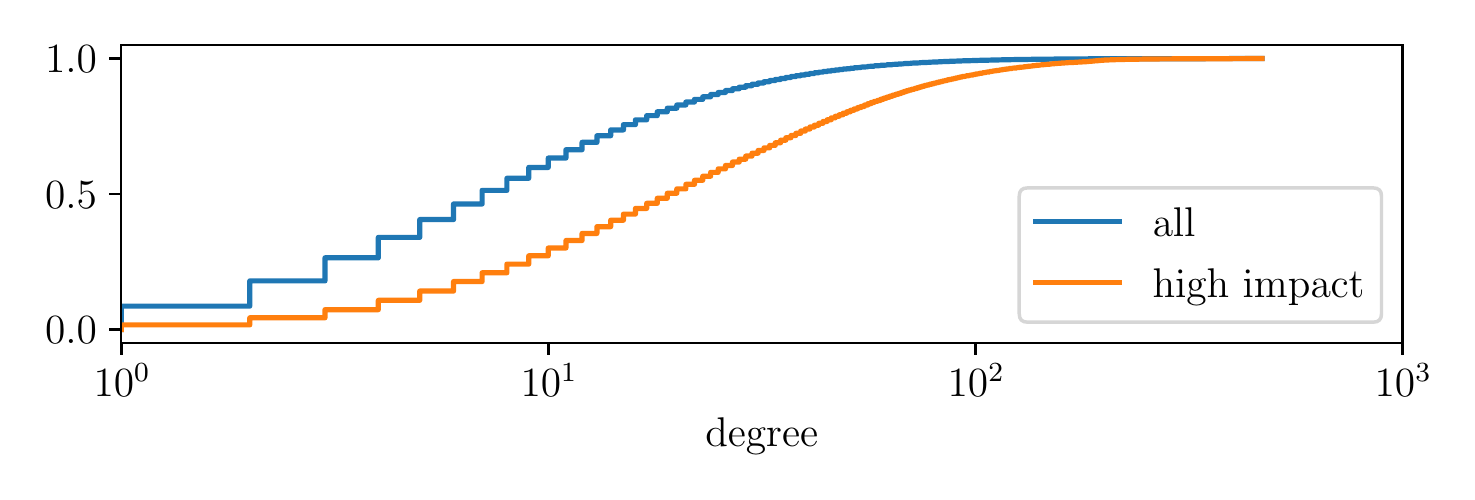}
\caption{CDFs of node degree (co-author count) for the full network and the
high-impact author network.}\label{f-scholar-coauthors} \end{figure}

Figure~\ref{f-scholar-coauthors} shows the CDF of node degrees for the
full network and the subgraph induced by high-impact authors. The number of
collaborators an author has is moderately correlated with h-index (the Pearson
correlation coefficient is 0.44); citation count is strongly correlated with
h-index (0.77).

The data includes 280,819 unique interests, with 337,514 authors interested in
at least one of the 500 most popular interests. The top five interests are
machine learning (listed by 38,220 authors), computer vision (15,123),
artificial intelligence (14,379), deep learning (8,984), and bioinformatics
(8,496).

\paragraph{Academic disciplines.}
We assigned the high-impact authors to one of five academic disciplines, namely
biology, physics, electrical engineering, computer science, and
artificial intelligence, in the following ad-hoc way. First, we manually
labeled the 500 most popular interests with the academic discipline to which
they belonged (interests that did not belong to any of the five disciplines
were left unlabeled). Then, we scanned each author's interests in the order
they chose to display them on their profile, assigning the author to the
discipline associated with their first labeled interest. For example, authors
interested in machine learning first and bioinformatics second were assigned to
artificial intelligence, while authors interested in bioinformatics first and
machine learning second were assigned to biology. Authors that did not have at
least one labeled interest were ignored. This left us with 16,569 high-impact
authors labeled with associated academic disciplines, out of 44,682.

\clearpage
\section{Preprocessing}
We compute graph distances between all pairs of authors. On the full network,
this yields approximately 174 billion distances, or roughly 696 GB of data,
too much to store in the memory of our machine. To make the embedding
tractable, we sample a small fraction of the graph distances uniformly at
random, choosing to retain 0.05 percent of the distances. This results in a
graph $\edges_{\text{all}}$ on 590,028 vertices, with 87,033,113 edges.
Computing all 174 billion distances and randomly subsampling them
took seven hours using six CPU cores. We repeat this preprocessing on the
high-impact network, retaining 10 percent of the graph distances. This yields
a graph $\edges_{\text{impact}}$ with 88,422,873 edges. Computing the graph
distances with PyMDE took less than one minute, using six CPU cores.

\begin{figure}
\begin{center}
\includegraphics{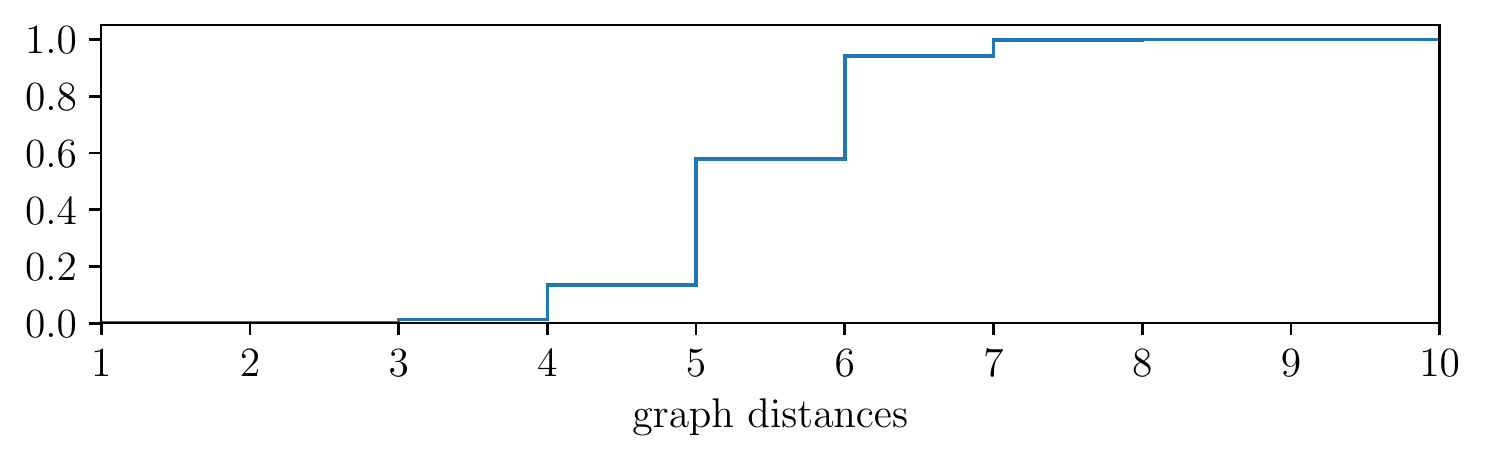}
\includegraphics{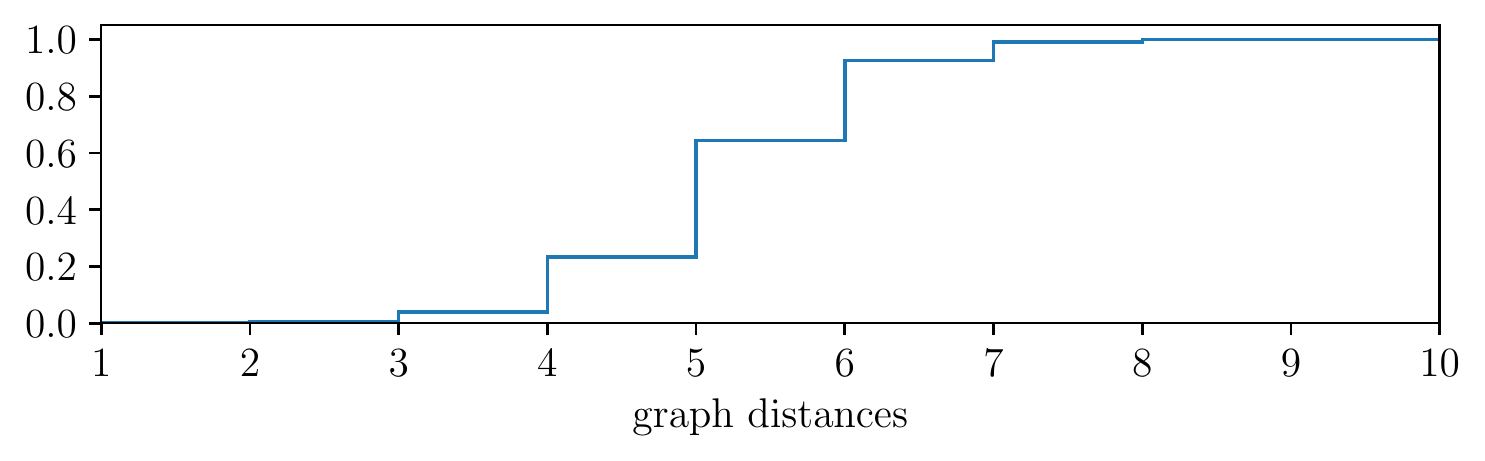}
\end{center}
\caption{CDFs of graph distances. \emph{Top.} All authors. \emph{Bottom.}
High-impact authors.}
\label{f-scholar-distance-CDF}
\end{figure}

The CDFs of graph distances are shown in figure~\ref{f-scholar-distance-CDF}.
The values 4, 5, and 6 account for around 90\% of the distances in both
networks. (We will see artifacts of these highly quantized values in some of
our embeddings.)

\section{Embedding}
We solve two unconstrained MDE problems, based on $\edges_{\text{all}}$ and
$\edges_{\text{impact}}$. The MDE problems are derived from original
deviations, with $\delta_{ij}$ being the graph distance between $i$ and $j$.
For both, we use the absolute loss \eqref{eq-absolute-loss} and embed into
$\reals^2$.

\subsection{All authors}
\begin{figure}
\includegraphics{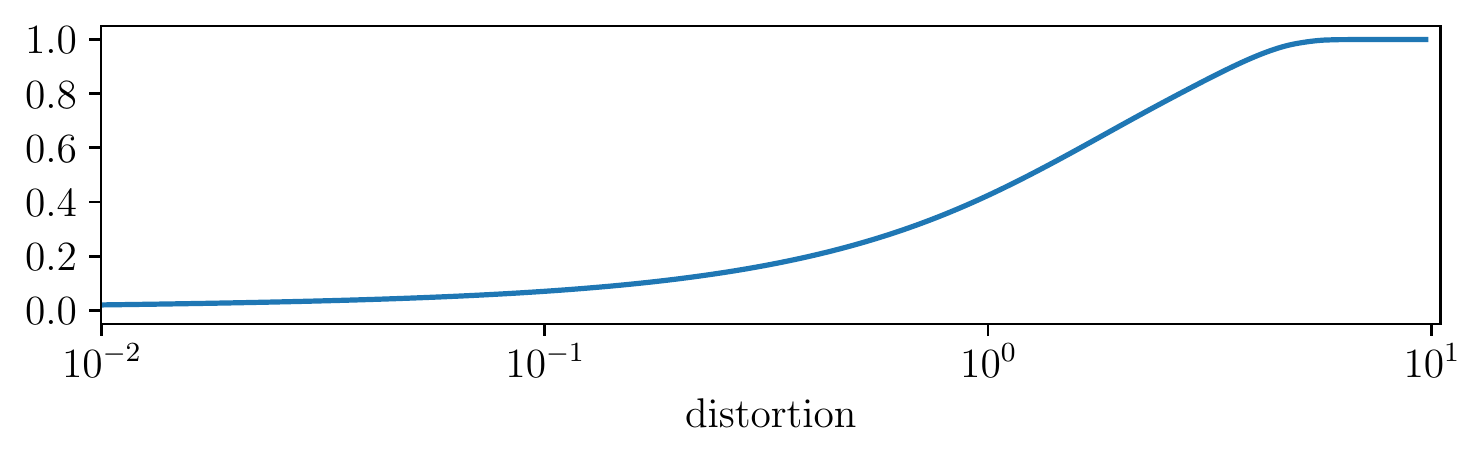}
\caption{CDF of distortions for the embedding on all authors.}
\label{f-scholar-distortion}
\end{figure}
The embedding on $\edges_{\text{all}}$ has distortion 1.58. The embedding was
computed in 300 iterations by PyMDE, to a residual norm of $1.9 \times
10^{-5}$. This took 71 seconds on our GPU.

Figure~\ref{f-scholar-distortion} shows a CDF of the the distortions. The five
largest distortions all belong to pairs whose true distance is 9, but whose
embedding distances are small (the pair with the largest distortion includes a
philosopher and an automatic control engineer). The authors in these
pairs are ``unimportant'' in that they have very few co-authors, and lie on the
periphery of the embedding.

\begin{figure}
\includegraphics{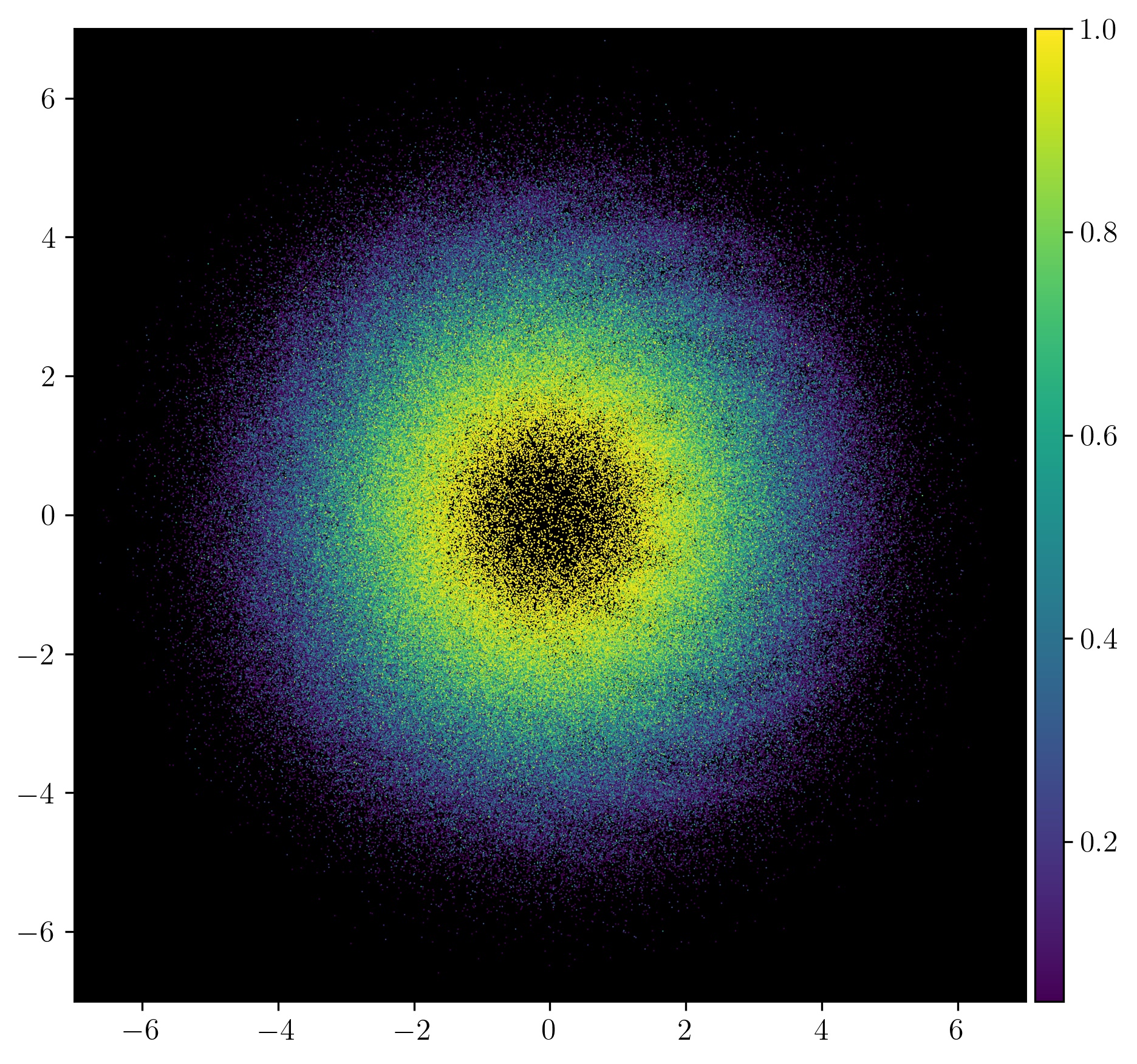}
\caption{\emph{Co-authorship network.} An embedding of a co-authorship network
from Google Scholar, colored by degree percentile. The brighter the color, the
more co-authors the author has. The network has 590,028
authors.}\label{f-coauthors-all}
\end{figure}

Figure~\ref{f-coauthors-all} shows the embedding. (Plots of embeddings in this
chapter have black backgrounds, to make various features of the embeddings
easier to see.) Each embedding vector is colored by its author's degree in the
co-authorship network; the brighter the color, the more co-authors the author
has. The embedding appears sensible: highly collaborative authors are near the
center, with the degree decreasing as one moves outward. The diameter of the
embedding is approximately 13 (measured in Euclidean norm), compared to the
original network's 10 (measured in graph distance).

\clearpage
\subsection{High-impact authors}
\begin{figure}
  \includegraphics{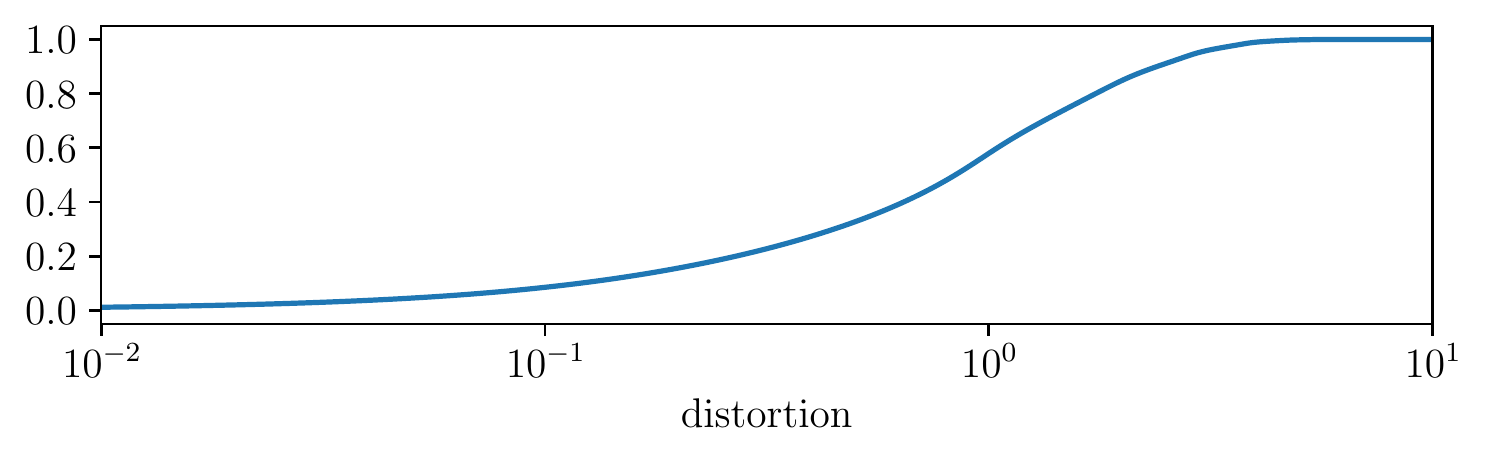}
  \caption{CDF of distortions for the embedding on high-impact authors.}\label{f-scholar-high-impact-distortion}
\end{figure}

An embedding on the graph $\edges_{\text{impact}}$ has average distortion 1.45
(figure~\ref{f-scholar-high-impact-distortion} shows a CDF of the distortions).
PyMDE computed this embedding on a GPU in 54 seconds, in 173 iterations.

\begin{figure}
\includegraphics{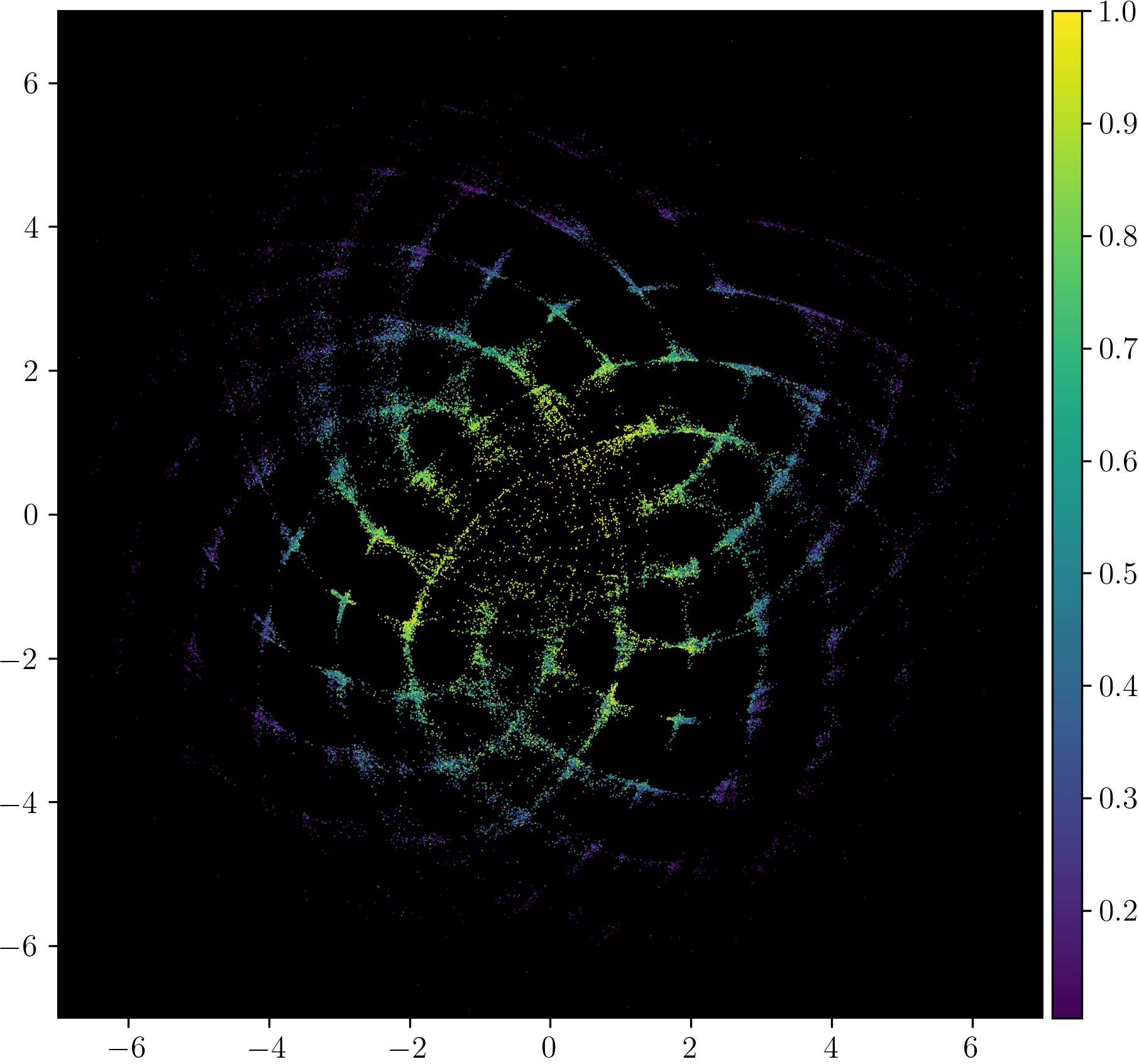}
\caption{\emph{High-impact authors.} An embedding of a co-authorship network
on authors with an h-index of 50 or higher, colored by degree percentile.}\label{f-coauthors-impact}
\end{figure}

Figure~\ref{f-coauthors-impact} shows the embedding, colored by degree.
Highly collaborative authors are near the center, with collaboration decreasing
radially from the center. The diameter is approximately 16, compared to the
original network's 13.

\begin{figure}
\includegraphics{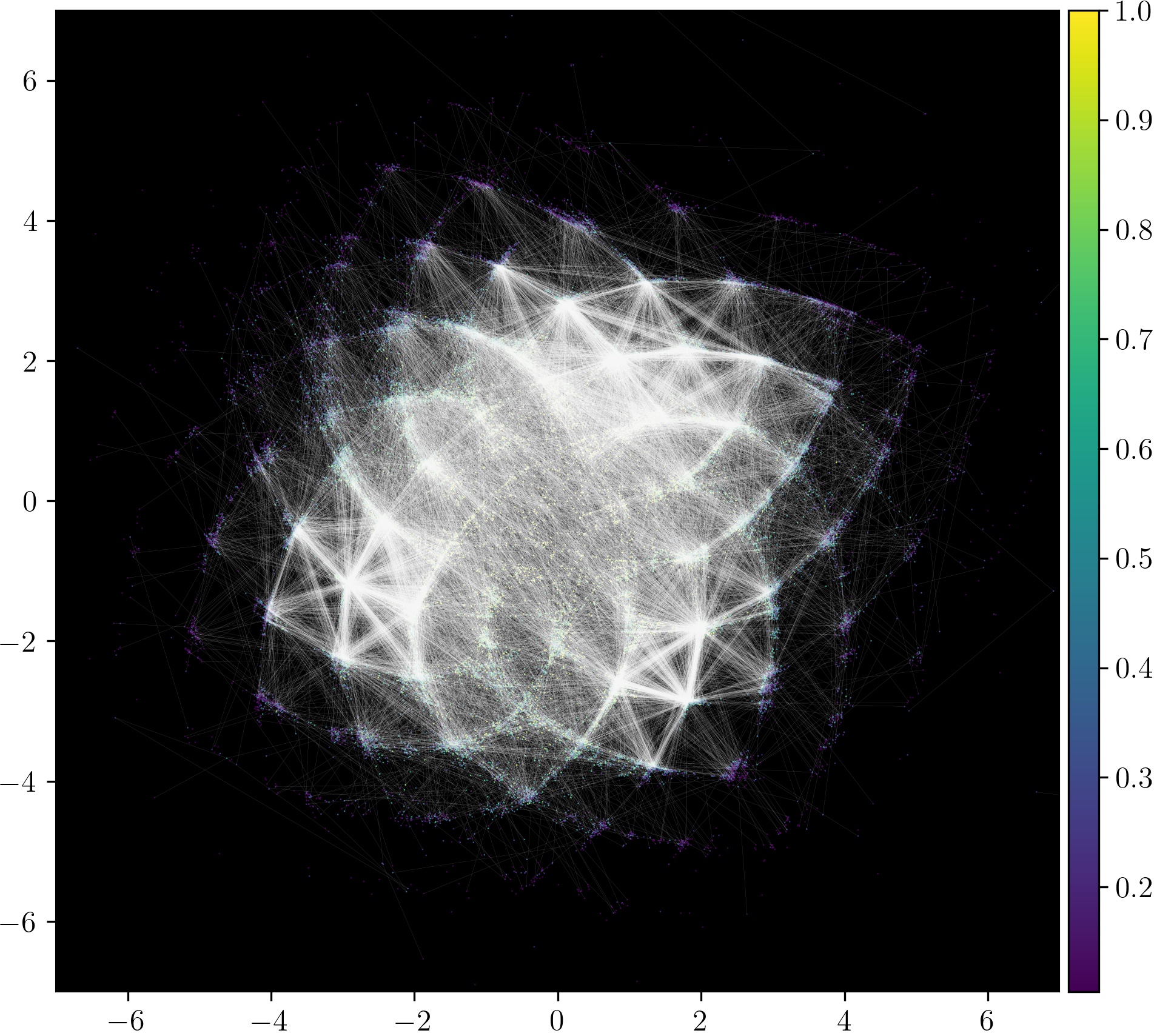}
\caption{\emph{High-impact authors, with links.} High-impact author embedding,
colored by degree percentile, with links between co-authors displayed as white
line segments.}\label{f-coauthors-impact-edges}
\end{figure}

The embedding has an interesting structure, with many intersecting arcs. These
arcs are likely due to the original distances taking on highly quantized values
(see figure~\ref{f-scholar-distance-CDF}). In
figure~\ref{f-coauthors-impact-edges} we overlay a large fraction of the
original high-impact network's edges onto the embedding (\ie, edges
$(i, j)$ for which $\delta_{ij} = 1$); we show 50,000 edges, sampled uniformly
at random from the original 210,681. The edges trace out the arcs, and connect
adjacent communities of highly collaborative authors.

\begin{figure}
\includegraphics{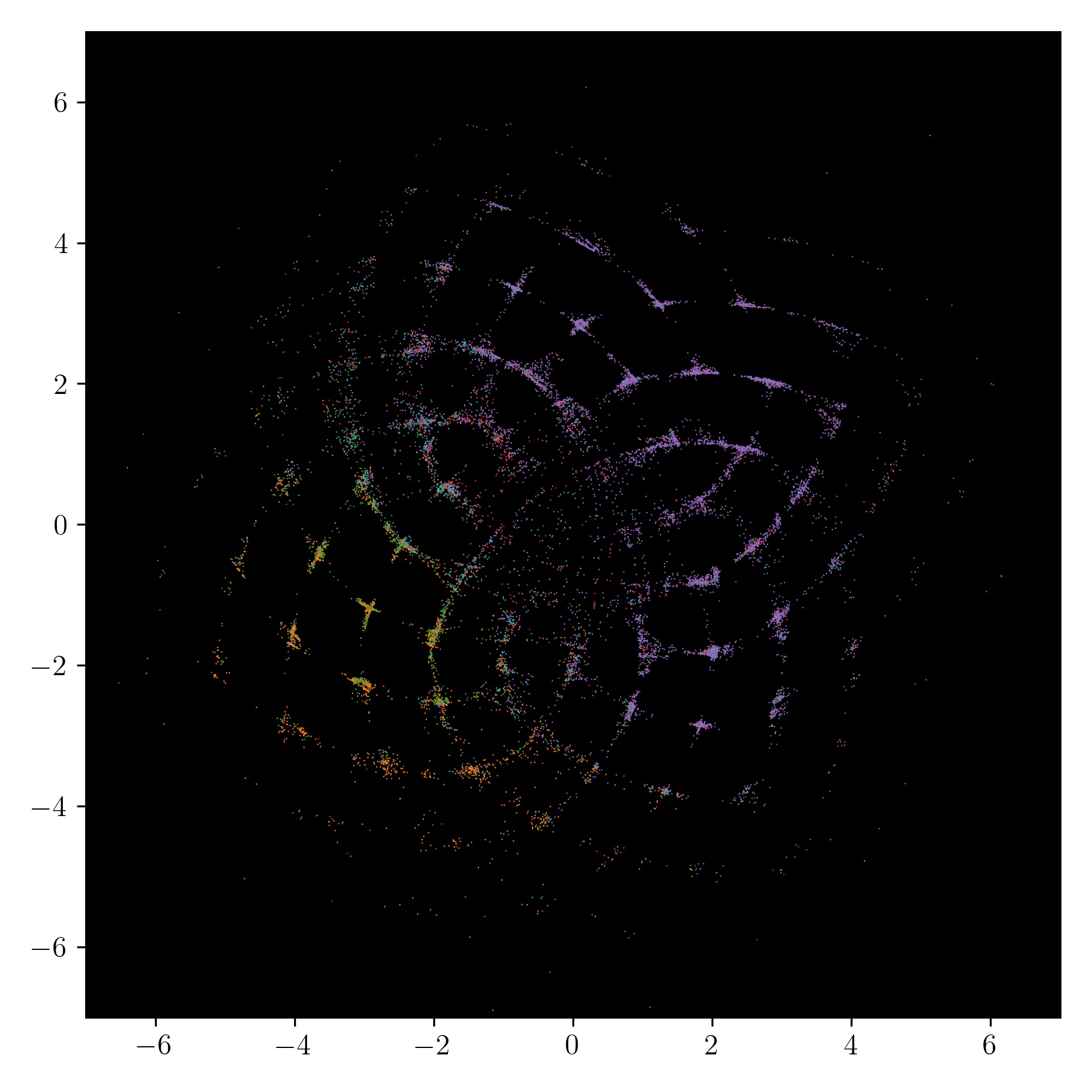}
\caption{\emph{Academic disciplines.} High-impact author embedding,
  colored by academic discipline. The disciplines are biology (\emph{purple}),
  physics (\emph{orange}), electrical engineering (\emph{green}), computer
  science (\emph{cyan}), and artificial intelligence (\emph{red}).
}\label{f-coauthors-topics}
\end{figure}

\paragraph{Academic disciplines.}
Figure~\ref{f-coauthors-topics} shows the embedding vectors colored
by academic discipline (authors that are not tagged with a discipline are
omitted). Purple represents biology, orange represents physics, green
represents electrical engineering, cyan represents computer science, and red
represents artificial intelligence.

Authors are grouped by academic discipline, and similar disciplines
are near each other. Additionally, it turns out nearby curves in an academic
discipline roughly correspond to sub-disciplines. For example, authors in the
purple curves toward the center of the embedding are largely interested in
ecology, while authors in the outer purple curves are more interested in
genomics and medicine.

Most highly-collaborative authors (authors near the center of the
embedding) study
computer science, artificial intelligence, or biology. This is consistent with
a previous study of Google Scholar \cite{chen2017building}, which found that
computer scientists and biologists are disproportionately important in the
co-authorship network, as measured by metrics such as degree and PageRank
\cite{page1999pagerank}.

\clearpage
\subsection{Comparison to other methods}\label{s-scholar-umap-tsne}
For comparison, we embed the high-impact co-authorship networks using
UMAP \cite{mcinnes2018umap} and t-SNE \cite{maaten2008visualizing}, 
using the umap-learn and the scikit-learn
\cite{pedregosa2011scikit} implementations. We give UMAP and t-SNE the distance
matrices containing the graph distances associated with
$\edges_{\text{impact}}$, and we use the default hyper-parameters set by the
software.

The embeddings are shown below in
figure~\ref{f-coauthors-umap-tsne}, colored by degree and academic discipline.
Neither method has preserved the global structure of the
network, but both have organized the embedding vectors by discipline. (As we
have seen, embeddings emphasizing local structure over global structure are
readily produced with PyMDE; in this chapter, however, we chose to emphasize
global structure instead.)

\begin{figure}[h]
\centering
\includegraphics{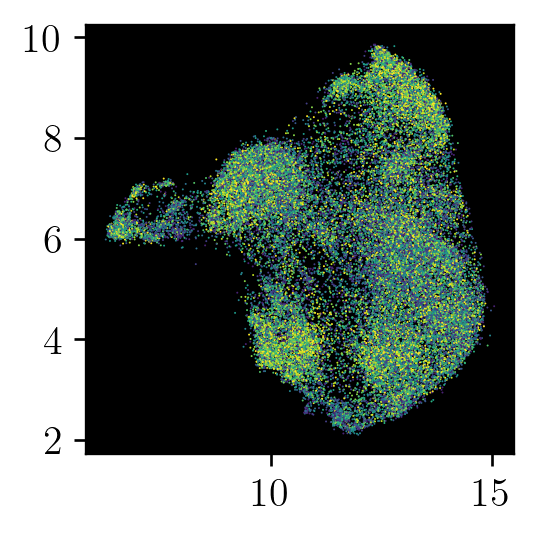}
\includegraphics{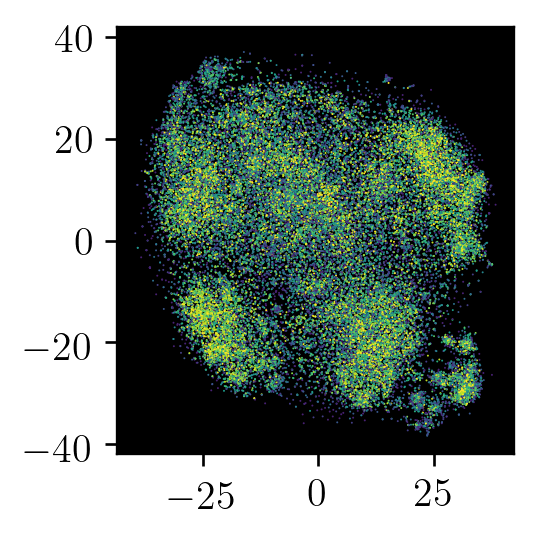} \\
\includegraphics{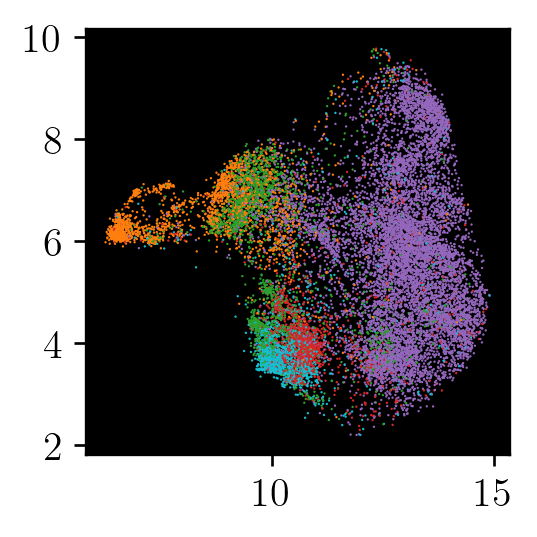}
\includegraphics{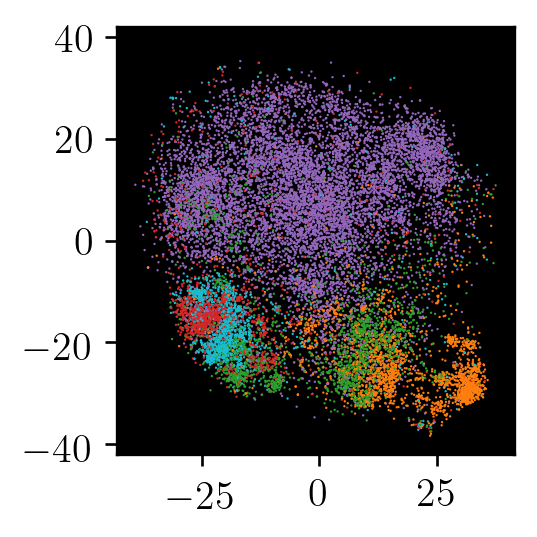}
\caption{UMAP (\emph{left}) and t-SNE (\emph{right})
embeddings of the high-impact author network, colored by degree
(\emph{top}) and academic discipline (\emph{bottom}).}
\label{f-coauthors-umap-tsne}
\end{figure}

\chapter{Counties}\label{c-zip-codes}
In this example, we embed demographic data for 3,220 United States
counties into $\reals^2$.  The data broadly describes the composition of the
counties, and contains information about gender, ethnicity, and income (among
other things).  
To make sure our embeddings make sense, we will use an additional
attribute for each county to validate the embeddings, which is the fraction who voted Democratic in the 2016
election.
(We do not use this attribute in forming the embeddings.)

\section{Data} 
Our raw data comes from the 2013--2017 American Community Survey (ACS)
5-Year Estimates \cite{ACS}, which is a longitudinal survey run by the United
States Census Bureau that records demographic information about
the $n=3,220$ US counties.  We will use 34 of the demographic features
to create our embedding of US counties.

\begin{table}
  \centering
  \begin{tabular}[]{p{0.65\linewidth} p{0.35\linewidth}}
    \toprule
    Feature & Transformation \\
    \midrule
    \midrule
    Number of people & Log-transform \\
    Number of voting age citizens & Log-transform \\        
    Number of men & Normalize \\
    Number of women & Normalize \\    
    Fraction Hispanic & None \\
    Fraction White & None \\
    Fraction Black & None \\
    Fraction Native American or Alaskan & None \\
    Fraction Asian & None \\
    Fraction Native Hawaiian or Pacific Islander & None \\
    Median household income & Log-transform \\
    Median household income standard deviation & Log-transform \\
    Household income per capita & Log-transform \\
    Household income per capita standard deviation & Log-transform \\
    Fraction below the poverty threshold & None \\
    Fraction of children below the poverty threshold & None \\
    Fraction working in management, business, science, or the arts & None \\
    Fraction working in a service job & None \\
    Fraction working in an office or sales job & None \\
    Fraction working in natural resources, construction, and maintenance & None \\
    Fraction working in production, transportation, and material movement & None \\
    Fraction working in the private sector & None \\                
    Fraction working in the public sector & None \\                 
    Fraction working for themselves & None \\    
    Fraction doing unpaid family work & None \\
    Number that have a job (are employed) & Normalize \\                
    Fraction unemployed & None \\
    Fraction commuting to work alone in a car, van, or truck & None \\    
    Fraction carpooling in a car, van, or truck & None \\        
    Fraction commuting via public transportation & None \\            
    Fraction walking to work & None \\            
    Fraction commuting via other means & None \\            
    Fraction working at home & None \\            
    Average commute time in hours & None \\            
    
    \bottomrule
  \end{tabular}
  \caption{List of the features in the ACS data set, 
along with the initial preprocessing transformation we applied.}\label{t-voting-feats}
\end{table}

\section{Preprocessing} 
We carry out some standard
preprocessing of the raw features, listed in table \ref{t-voting-feats}.
For positive features that range over a very wide scale, such as household income,
we apply a log-transform, \ie, replace the raw feature with its logarithm.
We normalize raw features that are counts, \ie, divide them by the county
population.  (We do not transform features that are given as fractions or 
proportions.)
After these transformations, we standardize each feature
to have zero mean and standard deviation one over the counties.

Using these preprocessed features,
we generate a $k$-nearest (Euclidean) neighbor graph, with $k=15$,
which gives a graph $\esim$ with $75,250$ edges.
We then sample, at random, another $75,250$ pairs of counties not in $\esim$,
to obtain a graph of dissimilar counties, $\edis$.
Our final graph $\edges = \esim \cup \edis$ has $150,500$ edges,
corresponding to an average degree of around $47$.
We assign weight $w_{ij} = 2$ for $(i,j) \in \esim$ when counties $i$ and $j$ are both neighbors of each other.  We assign weight $w_{ij} = 1$ for $(i,j) \in \esim$ when $i$ is a neighbor of $j$ but $j$ is not a neighbor of $i$ (or vice versa).  We assign weight $w_{ij} = -1$ for $(i,j) \in \edis$.

\section{Embedding}
\subsection{PCA embedding}
\begin{figure}
\centering
\includegraphics{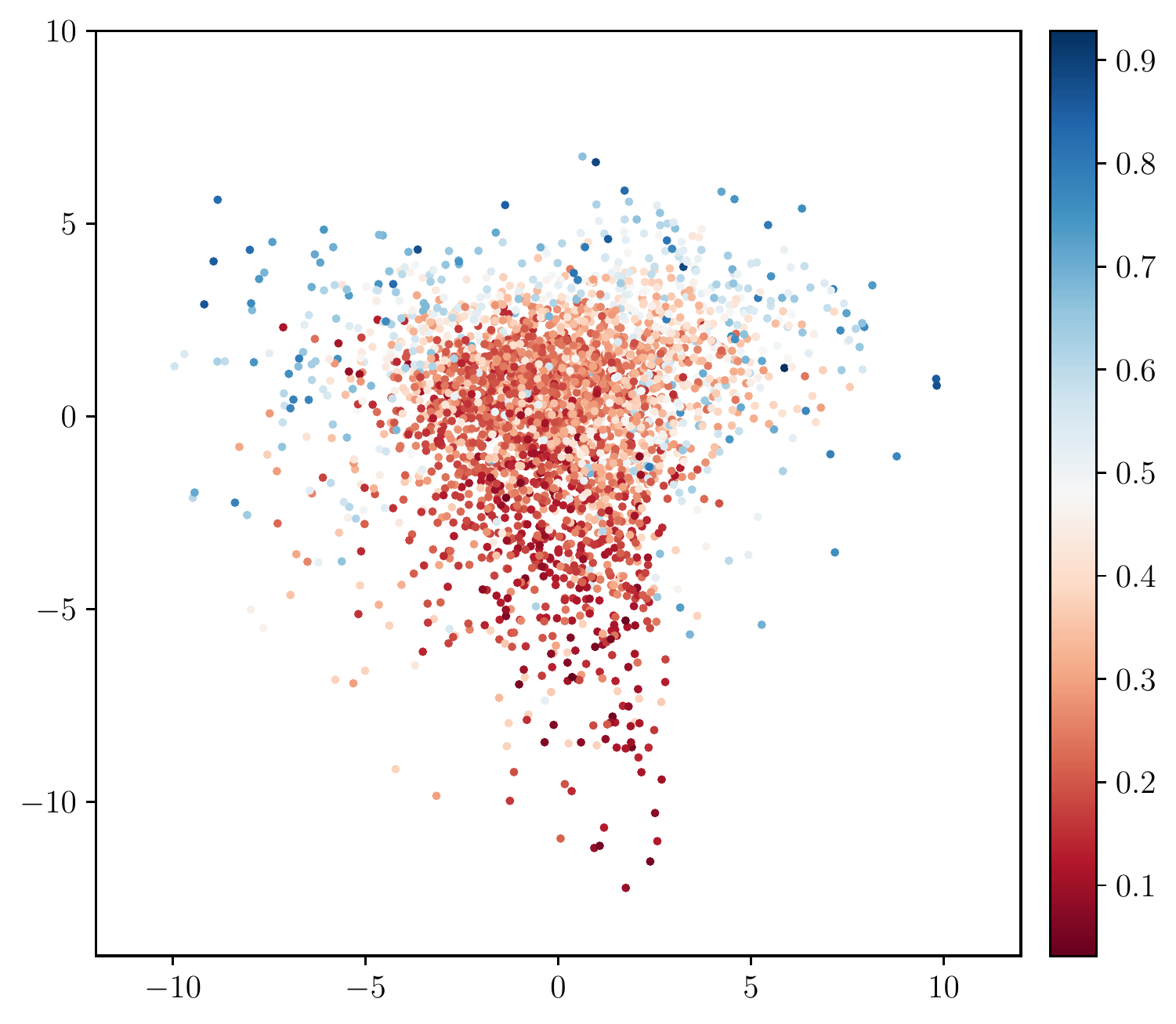} 
\caption{PCA
embedding of the election data, colored by the fraction of voters in each county
who voted Democratic in the 2016 United States presidential election, from red
(0\%) to blue (100\%).} \label{f-voting-pca}
\end{figure}

We consider PCA first.  The PCA embedding for $m=2$ 
is shown in figure \ref{f-voting-pca},
with each county shown as a colored dot, with colors depending on the fraction
of voters who voted Democratic in 2016, from red (0\%) to blue (100\%).
The embedding clearly captures some of the structure of the voting patterns,
but we can see many instances where counties with
very different voting appear near each other in the embedding.
(The blue counties are urban, and far more populous than the many
rural red counties seen in the embedding.  The total number of people who voted
Democratic exceeded those who voted Republican.)  

\subsection{Unconstrained embedding}
Next we solve an unconstrained (centered) MDE problem based on the graph
$\edges = \esim \cup \edis$. We use the log-one-plus penalty
\eqref{eq-log1p-penalty} ($\alpha=2$) for $\psim$ and the logarithmic penalty
\eqref{eq-log-penalty} for $\pdis$ ($\alpha=0.5$). We choose embedding
dimension $m=2$.

\begin{figure}
\centering  
\includegraphics{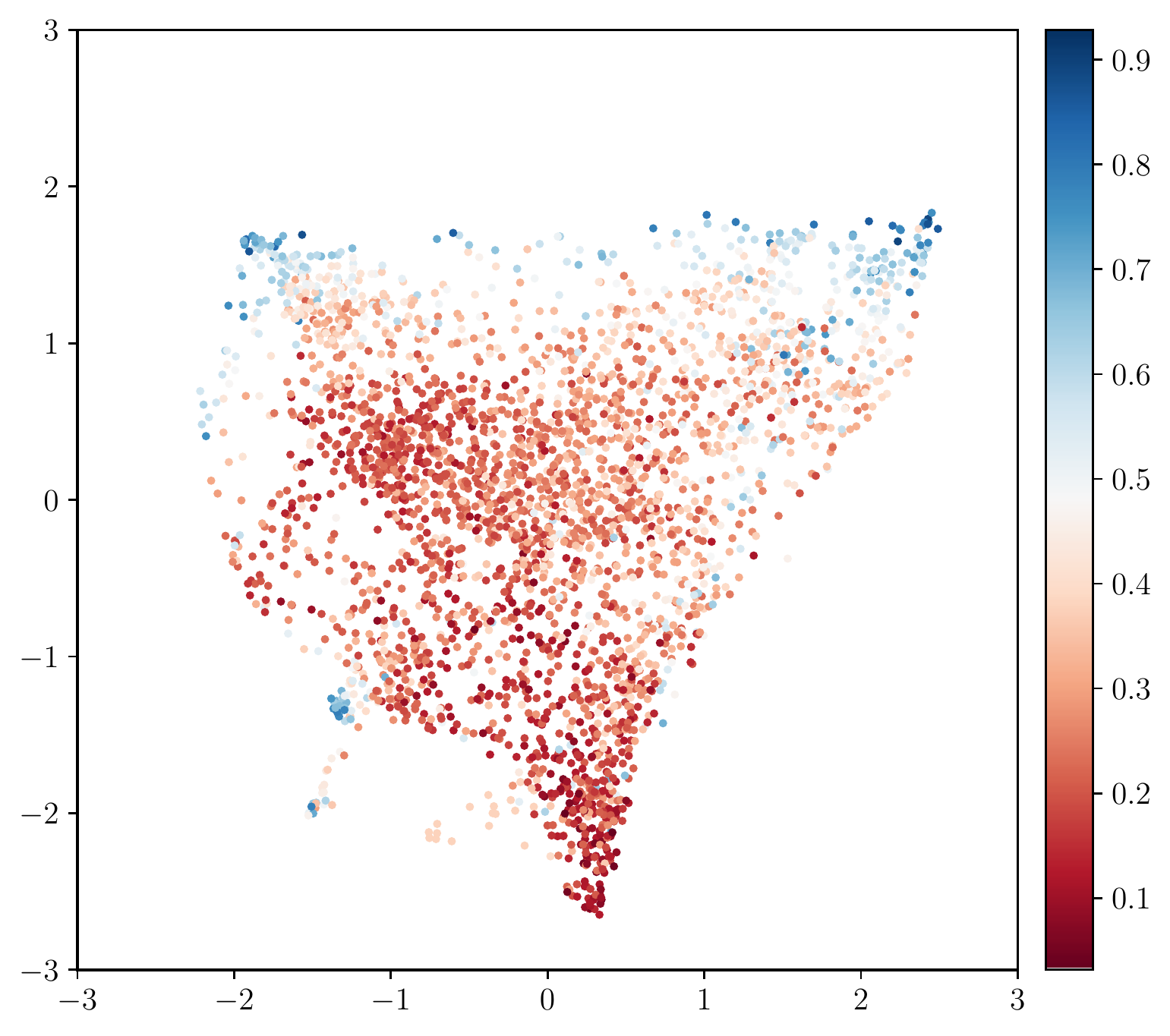}
\caption{\emph{Embedding.} A centered embedding of the
election data, with log-one-plus attractive and
logarithmic repulsive penalty functions.} \label{f-voting-log1p}
\end{figure}

\begin{figure}
\centering
\includegraphics{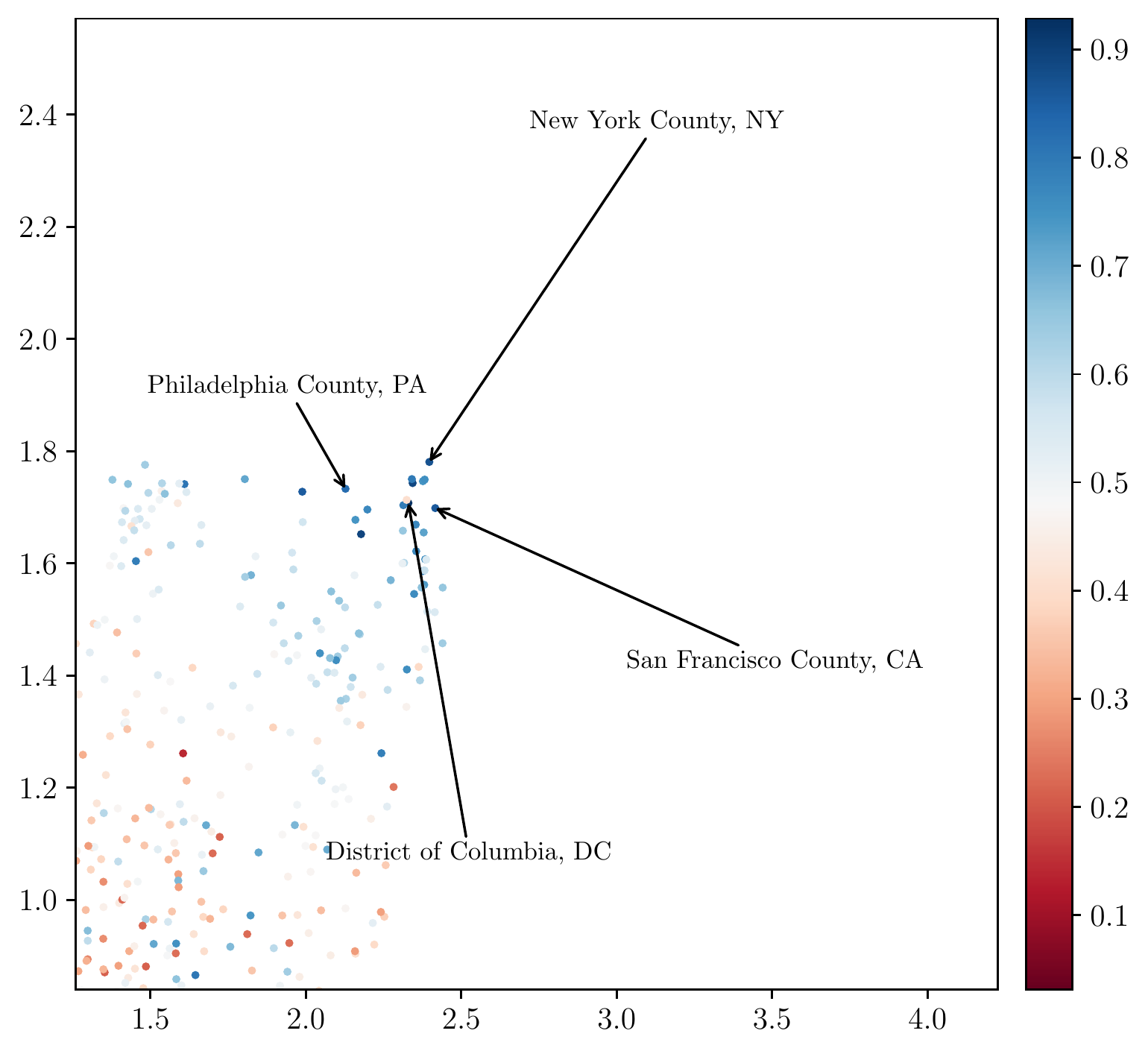}
\caption{\emph{Upper right corner.}  A closer look at the upper right corner of
the embedding shown in figure \ref{f-voting-log1p}.  
Several densely populated urban counties appear.} \label{f-voting-dems}
\end{figure}

\begin{figure}
\centering  
\includegraphics{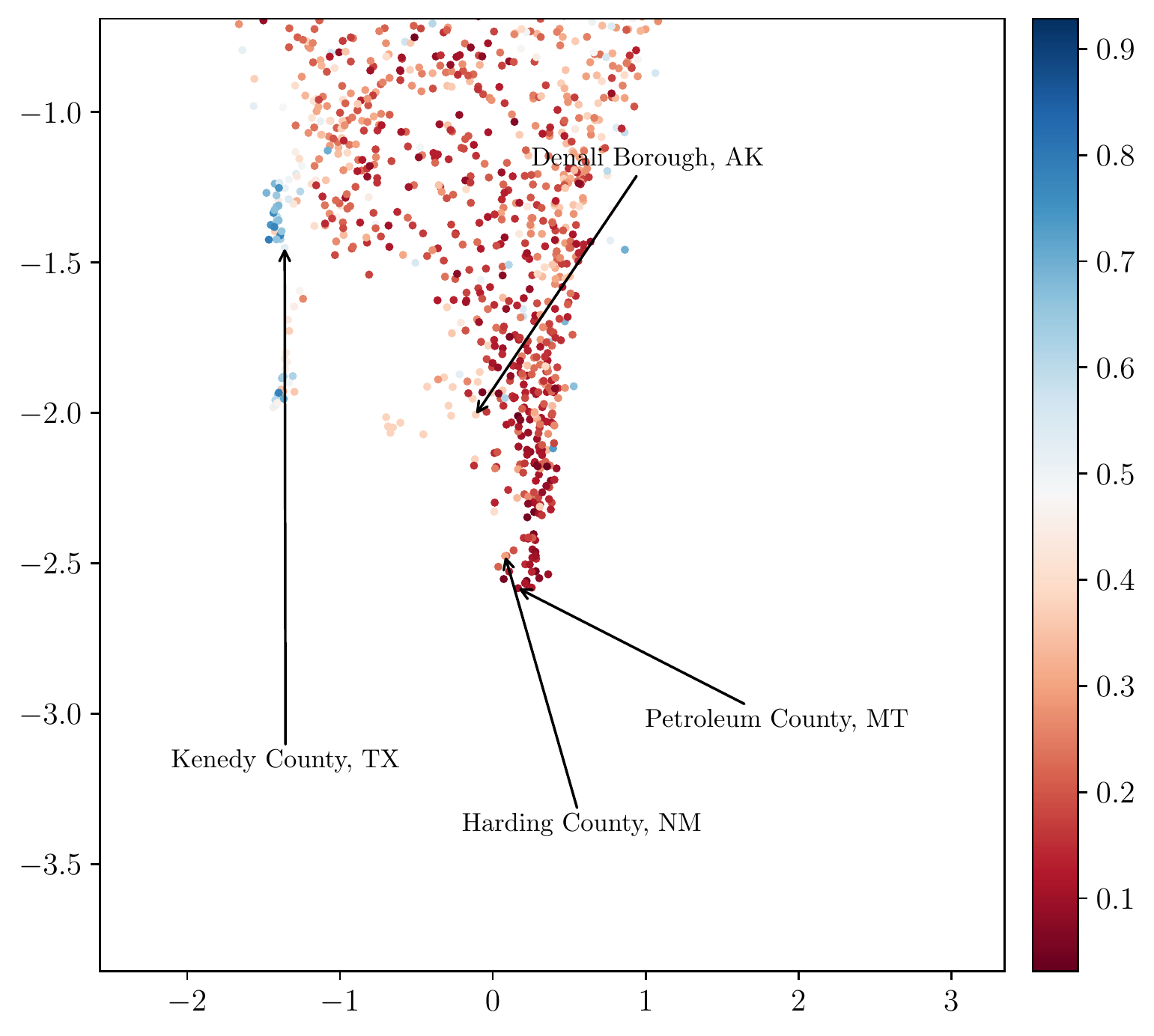}
\caption{\emph{Bottom.}  A closer look at the bottom of
the embedding shown in figure \ref{f-voting-log1p}.  Several
sparsely populated rural counties appear.} \label{f-voting-repubs}
\end{figure}

Figure \ref{f-voting-log1p} shows the results.  Compared to the PCA embedding,
we see a better separation of regions corresponding to heavily Democratic, 
heavily Republican, and mixed voting.  
Figures \ref{f-voting-dems} and
\ref{f-voting-repubs} take a closer look at the upper right and bottom of the plot, labeling some representative counties.  
We can also see a few swing counties, \ie,
counties that historically vote Democratic, but recently lean Republican (or
vice versa), near the strongly Democratic (or Republican) counties; 
these counties are colored light red (or blue), because they have weak majorities.  

\begin{figure}
\centering
\includegraphics{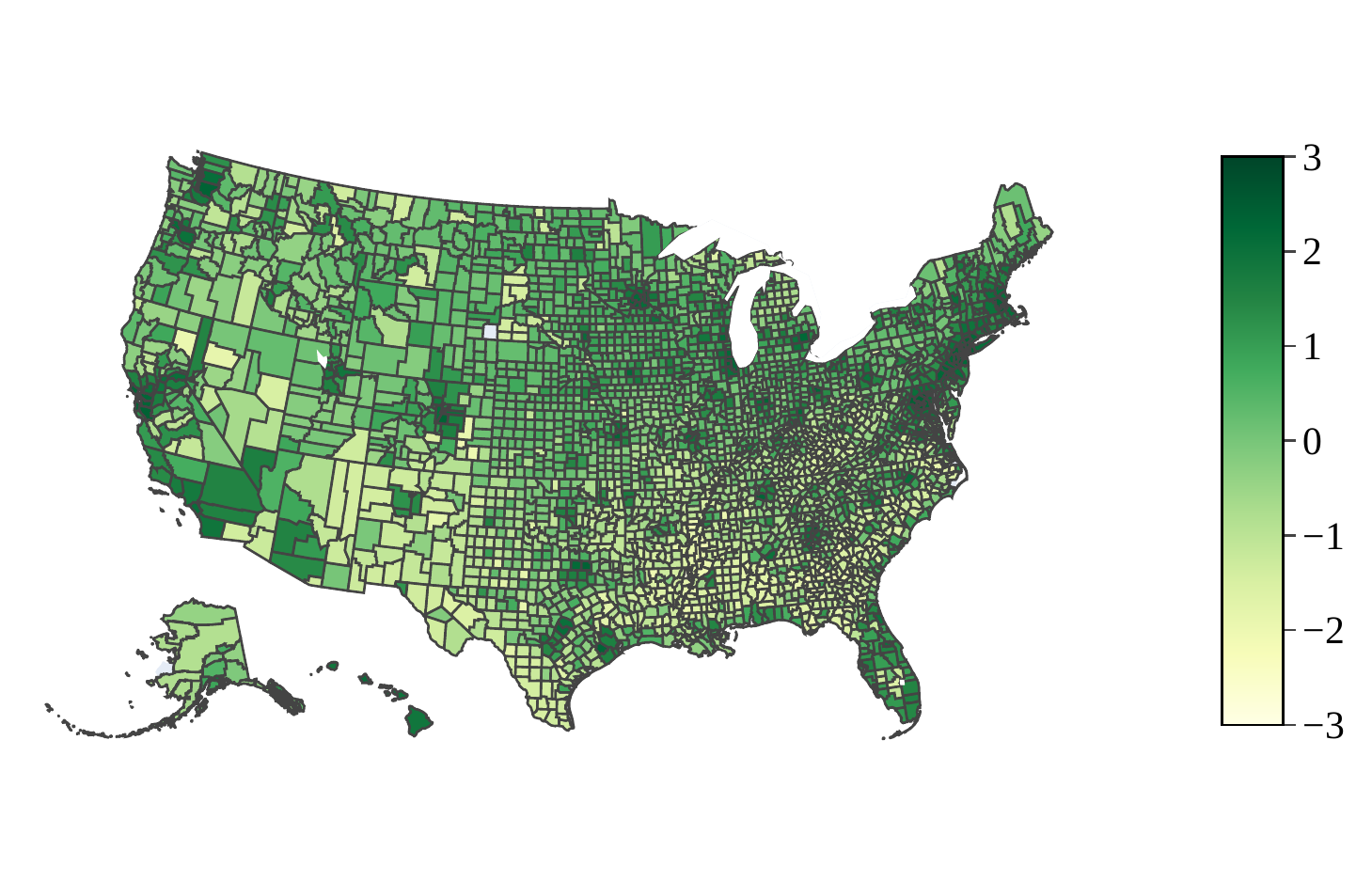}

\includegraphics{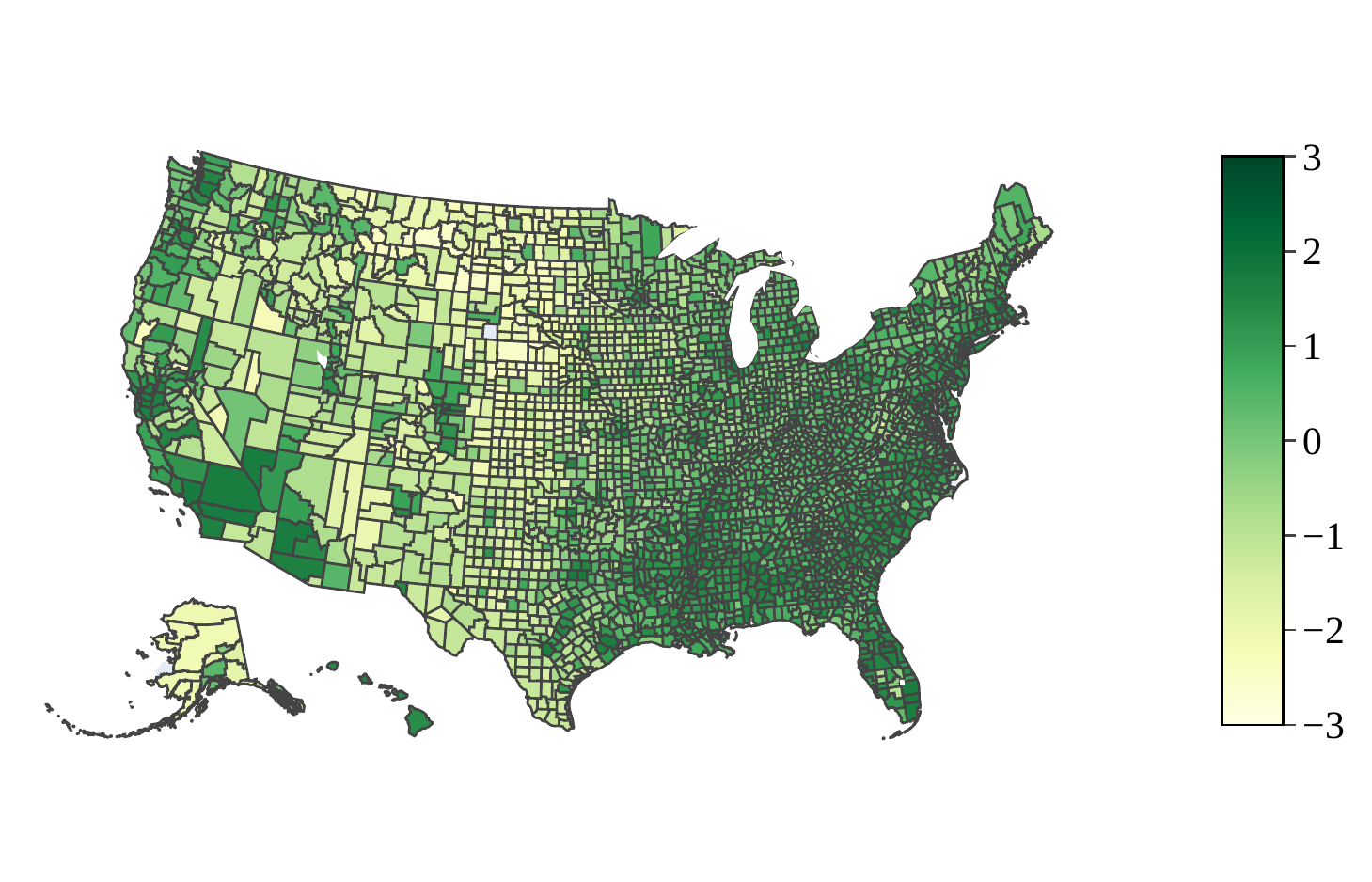}
\caption{\emph{Embedding by coordinate.}  Counties colored by their
first (\emph{top}) and second (\emph{bottom}) coordinate of the embedding from figure \ref{f-voting-log1p}.}
\label{f-voting-usa-coords}
\end{figure}

\begin{figure}
\centering
\includegraphics{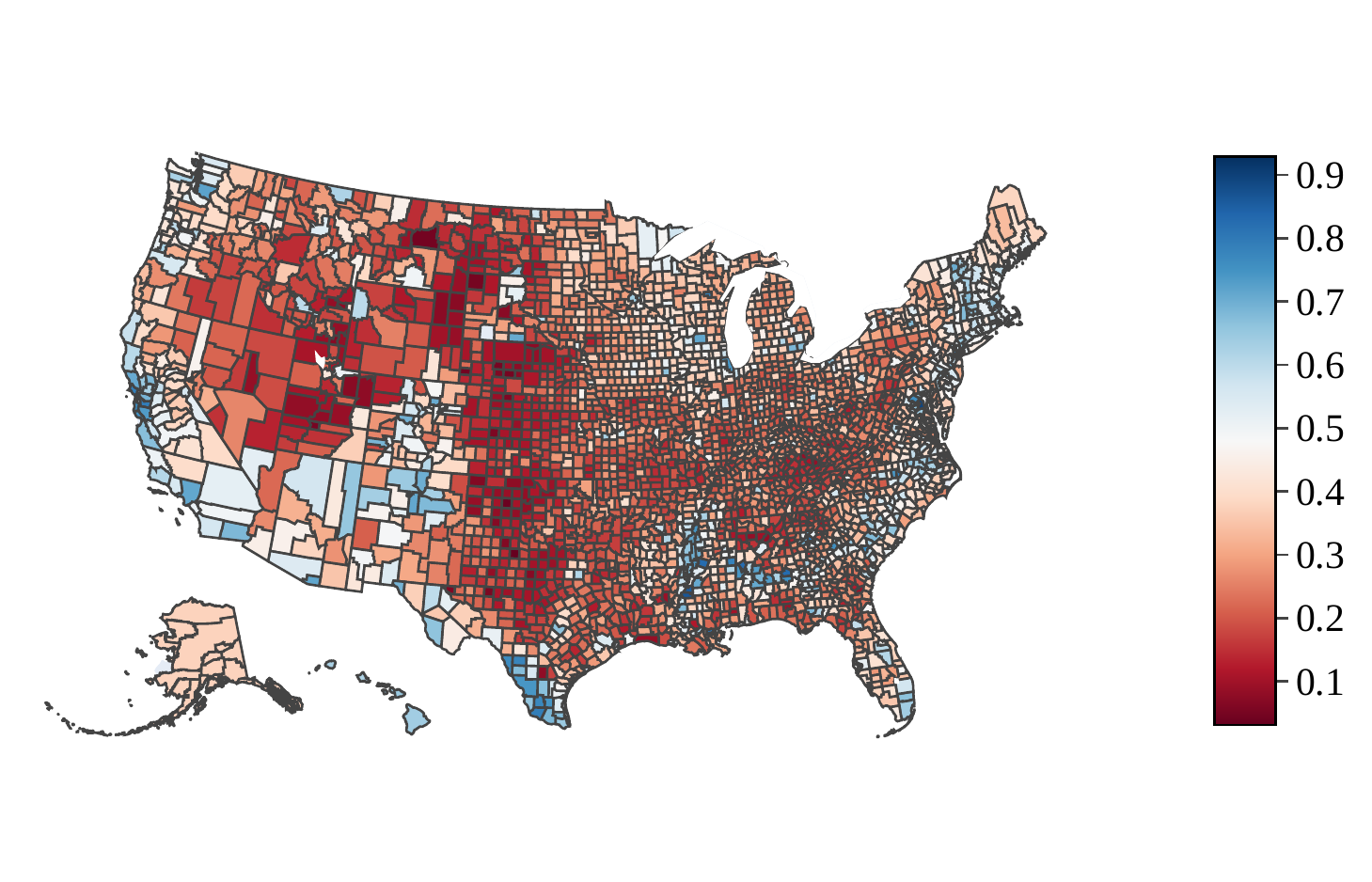}
\caption{\emph{Election data.}  Counties colored by the fraction of people
voting for the Democratic candidate in the 2016 election, from red (0\%) to blue
(100\%).  
(The urban blue counties are far more densely populated than the rural 
red counties; in total there were more Democratic than Republican votes.)}
\label{f-voting-usa-true}
\end{figure}

We give two alternative visualizations of the embedding in figure
\ref{f-voting-usa-coords}.  
In these visualizations, we color each county
according to its value in each of the two embedding coordinates.
In figure \ref{f-voting-usa-true}, we show the counties
colored with the fraction of people who actually voted Democratic in the 2016 election,
as a point of reference.  

\clearpage
\subsection{Comparison to other methods}
\begin{figure}
\centering
\includegraphics{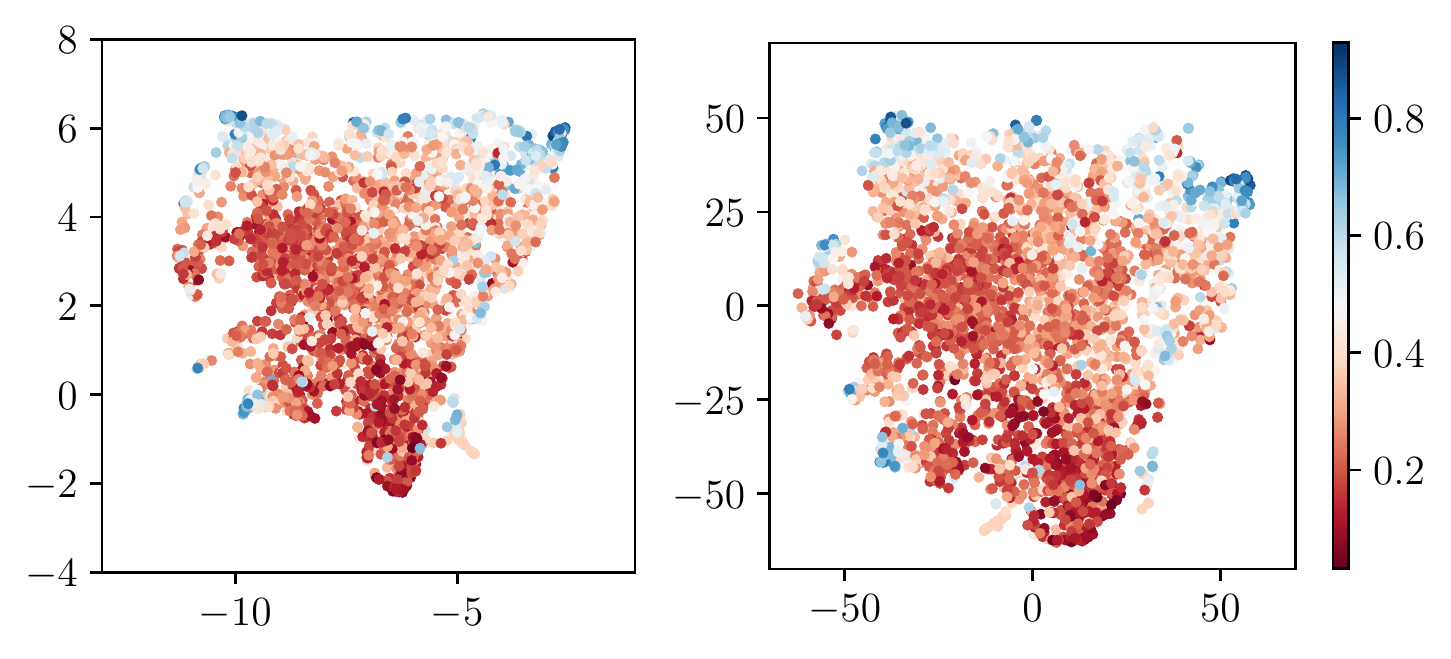}
\caption{UMAP \emph{(left)} and t-SNE \emph{(right)} embeddings of the ACS data.}
\label{f-voting-umap-tsne}
\end{figure}

We compare the embeddings generated by UMAP and t-SNE, shown in figure
\ref{f-voting-umap-tsne}.  To make it easier to compare the embeddings, we
orthogonally transformed these two embeddings with respect to the unconstrained
embedding (as described in \S\ref{s-orth-procrustes}), before plotting them.
The embeddings broadly resemble the unconstrained embedding, though they appear to
be a little bit more dispersed.

\chapter{Population Genetics}\label{c-pop-gen}
\begin{figure} \includegraphics{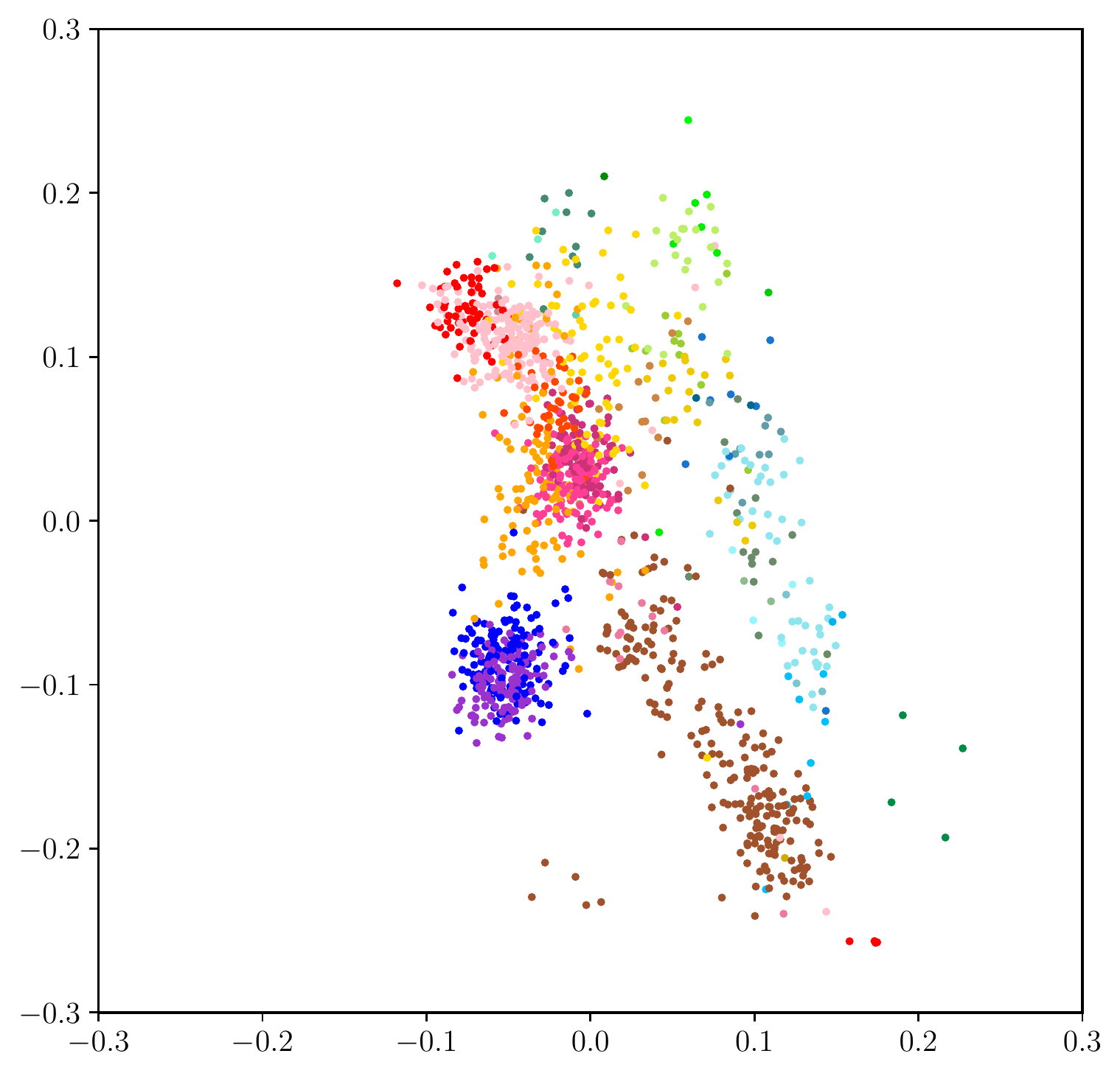}
\caption{\emph{PCA embedding of population genetics data.}  
Points are colored by country using the legend shown in the map 
of Europe in figure~\ref{f-popgen-map}.}
\label{f-popgen-clean-pca}
\end{figure}

\begin{figure}
\centering
\includegraphics{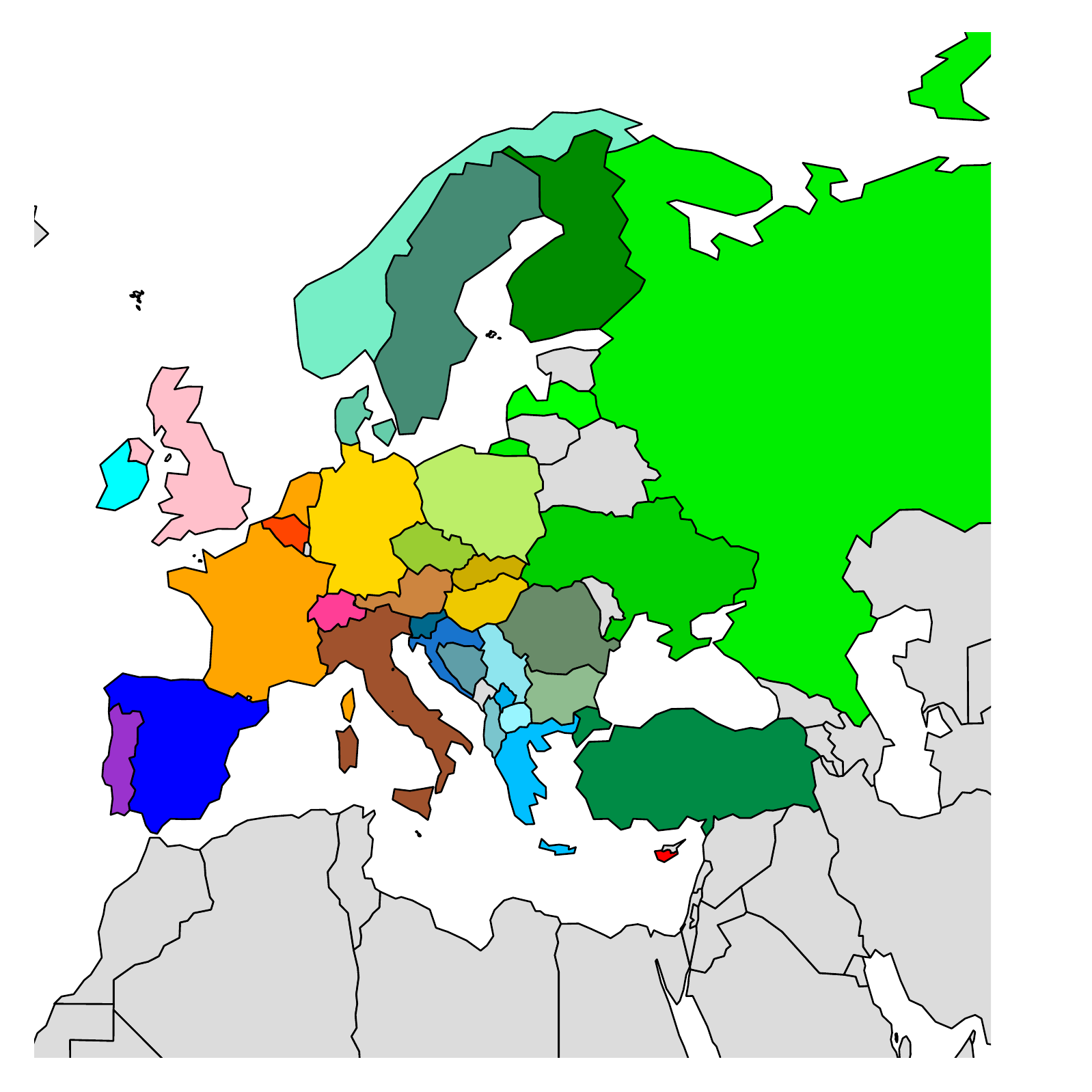}
\caption{Map of Europe, showing the country color legend for the 
population genetics data,
taken from Novembre et al.; countries with no data points are shown in gray.}
\label{f-popgen-map}
\end{figure}

Our next example is from population genetics.
We consider the well-known study carried out by Novembre et
al.~\cite{novembre2008genes} of high-dimensional single nucleotide polymorphism
(SNP) data associated with 3,000 individuals thought to be of European ancestry.
After carefully cleaning the data in order to remove outliers, the authors
embed the data into $\reals^2$ via PCA, obtaining the embedding
shown in figure \ref{f-popgen-clean-pca}.   Interestingly, the embedding bears a
striking resemblance to the map of Europe, shown in
figure \ref{f-popgen-map},
suggesting a close relationship between geography and genetic variation.
(In these figures we use the same color legend used in the original 
Novembre et al.\ paper.)
This resemblance does not emerge if the outliers are not first removed
from the original data.

Follow-on work by Diakonikolas et al.~\cite{diakonikolas2017being}
proposed a method for robust dimensionality reduction, showing good results on a
version of the data from the original Novembre et al.\ study 
that was intentionally corrupted with some bad data.
After appropriately tuning a number of parameters, they were able to 
recover the resemblance to a map of Europe despite the bad data.
We will explore this issue of robustness to poor data using distortion 
functions that are robust, \ie, allow for some pairs of similar items to have 
large distance.

\section{Data}
We create two sets of data: one clean, and one intentionally corrupted,
following \cite{diakonikolas2017being}.
We follow the same experiment setup as in
\cite{novembre2008genes,diakonikolas2017being}, and work with the SNP data
associated with 3,000 individuals thought to be of European ancestry, coming from the
Population Reference Sample project \citep{nelson2008population}. Novembre et
al.~and Diakonikolas et al.~both start off by pruning the original data set down
to the 1,387 individuals most likely to have European ancestry (explained in the
supplementary material of \cite{novembre2008genes}),
and then project the data onto its top 20 principal
components, obtaining the (clean) samples $y_i \in \reals^{20}$,
$i=1,\ldots,1387$.  

To create a corrupted set of data, we follow the method of
Diakonikolas et al., who randomly rotate the data and then inject $154$ additional
points $y_i$, $i=1388, \ldots, 1541$, 
where the first ten entries of the $y_i$ are i.i.d.~following a
discrete uniform distribution on $\{0,1,2\}$, and the last ten entries of the
$y_i$ are i.i.d.~following a discrete uniform distribution on $\{1/12,1/8\}$.
Thus the clean data is $y_1, \ldots, y_{1387}$, and the corrupted
data is $y_{1388}, \ldots, y_{1541}$.

We also have the country of origin for each of the data points,
given as one of 34 European countries.  We use this attribute to 
check our embeddings, but not to construct our embeddings.
In plots, we color the synthesized (corrupted) data points black.

\section{Preprocessing}
For both the clean and corrupted raw data sets we construct a set of pairs
of similar people $\esim$ as the $k$-nearest (Euclidean) neighbors, with $k=15$.
For the clean data, we have $|\esim|=16,375$, and for the corrupted
data we have $|\esim| = 17,842$.
We also treat the remaining pairs of people as dissimilar, so that $|\edis|=944,816$ for the clean data, and $|\edis|=1,168,728$ for the corrupted data.
We use weights $w_{ij} = 2$ for $(i,j) \in \esim$ when individuals $i$ and $j$ are both neighbors of each other.  We use weights $w_{ij} = 1$ for $(i,j) \in \esim$ when $i$ is a neighbor of $j$ but $j$ is not a neighbor of $i$ (or vice versa).  We use weights $w_{ij} = -1$ for $(i,j) \in \edis$.

\section{Embedding}

\subsection{PCA embedding}
\begin{figure}
\centering
\includegraphics{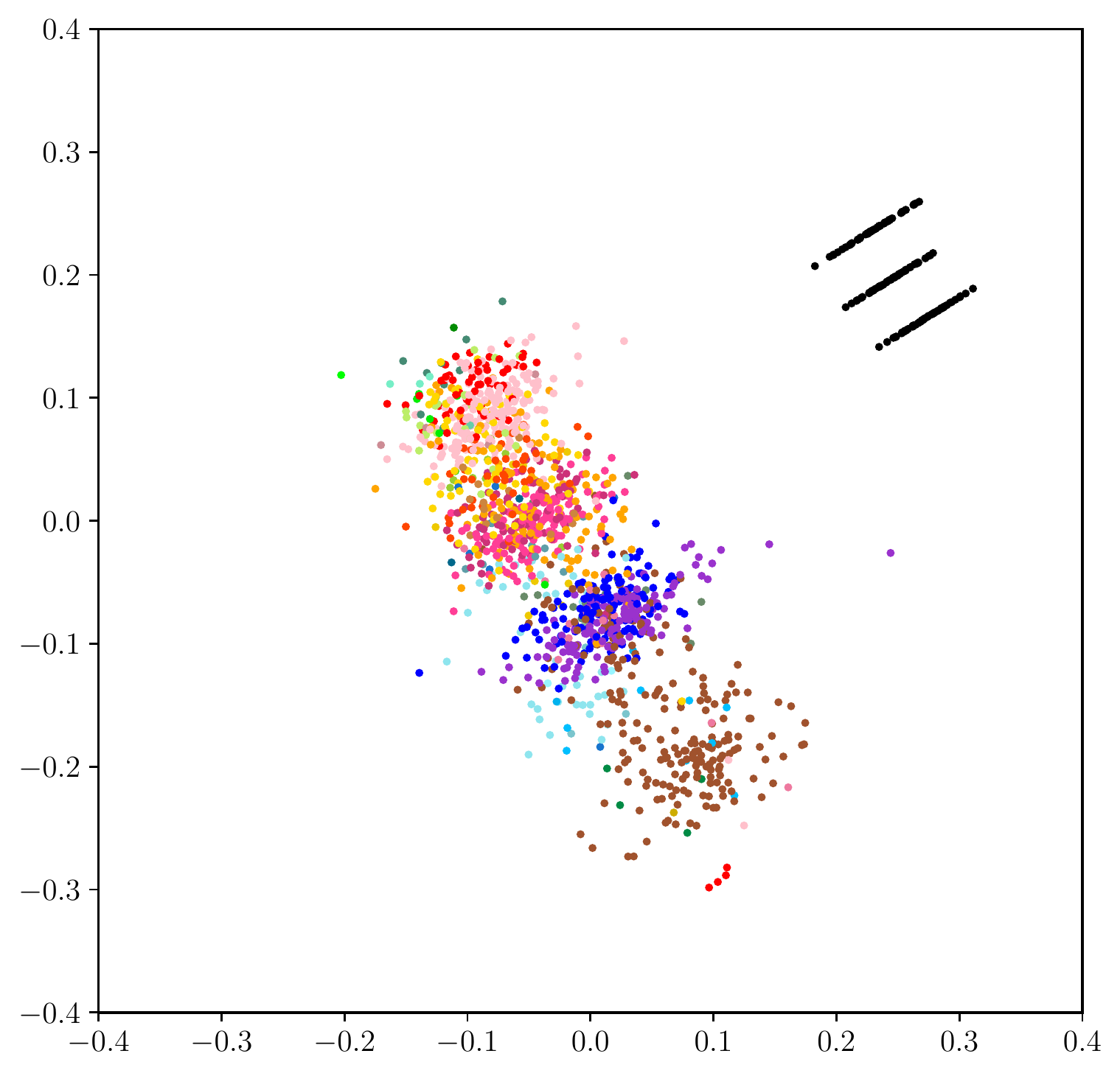}
\caption{\emph{PCA embedding of corrupted data.}
The resemblance to a map of Europe is substantially reduced.}
\label{f-popgen-pca}
\end{figure}

\begin{figure}
\centering
\includegraphics{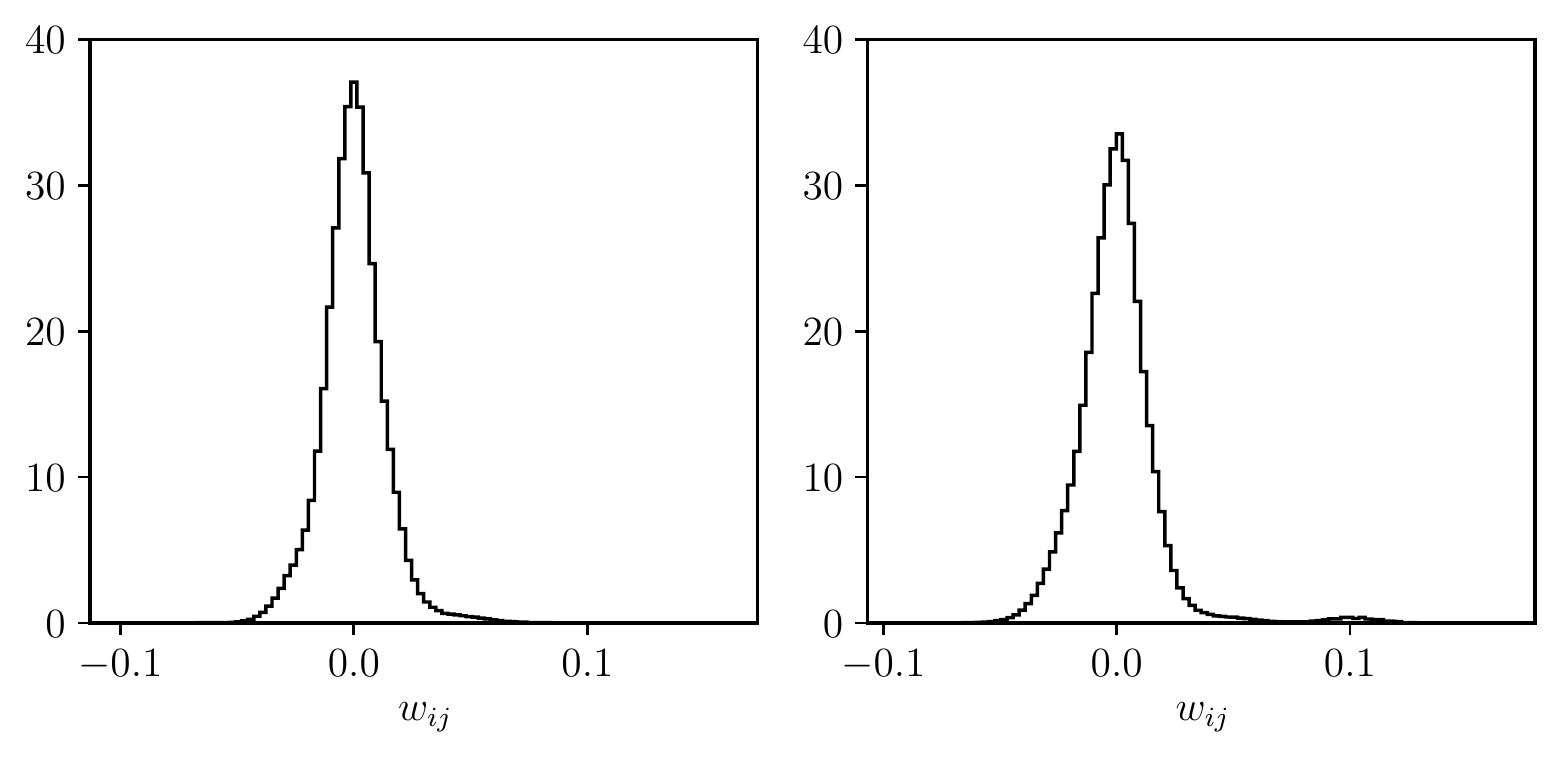}
\caption{\emph{PCA weights.}
Distribution of the weights used by PCA, when expressed as an MDE problem, for the clean \emph{(left)} and corrupted \emph{(right)} data.}
\label{f-popgen-hist}
\end{figure}

We start with PCA, exactly following Novembre et al.,
which on the clean data 
results in the embedding shown in figure~\ref{f-popgen-clean-pca}.  In the language of 
this monograph, this embedding uses quadratic distortion for 
$(i,j) \in \mathcal E$, weights $w_{ij} = y_i^T y_j$ (after centering the data), and a standardization constraint.  The distribution of the weights is shown in figure \ref{f-popgen-hist}.  
When PCA embedding is used on the corrupted data, we obtain the 
embedding in figure~\ref{f-popgen-pca}.  
In this embedding only some of the resemblance to the map of Europe is preserved.
The synthesized points are embedded on the right side.  (If those points are manually
removed, and we PCA run again, we recover the original embedding.)

\subsection{Unconstrained embeddings}
\begin{figure}
\centering
\includegraphics{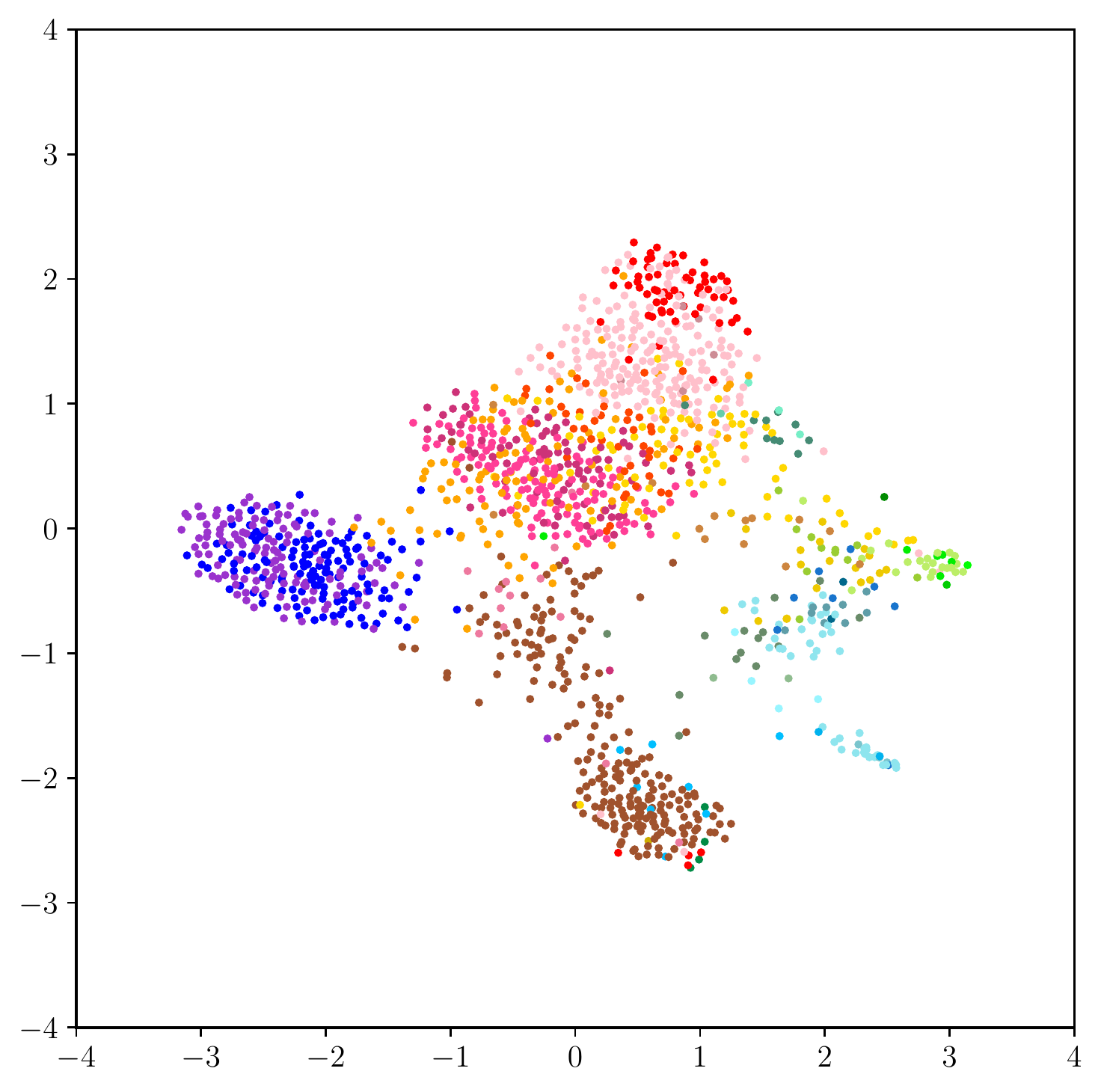}
\caption{\emph{Embedding of clean data.}  The embedding resembles a map of Europe.}
\label{f-popgen-clean-hub}
\end{figure}

\begin{figure}
\centering
\includegraphics{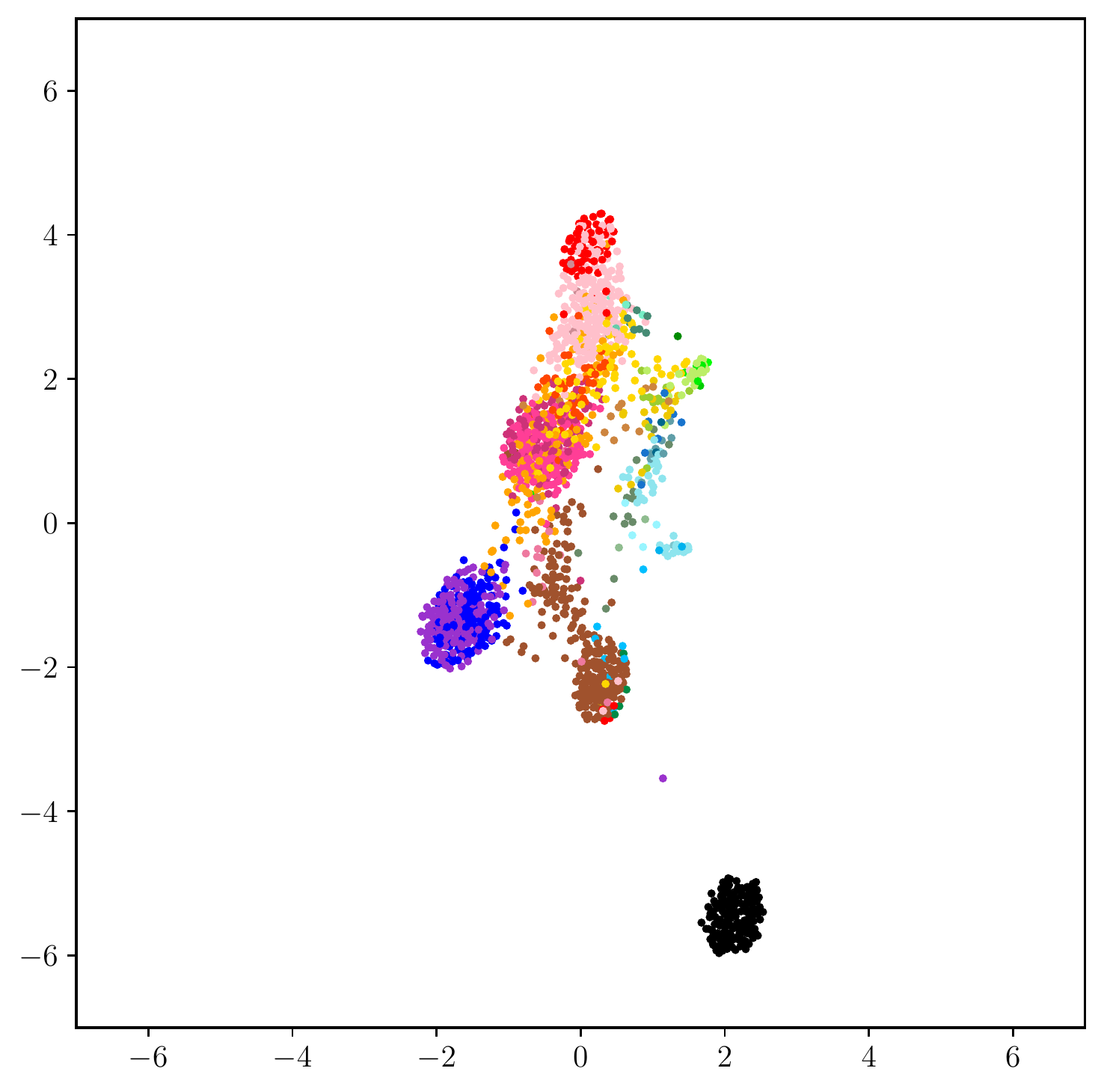}
\caption{\emph{Embedding of corrupted data.}  The embedding is able to recover the similarity to the map of Europe.}
\label{f-popgen-hub}
\end{figure}

We embed both the clean and corrupted data unconstrained (centered), 
based on the graph $\edges = \esim \cup \edis$.  We use a Huber attractive penalty
\eqref{eq-huber-penalty}, and a logarithmic repulsive penalty
\eqref{eq-log-penalty} ($\alpha=1$) to enforce spreading. 
The embedding of the clean data is shown in figure \ref{f-popgen-clean-hub},
and for the corrupted data in figure \ref{f-popgen-hub}.  We can see that this
embedding is able to recover the similarity to the map of Europe, despite the
corrupted data.  In fact, this embedding appears virtually identical to the
embedding of the clean data in figure \ref{f-popgen-clean-hub}.

\subsection{Comparison to other methods}
\begin{figure}
\centering
\includegraphics{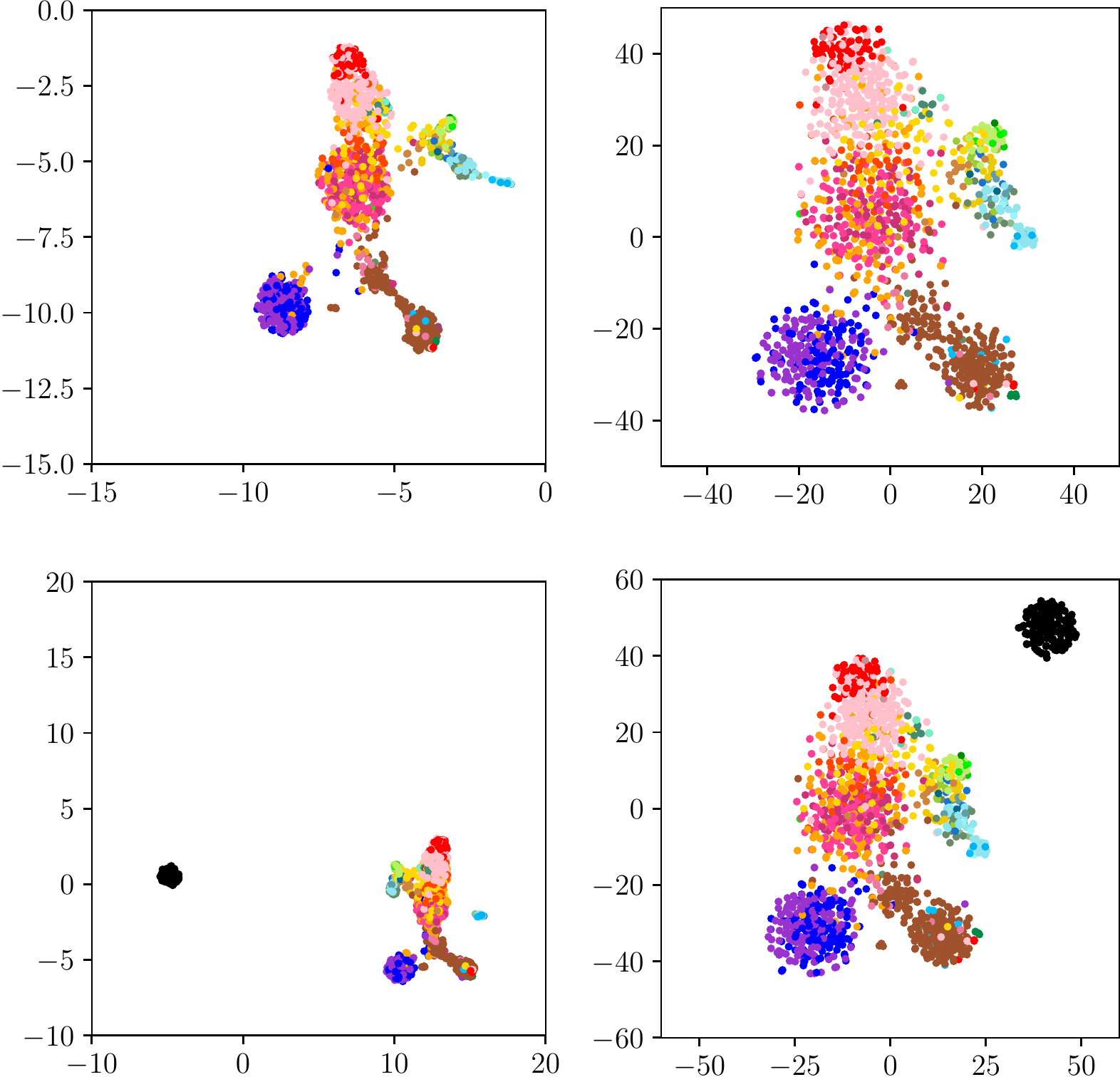}
\caption{UMAP \emph{(left)} and t-SNE \emph{(right)} embeddings of clean \emph{(top)}
and corrupted data \emph{(bottom)}.}
\label{f-tsne-umap-corrupted}
\end{figure}

We show the UMAP and t-SNE embeddings on the clean and corrupted data in figure
\ref{f-tsne-umap-corrupted}.  These embeddings were scaled and 
orthogonally transformed
with respect to the map given in figure \ref{f-popgen-map}, following the
method described in \S\ref{s-orth-procrustes}.  The embeddings
with the corrupted data are better than the PCA embedding.

\chapter{Single-Cell Genomics}\label{c-scrna-seq}
For our last example we embed several thousand single-cell mRNA (scRNA)
transcriptomes. We use data that accompanies the recent publication
\cite{wilk2020single}; these data are sequences of human peripheral blood
mononuclear cells (PBMCs), collected from seven patients afflicted with severe
COVID-19 infections and six healthy controls. We embed into $\reals^3$
and visualize the embedding vectors, finding that the vectors are organized by
donor health status and cell type.

\section{Data}
The dataset from \cite{wilk2020single} includes $n=44,721$ scRNA
transcriptomes of PBMCs, with 16,627 cells from healthy donors and 28,094 from
donors infected with COVID-19. Each cell is represented by a sparse vector in
$\reals^{26361}$, and each component gives the expression level of a gene. The
authors of \cite{wilk2020single} projected the data matrix onto its top $50$
principal components, yielding a matrix $Y \in \reals^{44721 \times 50}$ 
which they further studied.

In constructing our embedding, we use only the reduced gene expression data,
\ie, the entries of $Y$. Each cell is tagged with many held-out
attributes, and we use two of these to informally
validate our embedding. The first attribute says whether a cell came from a
donor infected with COVID-19, and the second gives a classification of the
cells into (manually labeled) sub-families.

\section{Preprocessing}
We use distortion functions derived from weights. From the rows of $Y$
we construct a $k$-nearest (Euclidean) neighbor graph, with $k=15$;
this gives a graph $\esim$ with $515,376$ edges. We then sample,
uniformly at random, an additional $|\esim|$ pairs not in $\esim$ to obtain a
graph $\edis$ of dissimilar cells. The final graph $\edges = \esim \cup \edis$
has $p = 1,030,752$ edges. For the weights, we choose $w_{ij} = +2$ if $i$ and
$j$ are both neighbors of each other; $w_{ij} = +1$ if $i$ is a neighbor of $j$
but $j$ is not a neighbor of $i$ (or vice versa); and $w_{ij} = -1$ for $(i, j)
\in \edis$.

\section{Embedding}
\begin{figure}
\centering
\includegraphics{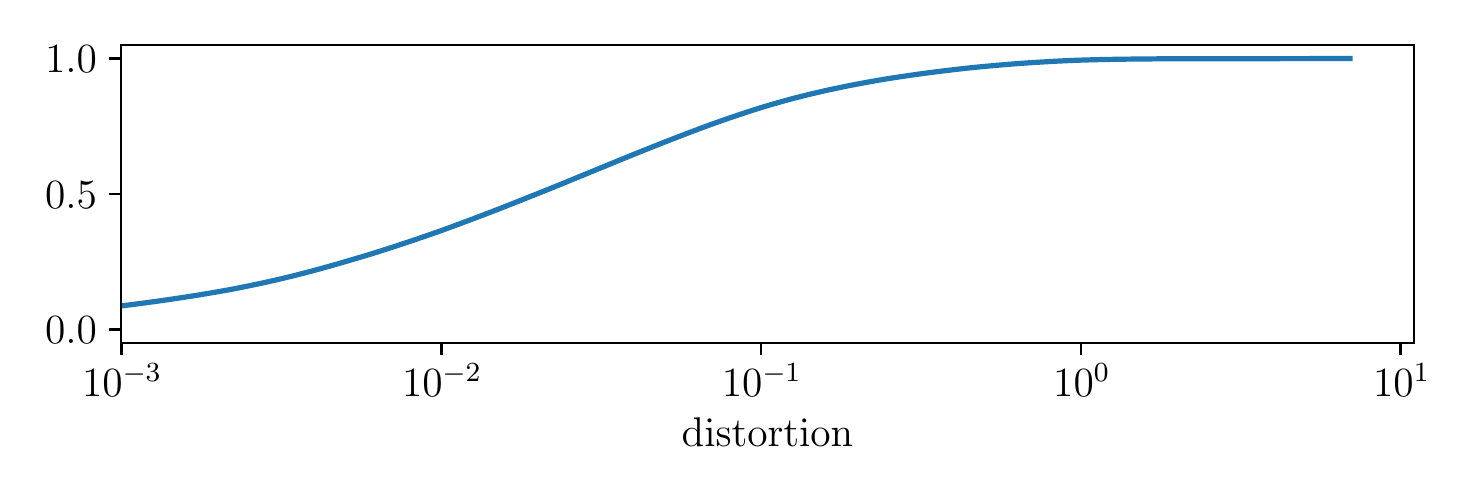}
\caption{CDF of distortions for the embedding of scRNA data.}\label{f-scrna-distortions}
\end{figure}

We form a standardized MDE problem with embedding dimension $m=3$, based
on the graph $\edges = \esim \cup \edis$. We use the log-one-plus 
penalty \eqref{eq-log1p-penalty} ($\alpha = 1.5$) for $\psim$ and the
logarithmic penalty \eqref{eq-log-penalty} ($\alpha = 1$) for $\pdis$.

\begin{figure}
\centering
\includegraphics{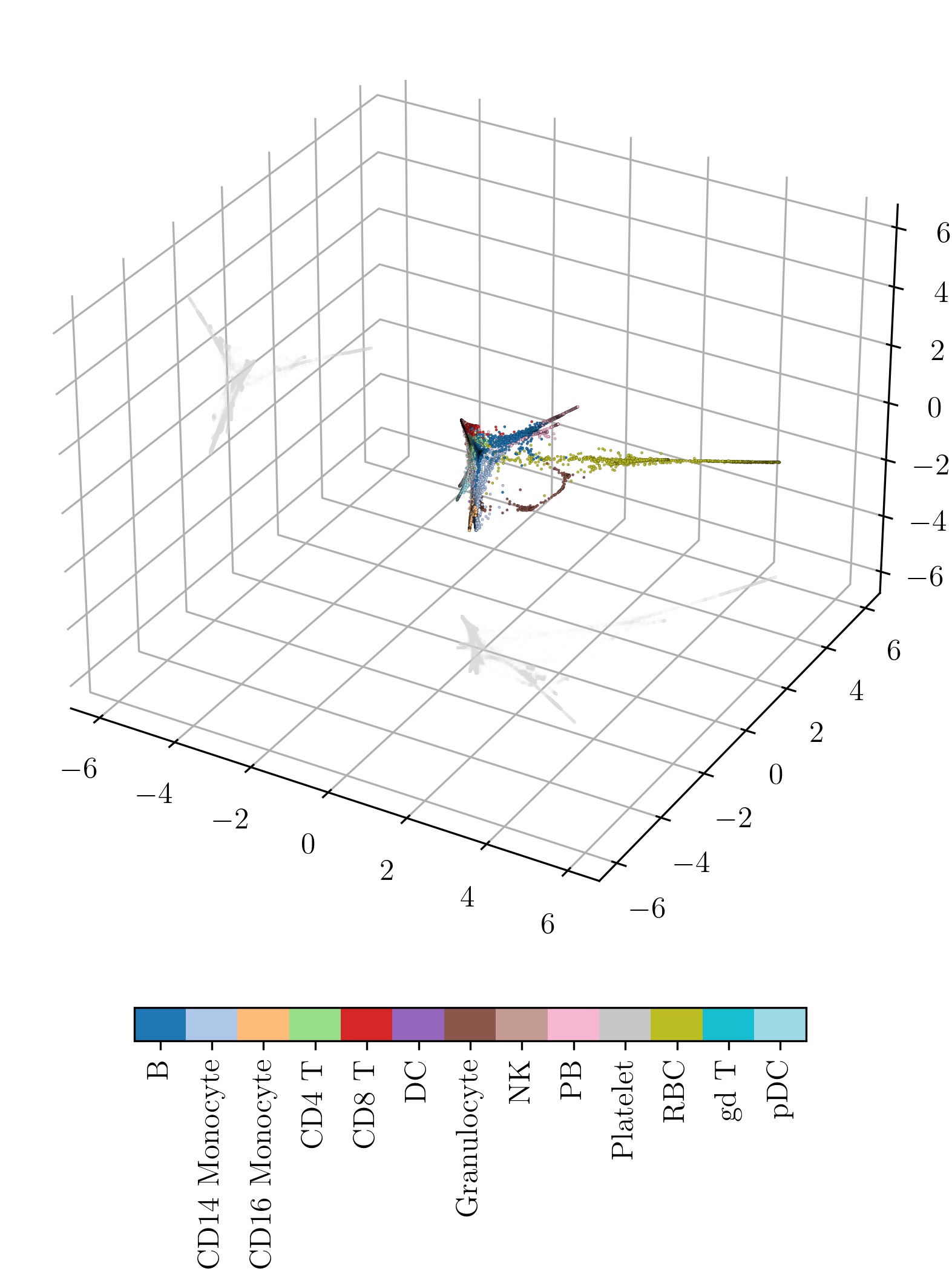}
\caption{\emph{Embedding based on} $\esim$. An embedding of scRNA
  transcriptomes of PBMCs from patients with severe COVID-19
  infections and healthy controls \cite{wilk2020single}, obtained by solving a
  quadratic MDE problem on $\esim$ and colored by cell
  type.}\label{f-scrna-quadratic}
\end{figure}

We solved the MDE problem with PyMDE, initializing the solve with a embedding
obtained by solving a quadratic MDE problem on $\esim$. The quadratic MDE
problem was solved in 7 seconds (249 iterations) on our CPU, and 2 seconds (185
iterations) on our GPU. Solving the MDE problem took 27 seconds (300
iterations) on our CPU, and 6 seconds (300 iterations) on our GPU. The
embedding has average distortion 0.13; a CDF of distortions is shown
in figure~\ref{f-scrna-distortions}. The embedding used for
initialization is shown in figure~\ref{f-scrna-quadratic}, with embedding
vectors colored by the cell-type attribute.

\begin{figure}
\centering
\includegraphics{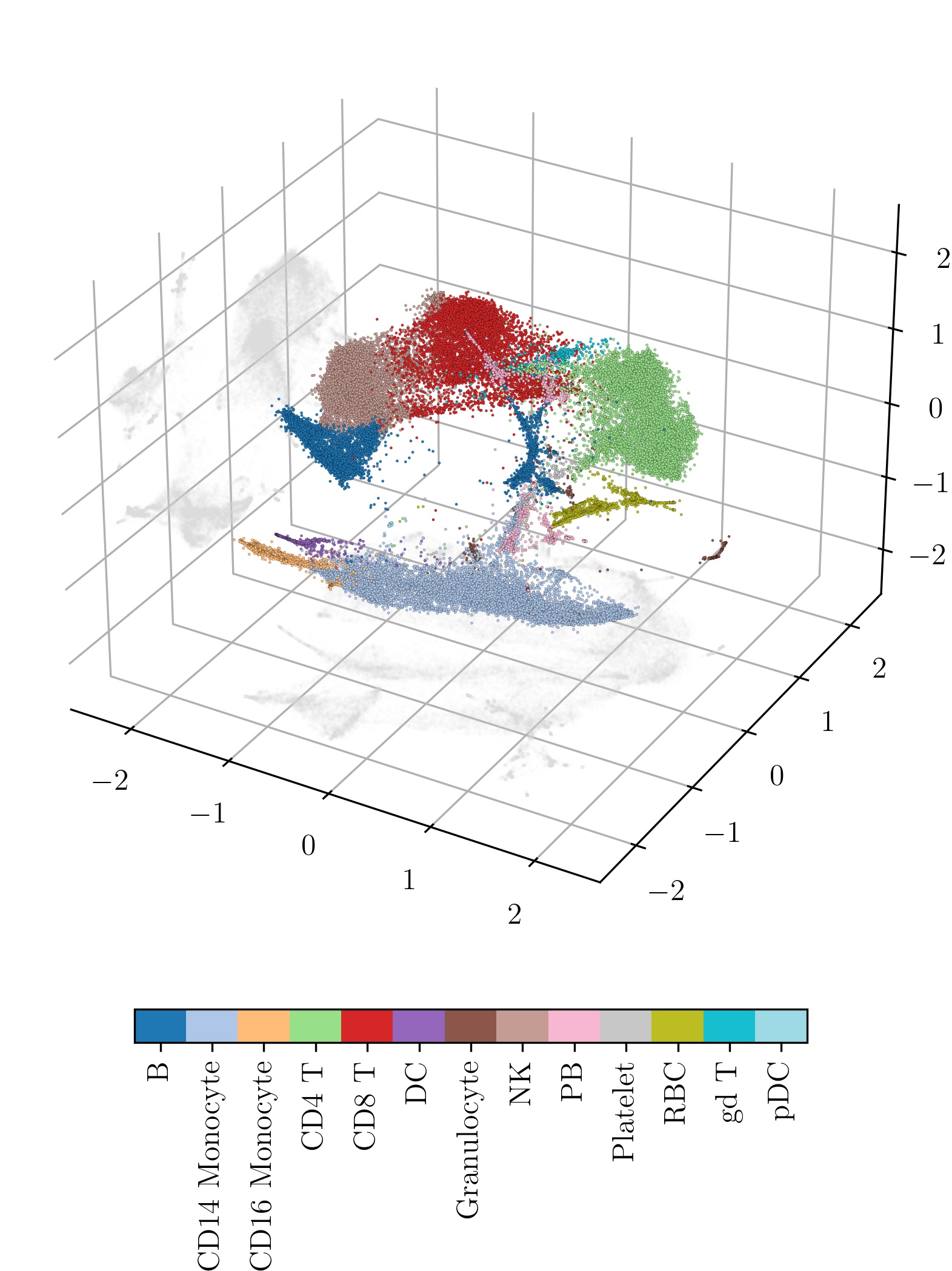}
\caption{\emph{Embedding based on } $\esim \cup \edis$. Embedding of scRNA
  transcriptomes, colored by cell type.
}\label{f-scrna-embedding}
\end{figure}

\begin{figure}
\centering
\includegraphics{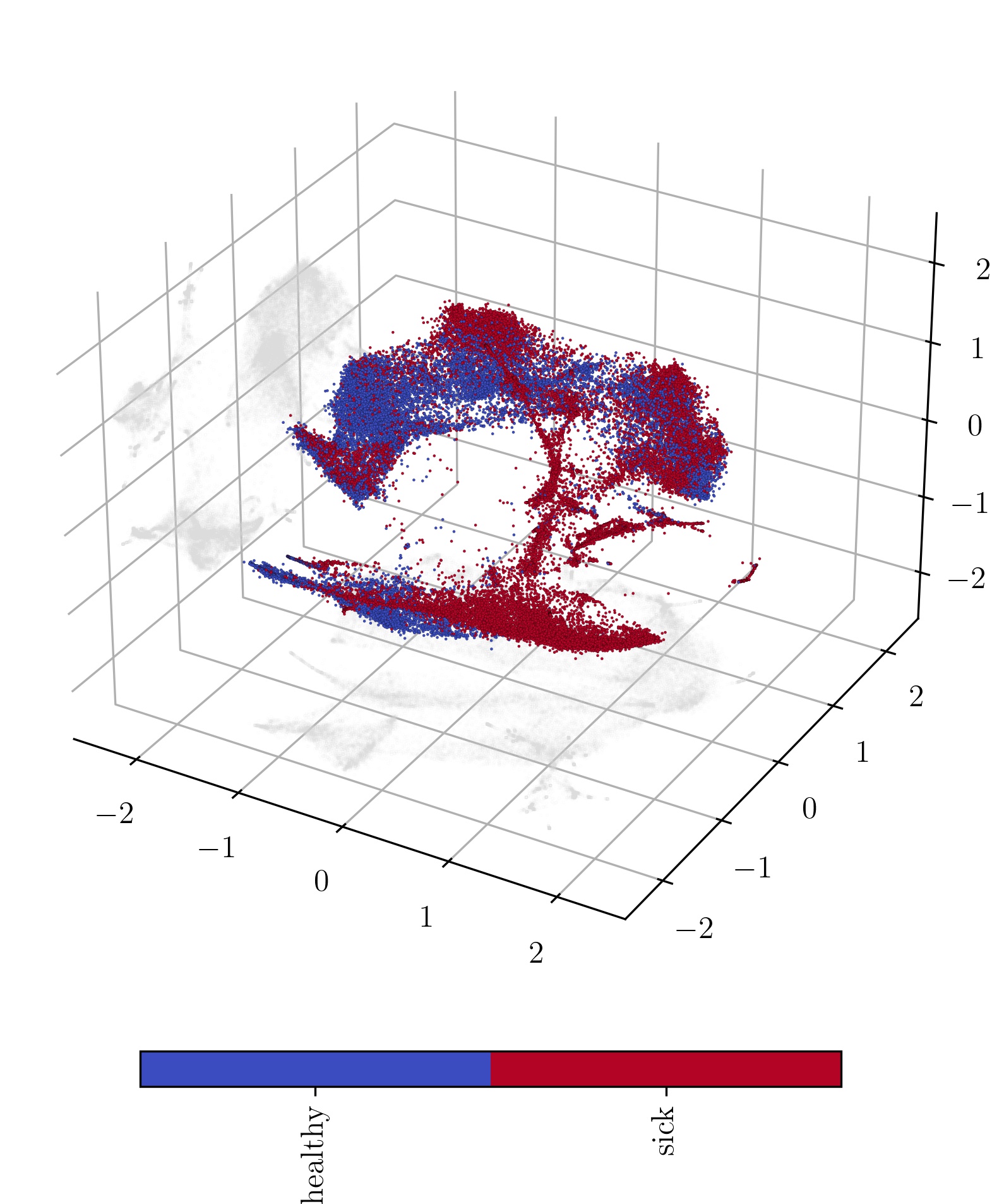}
\caption{\emph{Health status.} Embedding of
scRNA data, colored by health status.}\label{f-scrna-health}
\end{figure}

The embedding is shown in figure~\ref{f-scrna-embedding}, colored by cell type;
similar cells end up near each other in the embedding. The embedding is plotted
again in figure~\ref{f-scrna-health}, this time colored by donor health status.
Cells from healthy donors are nearer to each other than to cells from infected
donors, and the analogous statement is true for cells from infected donors.

As a sanity check, we partition $\esim$ into three sets: 
edges between healthy donors, edges between infected patients,
and edges between infected patients and healthy donors. The
average distortion of the embedding, restricted to each of these
sets, is $0.19$, $0.18$, and $0.21$, respectively. This means 
pairs linking healthy donors to infected donors are more heavily distorted on
average than pairs linking donors with the same health status.

\begin{figure}
\centering
\includegraphics{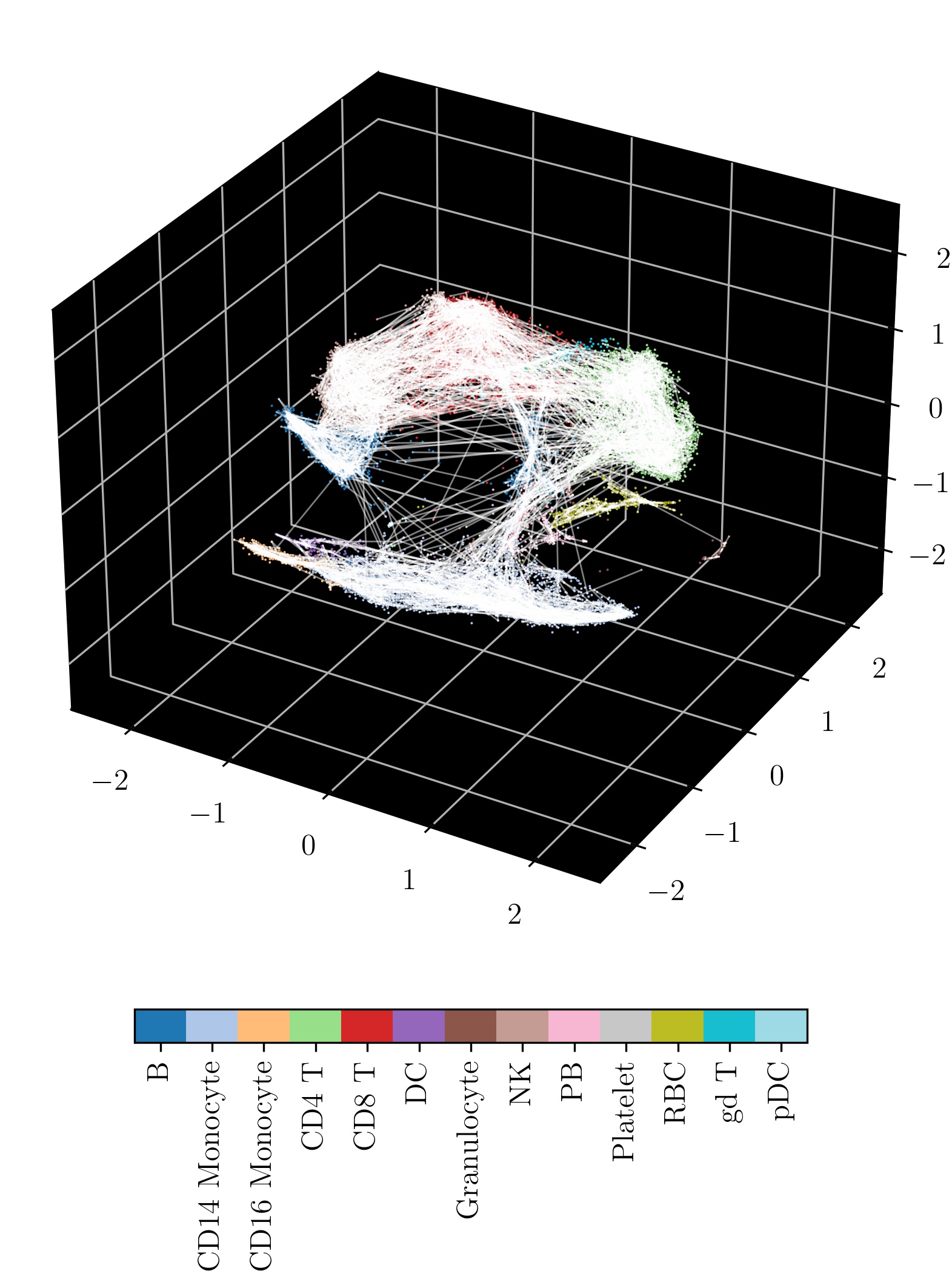}
\caption{\emph{With edges.} Embedding of scRNA data, with edges between
neighboring cells displayed as white line
segments.}\label{f-scrna-embedding-edges}
\end{figure}

\begin{figure}
\centering
\includegraphics{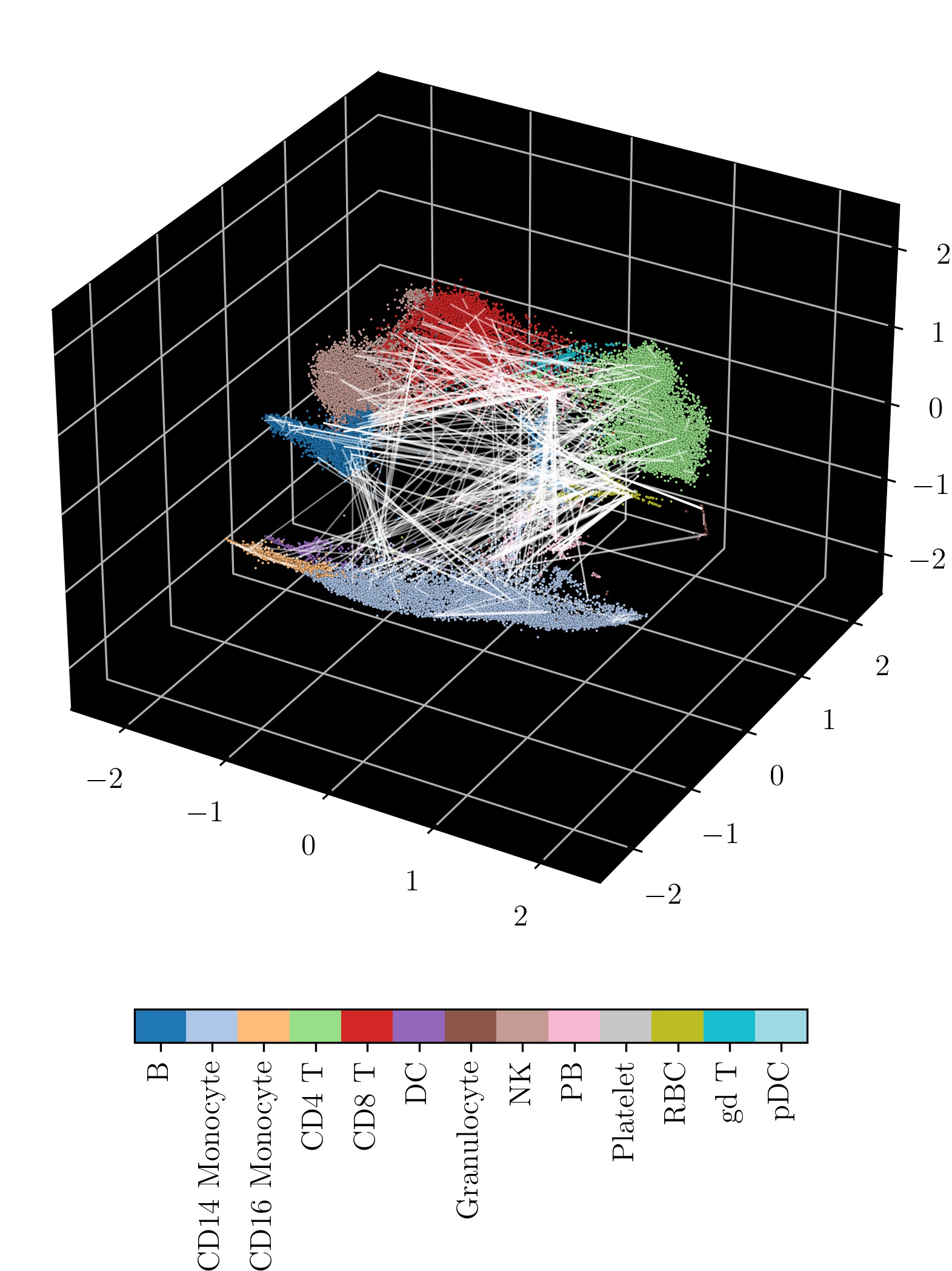}
\caption{\emph{High distortion pairs.} Embedding of scRNA data, with the 1000
most heavily distorted pairs displayed as white line
segments.}\label{f-scrna-embedding-distorted}
\end{figure}

Figure~\ref{f-scrna-embedding-edges} shows the embedding with roughly 10
percent of the pairs in $\esim$ overlaid as white line segments (the pairs were
sampled uniformly at random). Cells near each other are highly connected to
each other, while distant cells are not. Finally, the 1000 pairs from $\esim$
with the highest distortions are shown in
figure~\ref{f-scrna-embedding-distorted}; most of these pairs contain different
types of cells.

\pagebreak
\subsection{Comparison to other methods}
\begin{figure}
\centering
\includegraphics{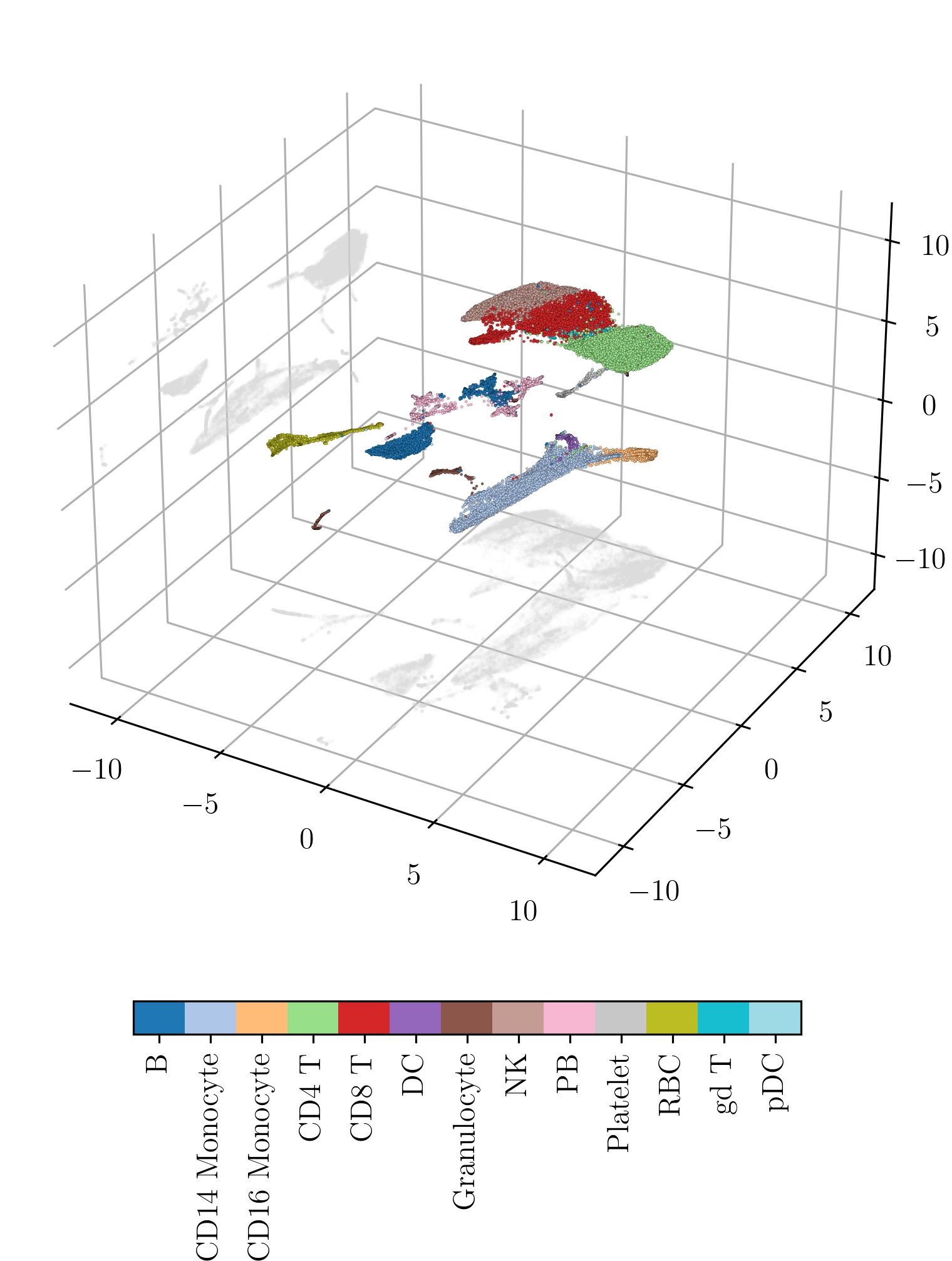}
\caption{\emph{UMAP.} Embedding of scRNA data, produced by UMAP.}
\label{f-scrna-umap}
\end{figure}

\begin{figure}
\centering
\includegraphics{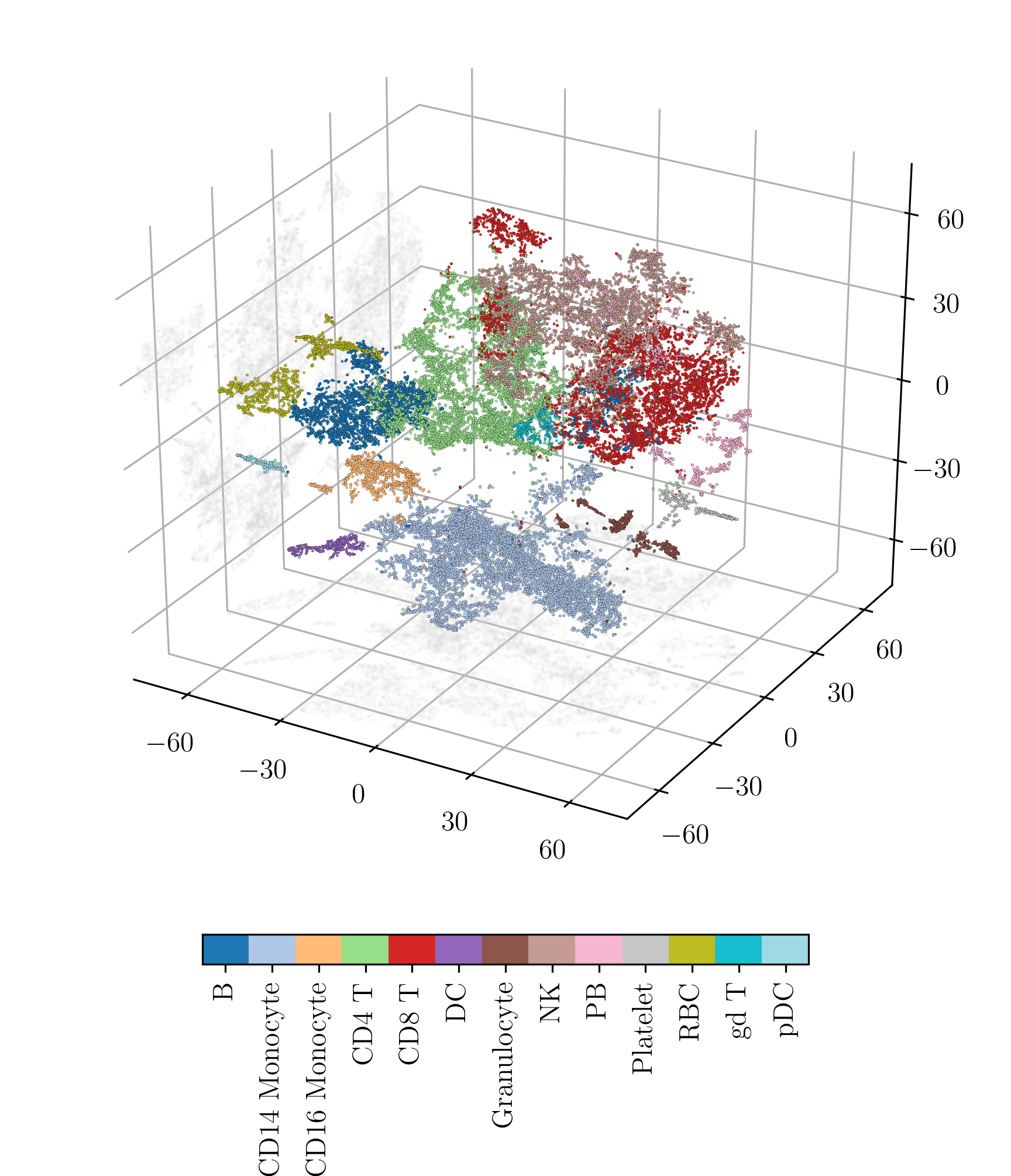}
\caption{\emph{t-SNE.} Embedding of scRNA data, produced by t-SNE.}
\label{f-scrna-tsne}
\end{figure}

Figure~\ref{f-scrna-umap} shows an embedding of the scRNA data produced by
UMAP, which took 11 seconds, and figure~\ref{f-scrna-tsne} shows an
embedding made using openTSNE, which took 96 seconds, using the value
\texttt{bh} for the parameter \texttt{negative\_gradient\_method} (the
default value resulted in an error). The embeddings were aligned to the
orientation of our embedding by solving a Procrustes problem, as described in
\S\ref{s-orth-procrustes}.

\makeatletter
\def\toclevel@chapter{-1}
\makeatother
\addtocontents{toc}{\protect\vspace*{\baselineskip}\protect}
\chapter{Conclusions}
The task of vector embedding, \ie, 
assigning each of a set of items an associated vector, is an old and
well-established problem, with many methods proposed over the past 100 years.
In this monograph we have formalized the problem as one of choosing the vectors
so as to minimize an average or total distortion, where 
distortion functions specified for pairs of items express our preferences
about the Euclidean distance between the associated vectors.
Roughly speaking, we want vectors associated with similar items to be 
near each other,
and vectors associated with dissimilar items to not be near each other.

The distortion functions, which express our prior knowledge about 
similarity and dissimilarity of pairs of items,
can be constructed in several different but related ways.
Weights on edges give us a measure of similarity (when positive)
and dissimilarity (when negative).
Alternatively, we can start with deviations between pairs of items,
with small deviation meaning high similarity.
Similarity and dissimilarity can also be expressed by one or more graphs on
the items, for example one specifying similar pairs and another specifying 
dissimilar pairs. There is some art in making the distortion functions, though
we find that a few simple preprocessing steps, such as the ones described in
this monograph, work well in practice.

Our framework includes a large number of well-known previously developed
methods
as special cases, including PCA, Laplacian embedding, UMAP, multi-dimensional
scaling, and many others. Some other existing embedding methods cannot be represented as
MDE problems, but MDE-based methods can produce similar embeddings. Our
framework can also be used to create new types of embeddings, depending on the
choice of distortion functions, the constraint set, and the preprocessing of
original data.

The quality of an embedding ultimately depends on whether
it is suitable for the downstream application. Nonetheless we can use our
framework to validate, or at least sanity-check, embeddings. For example,
we can examine pairs with abnormally high distortion. We can then see if these
pairs contain anomalous items, or whether their distortion functions
inaccurately conveyed their similarity or lack thereof.

Our examples in part~\ref{p-examples} focused on embedding into two or three
dimensions. This allowed us to plot the embeddings, which we judged by whether
they led to insight into the data, as well as by aesthetics. But we can just as
well embed into dimensions larger than two or three, as might be done when
developing a feature mapping on the items for downstream machine learning
tasks. When the original data records are high-dimensional vectors (say,
several thousand dimensions), we can embed them into five or ten dimensions
and use the embedding vectors as the features; this makes the vector
representation of the data records much smaller, and the fitting problem more
tractable. It can also improve the machine learning, since the embedding
depends on whatever data we used to carry it out, and in some sense inherits
its structure.

It is practical to exactly solve MDE problems only in some special cases, such
as when the distortion functions are quadratic and the embedding vectors are
required to be standardized. For other cases, we have introduced an efficient
local optimization method that produces good embeddings, while placing
few assumptions on the distortion functions and constraint set. We have
shown in particular how to reliably compute embeddings with a standardization
constraint, even when the objective function is not quadratic.

Our optimization method (and software) scales to very large problems, with
embedding dimensions much greater than two or three, and our experiments
show that it is competitive in runtime to more specialized algorithms for
specific embedding methods. The framework of MDE problems, coupled with our
solution method and software, makes it possible for practitioners to rapidly
experiment with new kinds of embeddings in a principled way, without
sacrificing performance.

\clearpage
\subsection*{Acknowledgments}

We thank Amr Alexandari, Guillermo Angeris, Shane Barratt, Steven
Diamond, Mihail Eric, Andrew Knyazev, Benoit Rostykus, Sabera
Talukder, Jian Tang, Jonathan Tuck, Junzi Zhang, as well as several anonymous
reviewers, for their comments on the
manuscript; Yifan Lu and Lawrence Saul, for their careful readings and many
thoughtful suggestions; and Delenn Chin and Dmitry Kobak, for their detailed
feedback on both the manuscript and early versions of the software package.

\printbibliography[heading=bibintoc]
\end{document}